\documentclass{article}

\usepackage[preprint]{neurips_2023}

\usepackage{bbm}
\usepackage[table]{xcolor}
\definecolor{darkred}{rgb}{0.6,0.0,0.0}
\definecolor{darkgreen}{rgb}{0.0,0.8,0.0}
\definecolor{darkblue}{rgb}{0.0,0.0,0.5}
\usepackage{soul}
\usepackage{mdframed}
\usepackage{tcolorbox}
\usepackage{multirow}
\usepackage{booktabs}
\usepackage{threeparttable}
\usepackage{longtable}
\usepackage{enumitem}
\usepackage[ruled,vlined,linesnumbered]{algorithm2e}
\let\oldnl\nl
\newcommand{\nonl}{\renewcommand{\nl}{\let\nl\oldnl}}
\usepackage{titletoc}
\usepackage{hyperref}
\hypersetup{
	colorlinks=true,
	linkcolor={darkblue},
	citecolor={darkblue},
	urlcolor={black}
}
\usepackage{graphicx}
\usepackage{wrapfig}
\usepackage{subcaption}

\usepackage{amsmath,amsfonts,bm, amssymb}
\usepackage{amsthm}
\usepackage{mathtools}
\allowdisplaybreaks

\def\1{\bm{1}}

\def\rve{{\mathbf{e}}}

\def\rvg{{\mathbf{g}}}

\def\rvx{{\mathbf{x}}}
\def\rvy{{\mathbf{y}}}

\def\vp{{\bm{p}}}
\def\vq{{\bm{q}}}

\def\vx{{\bm{x}}}
\def\vy{{\bm{y}}}
\def\vz{{\bm{z}}}

\def\mA{{\bm{A}}}

\def\mX{{\bm{X}}}
\def\mY{{\bm{Y}}}

\DeclareMathAlphabet{\mathsfit}{\encodingdefault}{\sfdefault}{m}{sl}
\SetMathAlphabet{\mathsfit}{bold}{\encodingdefault}{\sfdefault}{bx}{n}

\def\gB{{\mathcal{B}}}

\def\gD{{\mathcal{D}}}
\def\gE{{\mathcal{E}}}

\def\gO{{\mathcal{O}}}

\def\gS{{\mathcal{S}}}

\newcommand{\E}{\mathbb{E}}

\newcommand{\R}{\mathbb{R}}

\newcommand{\Var}{\mathrm{Var}}

\newcommand{\Cov}{\mathrm{Cov}}

\newcommand\inp[2]{\left\langle #1, #2 \right\rangle}
\newcommand{\norm}[1]{\left\lVert#1\right\rVert}
\newcommand{\normsq}[1]{\left\Vert #1 \right\Vert^2}
\newcommand{\Norm}[1]{\lVert#1\rVert}

\newcommand{\sqn}[1]{{\left\lVert#1\right\rVert}^2}
\newcommand{\abs}[1]{\left\lvert#1\right\rvert}

\newcommand{\Term}[2]{\text{Term}_{#1}\ \text{in}\ #2}

\newtheorem{assumption}{Assumption}

\newtheorem{theorem}{Theorem}
\newtheorem{corollary}{Corollary}
\newtheorem{lemma}{Lemma}

\usepackage[noabbrev,sort,capitalize]{cleveref}

\crefname{assumption}{Assumption}{Assumptions}

\title{Convergence Analysis of Sequential Federated Learning on Heterogeneous Data\thanks{The third version is refined based on the corresponding journal version \citep{li2025sharp}. The second version fixes some mistakes in the proofs of the non-convex case in Theorems~1 and 2. The original version was at 37th Conference on Neural Information Processing Systems (NeurIPS 2023).}}

\author{%
  Yipeng Li and Xinchen Lyu\thanks{Xinchen Lyu is the corresponding author.}\\
  National Engineering Research Center for Mobile Network Technologies\\
  Beijing University of Posts and Telecommunications\\
  Beijing, 100876, China\\
  \texttt{\{liyipeng, lvxinchen\}@bupt.edu.cn} \\
}

\begin{document}

\maketitle

\begin{abstract}
	There are two categories of methods in Federated Learning (FL) for joint training across multiple clients: (i) parallel FL (PFL), where clients train models in a parallel manner; and (ii) sequential FL (SFL), where clients train models in a sequential manner. In contrast to that of PFL, the convergence theory of SFL on heterogeneous data is still lacking. In this paper, we establish the convergence guarantees of SFL for strongly/general/non-convex objectives on heterogeneous data. The convergence guarantees of SFL are better than that of PFL on heterogeneous data with both full and partial client participation. Experimental results validate the counterintuitive analysis result that SFL outperforms PFL on extremely heterogeneous data in cross-device settings.
\end{abstract}

\enlargethispage{2\baselineskip}
\section{Introduction}
Federated Learning (FL) \citep{mcmahan2017communication} is a popular distributed machine learning paradigm, where multiple clients collaborate to train a global model. To preserve data privacy and security, data must be kept in clients locally and cannot be shared with others, causing one severe and persistent issue, namely ``data heterogeneity''. In cross-device FL, where data is generated and kept in massively distributed resource-constrained devices (e.g., IoT devices), the negative impact of data heterogeneity would be further exacerbated \citep{jhunjhunwala2023fedexp}.

\begin{figure}[htbp]
	\centering
	\includegraphics[width=0.84\linewidth]{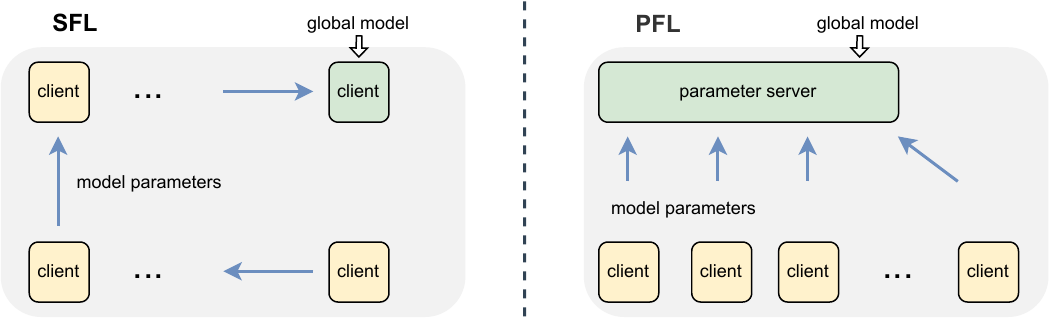}
	\caption{Illustration of SFL and PFL.}
	\label{fig:SFL}
\end{figure}

There are two categories of methods in FL to enable distributed training across multiple clients \citep{qu2022rethinking}: (i) parallel FL (PFL), where models are trained in a parallel manner across clients with synchronization at intervals, e.g., Federated Averaging (\texttt{FedAvg}) \citep{mcmahan2017communication}; and (ii) sequential FL (SFL), where models are trained in a sequential manner across clients, e.g., Cyclic Weight Transfer (\texttt{CWT}) \citep{chang2018distributed}. A simple illustration of SFL and PFL is presented in Figure~\ref{fig:SFL}. However, both categories of methods suffer from the ``client drift'' \citep{karimireddy2020scaffold}, i.e., the local updates on heterogeneous clients would drift away from the right direction, resulting in performance degradation.

Recently, SFL (more generally, the sequential training manner) has attracted much attention in the FL community. Specifically, SFL demonstrates advantages on training speed (in terms of training rounds) \citep{zaccone2022speeding} and small datasets \citep{kamp2023federated}, and both are crucial for cross-device FL. Furthermore, the sequential manner has played a great role in Split learning (SL) \citep{gupta2018distributed,thapa2022splitfed}, an emerging distributed learning technology at the edge side \citep{zhou2019edge}, where the full model is split into client-side and server-side portions to alleviate the excessive computation overhead for resource-constrained devices. Appendix~\ref{sec:app:applicability} will show that the convergence theory in this work is also applicable to SL.

Convergence theory is critical for analyzing the learning performance of algorithms on heterogeneous data in FL. So far, there are numerous works to analyze the convergence of PFL \citep{li2020convergence, khaled2020tighter, koloskova2020unified} on heterogeneous data. However, the convergence theory of SFL on heterogeneous data, given the complexity of its sequential training manner, has not been well investigated in the literature, with only limited preliminary empirical studies \citep{gao2020end, gao2021evaluation}. This paper aims to establish the convergence guarantees for SFL and compare the convergence results of PFL and SFL.

\textbf{Setup.} The basic FL problem is to minimize a global objective function:
\begin{align*}
	\min_{\rvx\in \R^d} \left\{ F(\rvx) \coloneqq \frac{1}{M}\sum_{m=1}^M \left(F_m(\rvx)\coloneqq \E_{\xi \sim \gD_m}[f_m(\rvx; \xi)] \right) \right\},
\end{align*}
where $F_m$, $f_m$ and $\gD_m$ denote the local objective function, the loss function and the local dataset of client $m\in \{1,2,\ldots,M\}$, respectively. 

\textbf{Update rule of SFL (Algorithm~\ref{algorithm1}).}  At the beginning of each training round $r$, the indices $\pi_1^{(r)}, \pi_2^{(r)}, \ldots, \pi_M^{(r)}$ are sampled uniformly without replacement from $\{1,2,\ldots,M\}$ to determine the clients' training order. The permutations are sampled independently across training rounds. Within a round, each client (i) initializes its model with the latest parameters from its previous client, (ii) performs $K$ steps of local updates over its local dataset, and (iii) passes the updated parameter to the next client. This process continues until all clients finish their local training. Let $\rvx_{m,k}^{(r)}$ denote the local parameter of the $m$-th client (i.e., client $\pi_m$) after $k$ local steps in the $r$-th round; let $\rvg_{\pi_m^{(r)}}(\rvx_{m,k}^{(r)})\coloneqq \nabla f_{\pi_m^{(r)}}(\rvx_{m,k}^{(r)};\xi_{m,k}^{(r)})$ denote the stochastic gradient of $F_{\pi_m}$ evaluated at $\rvx_{m,k}^{(r)}$ using the data sample $\xi_{m,k}^{(r)}$; let $\rvx^{(r)}$ denote the global parameter in the $r$-th round. Let $\eta$ denote the learning rate. With SGD (stochastic gradient descent) as the local solver, the update rule of SFL is as follows:
\begin{align*}
	&\text{Local update}: \rvx_{m,k+1}^{(r)} = \rvx_{m,k}^{(r)} - \eta \rvg_{\pi_m^{(r)}}(\rvx_{m,k}^{(r)}),\quad \text{initializing as}\; \rvx_{m,0}^{(r)} =
	\begin{cases}
		\rvx^{(r)}\ , &m=1\\
		\rvx_{m-1,K}^{(r)}\ , &m>1
	\end{cases}\\
	&\text{Global model}: \rvx^{(r+1)} = \rvx_{M,K}^{(r)}.
\end{align*}

\textbf{Update rule of PFL (Algorithm~\ref{algorithm2}).} Within a round, each client (i) initializes its model with the global parameter, (ii) performs $K$ steps of local updates, and (iii) sends the updated parameter to the central server. The server will aggregate the local parameters to generate the global parameter. Let $\rvx_{m,k}^{(r)}$ denote the local parameter of client $m$ after $k$ local steps in the $r$-th round; let $\rvg_{m}(\rvx_{m,k}^{(r)})\coloneqq \nabla f_{m}(\rvx_{m,k}^{(r)};\xi_{m,k}^{(r)})$ denote the stochastic gradient of $F_m$ evaluated at $\rvx_{m,k}^{(r)}$ using the data sample $\xi_{m,k}^{(r)}$; let $\rvx^{(r)}$ denote the global parameter in the $r$-th round. Let $\eta$ denote the learning rate. With SGD as the local solver, the update rule of PFL is as follows:
\begin{align*}
	&\text{Local update}:\ \rvx_{m,k+1}^{(r)} = \rvx_{m,k}^{(r)} - \eta \rvg_m(\rvx_{m,k}^{(r)}),\quad \text{initializing}\ \rvx_{m,0}^{(r)} = \rvx^{(r)},\forall m \in [M]\\
	&\text{Global model}:\ \rvx^{(r+1)} = \frac{1}{M} \sum_{m=1}^M \rvx_{m,K}^{(r)}.
\end{align*}

For SFL, the first subscript of $\rvx_{m,k}^{(r)}$ denotes the client's \emph{position} in the
sequential training order, so $\rvx_{m,k}^{(r)}$ is the local parameter at
client $\pi_m$ after $k$ local steps. In contrast, for PFL, the first subscript of $\rvx_{m,k}^{(r)}$ denotes the \emph{physical client identity}: $\rvx_{m,k}^{(r)}$ is the potential local iterate of client $m$, initialized by $\rvx_{m,0}^{(r)}=\rvx^{(r)}$.

\begin{minipage}[t]{0.5\linewidth}
	\IncMargin{1em}
	\begin{algorithm}[H]
		\DontPrintSemicolon
		\caption{Sequential FL}
		\label{algorithm1}
		\KwOut{$\{\rvx^{(r)}\}_{r=0}^R$}
		\For{training round $r = 0, 1,\ldots, R-1$}{
			Sample a permutation $\pi_1^{(r)}, \pi_2^{(r)}, \ldots, \pi_{M}^{(r)}$ of $\{1,2,\ldots,M\}$\;
			\For{$m = 1,2,\ldots,M$ {\bf\textcolor{red}{in sequence}}}{
				$\rvx_{m,0}^{(r)} =
				\begin{cases}
					\rvx^{(r)}\ , &m=1\\
					\rvx_{m-1,K}^{(r)}\ , &m>1
				\end{cases}$\;
				\For{local step $k = 0,\ldots, K-1$}{
					$\rvx_{m,k+1}^{(r)} = \rvx_{m,k}^{(r)} - \eta \rvg_{\pi_m^{(r)}}(\rvx_{m,k}^{(r)})$\;
				}
			}
			Global model: $\rvx^{(r+1)} = \rvx_{M,K}^{(r)}$\;
		}
	\end{algorithm}
	\DecMargin{1em}
\end{minipage}
\hfill
\begin{minipage}[t]{0.5\linewidth}
	\IncMargin{1em}
	\begin{algorithm}[H]
		\DontPrintSemicolon
		\caption{Parallel FL}
		\label{algorithm2}
		\KwOut{$\{\rvx^{(r)}\}_{r=0}^R$}
		\For{training round $r = 0,1,\ldots, R-1$}{
			\For{$m = 1,2,\ldots,M$ {\bf \textcolor{red}{in parallel}}}{
				$\rvx_{m,0}^{(r)} = \rvx^{(r)}$\;
				\For{local step $k = 0,\ldots, K-1$}{
					$\rvx_{m,k+1}^{(r)} = \rvx_{m,k}^{(r)} - \eta \rvg_{m}(\rvx_{m,k}^{(r)})$\;
				}
			}
			Global model: $\displaystyle \rvx^{(r+1)} = \frac{1}{M} \sum_{m=1}^M \rvx_{m,K}^{(r)}$\;
		}
	\end{algorithm}
	\DecMargin{1em}
\end{minipage}

\section{Contributions}
\textbf{Brief literature review.} The most relevant work is the convergence of PFL and Random Reshuffling (SGD-RR). There are a wealth of works that have analyzed the convergence of PFL on data heterogeneity \citep{li2020convergence, khaled2020tighter, karimireddy2020scaffold, koloskova2020unified, woodworth2020minibatch}, system heterogeneity \citep{wang2020tackling}, partial client participation \citep{li2020convergence, yang2021achieving, wang2022unified} and other variants \citep{karimireddy2020scaffold, reddi2021adaptive}. In this work, we compare the convergence bounds between PFL and SFL (see \cref{subsec:PFL vs SFL}) on heterogeneous data. 

SGD-RR (where data samples are sampled without replacement) is deemed to be more practical than SGD (where data samples are sampled with replacement), and thus attracts more attention recently. \citet{gurbuzbalaban2021random}, \citet{haochen2019random}, \citet{nagaraj2019sgd}, \citet{ahn2020sgd} and \citet{mishchenko2020random} have proved the upper bounds. \cite{safran2020good}, \citet{safran2021random}, \citet{rajput2020closing} and \citet{cha2023tighter} have proved the lower bounds of SGD-RR. In particular, the lower bounds in \cite{cha2023tighter} are shown to match the upper bounds in \cite{mishchenko2020random}. In this work, we use the bounds of SGD-RR to examine the tightness of that of SFL (see \cref{subsec:analysis on SFL}).

Recently, the shuffling-based method has been applied to FL \citep{mishchenko2022proximal,horvath2022fedshuffle,yun2022minibatch,wang2022unified,cho2023convergence,malinovsky2023federated}. In particular, \citet{wang2022unified}, \citet{cho2023convergence} and \cite{malinovsky2023federated} analyzed the convergence of FL with cyclic (or, more generally, regularized) client participation, which can be seen as an extension of SFL. In \citet{wang2022unified} and \citet{cho2023convergence}, when it reduces to SFL, the client training order is deterministic (not random), and thus the analyses cannot be directly extended to our setting. In \cite{malinovsky2023federated}, although their bounds are slightly tighter on the optimization term with SGD-RR as the local solver, their analysis is limited to the case where the number of local steps equals the size of the local data set. In addition, \cite{cho2023convergence} considered upper bounds for PL objective functions and \cite{malinovsky2023federated} considered upper bounds for strongly convex objective functions,\footnote{PL condition can be seen as a non-convex generalization of strong convexity.} while we consider upper bounds for both convex and non-convex cases.

\textbf{Challenges.} In the case of homogeneous data, the task of deriving the convergence guarantees for SFL is relatively straightforward, as SFL reduces to SGD (Stochastic Gradient Descent). However, in the case of heterogeneous data, the task becomes more challenging than existing works, including PFL and SGD-RR (Random Reshuffling), primarily due to the following reasons:
\begin{itemize}
	\item Sequential and shuffling training manner across clients (vs. PFL). In PFL, the local updates at each client only depend on the randomness of the current client within each training round. However, in SFL, the local updates additionally depend on the randomness of all previous clients.
	\item Multiple local update steps at each client (vs. SGD-RR). In contrast to its with-replacement sibling SGD, SGD-RR samples data samples ``without replacement'' and then performs one step of GD (Gradient Descent) on each data sample. Similarly, SFL samples clients without replacement and then performs multiple steps of SGD at each client. In fact, SGD-RR can be regarded as a special case of SFL.
\end{itemize}

\textbf{Contributions.} The main contributions are as follows:
\begin{itemize}
	\item We derive convergence guarantees of SFL for strongly convex, general convex and non-convex objectives on heterogeneous data with the standard assumptions in FL in \cref{subsec:analysis on SFL}.
	\item We compare the convergence guarantees of PFL and SFL, and find a \textit{counterintuitive} comparison result that the guarantee of SFL is better than that of PFL (with both full participation and partial participation) in terms of training rounds on heterogeneous data in \cref{subsec:PFL vs SFL}.
	\item We validate our comparison result with simulations on quadratic functions (\cref{subsec:simulation}) and experiments on real datasets (\cref{subsec:cross-device exp}). The experimental results exhibit that SFL outperforms PFL on extremely heterogeneous data in cross-device settings.
\end{itemize}

\section{Convergence theory}\label{sec:convergence}

\textbf{Notations.} We use $\gO(\cdot)$, $\Omega(\cdot)$ to denote upper and lower bounds up to numerical constant factors, respectively, with $\tilde \gO(\cdot)$ and $\tilde\Omega(\cdot)$ additionally hiding logarithmic factors. In addition, we use $\gtrsim$ to denote ``greater than'' up to some absolute constants and polylogarithmic factors, and $\lesssim$ and $\asymp$ are defined likewise. We use $\norm{\cdot}$ to denote the standard Euclidean norm for both vectors and matrices.

We consider three typical cases for convergence theory, i.e., the strongly convex case, the general convex case and the non-convex case, where all local objectives $F_1,F_2,\ldots, F_M$ are \emph{$\mu$-strongly convex}, \emph{general convex} ($\mu=0$) and \emph{non-convex}.

\subsection{Assumptions}\label{subsec:assumption}
\vspace{-1ex}
We assume that (i) $F$ is lower bounded by $F^\ast$ for all cases and there exists a minimizer $\rvx^\ast$ such that $F(\rvx^\ast)=F^\ast$ for strongly and general convex cases; (ii) each local objective function is $L$-smooth (Assumption~\ref{asm:smoothness}). Furthermore, we need to make assumptions on the diversities: (iii) the assumptions on the stochasticity bounding the diversity of $\{f_m(\cdot;\xi_m^i)\}_i$ with respect to $i$ inside each client $m$ (Assumption~\ref{asm:stochasticity}); (iv) the assumptions on the heterogeneity bounding the diversity of local objectives $\{F_m\}$ with respect to $m$ across clients (Assumptions~\ref{asm:heterogeneity1} and \ref{asm:heterogeneity2}).

\begin{assumption}\label{asm:smoothness}
	Each local objective function $F_m$ is $L$-smooth, $m \in \{1,2,\ldots,M\}$. That is, there exists a constant $L>0$ such that for any $m \in \{1,2,\ldots,M\}$,
	\begin{align*}
		\norm{\nabla F_m(\rvx) - \nabla F_m(\rvy)} \leq L \norm{\rvx - \rvy},\quad \forall \rvx,\rvy\in \R^d.
	\end{align*}
\end{assumption}

We use Assumption~\ref{asm:stochasticity} to bound the stochasticity, where $\sigma$ measures the level of stochasticity (e.g., \citet{stich2019error}, \citet{karimireddy2020scaffold} and \citet{wang2024lightweight}).

\begin{assumption}\label{asm:stochasticity}

	For each client $m$, the stochastic gradient is conditionally unbiased and has bounded variance:
	\begin{align*}
		\E_{\xi\sim\gD_m}
		\left[
		\nabla f_m(\rvx;\xi)
		\mid \rvx
		\right]
		= \nabla F_m(\rvx), \quad \text{and}\quad 
		\E_{\xi\sim\gD_m}
		\left[
		\norm{\nabla f_m(\rvx;\xi)-\nabla F_m(\rvx)}^2
		\mid \rvx
		\right]
		\leq \sigma^2,\forall \rvx.
	\end{align*}
	The stochastic gradient noise $\nabla f_m(\rvx;\xi)-\nabla F_m(\rvx)$, equivalently denoted by $\rvg_m(\rvx)-\nabla F_m(\rvx)$, is independent across training rounds, clients, and local update steps.
\end{assumption}

Now we make assumptions on the diversity of the local objective functions in Assumption~\ref{asm:heterogeneity1} and Assumption~\ref{asm:heterogeneity2}, also known as the heterogeneity in FL \citep{woodworth2020minibatch}. Assumption~\ref{asm:heterogeneity1} is made for non-convex cases, where the constants $\beta$ and $\zeta$ measure the heterogeneity of the local objective functions. If the local objective functions are strongly and general convex, we use one weaker assumption \ref{asm:heterogeneity2} as \cite{koloskova2020unified}, which bounds the diversity only at the optima. Notably, if all local objective functions are identical (i.e., there is no heterogeneity), one can take \(\zeta_\ast=\beta=\zeta=0\) in these assumptions. However, the converse does not necessarily hold, since these quantities characterize only first-order heterogeneity \citep{karimireddy2020scaffold}.

\let\origtheassumption\theassumption
\let\origtheHassumption\theHassumption

\edef\oldassumption{\the\numexpr\value{assumption}+1}
\setcounter{assumption}{0}

\renewcommand{\theassumption}{\oldassumption\alph{assumption}}
\renewcommand{\theHassumption}{\oldassumption.\alph{assumption}}

\begin{assumption}\label{asm:heterogeneity1}
	There exist constants $\beta^2$ and $\zeta^2$ such that
	\begin{align*}
		\textstyle
		\frac{1}{M}\sum_{m=1}^M \norm{\nabla F_m(\rvx)-\nabla F(\rvx)}^2 \leq \beta^2\norm{\nabla F(\rvx)}^2 + \zeta^2.
	\end{align*}
\end{assumption}
\begin{assumption}\label{asm:heterogeneity2}
	Let $\rvx^\ast\in \R^d$ be a minimizer of the global objective function $F$. Define 
	\begin{align*}\textstyle
		\zeta_\ast^2 \coloneqq \frac{1}{M}\sum_{m=1}^M \norm{\nabla F_m(\rvx^\ast)}^2,
	\end{align*}
	where $\zeta_\ast^2$ is assumed to be bounded.
\end{assumption}

\let\theassumption\origtheassumption
\let\theHassumption\origtheHassumption

\subsection{Convergence analysis of SFL}\label{subsec:analysis on SFL}

\begin{theorem}\label{thm:SFL}
	Define $\bar\rvx^{(R)}$ as an iterate constructed from the iterates
	$\{\rvx^{(r)}\}_{r=0}^{R}$ generated by \cref{algorithm1}, with its
	specific construction given for each setting below. Define the effective learning rate $\tilde\eta \coloneqq \eta MK$. Define $D\coloneqq\Norm{\rvx^{(0)}-\rvx^\ast}$ for the convex cases and $A \coloneqq F(\rvx^{(0)}) - F^\ast$ for the non-convex case. Then, we have the following upper bounds:
	\setlist[itemize]{label=}
	\begin{itemize}[leftmargin=0.5em]
		\item \textbf{Strongly convex}: Under \cref{asm:smoothness,asm:stochasticity,asm:heterogeneity2}, there exists $\tilde\eta \leq \frac{1}{6L}$, such that for $M\geq 2$, $K\geq 1$ and $R\geq 6\kappa$ ($\kappa \coloneqq L/\mu$),
		\begin{flalign*}
			\E\left[F(\bar\rvx^{(R)})-F(\rvx^\ast)\right] \leq \frac{9}{2}\mu D^2 \exp\left(-\frac{\mu\tilde\eta R}{2} \right)+\frac{12\tilde{\eta}\sigma^2}{MK}+\frac{18L\tilde{\eta}^2\sigma^2}{MK}+\frac{18L\tilde{\eta}^2\zeta_\ast^2}{M},\label{eq:thm:strongly convex} &&
		\end{flalign*}
		where $\bar\rvx^{(R)} = \frac{1}{\sum_{r=0}^Rw_r} \sum_{r=0}^{R}w_r\rvx^{(r)}$ with $w_r=(1-\frac{\mu\tilde\eta}{2})^{-(r+1)}$.
		\item \textbf{General convex}: Under \cref{asm:smoothness,asm:stochasticity,asm:heterogeneity2}, there exists $\tilde\eta \leq \frac{1}{6L}$, such that for $M\geq 2$, $K\geq 1$ and $R\geq 1$,
		\begin{flalign*}
			\E\left[F(\bar\rvx^{(R)})-F(\rvx^\ast)\right] \leq \frac{3D^2}{\tilde\eta R}+\frac{12\tilde{\eta}\sigma^2}{MK}+\frac{18L\tilde{\eta}^2\sigma^2}{MK}+\frac{18L\tilde{\eta}^2\zeta_\ast^2}{M}, &&
		\end{flalign*}
		where $\bar\rvx^{(R)} = \frac{1}{\sum_{r=0}^Rw_r} \sum_{r=0}^{R}w_r\rvx^{(r)}$ with $w_r=1$.
		\item \textbf{Non-convex}: Under Assumptions~\ref{asm:smoothness},~\ref{asm:stochasticity} and~\ref{asm:heterogeneity1}, there exists $\tilde\eta\leq \frac{1}{6L(1+\beta^2/M)}$, such that for $M\geq 2$, $K\geq 1$ and $R\geq 1$,
		\begin{flalign*}
			\E\left[\Norm{\nabla F(\bar \rvx^{(R)})}^2\right] \leq \frac{10A}{\tilde\eta R} + \frac{20L\tilde\eta\sigma^2}{MK} + \frac{75L^2\tilde\eta^2\sigma^2}{4MK} + \frac{75L^2\tilde\eta^2\zeta^2}{4M}, &&
		\end{flalign*}
		where $\bar\rvx^{(R)}$ is sampled uniformly at random from $\{\rvx^{(r)}\}_{r=0}^{R}$, and the expectation is taken over both the algorithmic randomness and the additional randomness introduced by this sampling procedure.
	\end{itemize}
\end{theorem}

As in \cite{karimireddy2020scaffold} and \citet{wang2020tackling}, we use the \textit{effective learning rate} $\tilde\eta$ in \cref{thm:SFL}. All these upper bounds consist of two parts: the optimization part (the first term) and the error part (the last three terms). Setting $\tilde\eta$ larger can make the optimization part vanish at a higher rate, yet cause the error part to be larger. This implies that we need to choose an appropriate $\tilde\eta$ to achieve a balance between these two parts, which is actually done in Corollary~\ref{cor:SFL}. Here we tune the learning rate with a prior knowledge of the total training rounds $R$, as done in the previous works \citep{karimireddy2020scaffold,koloskova2020unified,reddi2021adaptive}.

\begin{corollary}\label{cor:SFL}
	By choosing an appropriate learning rate (see the proofs of Theorem~\ref{thm:SFL} in Appendices~\ref{subsec:SFL strongly convex},~\ref{subsec:SFL general convex},~\ref{subsec:SFL non-convex}) for the results of Theorem~\ref{thm:SFL}, we can obtain the upper bounds of SFL:
	\setlist[itemize]{label=}
	\begin{itemize}[leftmargin=0.5em]
		\item \textbf{Strongly convex}: When $\tilde\eta=\eta MK \asymp  \min \{\frac{1}{L},\frac{1}{\mu R} \}$ for Theorem~\ref{thm:SFL}, then for $R\gtrsim \kappa$
		\begin{flalign*}
			\E\left[F(\bar\rvx^{(R)})-F(\rvx^\ast)\right] = \tilde\gO\left(\frac{\sigma^2}{\mu MKR} + \frac{L\sigma^2}{\mu^2MKR^2} + \frac{L\zeta_\ast^2}{\mu^2MR^2} + \mu D^2 \exp\left(\frac{-\mu R}{L}\right)\right).&&
		\end{flalign*}
		\item \textbf{General convex}: When $\tilde\eta=\eta MK \asymp  \min \{\frac{1}{L},\frac{D}{c_1^{1/2}R^{1/2}},\frac{D^{2/3}}{c_2^{1/3}R^{1/3}} \}$ with $c_1 \asymp \frac{\sigma^2}{MK}$ and $c_2 \asymp \frac{L\sigma^2}{MK} + \frac{L\zeta_\ast^2}{M}$ for Theorem~\ref{thm:SFL}, then
		\begin{flalign*}
			\E\left[F(\bar\rvx^{(R)})-F(\rvx^\ast)\right] = \gO\left(\frac{\sigma D}{\sqrt{MKR}} + \frac{\left(L\sigma^2D^4\right)^{1/3}}{(MK)^{1/3}R^{2/3}} + \frac{\left(L\zeta_\ast^2D^4\right)^{1/3}}{M^{1/3}R^{2/3}} + \frac{LD^2}{R}\right). &&
		\end{flalign*}
		\item \textbf{Non-convex}: When $\tilde\eta=\eta MK \asymp  \min \{\frac{1}{L(1+\beta^2/M)},\frac{A^{1/2}}{c_1^{1/2}R^{1/2}},\frac{A^{1/3}}{c_2^{1/3}R^{1/3}} \}$ with $c_1 \asymp \frac{L\sigma^2}{MK}$ and $c_2 \asymp \frac{L^2\sigma^2}{MK} + \frac{L^2\zeta^2}{M}$ for Theorem~\ref{thm:SFL}, then
		\begin{flalign*}
			\E\left[\Norm{\nabla F(\bar \rvx^{(R)})}^2\right] = \gO\left(\frac{ \left(L\sigma^2A\right)^{1/2}}{\sqrt{MKR}} + \frac{\left(L^2\sigma^2A^2\right)^{1/3}}{(MK)^{1/3}R^{2/3}} + \frac{\left(L^2\zeta^2A^2\right)^{1/3}}{M^{1/3}R^{2/3}} + \frac{LA(1+\frac{\beta^2}{M})}{R}\right). &&
		\end{flalign*}
	\end{itemize}
\end{corollary}

By Corollary~\ref{cor:SFL}, for sufficiently large $R$, the convergence rate is determined by the first term for all cases, resulting in convergence rates of $\tilde\gO(1/MKR)$ for strongly convex cases, $\gO(1/\sqrt{MKR})$ for general convex cases and $\gO(1/\sqrt{MKR})$ for non-convex cases.

Recall that SGD-RR can be seen as one special case of SFL, where one step of GD is performed on each local objective $F_m$, which implies $K=1$ and $\sigma=0$. We now compare the upper bounds of SFL with those of SGD-RR to examine the tightness. As shown in \cite{mishchenko2020random}'s Corollaries~1, 2 and 3, the upper bounds of SGD-RR are $\tilde\gO\left({\color{blue}\frac{L}{\mu}}\left(\frac{L\zeta_\ast^2}{\mu^2MR^2} + \mu D^2\exp\left(\frac{-\mu {\color{red}\boldsymbol{M}}R}{L}\right)\right)\right)$, $\gO\left(\frac{\left(L\zeta_\ast^2D^4\right)^{1/3}}{M^{1/3}R^{2/3}} + \frac{LD^2}{R}\right)$, $\gO\left(\frac{\left(L^2\zeta^2A^2\right)^{1/3}}{M^{1/3}R^{2/3}} + \frac{L A}{R}\right)$ for the strongly convex, general convex and non-convex cases, respectively. We see that our bounds match those of SGD-RR in the general convex, and non-convex cases. For the strongly convex case, the bound of SGD-RR shows an advantage on the optimization term (marked in red). This advantage is due to the advanced technique of Shuffling Variance introduced by \citet[][Definition~2]{mishchenko2020random}. We leave the investigation on introducing this advanced technique to SFL for future work.

Corollary~\ref{cor:SFL} yields two implications regarding the effect of local updates: (i) It can be seen that local updates can help the convergence with proper learning rate choices (small enough). Yet this increases the total steps (iterations), leading to a higher computation cost. (ii) Excessive local updates do not benefit the dominant term of the convergence rate. Take the strongly convex case as an example. When $\frac{\sigma^2}{\mu MKR} \leq \frac{L\zeta_\ast^2}{\mu^2MR^2}$, the latter turns dominant, which is unaffected by $K$. In other words, when the value of $K$ exceeds $\tilde\Omega\left(\sigma^2/\zeta_\ast^2\cdot \mu/L \cdot R\right)$, increasing local updates will no longer benefit the dominant term of the convergence rate. Note that the maximum value of $K$ is affected by $\sigma^2/\zeta_\ast^2$, $\mu/L$ and $R$. This analysis follows \cite{reddi2021adaptive} and \citet{khaled2020tighter}.

\subsection{PFL vs. SFL on heterogeneous data}\label{subsec:PFL vs SFL}

\begin{table}[phtb]
	\centering
	\renewcommand{\arraystretch}{1}
	\caption{Upper bounds of PFL and SFL with numerical constants and polylogarithmic factors omitted. Main differences are marked in red fonts.}
	\label{tab:comparison}
	\resizebox{\linewidth}{!}{
		\begin{threeparttable}[b]
			\begin{tabular}{p{13.5em}l}
				\toprule
				Method &Upper Bound\\\midrule
				\multicolumn{2}{c}{Strongly Convex}\\
				\midrule
				PFL \vspace{-2ex}\\
				\quad\ \citep{karimireddy2020scaffold} &$\frac{\sigma^2}{\mu MKR} + \frac{L\sigma^2}{\mu^2KR^2} + \frac{L\zeta^2}{\mu^2 R^2} + \mu D^2 \exp\left(\frac{-\mu R}{L}\right)
				$ \tnote{\color{black}(1)}\\[1.5ex]
				\quad\ \citep{koloskova2020unified} &$\frac{\sigma_\ast^2}{\mu MKR} + \frac{L\sigma_\ast^2}{\mu^2KR^2} + \frac{L\zeta_\ast^2}{\mu^2 R^2} + LKD^2 \exp\left(\frac{-\mu R}{L}\right)
				$ \tnote{\color{black}(2)}\\[1.5ex]
				\quad\ (Theorem~\ref{thm:PFL}) &$\frac{\sigma^2}{\mu MKR} + \frac{L\sigma^2}{\mu^2KR^2} + \frac{L\zeta_\ast^2}{\mu^2 R^2} + \mu D^2 \exp\left(\frac{-\mu R}{L}\right)
				$\\[1.5ex]
				SFL (Theorem~\ref{thm:SFL}) &$\frac{\sigma^2}{\mu MKR} + \frac{L\sigma^2}{\mu^2{\color{red}\boldsymbol{M}}KR^2} + \frac{L\zeta_\ast^2}{\mu^2{\color{red}\boldsymbol{M}}R^2} + \mu D^2 \exp\left(\frac{-\mu R}{L}\right)$\\\midrule
				
				\multicolumn{2}{c}{Convex}\\
				\midrule
				PFL \vspace{-2ex}\\
				\quad\ \citep{karimireddy2020scaffold} &$\frac{\sigma D}{\sqrt{MKR}} + \frac{\left(L\sigma^2D^4\right)^{1/3}}{K^{1/3}R^{2/3}} + \frac{\left(L\zeta^2D^4\right)^{1/3}}{R^{2/3}} + \frac{LD^2}{R}
				$\\[1.5ex]
				\quad\ \citep{koloskova2020unified} &$\frac{\sigma_\ast D}{\sqrt{MKR}} + \frac{\left(L\sigma_\ast^2D^4\right)^{1/3}}{K^{1/3}R^{2/3}} + \frac{\left(L\zeta_\ast^2D^4\right)^{1/3}}{R^{2/3}} + \frac{LD^2}{R}
				$\\[1.5ex]
				SFL (Theorem~\ref{thm:SFL}) &$\frac{\sigma D}{\sqrt{MKR}} + \frac{\left(L\sigma^2D^4\right)^{1/3}}{{\color{red}\boldsymbol{M^{1/3}}}K^{1/3}R^{2/3}} + \frac{\left(L\zeta_\ast^2D^4\right)^{1/3}}{{\color{red}\boldsymbol{M^{1/3}}}R^{2/3}} + \frac{LD^2}{R}$\\\midrule
				
				\multicolumn{2}{c}{Non-convex}\\
				\midrule
				PFL \citep{karimireddy2020scaffold, koloskova2020unified} &$\frac{ \left(L\sigma^2A\right)^{1/2}}{\sqrt{MKR}} + \frac{\left(L^2\sigma^2A^2\right)^{1/3}}{K^{1/3}R^{2/3}} + \frac{\left(L^2\zeta^2A^2\right)^{1/3}}{R^{2/3}} + \frac{LA}{R}
				$ \tnote{\color{black}(3)}\\
				SFL (Theorem~\ref{thm:SFL}) &$\frac{ \left(L\sigma^2A\right)^{1/2}}{\sqrt{MKR}} + \frac{\left(L^2\sigma^2A^2\right)^{1/3}}{{\color{red}\boldsymbol{M^{1/3}}}K^{1/3}R^{2/3}}+\frac{\left(L^2\zeta^2A^2\right)^{1/3}}{{\color{red}\boldsymbol{M^{1/3}}}R^{2/3}} + \frac{LA}{R}$\tnote{\color{black}(4)}\\
				\bottomrule
			\end{tabular}
			\begin{tablenotes}
				\begin{footnotesize}
					\item [\tnote{\color{black}(1)}] (i) We use $\frac{3L\eta^3 K^3 \sigma^2}{K}$ (see the last inequality of the proof of their Lemma~8) while \cite{karimireddy2020scaffold} use $\frac{\eta^2 K^2 \sigma^2}{2K}$ with $\eta \leq \frac{1}{8LK}$ (their Lemma~8), which causes the difference between their original bounds and our recovered bounds. (ii) This difference also exists in the other two cases. (iii) Their Assumption~A1 is essentially equivalent to Assumption~\ref{asm:heterogeneity1}. For simplicity, we let $B=1$ in their Assumption~A1 for all three cases.
					\item [\tnote{\color{black}(2)}] \cite{koloskova2020unified} use $\sigma_\ast^2 \coloneqq \frac{1}{M}\sum_{m=1}^M \E \left[\norm{\nabla f_m(\rvx^\ast;\xi)-\nabla F_m(\rvx^\ast)}^2\right]$ to bound the stochasticity, which is weaker than Assumption~\ref{asm:stochasticity}.
					\item [\tnote{\color{black}(3)}] We let $P=1, M=0$ in \cite{koloskova2020unified}'s Assumption~3b.
					\item [\tnote{\color{black}(4)}] We let $\beta=0$ in Assumption~\ref{asm:heterogeneity1}.
				\end{footnotesize}
			\end{tablenotes}
	\end{threeparttable}}
\end{table}

Unless otherwise stated, our comparisons are in terms of training rounds, which is also adopted in \cite{gao2020end,gao2021evaluation}. This comparison (running for the same total training rounds $R$) is fair considering the same total computation cost for both methods.

For fair comparison, we slightly improve the upper bounds of PFL for strongly convex cases by combining the works of \cite{karimireddy2020scaffold} and \citet{koloskova2020unified}. In addition, we note that the proofs in \cite{koloskova2020unified}, developed for a unified theory of Decentralized SGD, differ somewhat from those in other works focusing on PFL. Therefore, we reproduce the PFL bounds for the general convex and non-convex cases based on \cite{karimireddy2020scaffold} and \citet{koloskova2020unified}. The resulting bounds are presented in Theorem~\ref{thm:PFL} in Appendix~\ref{sec:proof PFL}. \cref{tab:comparison} summarizes the convergence results for PFL and SFL.

We note that \citet{woodworth2020minibatch} provided a better bound on the optimization term for PFL under a stronger assumption than \cref{asm:heterogeneity1,asm:heterogeneity2} \citep[Equation~(12)]{woodworth2020minibatch} in the convex cases. Detailed comparisons with \citet{woodworth2020minibatch} can be found in \citet{li2025sharp}.

As shown in Table~\ref{tab:comparison}, the upper bound of SFL is better than that of PFL, with an advantage of $1/M$ on the second and third terms (marked in red). This benefits from its sequential and shuffling-based training manner of SFL.

In the more challenging cross-device settings, only a small fraction of clients participate in each training round. Following \cite{karimireddy2020scaffold} and \citet{yang2021achieving}, we provide the upper bounds for PFL and SFL with partial client participation as follows:
\begin{align*}
	\text{PFL:} \quad &\tilde\gO\left(\frac{\sigma^2}{\mu SKR} + {\color{black}\frac{\zeta_\ast^2}{\mu R}\frac{M-S}{S(M-1)}} + \frac{L\sigma^2}{\mu^2KR^2} + \frac{L\zeta_\ast^2}{\mu^2 R^2} + \mu D^2 \exp\left(\frac{-\mu R}{L}\right)\right), \\
	\text{SFL:} \quad&\tilde\gO\left(\frac{\sigma^2}{\mu SKR}+{\color{black}\frac{\zeta_\ast^2}{\mu R}\frac{(M-S)}{S(M-1)}} + \frac{L\sigma^2}{\mu^2{\color{red}\boldsymbol{S}}KR^2} + \frac{L\zeta_\ast^2}{\mu^2{\color{red}\boldsymbol{S}}R^2}+\mu D^2 \exp\left(\frac{-\mu R}{L}\right)\right),
\end{align*}
where a subset of clients $\gS$ (its size is $\abs{\gS}=S$) are selected randomly without replacement in each training round. There are additional terms (the second terms) for both PFL and SFL, which is due to partial client participation and random sampling \citep{yang2021achieving}. \textit{It can be seen that the advantage of $1/S$ (marked in red) of SFL still exists, similar to the full client participation setup.}

\section{Experiments}\label{sec:exp}
We run experiments on quadratic functions (\cref{subsec:simulation}) and real datasets (\cref{subsec:cross-device exp}) to validate our theory. The main findings are (i) in extremely heterogeneous settings, SFL performs better than PFL, (ii) while in moderately heterogeneous settings, this may not be the case. The code is available at \url{https://github.com/liyipeng00/convergence}.

\subsection{Experiments on quadratic functions}\label{subsec:simulation}
According to Table~\ref{tab:comparison}, SFL outperforms PFL on heterogeneous settings (in the worst case). Here we show that the counterintuitive result can appear even for simple one-dimensional quadratic functions.

\begin{figure}[htbp]
	\vspace{-2ex}
	\centering
	\includegraphics[width=0.32\linewidth]{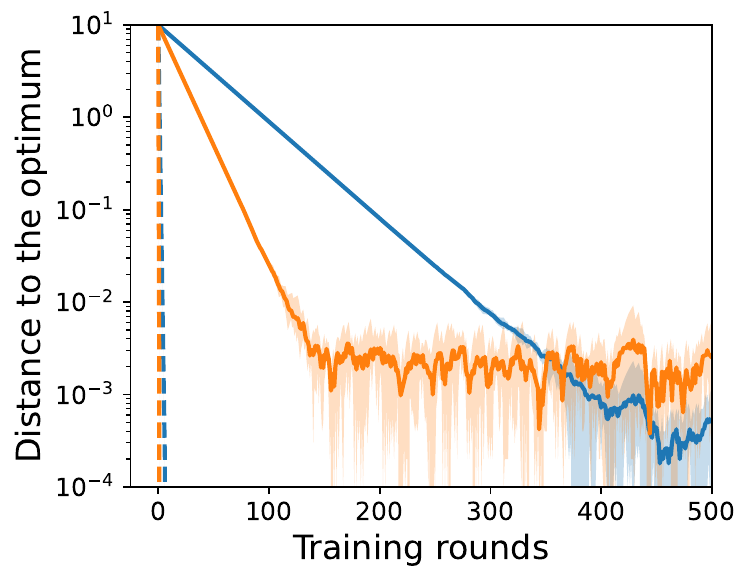}\hfill	\includegraphics[width=0.32\linewidth]{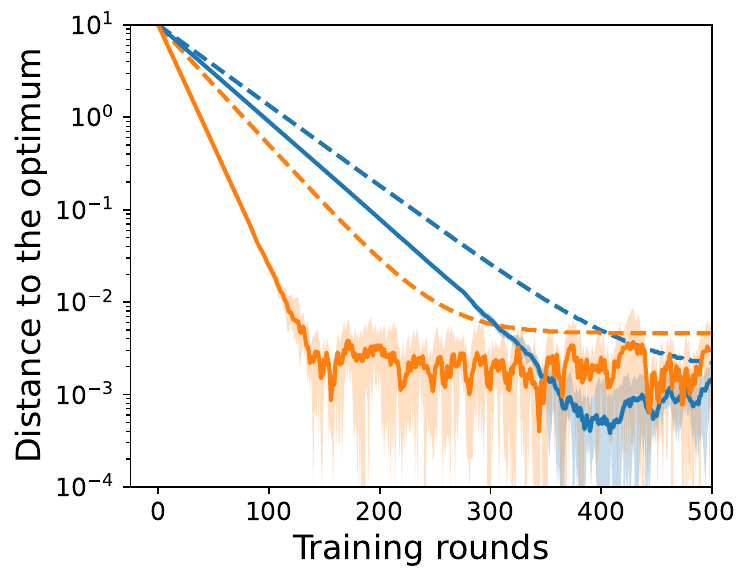}\hfill
	\includegraphics[width=0.32\linewidth]{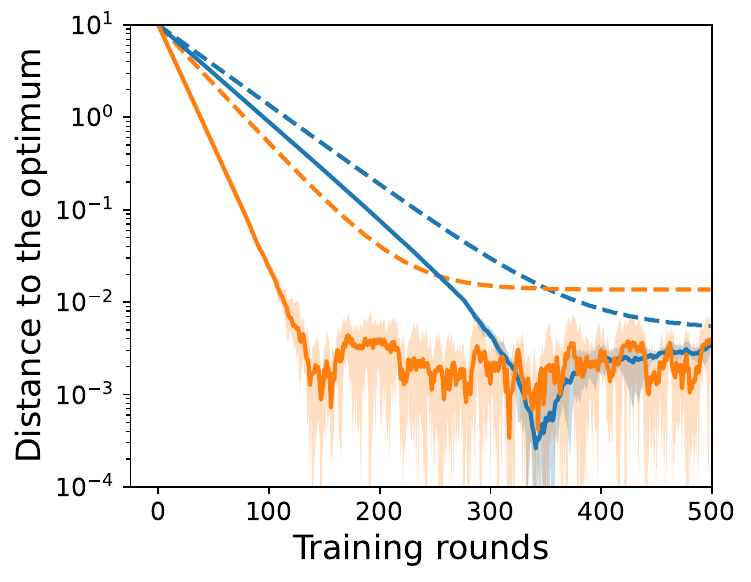}
	\caption{Simulations on quadratic functions. It displays the experimental results from Group 1 to Group 3 in Table~\ref{tab:simulation settings} from left to right. Shaded areas show the min-max values.}
	\label{fig:simulation}
\end{figure}

\textbf{Results of simulated experiments.} As shown in Table~\ref{tab:simulation settings}, we use four groups of experiments with various degrees of heterogeneity. To further catch the heterogeneity, in addition to Assumption~\ref{asm:heterogeneity2}, we also use bounded Hessian heterogeneity in \cite{karimireddy2020scaffold}:
\[
\max_{m} \norm{\nabla^2 F_m (\rvx) - \nabla^2 F(\rvx)}\leq \delta .
\]
Choosing larger values of $\zeta_\ast$ and $\delta$ means higher heterogeneity. The experimental results of Table~\ref{tab:simulation settings} are shown in Figure~\ref{fig:simulation}. When $\zeta_\ast=1$ and $\delta=0$, the heterogeneity has no bad effect on the performance of PFL while hurts that of SFL significantly (Group 1). When the heterogeneity continues to increase to $\delta>0$, SFL outperforms PFL with a faster rate and better result (Groups 2 and 3). This in fact tells us that the comparison between PFL and SFL can be associated with the data heterogeneity, and SFL outperforms PFL when meeting high data heterogeneity, which coincides with our theoretical conclusion.

\textbf{Limitation and intuitive explanation.} The bounds (see Table~\ref{tab:comparison}) above suggest that SFL outperforms PFL regardless of heterogeneity (the value of $\zeta_\ast$), while the simulated results show that it only holds in extremely heterogeneous settings. This inconsistency is because existing theoretical works \citep{karimireddy2020scaffold,koloskova2020unified} with Assumptions~\ref{asm:heterogeneity1} and \ref{asm:heterogeneity2} may underestimate the capacity of PFL, where the function of global aggregation is omitted. In particular, \cite{wang2024unreasonable} have provided rigorous analyses showing that PFL performs much better than the bounds suggest in moderately heterogeneous settings. Hence, the comparison turns vacuous under this condition. Intuitively, PFL updates the global model less frequently with more accurate gradients (with the global aggregation). In contrast, SFL updates the global model more frequently with less accurate gradients. In extremely heterogeneous settings (gradients of both are inaccurate), the benefits of frequent updates become dominant, and thus SFL outperforms PFL. In moderately heterogeneous settings, it's the opposite.

\begin{table}[h!]
	\renewcommand{\arraystretch}{1}
	\centering
	\caption{Settings of simulated experiments. Each group has two local objectives (i.e., $M=2$) and shares the same global objective. The heterogeneity increases from Group 1 to Group 3.}
	\label{tab:simulation settings}
		\begin{tabular}{ccccc} 
			\toprule
			 &Group 1 &Group 2 &Group 3\\
			\midrule
			Settings
				     &$\begin{cases}F_1(x)=\frac{1}{2}x^2 + x\\ F_2(x)=\frac{1}{2}x^2 - x\end{cases}$ 
				     &$\begin{cases}F_1(x)=\frac{2}{3}x^2 + x\\ F_2(x)=\frac{1}{3}x^2 - x\end{cases}$
				     &$\begin{cases}F_1(x)=x^2 + x\\ F_2(x) = -x\end{cases}$\\[2ex]
			$\zeta_\ast, \delta$  
			&$\zeta_\ast=1, \delta=0$ &$\zeta_\ast=1, \delta=\frac{1}{3}$ &$\zeta_\ast=1, \delta=1$\\
			\bottomrule
	\end{tabular}
\end{table}

\subsection{Experiments on real datasets}\label{subsec:cross-device exp}
\textbf{Extended Dirichlet strategy.} This is to generate arbitrarily heterogeneous data across clients by extending the popular Dirichlet-based data partition strategy \citep{yurochkin2019bayesian,hsu2019measuring}. The difference is to add a step of allocating classes (labels) to determine the number of classes per client (denoted by $C$) before allocating samples via Dirichlet distribution (with concentrate parameter $\alpha$). Thus, the extended strategy can be denoted by $\text{ExDir}(C,\alpha)$. The implementation is as follows (with more details deferred to Appendix~\ref{subsec:app:exdir}):
\begin{itemize}[leftmargin=2em]
	\item Allocating classes: We randomly allocate $C$ different classes to each client. After assigning the classes, we can obtain the prior distribution $\vq_c$ for each class $c$.
	\item Allocating samples: For each class $c$, we draw $\vp_c\sim \text{Dir}(\alpha \vq_c)$ and then allocate a $\vp_{c,m}$ proportion of the samples of class $c$ to client $m$. For example, $\vq_c = [1, 1, 0, 0, \ldots,]$ means that the samples of class $c$ are only allocated to the first 2 clients.
\end{itemize}

\textbf{Experiments in cross-device settings.} We next validate the theory in cross-device settings \citep{kairouz2021advances} with partial client participation on real datasets. 

\textit{Setup.} We consider the common CV tasks training VGGs \citep{simonyan2014very} and ResNets \citep{he2016deep} on CIFAR-10 \citep{krizhevsky2009learning} and CINIC-10 \citep{darlow2018cinic10}. Specifically, we use the models VGG-9 \citep{lin2020ensemble} and ResNet-18 \citep{acar2021federated}. We partition the training sets of CIFAR-10 into 500 clients / CINIC-10 into 1000 clients by $\text{ExDir}(1,10.0)$ and $\text{ExDir}(2,10.0)$; and spare the test sets for computing test accuracy. As both partitions share the same parameter $\alpha=10.0$, we use $C=1$ (where each client owns samples from one class) and $C=2$ (where each client owns samples from two classes) to represent them, respectively. Note that these two partitions are not rare in FL \citep{li2022federated}. They are called \emph{extremely heterogeneous} data and \emph{moderately heterogeneous} data respectively in this paper. We fix the number of participating clients to 10 and the mini-batch size to 20. The local solver is SGD with learning rate being constant, momentum being 0 and weight decay being 1e-4. We apply gradient clipping to both algorithms (Appendix~\ref{subsec:app:gradient clipping}) and tune the learning rate by grid search (Appendix~\ref{subsec:app:grid search}).

\begin{figure}[htbp]
	\centering
	\begin{minipage}{0.44\textwidth}
		\centering
		\includegraphics[width=\linewidth]{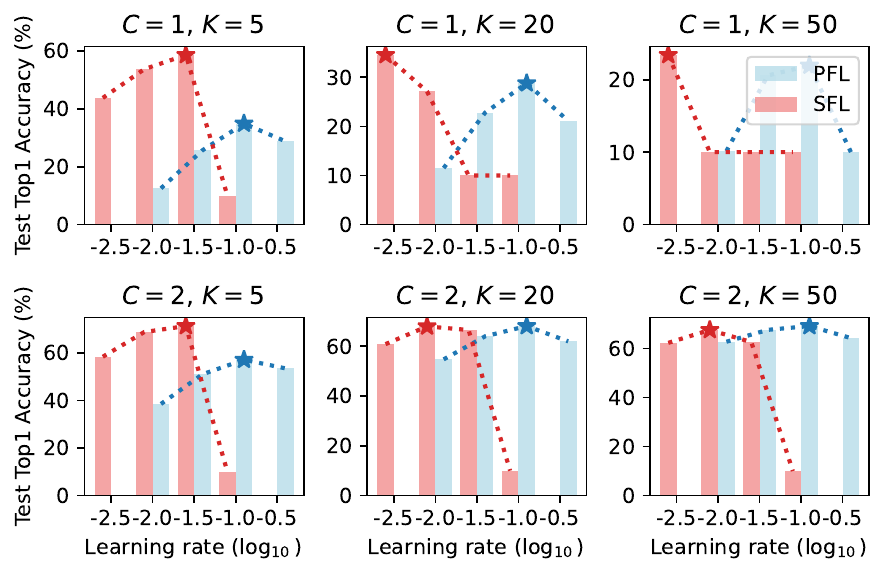}
		\caption{Test accuracies after training VGG-9 on CIFAR-10 for 1000 training rounds with different learning rates.}
		\label{fig:best learning rate}
	\end{minipage}
	\hfill
	\begin{minipage}{0.52\textwidth}
		\centering
		\includegraphics[width=0.48\linewidth]{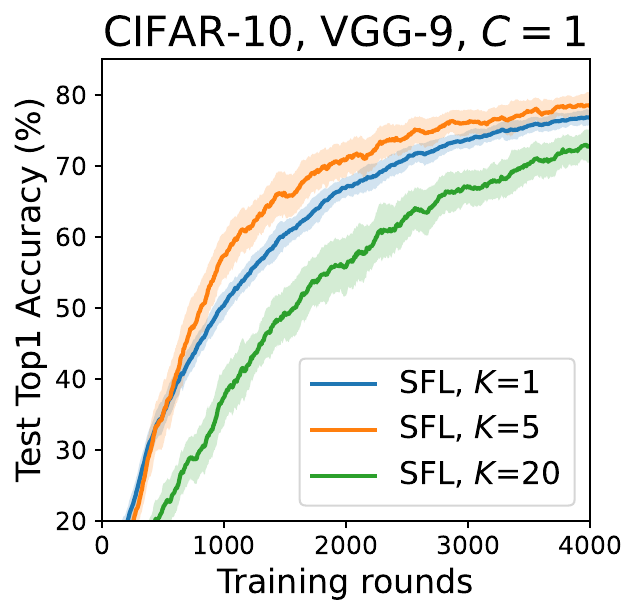}
		\includegraphics[width=0.48\linewidth]{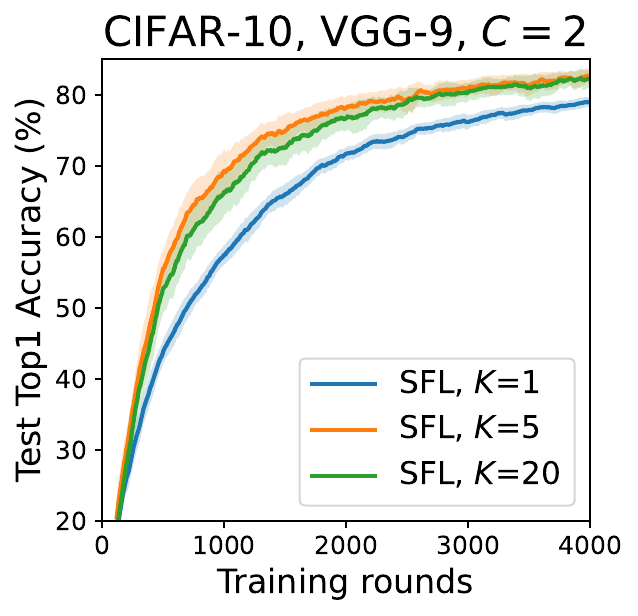}
		\caption{Effect of local steps.}
		\label{fig:local steps}
	\end{minipage}
\end{figure}

\textit{The best learning rate of SFL is smaller than that of PFL.} We have the following observations from Figure~\ref{fig:best learning rate}: (i) the best learning rates of SFL is smaller than that of PFL (by comparing PFL and SFL), and (ii) the best learning rate of SFL becomes smaller as data heterogeneity increases (by comparing the top row and bottom row). These observations are critical for hyperparameter selection.

\textit{Effect of local steps.} Figure~\ref{fig:local steps} is aimed to study the effects of local steps. In both plots, it can be seen that the performance of SFL improves as $K$ increases from 1 to 5. This validates the theoretical conclusion that local steps can help the convergence of SFL even on heterogeneous data. Then, the performance of SFL deteriorates as $K$ increases from 5 to 10, whereas the upper bound of SFL always diminishes as long as $K$ increases. This is because when $K$ exceeds one threshold, the dominant term of the upper bound will be immune to its change as stated in \cref{subsec:analysis on SFL}. Then, considering ``catastrophic forgetting'' \citep{kirkpatrick2017overcoming,sheller2019multi} problems in SFL, it can be expected to see such phenomenon.

\textit{SFL outperforms PFL on extremely heterogeneous data.} The test accuracy results for various tasks are collected in Table~\ref{tab:cross-device settings}. When $C=1$ (extremely heterogeneous), the performance of SFL is better than that of PFL across all tried settings. When $C=2$ (moderately heterogeneous), PFL can achieve the close or even slightly better performance than SFL in some cases (e.g., CIFAR-10/$C=2$/$K=50$). This is consistent with our observation and analysis in \cref{subsec:simulation}. Notably, on the more complicated dataset CINIC-10, SFL shows better for all settings, which may be due to higher heterogeneity.

\begin{table}[ht]
	\renewcommand{\arraystretch}{1}
	\centering
	\caption{Test accuracy results in cross-device settings. We run PFL and SFL for 4000 training rounds on CIFAR-10 and CINIC-10. Results are computed across three random seeds and the last 100 training rounds. The better results (with larger than 2\% test accuracy gain) between PFL and SFL in each setting are marked in bold.}
	\label{tab:cross-device settings}
	\setlength{\tabcolsep}{0.15em}{
		\resizebox{\linewidth}{!}{
			\begin{tabular}{llllll@{\hspace{1em}}lll}
				\toprule
				\multicolumn{3}{c}{\multirow{1}{*}{Setup}} &\multicolumn{3}{c}{$C=1$} &\multicolumn{3}{c}{$C=2$} \\\midrule
				Dataset &Model &Method  &$K=5$ &$K=20$ &$K=50$ &$K=5$ &$K=20$ &$K=50$ \\\midrule
				\multirow{4}{*}{CIFAR-10} 
				&\multirow{2}{*}{VGG-9} 
				&PFL &67.61\tiny{$\pm$4.02} &62.00\tiny{$\pm$4.90} &45.77\tiny{$\pm$5.91} &78.42\tiny{$\pm$1.47} &78.88\tiny{$\pm$1.35} &78.01\tiny{$\pm$1.50} \\
				& &SFL &\textbf{78.43}\tiny{$\pm$2.46} &\textbf{72.61}\tiny{$\pm$3.27} &\textbf{68.86}\tiny{$\pm$4.19} &\textbf{82.56}\tiny{$\pm$1.68} &\textbf{82.18}\tiny{$\pm$1.97} &79.67\tiny{$\pm$2.30} \\[2ex]
				&\multirow{2}{*}{ResNet-18} 
				&PFL &52.12\tiny{$\pm$6.09} &44.58\tiny{$\pm$4.79} &34.29\tiny{$\pm$4.99} &80.27\tiny{$\pm$1.52} &82.27\tiny{$\pm$1.55} &79.88\tiny{$\pm$2.18} \\
				& &SFL &\textbf{83.44}\tiny{$\pm$1.83} &\textbf{76.97}\tiny{$\pm$4.82} &\textbf{68.91}\tiny{$\pm$4.29} &\textbf{87.16}\tiny{$\pm$1.34} &\textbf{84.90}\tiny{$\pm$3.53} &79.38\tiny{$\pm$4.49} \\
				\midrule
				\multirow{4}{*}{CINIC-10} 
				&\multirow{2}{*}{VGG-9} 
				&PFL &52.61\tiny{$\pm$3.19} &45.98\tiny{$\pm$4.29} &34.08\tiny{$\pm$4.77} &55.84\tiny{$\pm$0.55} &53.41\tiny{$\pm$0.62} &52.04\tiny{$\pm$0.79} \\
				& &SFL &\textbf{59.11}\tiny{$\pm$0.74} &\textbf{58.71}\tiny{$\pm$0.98} &\textbf{56.67}\tiny{$\pm$1.18} &\textbf{60.82}\tiny{$\pm$0.61} &\textbf{59.78}\tiny{$\pm$0.79} &\textbf{56.87}\tiny{$\pm$1.42} \\[2ex]
				&\multirow{2}{*}{ResNet-18} 
				&PFL &41.12\tiny{$\pm$4.28} &33.19\tiny{$\pm$4.73} &24.71\tiny{$\pm$4.89} &57.70\tiny{$\pm$1.04} &55.59\tiny{$\pm$1.32} &46.99\tiny{$\pm$1.73} \\
				& &SFL &\textbf{60.36}\tiny{$\pm$1.37} &\textbf{51.84}\tiny{$\pm$2.15} &\textbf{44.95}\tiny{$\pm$2.97} &\textbf{64.17}\tiny{$\pm$1.06} &\textbf{58.05}\tiny{$\pm$2.54} &\textbf{56.28}\tiny{$\pm$2.32} \\
				\bottomrule
	\end{tabular}}}
\end{table}

\section{Conclusion}
In this paper, we have derived the convergence guarantees of SFL for strongly convex, general convex and non-convex objectives on heterogeneous data. Furthermore, we have compared SFL against PFL, showing that the guarantee of SFL is better than PFL on heterogeneous data. Experimental results validate that SFL outperforms PFL on extremely heterogeneous data in cross-device settings. 

Future directions include (i) lower bounds for SFL (this work focuses on the upper bounds of SFL), (ii) other potential factors that may affect the performance of PFL and SFL (this work focuses on data heterogeneity) and (iii) new algorithms to facilitate our findings (no new algorithm in this work).

\section*{Acknowledgments}
This work was supported in part by the National Key Research and Development Program of China under Grant 2021YFB2900302, in part by the National Science Foundation of China under Grant 62001048, and in part by the Fundamental Research Funds for the Central Universities under Grant 2242022k60006.

We thank the reviewers in NeurIPS 2023 for the insightful suggestions. We also thank Sai Praneeth Karimireddy and Ahmed Khaled for their kind help on the upper bounds of PFL.

\bibliographystyle{plainnat}
\bibliography{refs,refs2}

\clearpage
\begin{center}
	\LARGE {Appendix}
\end{center}
\appendix
\vskip 3ex\hrule\vskip 2ex
{
	\hypersetup{linktoc=page}
	\parskip=0.5ex
	\startcontents[sections]
	\printcontents[sections]{l}{1}{\setcounter{tocdepth}{3}}
}
\vskip 3ex\hrule\vskip 5ex
\clearpage

\section{Applicable to Split Learning}\label{sec:app:applicability}
\begin{figure}[htbp]
	\centering
	\includegraphics[width=0.8\linewidth]{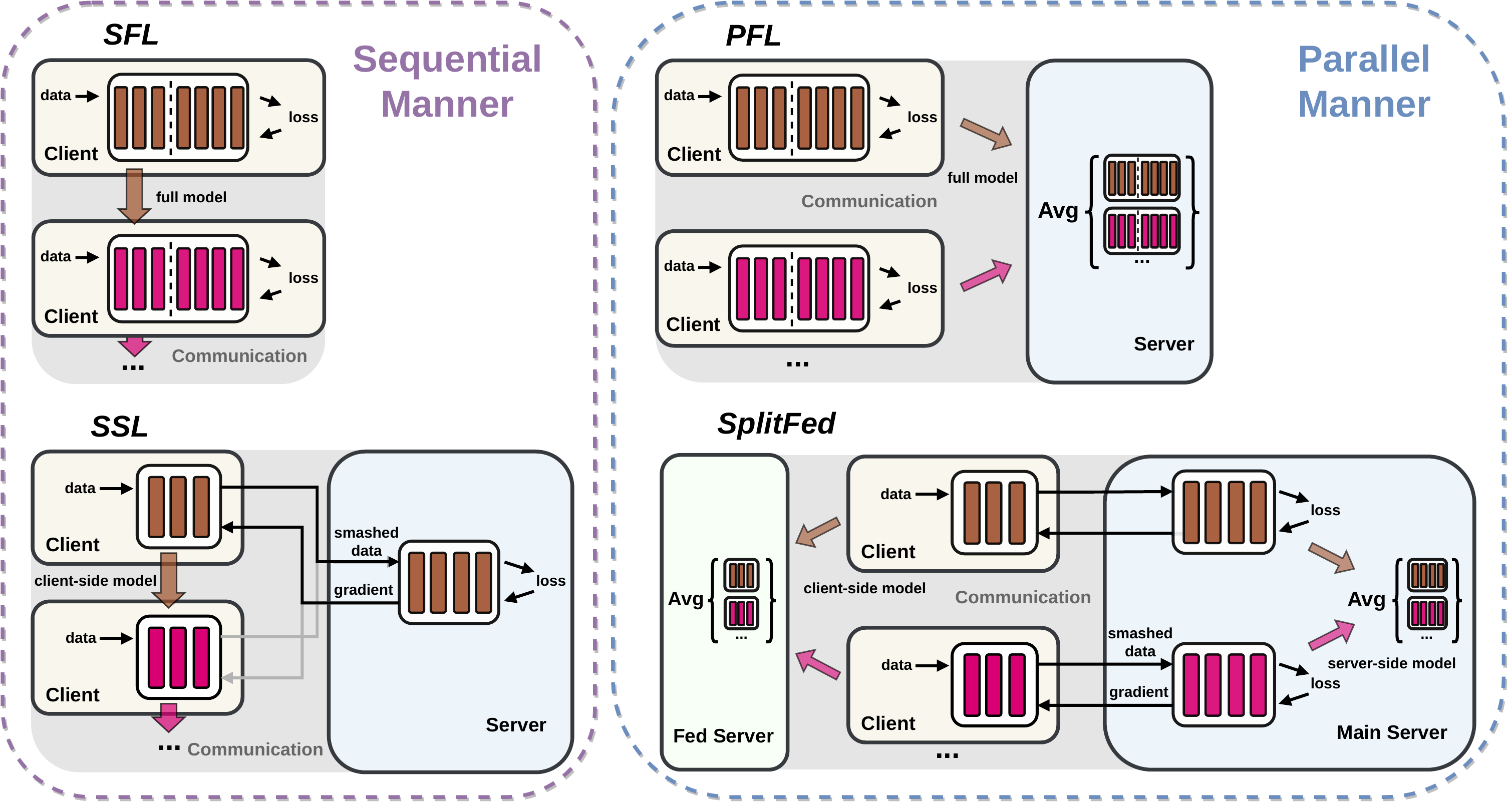}
	\caption{Overviews of paradigms in FL and SL. The top row shows the FL algorithms, SFL and PFL. The bottom row shows the SL algorithms, SSL and SplitFed.}
	\label{fig:app:overview}
	\vspace{-2ex}
\end{figure}

Split Learning is proposed to address the computation bottleneck of resource-constrained devices, where the full model is split into two parts: the \textit{client-side model} (front-part) and the \textit{server-side model} (back-part). There are two typical algorithms in SL, i.e., Sequential Split Learning (SSL)\footnote{In SSL, client-side model parameters can be synchronized in two modes, the \textit{peer-to-peer mode} and \textit{centralized mode}. In the peer-to-peer mode, parameters are sent to the next client directly, while in the centralized mode, parameters are relayed to the next client through the server. This paper considers the peer-to-peer mode.} \citep{gupta2018distributed} and Split Federated Learning (SplitFed)\footnote{There are two versions of SplitFed and the first version is considered in this paper by default.} \citep{thapa2022splitfed}. The overviews of these four paradigms are illustrated in Figure~\ref{fig:app:overview}

\textbf{Training process of SSL.} Each client keeps one client-side model and the server keeps one server-side model. Thus, each client and the server can collaborate to complete the local training of a full model (over the local data kept in clients). Each local step of local training includes the following operations: 1) the client executes the forward pass on the local data, and sends the activations of the cut-layer (called \textit{smashed data}) and labels to the server; 2) the server executes the forward pass after receiving the smashed data and computes the loss with received labels; 3) the server executes the backward pass and sends the gradients of the smashed data to the client; 4) the client executes the backward pass with the received gradients. After finishing the local training, the client sends the updated parameters of its client-side model to the next client. This process continues until all clients complete their local training. See the bottom-left subfigure and Algorithms~\ref{alg1:app:SSL}, \ref{alg2:app:SSL}. For clarity, we adjust the superscripts and subscripts in this section, and provide the required notations in Table~\ref{table:app:notations of SL}.

\textbf{Training process of SplitFed.} Each client keeps one client-side model and the server (named as \textit{main server}) keeps multiple server-side models, whose quantity is the same as the number of clients. Thus, each client and its corresponding server-side model in the main server can complete the local training of a full model in parallel. The local training operations of each client in SplitFed are identical to that in SSL. After the local training with the server, clients send the updated parameters to the fed server, one server introduced in SplitFed to achieve the aggregation of client-side models. The fed server aggregates the received parameters and sends the aggregated (averaged) parameters to the clients. The main server also aggregates the parameters of server-side models it kept and updates them accordingly. See the bottom-right subfigure and \cite{thapa2021advancements}'s Algorithm~2.

\textbf{Applicable to SL.} According to the complete training process of SSL and SplitFed, we can conclude the relationships between SFL and SSL, and, PFL and SplitFed as follows:
\begin{itemize}[leftmargin=2em]
	\item SSL and SplitFed can be viewed as the practical implementations of SFL and PFL respectively in the context of SL.
	\item SSL and SplitFed share the same update rules with SFL (Algorithm~\ref{algorithm1}) and PFL (Algorithm~\ref{algorithm2}) respectively, and hence, the same convergence results.
\end{itemize}
\begin{table}[ht]
	\renewcommand{\arraystretch}{1.2}
	\centering
	\caption{Additional notations for Section~\ref{sec:app:applicability}.}
	\label{table:app:notations of SL}
	\begin{tabular}{cl}
		\toprule
		Symbol &Description\\ \midrule
		$\tau_m, k$ &number, index of local update steps (when training) with client $\pi_m$\\
		$\rvx_{m}^{(r,k)}$/$\rvx_{c,m}^{(r,k)}$/$\rvx_{s,m}^{(r,k)}$ &\begin{tabular}[c]{@{}l@{}}full/client-side/server-side local model parameters ($\rvx_{m}^{(r,k)} = [\rvx_{c,m}^{(r,k)};\rvx_{s,m}^{(r,k)}]$)\\ \quad after $k$ local updates with client $\pi_m$ in the $r$-th round \\
		\end{tabular} \\
		$\mX_m^{(r,k)}$/$\mY_m^{(r,k)}$/$\hat\mY_m^{(r,k)}$ &\begin{tabular}[c]{@{}l@{}}features/labels/predictors \\
			\quad after $k$ local updates with client $\pi_m$ in the $r$-th round\end{tabular} \\
		$\rvx^{(r)}$/$\rvx_{c}^{(r)}$/$\rvx_{s}^{(r)}$ &full/client-side/server-side global model parameters in the $r$-th round \\
		$\mA_m^{(r,k)}$ &\begin{tabular}[c]{@{}l@{}}smashed data (activation of the cut layer) \\
			\quad after $k$ local updates with client $\pi_m$ in the $r$-th round \end{tabular}\\
		$\ell_{\pi_m}$ &loss function with client $\pi_m$\\
		$\nabla \ell_{\pi_m} (\rvx_{s,m}^{(r,k)}; \mA_m^{(r,k)})$ &gradients of the loss regarding $\rvx_{s,m}^{(r,k)}$ on input $\mA_m^{(r,k)}$\\[0.5ex]
		$\nabla \ell_{\pi_m} (\mA_m^{(r,k)};\rvx_{s,m}^{(r,k)})$ &gradients of the loss regarding $\mA_m^{(r,k)}$ on parameters $\rvx_{s,m}^{(r,k)}$ \\[0.5ex]	
		$\nabla \ell_{\pi_m} (\rvx_{c,m}^{(r,k)}; \mX_m^{(r,k)})$ &gradients of the loss regarding $\rvx_{c,m}^{(r,k)}$ on input $\mX_m^{(r,k)}$\\[0.5ex]
		\bottomrule
	\end{tabular}
\end{table}
\IncMargin{1em}	
\begin{algorithm}[H]
	\DontPrintSemicolon
	\caption{Sequential Split Learning (Server-side operations)}
	\label{alg1:app:SSL}
	\BlankLine
	\nonl\textbf{Main Server} executes:
	\BlankLine
	Initialize server-side global parameters $\rvx_s^{(0)}$\;
	\For{round $r = 0,\ldots, R-1$}{
		Sample a permutation $\pi_1, \pi_2, \ldots, \pi_{M}$ of $\{1,2,\ldots,M\}$ as clients' update order\;
		
		\For{$m = 1,2,\ldots,M$ {\bf \textcolor{red}{in sequence}}}{
			Initialize server-side local parameters: $\rvx_{s,m}^{(r,0)} \gets
			\begin{cases}
				\rvx_s^{(r)}\ , &m=1\\
				\rvx_{s,m-1}^{(r,\tau_{m-1})}\ , &m>1
			\end{cases}$\;
			\For{local update step $k = 0,\ldots, \tau_m-1$}{
				Receive ($\mA_m^{(r,k)}$, $\mY_m^{(r,k)}$) from client $m$\tcp*[r]{Com.}
				Execute forward passes with smashed data $\mA_m^{(r,k)}$\;
				Calculate the loss with ($\hat\mY_m^{(r,k)}$, $\mY_m^{(r,k)}$)\;
				Execute backward passes and compute $\nabla \ell_{\pi_m} (\rvx_{s,m}^{(r,k)}; \mA_m^{(r,k)})$\;
				Send $\nabla \ell_{\pi_m} (\mA_m^{(r,k)};\rvx_{s,m}^{(r,k)})$ to client $m$\tcp*[r]{Com.}
				Update server-side parameters:
				$\rvx_{s,m}^{(r,k+1)} \gets \rvx_{s,m}^{(r,k)} - \eta \nabla \ell_{\pi_m} (\rvx_{s,m}^{(r,k)}; \mA_m^{(r,k)})$\;
			}
		}
	}
\end{algorithm}
\DecMargin{1em}
\IncMargin{1em}	
\begin{algorithm}[H]
	\DontPrintSemicolon
	\caption{Sequential Split Learning (Client-side operations)}
	\label{alg2:app:SSL}	
	\BlankLine
	\nonl\textbf{Client $\pi_m$} executes:
	\BlankLine
	Request the latest client-side parameters from the previous client\tcp*[r]{Com.}
	Initialize client-side parameters: $\rvx_{c,m}^{(r,0)} \gets
	\begin{cases}
		\rvx_c^{(r)}\ , & m=1\\
		\rvx_{c,m-1}^{(r,\tau_{m-1})}\ , & m>1
	\end{cases}$\;
	\For{local update step $k = 0,\ldots, \tau_m-1$}{
		Execute forward passes with data features $\mX_m^{(r,k)}$
		
		Send ($\mA_m^{(r,k)}$, $\mY_m^{(r,k)}$) to the server\tcp*[r]{Com.}
		Receive $\nabla \ell_{\pi_m} (\mA_m^{(r,k)};\rvx_{s,m}^{(r,k)})$\tcp*[r]{Com.}
		Execute backward passes and compute $\nabla \ell_{\pi_m} (\rvx_{c,m}^{(r,k)}; \mX_m^{(r,k)})$\;
		Update client-side parameters: $\rvx_{c,m}^{(r,k+1)} \gets \rvx_{c,m}^{(r,k)} - \eta \nabla \ell_{\pi_m} (\rvx_{c,m}^{(r,k)}; \mX_m^{(r,k)})$
	}
\end{algorithm}
\DecMargin{1em}

\section{Related work}\label{sec:app:related work}
\textbf{Convergence of PFL.} The convergence of PFL (also known as Local SGD, \texttt{FedAvg}) has developed rapidly recently, with weaker assumptions, tighter bounds and more complex scenarios. \cite{zhou2018convergence}, \citet{stich2019local}, \citet{khaled2020tighter} and \citet{wang2021cooperative} analyzed the convergence of PFL on homogeneous data. \cite{li2020convergence} derived the convergence guarantees for PFL with the bounded gradients assumption on heterogeneous data. Yet this assumption has been shown too strong \citep{khaled2020tighter}. To further catch the heterogeneity, \cite{karimireddy2020scaffold, koloskova2020unified} assumed the variance of the gradients of local objectives is bounded either uniformly (Assumption~\ref{asm:heterogeneity1}) or on the optima (Assumption~\ref{asm:heterogeneity2}). Moreover, \cite{li2020convergence}, \citet{karimireddy2020scaffold} and \citet{yang2021achieving} also consider the convergence with partial client participation. The lower bounds of PFL are also studied in \citet{woodworth2020local}, \citet{woodworth2020minibatch} and \citet{yun2022minibatch}. There are other variants in PFL, which show a faster convergence rate than the vanilla one (Algorithm~\ref{algorithm2}), e.g., \texttt{SCAFFOLD} \citep{karimireddy2020scaffold}.

\textbf{Convergence of SGD-RR.} Random Reshuffling (SGD-RR) has attracted more attention recently, as it (where data samples are sampled without replacement) is more common in practice than its counterpart algorithms SGD (where data samples are sample with replacement). Early works \citep{gurbuzbalaban2021random, haochen2019random} prove the upper bounds for strongly convex and twice-smooth objectives. Subsequent works \citep{nagaraj2019sgd,ahn2020sgd,mishchenko2020random,nguyen2021unified,lu2022general} further prove upper bounds for strongly convex, convex and non-convex cases. The lower bounds of SGD-RR are also investigated in the quadratic case \citep{safran2020good, safran2021random} and the strongly convex case \citep{rajput2020closing, cha2023tighter}. In particular, the lower bounds in \cite{cha2023tighter} are shown to match the upper bounds in \cite{mishchenko2020random} for both strongly convex and general convex cases. These works have reached a consensus that SGD-RR is better than SGD at least when the number of epochs (passes over the data) is large enough. In this paper, we use the bounds of SGD-RR to examine the tightness of our bounds of SFL.
Works studying randomized incremental gradient methods are also relevant \citep{ram2009incremental, johansson2010randomized,ayache2021private,mao2020walkman,cyffers2022privacy}; these works consider a single update at each client and focus on random walks.

\textbf{Shuffling-based methods in FL.} Recently, shuffling-based methods have appeared in FL \citep{mishchenko2022proximal,yun2022minibatch, cho2023convergence}. \cite{mishchenko2022proximal} gave the convergence result of Federated Random Reshuffling (FedRR) as an application to Federated Learning of their theory for Proximal Random Reshuffling (ProxRR). \cite{yun2022minibatch} analyzed the convergence of Minibatch RR and Local RR, the variants of Minibatch SGD and Local SGD (Local SGD is equivalent to PFL in this work), where clients perform SGD-RR locally (in parallel) instead of SGD. Both FedRR and Local RR are different from SFL from the algorithm perspective. See \cite{yun2022minibatch}'s Appendix A for comparison.
The most relevant works are FL with cyclic client participation \citep{cho2023convergence} and FL with shuffling client participation \citep{malinovsky2023federated}.
\begin{figure}[htbp]
	\centering
	\includegraphics[width=0.6\linewidth]{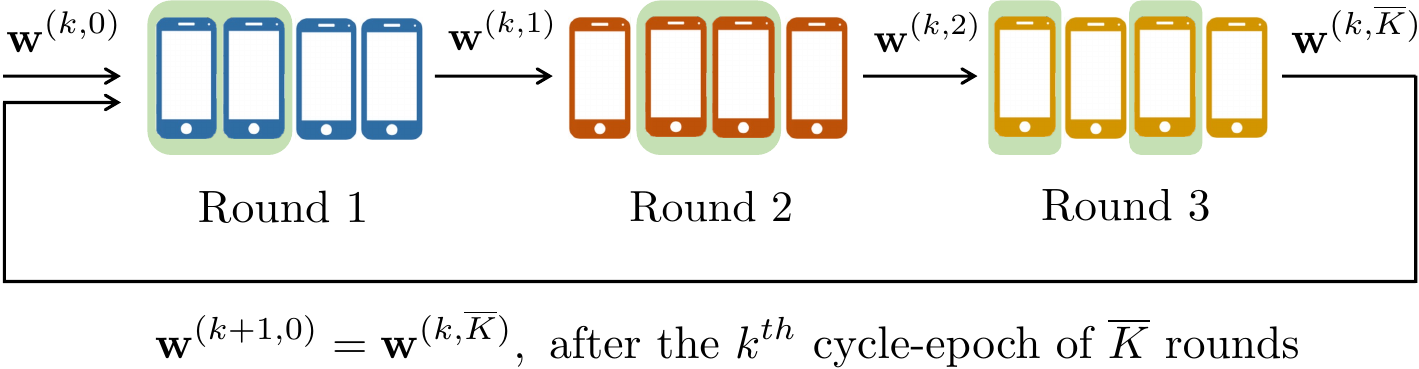}
	\caption{Illustration of FL with cyclic client participation with $M = 12$ clients divided into $\overline K = 3$ groups \citep{cho2023convergence}. In each training round, $N = 2$ clients are selected for training from the client group. All groups are traversed once in a cycle-epoch consisting of $\overline K$ training rounds.}
	\label{fig:cyclic participation}
\end{figure}

\textit{Discussion about FL with cyclic client participation.} \cite{cho2023convergence} consider the scenario where the total $M$ clients are divided into $\overline K$ non-overlapping client groups such that each group contains $M/\overline K$ clients. In each training round, the server selects a subset of $N$ clients from a group without replacement for training in this round. One example is shown in Figure~\ref{fig:cyclic participation}. As said in the paragraph ``Cyclic Client Participation (CyCP)'' in their Section 3 (Problem Formulation), the groups' training order of FL with cyclic client participation is pre-determined and fixed. In contrast, the clients' training order of SFL (precisely, Algorithm~\ref{algorithm1}) will be shuffled at the beginning of each round.
It can be verified by \cite{cho2023convergence}'s Theorem 2. Letting $\overline K=1$ and $N=M/\overline K=M$ and $\overline K=M$ and $N=M/\overline K=1$, we get the bounds for PFL and SFL, respectively:
\begin{align*}
	\text{PFL:}&\quad\tilde\gO\left( \frac{L\sigma^2}{\mu^2MKR} + \frac{L^2\zeta^2}{\mu^3 M^2R^2}\right)
\end{align*}
where we have omitted the optimization term and changed their notations to ours (change $\alpha$ to 0, $\gamma$ to $\zeta$, $\nu$ to $\zeta$, $T$ to $R$).
\begin{align*}
	\text{SFL:}&\quad\tilde\gO\left( \frac{L\sigma^2}{\mu^2KR} + \frac{L^2\zeta^2}{\mu^3 R^2} + \frac{L^2\zeta^2}{\mu^3 M^2R^2}\right)
\end{align*}
where we have omitted the optimization term and changed their notations to ours (change $\alpha$ to $\zeta$, $\gamma$ to $0$, $\nu$ to $\zeta$, $T$ to $MR$). As we can see, we do not see a clear advantage of SFL like ours.
\begin{figure}[htbp]
	\centering
	\includegraphics[width=0.8\linewidth]{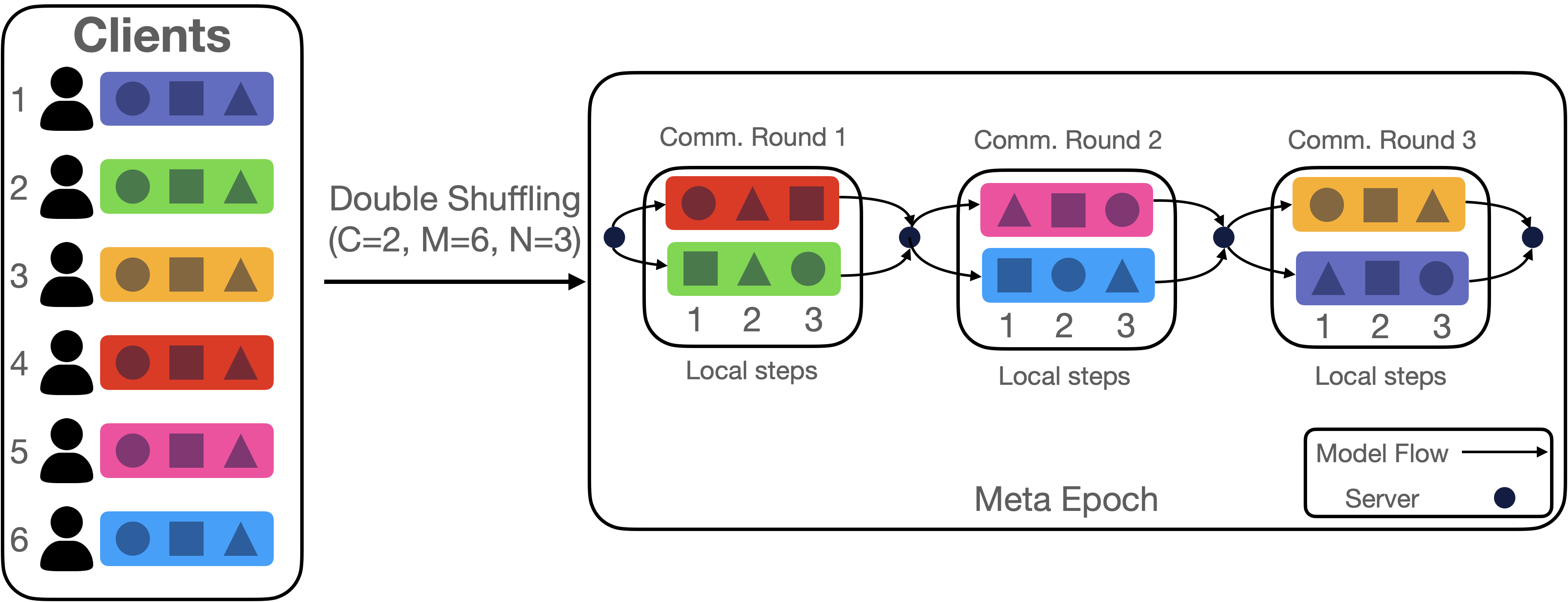}
	\caption{Visualization of FL with shuffling client participation for 6 clients, each with 3 datapoints. Two clients are sampled in each communication round \citep{malinovsky2023federated}.}
	\label{fig:shuffling participation}
\end{figure}

\textit{Discussion about FL with shuffling client participation.} At the beginning of each meta epoch, FL with shuffling client participation partition all $M$ clients into $M/C$ cohorts, each with size $C$. These cohorts are obtained using the without replacement sampling of clients. Each meta epoch contains $R=M/C$ communication rounds. At each communication round, clients in a cohort participate in the training process. One example is shown in Figure~\ref{fig:shuffling participation}. It is noteworthy that SGD-RR is used as the local solver in \cite{malinovsky2023federated} while SGD is used in this paper.
Letting $C=1$, $R=M$, $T=R$ in their Theorem 6.1, we can get the bounds for SFL:
\begin{align*}
\text{SFL \citep{malinovsky2023federated}:}&\quad \gO\left(\frac{L}{\mu}\cdot \left(LMK^2\eta^2\zeta_\ast^2+ \mu D^2 \exp\left(-\mu MKR\eta \right)\right) \right) \text{with} \ \eta\leq \frac{1}{L},\\
\text{SFL (Theorem~\ref{thm:SFL}):}&\quad \gO\left( LMK^2\eta^2\zeta_\ast^2+ \mu D^2 \exp\left(-\mu MKR\eta \right) \right) \text{with} \ \eta\leq \frac{1}{LMK},
\end{align*}
where we only considered the heterogeneity terms (let $\sigma_{\star}^2 = 0$ for \cite{malinovsky2023federated} and $\sigma=0$ for ours) and changed their notations to ours (change $\gamma$ to $\eta$, $N$ to $K$, $\tilde\sigma_{\star}$ to $\zeta_\ast$). Their bound almost matches ours, with some differences on the constants $\mu$, $L$ and restrictions on $\eta$, which is caused by using different local solvers. However, their results are limited to the case where the number of local steps equals the size of the local dataset. It is still uncertain whether their results for SGD-RR can be generalized to situations with varying numbers of local steps.

\section{Preliminaries}
\subsection{Basic identities and inequalities}
These identities and inequalities are mostly from \cite{zhou2018fenchel}, \cite{khaled2020tighter}, \citet{mishchenko2020random}, \citet{karimireddy2020scaffold} and \citet{garrigos2023handbook}.
For any random variable $\rvx$, the variance can be decomposed as
\begin{align}
	\E\left[\norm{\rvx-\E\left[\rvx\right]}^2\right] = \E\left[\norm{\rvx}^2\right] - \norm{\E\left[\rvx\right]}^2 \label{eq:variance decomposition}
\end{align}
In particular, its version for vectors with finite number of values gives 
\begin{align}
	\frac{1}{n} \sum_{i=1}^{n} \sqn{\vx_i - \bar{\vx}} = \frac{1}{n} \sum_{i=1}^{n} \norm{\vx_{i}}^2 - \norm{ \frac{1}{n} \sum_{i=1}^{n} \vx_i }^2\label{eq:variance decomposition discrete}
\end{align}
where vectors $\vx_{1}, \ldots, \vx_{n} \in \R^d$ are the values of $\rvx$ and their average is $\bar{\vx} = \frac{1}{n}\sum_{i=1}^n \vx_i$.

\textbf{Jensen's inequality.} For any convex function $h$ and any vectors $\vx_1,\dotsc, \vx_n$ we have
\begin{align}
	h\left(\frac{1}{n}\sum_{i=1}^n \vx_i\right) \leq \frac{1}{n}\sum_{i=1}^n h(\vx_i). \label{eq:jensen's inquality}
\end{align}
As a special case with $h(\vx)=\norm{\vx}^2$, we obtain
\begin{align}
	\sqn{\frac{1}{n}\sum_{i=1}^n \vx_i} \leq \frac{1}{n}\sum_{i=1}^n \norm{\vx_i}^2. \label{eq:jensen norm}
\end{align}

\textbf{Smoothness and general convexity, strong convexity.} There are some useful inequalities with respect to $L$-smoothness (Assumption~\ref{asm:smoothness}), convexity and $\mu$-strong convexity. Their proofs can be found in \cite{zhou2018fenchel} and \citet{garrigos2023handbook}.

\textit{Bregman Divergence} associated with function $h$ and arbitrary $\vx$, $\vy$ is denoted as
\begin{align*}
	D_h(\vx, \vy) \coloneqq h(\vx) - h(\vy) - \inp{\nabla h(\vy)}{\vx-\vy}
\end{align*}
When the function $h$ is convex, the divergence is non-negative. A more formal definition can be found in \cite{orabona2026online}. One corollary (Chen and Teboulle, 1993) called \textit{three-point-identity} is, 
\begin{align*}
D_h(\vz, \vx) + D_h(\vx,\vy) - D_h(\vz,\vy) = \inp{\nabla h(\vy) - \nabla h(\vx)}{\vz-\vx}
\end{align*}
where $\vx, \vy, \vz$ is three points in the domain.
Let $h$ be $L$-smooth. With the definition of Bregman divergence, a consequence of $L$-smoothness is
\begin{equation}
	D_h(\vx, \vy) = h(\vx) - h(\vy) - \inp{\nabla h(\vy)}{\vx-\vy} \leq \frac{L}{2}\norm{\vx-\vy}^2 \label{eq1:smooth}
\end{equation}
Further, If $h$ is $L$-smooth and lower bounded by $h_\ast$, then
\begin{equation}
	\sqn{\nabla h(\vx)} \leq 2 L \left(h(\vx) - h_\ast\right).\label{eq:smooth:grad bound}
\end{equation}
If $h$ is $L$-smooth and convex (The definition of convexity can be found in \cite{boyd2004convex}), then
\begin{align}
	D_h(\vx, \vy)\ge \frac{1}{2L}\norm{\nabla h(\vx) - \nabla h(\vy)}^2. \label{eq:smooth+convex:bregman lower bound}
\end{align}
The function $h: \R^d\to \R$ is $\mu$-strongly convex if and only if there exists a convex function $g: \R^d\to \R$ such that $h(\vx) = g(\vx) + \frac{\mu}{2}\norm{\vx}^2$.
If $h$ is $\mu$-strongly convex, it holds that
\begin{align}
	\frac{\mu}{2}\norm{\vx-\vy}^2 \leq D_h(\vx, \vy)\label{eq:strongly convex:bregman lower bound}
\end{align}
\subsection{Technical lemmas}
\begin{lemma}[\cite{karimireddy2020scaffold}]\label{lem:martingale difference property}
	Let $\{\xi_i\}_{i=1}^n$ be a sequence of random variables. And the random sequence $\{\rvx_i\}_{i=1}^n$ satisfy that $\rvx_i \in \R^d$ is a function of $\xi_i, \xi_{i-1}, \ldots, \xi_{1}$ for all $i$. Suppose that the conditional expectation is $\E_{\xi_i}\left[\left.\rvx_i\right|\xi_{i-1},\ldots\xi_{1}\right]=\rve_i$ (i.e., the vectors $\{\rvx_i-\rve_i\}$ form a martingale difference sequence with respect to $\{\xi_i\}$), and the variance is bounded by $\E_{\xi_i}\left[\norm{\rvx_i-\rve_i}^2\mid \xi_{i-1},\ldots\xi_{1}\right] \leq \sigma^2$. Then it holds that
	\begin{align}
		\E\left[\norm{\sum_{i=1}^n(\rvx_i-\rve_i)}^2\right] = \sum_{i=1}^n \E\left[\norm{\rvx_i-\rve_i}^2\right] \leq n\sigma^2
	\end{align}
\end{lemma}
\begin{proof}
	The above result can also be found in, e.g., \citet[Lemma 15]{stich2019error}, \citet[Lemma 4, separating mean and variance]{karimireddy2020scaffold} and \citet[Lemma 3]{even2022asynchronous}. It follows that
	\begin{align*}
	\E\left[\norm{\sum_{i=1}^n(\rvx_i-\rve_i)}^2\right] &= \sum_{i=1}^n\E\left[\norm{\rvx_i-\rve_i}^2\right] + \sum_{i=1}^n\sum_{j\neq i}^n\E\left[(\rvx_i-\rve_i)^\top(\rvx_j-\rve_j)\right]\\
	&= \sum_{i=1}^n\E\left[\norm{\rvx_i-\rve_i}^2\right]\\
	&= \sum_{i=1}^n\E\left[\E\left[\norm{\rvx_i-\rve_i}^2\mid \xi_{i-1},\ldots,\xi_{1} \right]\right]\\ &\leq n \sigma^2,
	\end{align*}
	where the second equality uses the orthogonality of martingale difference sequences (e.g., see Lemma~4.1 in Chapter~10 (Martingales) of \citet{gut2013probability}) and the third equality uses the law of total expectation (e.g., see Theorem~4.7.1 in \citet{degroot2012probability}).
\end{proof}
\begin{lemma}[\cite{karimireddy2020scaffold}]
	\label{lem:perturbed strong convexity}
	The following holds for any $L$-smooth and $\mu$-strongly convex function $h$, and any $\vx, \vy, \vz$ in the domain of $h$:
	\begin{align}
		\left\langle \nabla h(\vx), \vz-\vy\right\rangle \geq h(\vz) - h(\vy) +\frac{\mu}{4}\norm{\vy - \vz}^2  - L\norm{\vz - \vx}^2.
	\end{align}
\end{lemma}
\begin{proof}
	Using the \textit{three-point-identity}, we get
	\[
	\inp{\nabla h(\vx)}{\vz-\vy} = D_h(\vy,\vx) - D_h(\vz,\vx) + h(\vz) - h(\vy)
	\]
	Then, we get the following two inequalities using smoothness and strong convexity of $h$:
	\begin{align*}
		\inp{\nabla h(\vx)}{\vz-\vy} \geq \frac{\mu}{2}\norm{\vy - \vx}^2 - \frac{L}{2}\norm{\vz - \vx}^2 + h(\vz) - h(\vy)
	\end{align*}
	Further, using Jensen's inequality (i.e., $\norm{\vy -\vz}^2\leq 2(\norm{\vx -\vz}^2 + \norm{\vy -\vx}^2)$), we have
	\[
	\frac{\mu}{2}\norm{\vy - \vx}^2 \geq \frac{\mu}{4}\norm{\vy - \vz}^2 - \frac{\mu}{2}\norm{\vx - \vz}^2\,.
	\]
	Combining all the inequalities together we have
	\begin{align*}
		\inp{\nabla h(\vx)}{\vz - \vy} &\geq h(\vz) - h(\vy) +\frac{\mu}{4}\norm{\vy - \vz}^2  - \frac{L + \mu}{2}\norm{\vz - \vx}^2\\
		&\geq h(\vz) - h(\vy) +\frac{\mu}{4}\norm{\vy - \vz}^2 - L\norm{\vz - \vx}^2,
	\end{align*}
	which is the claim of this lemma.
\end{proof}
\begin{lemma}[Simple Random Sampling]\label{lem:simple random sampling}
	Let $\vx_1, \vx_2,\ldots, \vx_n$ be fixed units (e.g., vectors). The population mean and population variance are given as
	\begin{align*}
		\textstyle\overline \vx \coloneqq \frac{1}{n}\sum_{i=1}^n \vx_i 
		&&
		\textstyle\zeta^2 \coloneqq \frac{1}{n}\sum_{i=1}^n \norm{\vx_i-\overline \vx}^2
	\end{align*}
	Draw $s\in [n]=\{1,2,\ldots,n\}$ random units $\rvx_{\pi_1}, \rvx_{\pi_2},\ldots \rvx_{\pi_s}$ randomly from the population. There are two possible ways of simple random sampling, well known as ``sampling with replacement (SWR)'' and ``sampling without replacement (SWOR)''. For these two ways, the expectation and variance of the sample mean $\overline \rvx_\pi \coloneqq \frac{1}{s}\sum_{p=1}^s \rvx_{\pi_p}$ satisfies
	\begin{align}
		\text{SWR\hphantom{O}:}\quad\E[\overline \rvx_\pi] = \overline \vx,&&
		&\E\left[\norm{\overline \rvx_{\pi} - \overline \vx}^2\right] = \frac{\zeta^2}{s}\label{eq:lem:sampling with replacement}\\
		\text{SWOR:}\quad\E[\overline \rvx_\pi] = \overline \vx,&&
		&\E\left[\norm{\overline \rvx_{\pi} - \overline \vx}^2\right] = \frac{n-s}{s(n-1)}\zeta^2\label{eq:lem:sampling without replacement}
	\end{align}
\end{lemma}
\begin{proof}
	The proof of this lemma is mainly based on \cite{mishchenko2020random}'s Lemma 1 (A lemma for sampling without replacement) and \cite{wang2020tackling}'s Appendix G (Extension: Incorporating Client Sampling). Since the probability of each unit being selected equals $\frac{1}{n}$ in each draw, we can get the expectation and variance of any random unit $\rvx_{\pi_p}$ at the $p$-th draw:
	\begin{align*}
		&\E\left[\rvx_{\pi_p}\right] = \sum_{i=1}^n \vx_i\cdot \Pr(\rvx_{\pi_p}= \vx_i) = \sum_{i=1}^n \vx_i\cdot \frac{1}{n}= \overline \vx,\\
		&\Var(\rvx_{\pi_p}) = \E\left[\norm{\rvx_{\pi_p} - \overline \vx}^2\right] = \sum_{i=1}^n \norm{\vx_i - \overline \vx}^2 \cdot \Pr(\rvx_{\pi_p} = \vx_i) = \sum_{i=1}^n \norm{\vx_i - \overline \vx}^2 \cdot \frac{1}{n} = \zeta^2,
	\end{align*}
	where the preceding equations hold for both sampling ways. Thus, we can compute the expectations of the sample mean for both sampling ways as
	\begin{align*}
		\E\left[\overline \rvx_\pi\right] = \E \left[\frac{1}{s} \sum_{p=1}^s \rvx_{\pi_p}\right] = \frac{1}{s}\sum_{p=1}^s\E\left[\rvx_{\pi_p}\right] =\overline \vx,
	\end{align*}
	which indicates that the sample means for both ways are unbiased estimators of the population mean. The variance of the sample mean can be decomposed as
	\begin{align*}
		\E\norm{\overline \rvx_\pi - \overline \vx}^2 = \E\norm{\frac{1}{s}\sum_{p=1}^s(\rvx_{\pi_p} - \overline \vx)}^2 =\frac{1}{s^2}\sum_{p=1}^s\Var(\rvx_{\pi_p}) + \frac{1}{s^2}\sum_{p=1}^s\sum_{q\neq p}^s \Cov (\rvx_{\pi_p}, \rvx_{\pi_q})
	\end{align*}
	Next, we deal with these two ways separately: 
	\begin{itemize}
		\item SWR: It holds that $\Cov(\rvx_{\pi_p}, \rvx_{\pi_q})=0$, $\forall p\neq q$ since $\rvx_{\pi_p}$, $\rvx_{\pi_q}$ are independent for SWR. Thus, we can get $\E\norm{\overline \rvx_\pi - \overline \vx}^2 = \frac{1}{s^2}\sum_{p=1}^s\Var(\rvx_{\pi_p}) = \frac{\zeta^2}{s}$.
		\item SWOR: For $p\neq q$, we have
		\begin{align*}
			\Cov(\rvx_{\pi_p}, \rvx_{\pi_q}) &= \E \left[\left\langle \rvx_{\pi_p} - \overline \vx, \rvx_{\pi_q} - \overline \vx\right\rangle\right]\\
			&=\sum_{i=1}^n\sum_{j\neq i}^n\left\langle \vx_i - \overline \vx, \vx_j - \overline \vx\right\rangle\cdot \Pr(\rvx_{\pi_p}=\vx_i, \rvx_{\pi_q} = \vx_j).
		\end{align*}
		Since there are $n(n-1)$ possible combinations of $(\rvx_{\pi_p},\rvx_{\pi_q})$ and each has the same probability, we get $\Pr(\rvx_{\pi_p}=\vx_i, \rvx_{\pi_q} = \vx_j) = \frac{1}{n(n-1)}$. As a consequence, we have
		\begin{align}
			\Cov(\rvx_{\pi_p}, \rvx_{\pi_q}) &=\frac{1}{n(n-1)}\sum_{i=1}^n\sum_{j\neq i}^n\left[\left\langle \vx_i - \overline \vx, \vx_j - \overline \vx\right\rangle\right] \nonumber\\
			&= \frac{1}{n(n-1)}\norm{\sum_{i=1}^n\left( \vx_i - \overline \vx\right)}^2- \frac{1}{n(n-1)}\sum_{i=1}^n \norm{\vx_i-\overline \vx}^2 \nonumber\\
			&= -\frac{\zeta^2}{n-1}.\label{eq:proof:lem:simple random sampling}
		\end{align}
		Thus we have $\E\norm{\overline \rvx_\pi - \overline \vx}^2 = \frac{\zeta^2}{s} - \frac{s(s-1)}{s^2}\cdot\frac{\zeta^2}{n-1}=\frac{(n-s)}{s(n-1)}\zeta^2$.
	\end{itemize}
\end{proof}
\begin{lemma}\label{lem:sequential partial participation}
	Under the same conditions of Lemma~\ref{lem:simple random sampling}, use the way ``sampling without replacement'' and let $b_{m,k}(i)= \begin{cases}
		K-1,& i \leq m-1\\
		k-1,& i = m
	\end{cases}$. Then for $S\leq M$ ($M\geq 2$), it holds that
	\begin{align}
		\sum_{m=1}^S\sum_{k=0}^{K-1}\E\norm{\sum_{i=1}^{m}\sum_{j=0}^{b_{m,k}(i)}\left(\rvx_{\pi_i} - \overline \vx\right)}^2 \leq \frac{1}{2}S^2K^3\zeta^2
	\end{align}
\end{lemma}
\begin{proof}
	For clarity, we will use ``Term$_n$'' to denote the $n$-th term on the right hand side in some equation in the following proofs.
		
	Let us focus on the term in the following:
	\begin{align}
		&\E\norm{\sum_{i=1}^{m}\sum_{j=0}^{b_{m,k}(i)}\left(\rvx_{\pi_i} - \overline \vx\right)}^2 \nonumber\\
		&= \E\norm{K\sum_{i=1}^{m-1}\left(\rvx_{\pi_i} - \overline \vx\right) + k\left(\rvx_{\pi_m} - \overline \vx\right)}^2\nonumber\\
		&= K^2\E\norm{\sum_{i=1}^{m-1}\left(\rvx_{\pi_i} - \overline \vx\right)}^2 + k^2\E\norm{ \rvx_{\pi_m} - \overline \vx}^2 + 2Kk\E\left[\inp{\sum_{i=1}^{m-1}\left(\rvx_{\pi_i} - \overline \vx\right)}{\left(\rvx_{\pi_m} - \overline \vx\right)}\right]\label{eq1:proof:lem:sequential partial participation}\,.
	\end{align}
	We next bound the terms on the right hand side in Eq.~\eqref{eq1:proof:lem:sequential partial participation} one by one:
	\begin{align*}
		&\Term{1}{\eqref{eq1:proof:lem:sequential partial participation}} \overset{\eqref{eq:lem:sampling without replacement}}{=} \frac{(m-1)(M-(m-1))}{M-1}K^2\zeta^2 ,\\
		&\Term{2}{\eqref{eq1:proof:lem:sequential partial participation}} = k^2 \zeta^2\,,\\
		&\Term{3}{\eqref{eq1:proof:lem:sequential partial participation}} \overset{\eqref{eq:proof:lem:simple random sampling}}{=} -\frac{2(m-1)}{M-1}Kk\zeta^2 ,
	\end{align*}
	where we use \eqref{eq:proof:lem:simple random sampling} in the last equality, since $i \in \{1,2,\ldots,m-1\} \neq m$. With these three preceding equations, we get
	\begin{align*}
		\E\norm{\sum_{i=1}^{m}\sum_{j=0}^{b_{m,k}(i)}\left(\rvx_{\pi_i} - \overline \vx\right)}^2 = \frac{(m-1)(M-(m-1))}{M-1}K^2\zeta^2 + k^2\zeta^2 - \frac{2(m-1)}{M-1}Kk\zeta^2\,.
	\end{align*}
	Then summing the preceding terms over $m$ and $k$, we can get
	\begin{align*}
		&\sum_{m=1}^S\sum_{k=0}^{K-1}\E\norm{\sum_{i=1}^{m}\sum_{j=0}^{b_{m,k}(i)}\left(\rvx_{\pi_i} - \overline \vx\right)}^2\nonumber\\
		&=\frac{MK^3\zeta^2}{M-1}\sum_{m=1}^S(m-1)-\frac{K^3\zeta^2}{M-1}\sum_{m=1}^S(m-1)^2+ S\zeta^2\sum_{k=0}^{K-1}k^2-\frac{2K\zeta^2}{M-1}\sum_{m=1}^S (m-1)\sum_{k=0}^{K-1}k\,.
	\end{align*}
	Then applying the facts $\sum_{k=1}^{K-1}k = \frac{(K-1)K}{2}$ and $\sum_{k=1}^{K-1}k^2 = \frac{(K-1)K(2K-1)}{6}$, we can simplify the preceding equation as
	\begin{align}
		&\sum_{m=1}^S\sum_{k=0}^{K-1}\E\norm{\sum_{i=1}^{m}\sum_{j=0}^{b_{m,k}(i)}\left(\rvx_{\pi_i} - \overline \vx\right)}^2 \nonumber\\
		&= \zeta^2\left(\frac{1}{2}SK^2(SK-1) - \frac{1}{6}SK(K^2-1)-\frac{1}{M-1}(S-1)SK^2\left(\frac{1}{6}(2S-1)K-\frac{1}{2}\right)\right)\label{eq:lem:sequential partial participation:last equality}\\
		&\leq \frac{1}{2}S^2K^3\zeta^2,\nonumber
	\end{align}
	which is the claim of this lemma.
	
	\begin{mdframed}[backgroundcolor=gray!20, hidealllines=false, roundcorner=0pt]
	Details of Eq.~\eqref{eq:lem:sequential partial participation:last equality}. Let
	\begin{align*}
	T \coloneqq \sum_{m=1}^S\sum_{k=0}^{K-1}\E\norm{\sum_{i=1}^{m}\sum_{j=0}^{b_{m,k}(i)}\left(\rvx_{\pi_i} - \overline \vx\right)}^2\,.
	\end{align*}
	For convenience, drop all the $\zeta$'s for a while:
	\begin{align*}
		T
		&= \frac{MK^3}{M-1}\sum_{m=1}^S(m-1)-\frac{K^3}{M-1}\sum_{m=1}^S(m-1)^2+ S\sum_{k=0}^{K-1}k^2-\frac{2K}{M-1}\sum_{m=1}^S (m-1)\sum_{k=0}^{K-1}k\\
		&=\frac{MK^3}{M-1}\frac{(S-1)S}{2} - \frac{K^3}{M-1}\frac{(S-1)S(2S-1)}{6}\\ 
		&\quad+ S\frac{(K-1)K(2K-1)}{6} - \frac{2K}{M-1}\frac{(S-1)S}{2}\frac{(K-1)K}{2}.
	\end{align*}
	Separating the first term in to two terms, we have
	\begin{align*}
		T &= \frac{(S-1)SK^3}{2} +\frac{K^3}{M-1}\frac{(S-1)S}{2}- \frac{K^3}{M-1}\frac{(S-1)S(2S-1)}{6}\\
		&\quad  + S\frac{(K-1)K(2K-1)}{6} - \frac{2K}{M-1}\frac{(S-1)S}{2}\frac{(K-1)K}{2}\,.
	\end{align*}
	Collecting the terms containing $\frac{1}{M-1}$ together, we have
	\begin{align*}
		T &= \frac{(S-1)SK^3}{2} + S\frac{(K-1)K(2K-1)}{6}\\
		&\quad+\frac{K^3}{M-1}\frac{(S-1)S}{2}- \frac{K^3}{M-1}\frac{(S-1)S(2S-1)}{6}  - \frac{2K}{M-1}\frac{(S-1)S}{2}\frac{(K-1)K}{2}\,.
	\end{align*}
	For the first two terms, we have
	\begin{align*}
		\text{Term}_{1+2} &= \frac{S^2K^3}{2}- SK \left( \frac{3K^2}{6} - \frac{(K-1)(2K-1)}{6} \right)\\
		&=\frac{S^2K^3}{2} - \frac{SK(K^2+3K-1)}{6}\\
		&=\frac{1}{2}SK^2(SK-1) - \frac{1}{6}SK(K^2-1)\,.
	\end{align*}
	For the remaining three terms, we have
	\begin{align*}
		\text{Term}_{3+4+5} &= \frac{K^2}{M-1} \left( \frac{(S-1)SK}{2}-\frac{(S-1)S(2S-1)K}{6} - \frac{(S-1)S(K-1)}{2} \right)\\
		&= - \frac{K^2}{M-1} \left( \frac{(S-1)S(2S-1)K}{6} - \frac{(S-1)S}{2} \right)\\
		&= - \frac{(S-1)SK^2}{2(M-1)}\left( \frac{1}{3}(2S-1)K - 1 \right)\,.
	\end{align*}
	Now, we get
	\begin{align*}
		T = \frac{1}{2}SK^2(SK-1) - \frac{1}{6}SK(K^2-1) - \frac{1}{2}\frac{(S-1)S}{(M-1)}K^2\left( \frac{1}{3}(2S-1)K - 1 \right) \,.
	\end{align*}
	Recovering $\zeta$'s yields Eq.~\eqref{eq:lem:sequential partial participation:last equality}. 
	\end{mdframed}
\end{proof}
\section{Proofs of Theorem~\ref{thm:SFL}}\label{sec:proof SFL}
In this section, we provide the proofs of Theorem~\ref{thm:SFL} for the strongly convex, general convex and non-convex cases in \ref{subsec:SFL strongly convex}, \ref{subsec:SFL general convex} and \ref{subsec:SFL non-convex}, respectively.

In the following proofs, we consider the partial client participation setting. Specifically, in each training round $r$, given a complete permutation of the clients, $\pi^{(r)}=(\pi_1^{(r)},\pi_2^{(r)},\ldots,\pi_M^{(r)})$, only the first $S$ clients $\pi_1^{(r)},\pi_2^{(r)},\ldots,\pi_S^{(r)}$ participate in training. The permutation $\pi^{(r)}$ is sampled uniformly at random and independently across training rounds.

According to Algorithm~\ref{algorithm1}, the global update of SFL after one complete training round (with $S$ clients selected for training) is
\begin{align}
	\Delta^{(r)} = \rvx^{(r+1)}-\rvx^{(r)} =-\eta\sum_{m=1}^S\sum_{k=0}^{K-1}\rvg_{\pi_m^{(r)}}(\rvx_{m,k}^{(r)}).\label{eq:SFL:complete update}
\end{align}
The local update of SFL from $\rvx^{(r)}$ to $\rvx_{m,k}^{(r)}$ is
\begin{align}
	\rvx_{m,k}^{(r)}-\rvx^{(r)}=-\eta\sum_{i=1}^{m} \sum_{j=0}^{b_{m,k}(i)} \rvg_{\pi_i^{(r)}}(\rvx_{i,j}^{(r)}) \quad \text{with}\quad  b_{m,k}(i) \coloneqq \begin{cases}
		K-1,& i \leq m-1\\
		k-1,& i = m
	\end{cases}.\label{eq:SFL:partial update}
\end{align}
For brevity, we also define $\gB_{m,k} \coloneqq \sum_{i=1}^{m}\sum_{j=0}^{b_{m,k}(i)}1 = (m-1)K+k$.
Define $\E_r[\cdot]\coloneqq \E[\cdot \mid \rvx^{(r)}]$. Similar to the ``client drift'' in PFL \citep{karimireddy2020scaffold}, we define the client drift in SFL:
\begin{align}
	E_r \coloneqq \sum_{m=1}^S\sum_{k=0}^{K-1} \E_r\left[\Norm{\rvx_{m,k}^{(r)} - \rvx^{(r)}}^2\right].\label{eq:SFL:client drift}
\end{align}
For clarity, we will use ``Term$_n$'' to denote the $n$-th term on the right hand side in some equation in the following proofs.
\subsection{Strongly convex case}\label{subsec:SFL strongly convex}
\subsubsection{Finding the recursion}
\begin{lemma}\label{lem:SFL:strongly convex:recursion}
	Let Assumptions~\ref{asm:smoothness}, \ref{asm:stochasticity}, \ref{asm:heterogeneity2} hold and assume that all the local objectives are $\mu$-strongly convex. If the learning rate satisfies $\eta \leq \frac{1}{6LSK}$, then it holds that
	\begin{align}
		\E_r\left[\Norm{\rvx^{(r+1)}-\rvx^\ast}^2\right] &\leq \left(1-\frac{1}{2}\eta\mu SK\right)\Norm{\rvx^{(r)}-\rvx^\ast}^2+4SK\eta^2\sigma^2+4S^2K^2\eta^2\frac{M-S}{S(M-1)}\zeta_\ast^2 \nonumber\\
		&\quad-\frac{2}{3}SK\eta D_F(\rvx^{(r)},\rvx^\ast)+\frac{8}{3}L\eta\sum_{m=1}^S\sum_{k=0}^{K-1}\E_r\left[\Norm{\rvx_{m,k}^{(r)} - \rvx^{(r)}}^2\right]\label{eq:lem:SFL:strongly convex:recursion}
	\end{align}
\end{lemma}
\begin{proof}
	In the following, we focus on a single training round, and hence we drop the superscripts $r$ for a while, e.g., replacing $\rvx_{m,k}^{(r)}$ with $\rvx_{m,k}$ and replacing $\rvx^{(r)}$ with $\rvx$.
	
	We start from the following equation:
	\begin{align}
		\E_r\norm{\rvx+\Delta-\rvx^\ast}^2 =\norm{\rvx-\rvx^*}^2 + 2\E_r\left[\inp{\rvx-\rvx^*}{\Delta}\right]+\E_r\norm{\Delta}^2.\label{eq:SFL:convex:single round decomposition}
	\end{align}
	For $2\E_r\left[\inp{\rvx-\rvx^*}{\Delta}\right]$, substituting the global update $\Delta$, we obtain
	\begin{align*}
		&2\E_r\left[\inp{\rvx-\rvx^*}{\Delta}\right]\nonumber \\
		&=-2\eta\sum_{m=1}^S\sum_{k=0}^{K-1}\E_r\left[\inp{ \rvg_{\pi_m}(\rvx_{m,k})}{\rvx-\rvx^\ast}\right] \nonumber\\
		&=-2\eta\sum_{m=1}^S\sum_{k=0}^{K-1}\E_r\left[ \E_r\left[\inp{ \rvg_{\pi_m}(\rvx_{m,k})}{\rvx-\rvx^\ast}\mid \rvx_{m,k}\right] \right]\nonumber\\
		&=-2\eta\sum_{m=1}^S\sum_{k=0}^{K-1}\E_r\left[\inp{\nabla F_{\pi_m}(\rvx_{m,k})}{\rvx-\rvx^\ast}\right] \nonumber\\
		&\leq -2\eta\sum_{m=1}^S\sum_{k=0}^{K-1} \E_r\left[F_{\pi_m}(\rvx)-F_{\pi_m}(\rvx^\ast) + \frac{\mu}{4}\norm{\rvx-\rvx^\ast}^2 - L\norm{\rvx_{m,k}-\rvx}^2\right] \nonumber\\
		&\leq -2SK\eta D_F(\rvx,\rvx^\ast) - \frac{1}{2}\mu SK\eta\norm{\rvx-\rvx^\ast}^2 + 2L\eta\sum_{m=1}^S\sum_{k=0}^{K-1}\E_r\norm{\rvx_{m,k}-\rvx}^2,
	\end{align*}
	where the second equality uses the law of total expectation, the third equality uses \cref{asm:stochasticity}, and the first inequality uses Lemma~\ref{lem:perturbed strong convexity} with $\vx=\rvx_{m,k}$, $\vy=\rvx^\ast$, $\vz=\rvx$ and $h = F_{\pi_m}$.
	
	For $\E_r\norm{\Delta}^2$, substituting the global update $\Delta$, we obtain
	\begin{align}
		\E_r\norm{\Delta}^2
		&\leq 4\eta^2\E_r\norm{\sum_{m=1}^S\sum_{k=0}^{K-1} \left(\rvg_{\pi_m}(\rvx_{m,k}) - \nabla F_{\pi_m}(\rvx_{m,k})\right)}^2  \nonumber\\
		&\quad+ 4\eta^2\E_r\norm{\sum_{m=1}^S\sum_{k=0}^{K-1} \left(\nabla F_{\pi_m}(\rvx_{m,k}) - \nabla F_{\pi_m}(\rvx)\right)}^2 \nonumber\\
		&\quad + 4\eta^2\E_r\norm{\sum_{m=1}^S\sum_{k=0}^{K-1} \left(\nabla F_{\pi_m}(\rvx) - \nabla F_{\pi_m}(\rvx^\ast)\right)}^2  \nonumber\\
		&\quad+ 4\eta^2\E_r\norm{\sum_{m=1}^S\sum_{k=0}^{K-1} \nabla F_{\pi_m}(\rvx^\ast)}^2.\label{eq:recursion-strongly-convex-1}
	\end{align}
	
	For the first term on the right-hand side of
	\eqref{eq:recursion-strongly-convex-1}, we have
	\begin{align*}
		\Term{1}{\eqref{eq:recursion-strongly-convex-1}}
		&= 4\eta^2 \sum_{m=1}^S\sum_{k=0}^{K-1}
		\E_r\norm{\rvg_{\pi_m}(\rvx_{m,k})
			- \nabla F_{\pi_m}(\rvx_{m,k})}^2 \\
		&\leq 4\eta^2SK\sigma^2,
	\end{align*}
	where the equality uses \cref{lem:martingale difference property} and the inequality uses \cref{asm:stochasticity}. Here, we apply Lemma~\ref{lem:martingale difference property}
	by identifying the data sample $\xi_{m,k}$, the stochastic gradient $\rvg_{\pi_m}(\rvx_{m,k})$, and the gradient $\nabla F_{\pi_m}(\rvx_{m,k})$ with $\xi_i$, $\rvx_i$, and $\rve_i$ in the lemma, respectively.
	
	For the second term on the right hand side in \eqref{eq:recursion-strongly-convex-1},
	\begin{align*}
		\Term{2}{\eqref{eq:recursion-strongly-convex-1}}
		&\leq 4\eta^2SK\sum_{m=1}^S\sum_{k=0}^{K-1}\E_r\norm{\nabla F_{\pi_m}(\rvx_{m,k}) - \nabla F_{\pi_m}(\rvx)}^2\\
		&\leq 4L^2\eta^2SK\sum_{m=1}^S\sum_{k=0}^{K-1}\E_r\norm{\rvx_{m,k} - \rvx}^2,
	\end{align*}
	where the first inequality uses Jensen's inequality and the second inequality uses \cref{asm:smoothness}.
	
	For the third term on the right hand side in \eqref{eq:recursion-strongly-convex-1},
	\begin{align*}
		\Term{3}{\eqref{eq:recursion-strongly-convex-1}}
		&\leq 4\eta^2SK\sum_{m=1}^S\sum_{k=0}^{K-1}\E_r\norm{\nabla F_{\pi_m}(\rvx) - \nabla F_{\pi_m}(\rvx^\ast)}^2 \\
		&\overset{\eqref{eq:smooth+convex:bregman lower bound}}{\leq} 8L\eta^2SK\sum_{m=1}^S\sum_{k=0}^{K-1}\E_r\left[D_{F_{\pi_m}}(\rvx, \rvx^\ast)\right] \\
		&= 8L\eta^2S^2K^2D_{F}(\rvx,\rvx^\ast).
	\end{align*}
	
	For the fourth term on the right hand side in \eqref{eq:recursion-strongly-convex-1}, using \cref{lem:simple random sampling}, we obtain
	\begin{align*}
		\Term{4}{\eqref{eq:recursion-strongly-convex-1}} &\leq  4\eta^2S^2K^2\frac{M-S}{S(M-1)}\zeta_\ast^2.
	\end{align*}
	
	With the preceding bounds for the terms on the right hand side in \eqref{eq:recursion-strongly-convex-1}, we have
	\begin{align*}
		\E_r\norm{\Delta }^2 &\leq 4\eta^2SK\sigma^2+4\eta^2S^2K^2\frac{M-S}{S(M-1)}\zeta_\ast^2\\
		&\quad+ 8L\eta^2S^2K^2D_{F}(\rvx, \rvx^\ast)+4L^2\eta^2SK\sum_{m=1}^S\sum_{k=0}^{K-1}\E_r\left[\norm{\rvx_{m,k} - \rvx}^2\right] \,.
	\end{align*}
	Plugging the bounds of $2\E_r\left[\inp{\rvx-\rvx^*}{\Delta}\right]$ and $\E_r\norm{\Delta}^2$ into \eqref{eq:SFL:convex:single round decomposition}, and using $\eta \leq \frac{1}{6LSK}$, we obtain
	\begin{align*}
		\E_r\norm{\rvx+\Delta-\rvx^\ast}^2
		&\leq \left(1-\frac{1}{2}\eta\mu SK\right)\norm{\rvx-\rvx^\ast}^2+4SK\eta^2\sigma^2+4S^2K^2\eta^2\frac{M-S}{S(M-1)}\zeta_\ast^2 \nonumber\\
		&\quad-\frac{2}{3}SK\eta D_F(\rvx, \rvx^\ast)+\frac{8}{3}L\eta\sum_{m=1}^S\sum_{k=0}^{K-1}\E_r\left[\norm{\rvx_{m,k} - \rvx}^2\right].
	\end{align*}
	The claim follows after recovering the superscripts.
\end{proof}
\subsubsection{Bounding the client drift with Assumption~\ref{asm:heterogeneity2}}\label{subsubsec:SFL strongly convex:client drift}
\begin{lemma}\label{lem:SFL:strongly convex:drift}
	Let Assumptions~\ref{asm:smoothness}, \ref{asm:stochasticity}, \ref{asm:heterogeneity2} hold and assume that all the local objectives are $\mu$-strongly convex. If the learning rate satisfies $\eta \leq \frac{1}{6LSK}$, then the client drift is bounded:
	\begin{align}
		E_r\leq \frac{9}{4}S^2K^2\eta^2\sigma^2 + \frac{9}{4}S^2K^3\eta^2\zeta_\ast^2 + 3LS^3K^3\eta^2D_{F}(\rvx^{(r)},\rvx^\ast) \label{eq:lem:SFL:strongly convex:drift}
	\end{align}
\end{lemma}
\begin{proof}
	In the following, we focus on a single training round, and hence we drop the superscripts $r$ for a while, e.g., replacing $\rvx_{m,k}^{(r)}$ with $\rvx_{m,k}$ and replacing $\rvx^{(r)}$ with $\rvx$.
	
	Using \cref{eq:SFL:partial update}, we obtain
	\begin{align}
		\E_r\norm{\rvx_{m,k} - \rvx}^2 &\leq 4\eta^2\E_r\norm{\sum_{i=1}^{m}\sum_{j=0}^{b_{m,k}(i)} \left(\rvg_{\pi_i}(\rvx_{i,j}) - \nabla F_{\pi_i} (\rvx_{i,j})\right)}^2 \nonumber\\
		&\quad+ 4\eta^2\E_r\norm{\sum_{i=1}^{m}\sum_{j=0}^{b_{m,k}(i)}\left(\nabla F_{\pi_i} (\rvx_{i,j}) - \nabla F_{\pi_i} (\rvx)\right)}^2 \nonumber\\
		&\quad+4\eta^2\E_r \norm{\sum_{i=1}^{m}\sum_{j=0}^{b_{m,k}(i)}\left(\nabla F_{\pi_i} (\rvx) - \nabla F_{\pi_i} (\rvx^{\ast})\right)}^2 \nonumber\\
		&\quad + 4\eta^2\underbrace{\E_r\norm{\sum_{i=1}^{m}\sum_{j=0}^{b_{m,k}(i)}\nabla F_{\pi_i} (\rvx^{\ast})}^2}_{T_{m,k}}\,.\label{eq:drift-strongly-convex-1}
	\end{align}	
	
	For the first term on the right hand side in \eqref{eq:drift-strongly-convex-1},
	\begin{align*}
		\Term{1}{\eqref{eq:drift-strongly-convex-1}}&= 4\eta^2 \sum_{i=1}^m \sum_{j=0}^{b_{m,k}(i)}\E_r\norm{\rvg_{\pi_i}(\rvx_{i,j})-\nabla F_{\pi_i} (\rvx_{i,j})}^2\\ &\leq 4\eta^2\gB_{m,k}\sigma^2 ,
	\end{align*}
	where we use \cref{lem:martingale difference property} and \cref{asm:stochasticity}.
	
	For the second term on the right hand side in \eqref{eq:drift-strongly-convex-1},
	\begin{align*}
		\Term{2}{\eqref{eq:drift-strongly-convex-1}} &\leq 4\eta^2\gB_{m,k}\sum_{i=1}^{m}\sum_{j=0}^{b_{m,k}(i)}\E_r\norm{\nabla F_{\pi_i} (\rvx_{i,j}) - \nabla F_{\pi_i} (\rvx)}^2\\
		&\leq 4L^2\eta^2\gB_{m,k}\sum_{i=1}^{m}\sum_{j=0}^{b_{m,k}(i)}\E_r\norm{\rvx_{i,j} - \rvx}^2,
	\end{align*}
	where the first inequality uses Jensen's inequality and the second inequality uses \cref{asm:smoothness}.
	
	For the third term on the right hand side in \eqref{eq:drift-strongly-convex-1},
	\begin{align*}
		\Term{3}{\eqref{eq:drift-strongly-convex-1}} 
		&\leq 4\eta^2\gB_{m,k}\sum_{i=1}^{m}\sum_{j=0}^{b_{m,k}(i)}\E_r\norm{\nabla F_{\pi_i} (\rvx)- \nabla F_{\pi_i} (\rvx^{\ast})}^2\\
		&\overset{\eqref{eq:smooth+convex:bregman lower bound}}{\leq} 8L\eta^2\gB_{m,k}\sum_{i=1}^{m}\sum_{j=0}^{b_{m,k}(i)} \E_r\left[D_{F_{\pi_i}}(\rvx, \rvx^\ast) \right] \\
		&\leq 8L\eta^2 \gB_{m,k}^2 D_{F}(\rvx, \rvx^\ast),
	\end{align*}
	where the first inequality uses Jensen's inequality.
	
	Then, plugging the preceding bounds into \eqref{eq:drift-strongly-convex-1}, and then plugging the resulting bound for $\E_r\norm{\rvx_{m,k} - \rvx}^2$ into $E_r$, we obtain
	\begin{align*}
		E_r &\leq4\eta^2\sigma^2\sum_{m=1}^S\sum_{k=0}^{K-1}\gB_{m,k}+ 4L^2\eta^2\sum_{m=1}^S\sum_{k=0}^{K-1}\gB_{m,k}\sum_{i=1}^{m}\sum_{j=0}^{b_{m,k}(i)}\E_r\norm{\rvx_{i,j} - \rvx}^2\\ &\quad+8L\eta^2\sum_{m=1}^S\sum_{k=0}^{K-1}\gB_{m,k}^2D_{F}(\rvx,\rvx^\ast) + 4\eta^2\sum_{m=1}^S\sum_{k=0}^{K-1}T_{m,k}\,.
	\end{align*}
	Noting that the last term can be bounded by Lemma~\ref{lem:sequential partial participation} with $\rvx_{\pi_i}=\nabla F_{\pi_i}(\rvx^\ast)$ and $\overline\vx=\nabla F(\rvx^\ast)=0$ and using $\sum_{m=1}^S\sum_{k=0}^{K-1}\gB_{m,k} \leq \frac{1}{2}S^2K^2$ and $\sum_{m=1}^S\sum_{k=0}^{K-1}\gB_{m,k}^2 \leq \frac{1}{3}S^3K^3$, we can simplify the preceding inequality as
	\begin{align*}
		E_r\leq2S^2K^2\eta^2\sigma^2+2L^2S^2K^2\eta^2E_r+\frac{8}{3}LS^3K^3\eta^2D_{F}(\rvx,\rvx^\ast) + 2S^2K^3\eta^2\zeta_\ast^2\,.
	\end{align*}
	After rearranging the preceding inequality and using the condition $\eta \leq \frac{1}{6LSK}$, we get
	\begin{align*}
		E_r\leq \frac{9}{4}S^2K^2\eta^2\sigma^2 + \frac{9}{4}S^2K^3\eta^2\zeta_\ast^2 + 3LS^3K^3\eta^2D_{F}(\rvx,\rvx^\ast)\,.
	\end{align*}
	The claim follows after recovering the superscripts.
\end{proof}
\subsubsection{Tuning the learning rate}
For completeness, we reproduce \cite{karimireddy2020scaffold}'s Lemma 1 (linear convergence rate) based on \cite{stich2019unified} and \citet{stich2019error}.
\begin{lemma}[\cite{karimireddy2020scaffold}]\label{lem:stongly convex:tuning learning rate}
	We consider two non-negative sequences $\{r_t\}_{t\geq 0}$, $\{s_t\}_{t\geq 0}$, which satisfies the relation
	\begin{align}
		r_{t+1} \leq (1-a\gamma ) r_t - b\gamma s_t +  c \gamma^2,\label{eq1:lem:stongly convex:tuning learning rate}
	\end{align}
	for all $t \geq 0$ and 
	for parameters $b > 0$, $a, c \geq 0$ and the learning rate $\gamma$ with $0\leq \gamma \leq \frac{1}{d}$, for a parameter $d > a$.
	
	Then there exists a constant learning rate $\gamma \leq \frac{1}{d}$ and the weights $w_t = (1-a\gamma)^{-(t+1)}$ and $W_T := \sum_{t=0}^T w_t$ such that
	\begin{align*}
		\Psi_T \coloneqq \frac{1}{W_T} \sum_{t=0}^{T} s_t w_t \leq \frac{3ar_0}{b}(1-a\gamma)^{T+1} + \frac{c\gamma}{b} \leq \frac{3ar_0}{b} \exp \left(-a\gamma (T+1)\right) +\frac{c\gamma}{b}\,.
	\end{align*}
	
	By choosing $\gamma = \min \left\{\frac{1}{d}, \frac{\ln(\max\{2,a^2 r_0 T /c\})}{a T}\right\}$, for $T \geq \frac{d}{2a}$, we have
	\begin{align*}
		\Psi_T = \tilde \gO \left(  a r_0 \exp \left(-\frac{aT}{d} \right) + \frac{c}{aT} \right)\,.
	\end{align*}
\end{lemma}
\begin{proof}
	We start by rearranging~\eqref{eq1:lem:stongly convex:tuning learning rate} and multiplying both sides with $w_t$:
	\begin{align*}
		b s_t w_t \leq \frac{w_t (1-a\gamma) r_t}{\gamma} - \frac{w_t r_{t+1}}{\gamma}  + c \gamma w_t = \frac{w_{t-1} r_t}{\gamma} - \frac{w_t r_{t+1}}{\gamma}  + c \gamma w_t \,.
	\end{align*} 
	By summing from $t=0$ to $t=T$, we obtain a telescoping sum:
	\begin{align*}
		b \sum_{t=0}^{T} s_t w_t \leq \frac{1}{\gamma} \left(w_0 (1-a\gamma)r_0 - w_{T} r_{T+1}\right) + c \gamma\sum_{t=0}^T w_t, 
	\end{align*}
	Dividing both sides by $W_T \coloneqq \sum_{t=0}^T w_t$, it follows that
	\begin{align}
		\Psi_T \coloneqq \frac{1}{W_T} \sum_{t=0}^{T} s_t w_t \leq \frac{1}{b\gamma W_T} \left(w_0 (1-a\gamma)r_0 - w_{T} r_{T+1}\right) + \frac{c\gamma}{b},\label{eq:proof:lem:stongly convex:tuning learning rate}
	\end{align}
	Note that the proof of \cite{stich2019unified}'s Lemma 2 used $W_T \geq w_T = (1-a\gamma)^{-(T+1)}$ to estimate $W_T$. It is reasonable given that $w_T$ is extremely larger than all the other terms $w_t$ ($t < T$) when $T$ is large. Yet \cite{karimireddy2020scaffold} goes further, showing that $W_T$ can be estimated more precisely:
	\begin{align*}
		W_T = \sum_{t=0}^T w_t = (1-a\gamma)^{-(T+1)}\sum_{t=0}^T (1-a\gamma)^t = (1-a\gamma)^{-(T+1)}\left(\frac{1-(1-a\gamma)^{T+1}}{a\gamma}\right)
	\end{align*}
	When $(T+1)\geq \frac{1}{2a\gamma}$, $(1-a\gamma)^{T+1}\leq \exp(-a\gamma(T+1)) \leq e^{-\frac{1}{2}} \leq \frac{2}{3}$, so it follows that
	\begin{align*}
		W_T = (1-a\gamma)^{-(T+1)}\left(\frac{1-(1-a\gamma)^{T+1}}{a\gamma}\right) \geq \frac{(1-a\gamma)^{-(T+1)}}{3a\gamma}
	\end{align*}
	With the estimates
	\begin{itemize}[leftmargin=2em]
		\item $W_T = (1-a\gamma)^{-(T+1)}\sum_{t=0}^T (1-a \gamma)^t \leq \frac{w_{T}}{a \gamma}$ (here we use $a\gamma \leq \frac{a}{d} \leq 1$),
		\item and $W_T \geq \frac{(1-a\gamma)^{-(T+1)}}{3a\gamma}$,
	\end{itemize}
	we can further simplify \eqref{eq:proof:lem:stongly convex:tuning learning rate}:
	\begin{align*}
		\Psi_T \leq \frac{3ar_0}{b}(1-a\gamma)^{T+1} + \frac{c\gamma}{b} \leq \frac{3ar_0}{b} \exp \left(-a\gamma (T+1)\right) +\frac{c\gamma}{b}\,,
	\end{align*}
	which is the first result of this lemma. We use the additional constraint $(T+1) \geq \frac{1}{2a\gamma}$ in this part.
	
	For clarity, we consider a looser bound $\Psi_T \leq \frac{3ar_0}{b} \exp \left(-a\gamma T\right) +\frac{c\gamma}{b}$ and a stricter constraint $T \geq \frac{1}{2a\gamma}$ with respect to $T$. Choosing $\gamma = \min \left\{\frac{1}{d}, \frac{\ln(\max\{2,a^2 r_0 T /c\})}{a T}\right\}$, we can get the second result of this lemma. We can verify it with the following cases:
	\begin{itemize}[leftmargin=2em]
		\item Case 1: If $\frac{\ln(\max\{2,a^2 r_0 T /c\})}{a T}\leq \frac{1}{d}$, we choose $\gamma = \frac{\ln(\max\{2,a^2 r_0 T /c\})}{a T}$ and get
		\begin{align*}
			\Psi_T \leq \frac{3ar_0}{b} \exp\left( -\ln\left(\max\{2,a^2r_0T/c\}\right) \right) + \frac{c}{aT}\cdot\frac{1}{b}\ln\left(\max\{2,a^2r_0T/c\}\right)
		\end{align*}
		When $a^2r_0T/c \leq 2$, it follows that
		\begin{align*}
			\Psi_T &\leq ar_0 \cdot \frac{3}{2b} + \frac{c}{aT}\cdot\frac{1}{b}\ln\left(\max\{2,a^2r_0T/c\}\right)\\
			&\leq \frac{c}{aT} \cdot \frac{3}{b} + \frac{c}{aT} \cdot\frac{1}{b}\ln\left(\max\{2,a^2r_0T/c\}\right) \tag{$\because a r_0 \leq \frac{2c}{aT}$}\\
			&= \tilde \gO \left( \frac{c}{aT} \right)
		\end{align*}
		When $a^2r_0T/c > 2$, it follows that
		\begin{align*}
			\Psi_T &\leq \frac{c}{aT} \cdot \frac{3}{b} + \frac{c}{aT} \cdot\frac{1}{b}\ln\left(\max\{2,a^2r_0T/c\}\right) =\tilde \gO \left( \frac{c}{aT} \right)
		\end{align*}
		where the constant $b$ is omitted. Since $\ln(\max\{2,a^2 r_0 T /c\}) \geq 0.6$, the constraint $T \geq \frac{1}{2a\gamma}$ holds automatically.
		
		\item Case 2: If otherwise $\frac{\ln(\max\{2,a^2 r_0 T /c\})}{a T} \geq \frac{1}{d}$, then we choose $\gamma = \frac{1}{d}$ and get
		\begin{align*}
			\Psi_T &\leq{ar_0}\exp\left( -\frac{aT}{d} \right) \cdot \frac{3}{b} + \frac{c}{d}\cdot\frac{1}{b}\\
			&\leq ar_0\exp\left( -\frac{aT}{d} \right) \cdot \frac{3}{b} + \frac{c}{a T}\cdot\frac{1}{b}\ln(\max\{2,a^2 r_0 T /c\})\tag{$\because \frac{1}{d}\leq \frac{\ln(\max\{2,a^2 r_0 T /c\})}{a T}$}\\
			&= \tilde \gO \left( ar_0\exp\left( -\frac{aT}{d} \right) + \frac{c}{aT} \right)
		\end{align*}
		where the constant $b$ is omitted. In this case, we need the additional constraint $T \geq \frac{1}{2a\gamma} = \frac{d}{2a}$.
	\end{itemize}
	Combining these two cases, we get
	\begin{align*}
		\Psi_T = \tilde \gO \left(  a r_0 \exp \left(- \frac{aT}{d} \right) + \frac{c}{aT} \right),
	\end{align*}
	with an additional constraint $T\geq \frac{d}{2a}$. In other words, compared to \cite{stich2019unified}'s Lemma 2, this lemma achieves a tighter bound with a larger number of training rounds. 
	
	Notably, the optimization term $\gO\left(ar_0\exp\left( -\frac{aT}{d}\right)\right)$ and the additional constraint $T\geq \frac{d}{2a}$ arise only in Case 2 in which $\frac{\ln(\max\{2,a^2 r_0 T /c\})}{a T} \geq \frac{1}{d}$ and $\gamma$ is chosen as $\frac1d$.
\end{proof}
\subsubsection{Proof of strongly convex case of Theorem~\ref{thm:SFL} and Corollary~\ref{cor:SFL}}
\begin{proof}[Proof of the strongly convex case of Theorem~\ref{thm:SFL}]
	Substituting \eqref{eq:lem:SFL:strongly convex:drift} into \eqref{eq:lem:SFL:strongly convex:recursion}, taking unconditional expectation and using $\eta \leq \frac{1}{6LSK}$, we obtain
	\begin{align*}
		\E\left[\Norm{\rvx^{(r+1)}-\rvx^*}^2\right]
		&\leq \left(1-\frac{1}{2}\eta\mu SK\right)\E\left[\Norm{\rvx^{(r)}- \rvx^\ast}^2\right] - \frac{1}{3}SK\eta\E\left[D_F(\rvx^{(r)}, \rvx^\ast)\right] \nonumber\\
		&\quad+4SK\eta^2\sigma^2 +4S^2K^2\eta^2\frac{M-S}{S(M-1)}\zeta_\ast^2+6LS^2K^2\eta^3\sigma^2+6LS^2K^3\eta^3\zeta_\ast^2\,.
	\end{align*}
	Using $\tilde{\eta} = \eta SK$, we have
	\begin{align}
		\E\left[\Norm{\rvx^{(r+1)}-\rvx^*}^2\right] &\leq \left(1-\frac{1}{2} \tilde{\eta}\mu\right)\E\left[\Norm{\rvx^{(r)}- \rvx^\ast}^2\right] - \frac{\tilde{\eta}}{3}\E\left[D_F(\rvx^{(r)}, \rvx^\ast)\right]\nonumber\\
		&\quad+\frac{4\tilde{\eta}^2\sigma^2}{SK}+\frac{4\tilde{\eta}^2(M-S)\zeta_\ast^2}{S(M-1)}+\frac{6L\tilde{\eta}^3\sigma^2}{SK}+\frac{6L\tilde{\eta}^3\zeta_\ast^2}{S}\,.\label{eq:thm:strongly convex:simplified recursion}
	\end{align}
	Applying Lemma~\ref{lem:stongly convex:tuning learning rate} with $t=r$ ($T=R$), $\gamma=\tilde\eta$, $r_{t} = \E\left[\Norm{\rvx^{(r)}- \rvx^*}^2\right]$, $a = \frac{\mu}{2}$, $b=\frac{1}{3}$, $s_t = \E\left[D_F(\rvx^{(r)}, \rvx^\ast)\right]$, $w_t=(1-\tfrac{\mu\tilde\eta}{2})^{-(r+1)}$, $c_1 = \frac{4\sigma^2}{SK}+\frac{4(M-S)\zeta_\ast^2}{S(M-1)}$, $c_2 = \frac{6L\sigma^2}{SK} + \frac{6L\zeta_\ast^2}{S}$ and $\frac{1}{d}=\frac{1}{6L}$ ($\tilde\eta \leq\frac{1}{6L}$), we obtain
	\begin{align}
		&\E\left[F(\bar\rvx^{(R)})-F(\rvx^\ast)\right]\nonumber\\
		&\leq \frac{9}{2}\mu\Norm{\rvx^{(0)}- \rvx^*}^2 \exp\left(-\frac{1}{2}\mu\tilde\eta R\right)+\frac{12\tilde{\eta}\sigma^2}{SK}+\frac{12\tilde{\eta}(M-S)\zeta_\ast^2}{S(M-1)}+\frac{18L\tilde{\eta}^2\sigma^2}{SK}+\frac{18L\tilde{\eta}^2\zeta_\ast^2}{S}\,,\label{eq:thm:strongly convex:bound}
	\end{align}
	where $\bar\rvx^{(R)} = \frac{1}{\sum_{r=0}^Rw_r}\sum_{r=0}^Rw_r\rvx^{(r)}$. Note that there are no terms containing $\gamma^3$ in Lemma~\ref{lem:stongly convex:tuning learning rate}. As the terms containing $\gamma^3$ is not the determinant factor for the convergence rate, Lemma~\ref{lem:stongly convex:tuning learning rate} can also be applied to this case \citep{koloskova2020unified}. Thus, when $\tilde\eta \asymp\min \left\{\frac{1}{L}, \frac{1}{\mu R}\right\}$ and $R\geq 6\kappa$, $\E\left[F(\bar\rvx^{(R)})-F(\rvx^\ast)\right]$ is upper bounded by
	\begin{align}
		\tilde\gO\left(\mu D^2 \exp\left(-\frac{\mu R}{12L}\right)+\frac{\sigma^2}{\mu SKR}+\frac{(M-S)\zeta_\ast^2}{\mu SR(M-1)} + \frac{L\sigma^2}{\mu^2SKR^2} + \frac{L\zeta_\ast^2}{\mu^2SR^2}\right),\label{eq:thm:strongly convex:bound2}
	\end{align}
	where $D = \norm{\rvx^{(0)}- \rvx^\ast}$. Ineqs.~\eqref{eq:thm:strongly convex:bound},~\eqref{eq:thm:strongly convex:bound2} are the bounds for partial participation. When $S=M$, we get the claim of the strongly convex case of Theorem~\ref{thm:SFL} and Corollary~\ref{cor:SFL}.
\end{proof}
\subsection{General convex case}\label{subsec:SFL general convex}
\subsubsection{Tuning the learning rate}
\begin{lemma}[\cite{koloskova2020unified}]\label{lem:general convex:tuning learning rate}
	We consider two non-negative sequences $\{r_t\}_{t\geq 0}$, $\{s_t\}_{t\geq 0}$, which satisfies the relation
	\begin{align}
		r_{t+1} \leq r_t - b\gamma s_t +  c_1 \gamma^2 + c_2 \gamma^3,\label{eq1:lem:general convex:tuning learning rate}
	\end{align}
	for all $t \geq 0$ and 
	for parameters $b > 0$, $c_1, c_2 \geq 0$ and the learning rate $\gamma$ with $0\leq \gamma \leq \frac{1}{d}$.
	
	Then there exists a constant learning rate $\gamma  \leq \frac{1}{d}$ and the weights $w_t=1$ and $W_T \coloneqq \sum_{t=0}^Tw_t$ such that
	\begin{align}
		\Psi_T \coloneqq \frac{1}{W_T} \sum_{t=0}^T s_tw_t \leq \frac{r_0}{b\gamma(T+1)} + \frac{c_1\gamma}{b} + \frac{c_2\gamma^2}{b}\,.\label{eq:lem:general convex:tuning learning rate}
	\end{align}
	By choosing $\gamma = \min\left\{\left(\frac{r_0}{c_1T}\right)^\frac{1}{2}, \left(\frac{r_0}{c_2T}\right)^\frac{1}{3}, \frac{1}{d}\right\}$, we have
	\begin{align*}
		\Psi_T =\gO\left(\frac{dr_0}{T} +\frac{c_1^\frac{1}{2}r_0^\frac{1}{2}}{T^\frac{1}{2}} + \frac{c_2^\frac{1}{3}r_0^\frac{2}{3}}{T^\frac{2}{3}} \right).
	\end{align*}
\end{lemma}
	\begin{proof} 
	We start by rearranging Eq.~\eqref{eq1:lem:general convex:tuning learning rate} and multiplying both sides with $w_t=1$:
	\begin{align*}
		bs_t \leq \frac{r_{t}}{\gamma} - \frac{r_{t+1}}{\gamma} + c_1 \gamma + c_2 \gamma^2
	\end{align*}
	By summing from $t=0$ to $t=T$, we obtain a telescoping sum:
	\begin{align*}
		b\sum_{t=0}^Ts_t \leq \frac{1}{\gamma} (r_0 - r_{T+1}) + c_1 \gamma(T+1) + c_2 \gamma^2(T+1)
	\end{align*}
	Dividing both sides by $W_T\coloneqq \sum_{t=0}^T w_t = T+1$, it follows that
	\begin{align*}
		\Psi_T \coloneqq \frac{1}{W_T} \sum_{t=0}^T s_tw_t \leq \frac{r_0}{b\gamma(T+1)} + \frac{c_1\gamma}{b} + \frac{c_2\gamma^2}{b}
	\end{align*}
	which is the first result \eqref{eq:lem:general convex:tuning learning rate} of this lemma.
	
	For clarity, we consider a looser bound $\Psi_T \coloneqq \frac{1}{W_T} \sum_{t=0}^T s_tw_t \leq \frac{r_0}{b\gamma T} + \frac{c_1\gamma}{b} + \frac{c_2\gamma^2}{b}$ with respect to $T$.
	Letting $\frac{r_0}{b\gamma T} = \frac{c_1\gamma}{b}$ and $\frac{r_0}{b\gamma T} = \frac{c_2\gamma^2}{b}$, we can get two choices of $\gamma$, $\gamma = \left(\frac{r_0}{c_1T}\right)^\frac{1}{2}$ and $\gamma = \left(\frac{r_0}{c_2T}\right)^\frac{1}{3}$. Then by choosing $\gamma = \min\left\{\left(\frac{r_0}{c_1T}\right)^\frac{1}{2}, \left(\frac{r_0}{c_2T}\right)^\frac{1}{3}, \frac{1}{d}\right\}$, we can get the desired upper bound. We can verify it with the following cases:

	\begin{itemize}[leftmargin=2em]
		\item If $\gamma = \frac{1}{d}$, which implies $\gamma=\frac{1}{d}\leq \left(\frac{r_0}{c_1T}\right)^\frac{1}{2}$ and $\gamma=\frac{1}{d}\leq \left(\frac{r_0}{c_2T}\right)^\frac{1}{3}$, then
		\begin{align*}
			\Psi_T \leq \frac{dr_0}{bT} + \frac{c_1}{bd} + \frac{c_2}{bd^2} \leq \frac{dr_0}{bT} +\frac{c_1^\frac{1}{2}r_0^\frac{1}{2}}{bT^\frac{1}{2}} + \frac{c_2^\frac{1}{3}r_0^\frac{2}{3}}{bT^\frac{2}{3}} = \gO\left(\frac{dr_0}{T} +\frac{c_1^\frac{1}{2}r_0^\frac{1}{2}}{T^\frac{1}{2}} + \frac{c_2^\frac{1}{3}r_0^\frac{2}{3}}{T^\frac{2}{3}} \right)
		\end{align*}
		\item If $\gamma = \left(\frac{r_0}{c_1T}\right)^\frac{1}{2}$, which implies $\gamma = \left(\frac{r_0}{c_1T}\right)^\frac{1}{2} \leq \left(\frac{r_0}{c_2T}\right)^\frac{1}{3}$ and $\gamma = \left(\frac{r_0}{c_1T}\right)^\frac{1}{2} \leq \frac{1}{d}$, then
		\begin{align*}
			\Psi_T \leq \frac{2c_1^\frac{1}{2}r_0^\frac{1}{2}}{bT^\frac{1}{2}} + \frac{c_2}{b} \cdot\frac{r_0}{c_1T} \leq \frac{2c_1^\frac{1}{2}r_0^\frac{1}{2}}{bT^\frac{1}{2}} + \frac{c_2^\frac{1}{3}r_0^\frac{2}{3}}{bT^\frac{2}{3}} = \gO\left(\frac{c_1^\frac{1}{2}r_0^\frac{1}{2}}{T^\frac{1}{2}} + \frac{c_2^\frac{1}{3}r_0^\frac{2}{3}}{T^\frac{2}{3}} \right)
		\end{align*}
		\item If $\gamma = \left(\frac{r_0}{c_2T}\right)^\frac{1}{3}$, which implies that $\gamma = \left(\frac{r_0}{c_2T}\right)^\frac{1}{3} \leq \left(\frac{r_0}{c_1T}\right)^\frac{1}{2}$ and $\gamma = \left(\frac{r_0}{c_2T}\right)^\frac{1}{3} \leq \frac{1}{d}$, then
		\begin{align*}
			\Psi_T \leq \frac{2c_2^\frac{1}{3}r_0^\frac{2}{3}}{bT^\frac{2}{3}} + \frac{c_1}{b}\left(\frac{r_0}{c_2T}\right)^\frac{1}{3} \leq \frac{2c_2^\frac{1}{3}r_0^\frac{2}{3}}{bT^\frac{2}{3}} + \frac{c_1^\frac{1}{2}r_0^\frac{1}{2}}{bT^\frac{1}{2}} = \gO\left(\frac{c_1^\frac{1}{2}r_0^\frac{1}{2}}{T^\frac{1}{2}} + \frac{c_2^\frac{1}{3}r_0^\frac{2}{3}}{T^\frac{2}{3}} \right)
		\end{align*}
	\end{itemize}
	Combining these three cases, we get the second result of this lemma. Notably, only when choosing $\gamma = \frac{1}{d}$, the optimization term $\gO\left(\frac{dr_0}{T}\right)$ exists.
\end{proof}
\subsubsection{Proof of general convex case of Theorem~\ref{thm:SFL} and Corollary~\ref{cor:SFL}}
\begin{proof}[Proof of the general convex case of Theorem~\ref{thm:SFL}]
	Letting $\mu=0$ in Ineq.~\eqref{eq:thm:strongly convex:simplified recursion}, we get the recursion of the general convex case,
	\begin{align*}
		\E\left[\Norm{\rvx^{(r+1)}-\rvx^*}^2\right] &\leq \E\left[\Norm{\rvx^{(r)}- \rvx^\ast}^2\right] - \frac{\tilde{\eta}}{3}\E\left[D_F(\rvx^{(r)}, \rvx^\ast)\right]\nonumber\\
		&\quad+\frac{4\tilde{\eta}^2\sigma^2}{SK}+\frac{4\tilde{\eta}^2(M-S)\zeta_\ast^2}{S(M-1)}+\frac{6L\tilde{\eta}^3\sigma^2}{SK}+\frac{6L\tilde{\eta}^3\zeta_\ast^2}{S}\,.
	\end{align*}
	Applying Lemma~\ref{lem:general convex:tuning learning rate} with $t=r$ ($T=R$), $\gamma=\tilde\eta$, $r_{t} = \E\left[\Norm{\rvx^{(r)}- \rvx^*}^2\right]$, $b=\frac{1}{3}$, $s_t = \E\left[D_F(\rvx^{(r)}, \rvx^\ast)\right]$, $w_t=1$, $c_1 = \frac{4\sigma^2}{SK}+\frac{4(M-S)\zeta_\ast^2}{S(M-1)}$, $c_2 = \frac{6L\sigma^2}{SK} + \frac{6L\zeta_\ast^2}{S}$ and $\frac{1}{d}=\frac{1}{6L}$, we obtain
	\begin{align}
		&\E\left[F(\bar\rvx^{(R)})-F(\rvx^\ast)\right]\nonumber\\
		&\leq \frac{3\Norm{\rvx^{(0)}- \rvx^*}^2}{\tilde\eta R}+\frac{12\tilde{\eta}\sigma^2}{SK}+\frac{12\tilde{\eta}(M-S)\zeta_\ast^2}{S(M-1)}+\frac{18L\tilde{\eta}^2\sigma^2}{SK}+\frac{18L\tilde{\eta}^2\zeta_\ast^2}{S}\,,\label{eq:thm:convex:bound}
	\end{align}
	where $\bar\rvx^{(R)} = \frac{1}{\sum_{r=0}^Rw_r}\sum_{r=0}^Rw_r\rvx^{(r)}$. When $\tilde\eta \asymp\min \{\frac{1}{L}, \frac{D}{c_1^{1/2} R^{1/2}}, \frac{D^{2/3}}{c_2^{1/3}R^{1/3}}\}$, $\E\left[F(\bar\rvx^{(R)})-F(\rvx^\ast)\right]$ is bounded by
	\begin{align}
		\gO\left(\frac{\sigma D}{\sqrt{SKR}}+\sqrt{1-\frac{S}{M}}\cdot \frac{\zeta_\ast D}{\sqrt{SR}} + \frac{\left(L\sigma^2D^4\right)^{1/3}}{(SK)^{1/3}R^{2/3}} + \frac{\left(L\zeta_\ast^2D^4\right)^{1/3}}{S^{1/3}R^{2/3}} + \frac{LD^2}{R}\right)\,,\label{eq:thm:convex:bound2}
	\end{align}
	where $D = \norm{\rvx^{(0)}- \rvx^\ast}$. Ineqs.~\eqref{eq:thm:convex:bound},~\eqref{eq:thm:convex:bound2} are the bounds for partial participation. When $S=M$, we get the claim of the general convex case of Theorem~\ref{thm:SFL} and Corollary~\ref{cor:SFL}.
\end{proof}
\subsection{Non-convex case}\label{subsec:SFL non-convex}
This section provides the proof of Theorem~\ref{thm:SFL} for the non-convex case.
\subsubsection{Finding the recursion}
\begin{lemma}\label{lem:non-convex:recursion}
	Let Assumptions~\ref{asm:smoothness}, \ref{asm:stochasticity}, \ref{asm:heterogeneity1} hold. If $\eta \leq \frac{1}{6LSK (1+\beta^2/S)}$, then
	\begin{align}
		\E_r\left[F(\rvx^{(r+1)}) - F(\rvx^{(r)})\right]&\leq-\frac{1}{6}SK\eta\Norm{\nabla F(\rvx^{(r)})}^2 +  2L\eta^2SK\sigma^2 + 2L\eta^2S^2K^2\frac{M-S}{S(M-1)}\zeta^2 \nonumber\\
		&\quad+\frac{5}{6}\eta L^2\sum_{m=1}^S\sum_{k=0}^{K-1}\E_r\left[\Norm{\rvx_{m,k}^{(r)}-\rvx^{(r)}}^2\right].\label{eq:SFL:non-convex:recursion}
	\end{align}
\end{lemma}
\begin{proof}
	In the following, we focus on a single training round, and hence we drop the superscripts $r$ for a while, e.g., replacing $\rvx_{m,k}^{(r)}$ with $\rvx_{m,k}$ and replacing $\rvx^{(r)}$ with $\rvx$.
	
	Since $F$ is $L$-smooth, we start from the following equation:
	\begin{align}
		\E_r\left[F(\rvx+\Delta) - F(\rvx)\right]\leq \E_r\left[\inp{\nabla F(\rvx)}{\Delta}\right] + \frac{L}{2}\E_r\norm{\Delta}^2.\label{eq:SFL:non-convex:decomposition}
	\end{align}
	For $\E_r\left[\inp{\nabla F(\rvx)}{\Delta}\right] $, substituting the global update $\Delta$ with \eqref{eq:SFL:complete update}, we obtain
	\begin{align*}
		&\E_r\left[\inp{\nabla F(\rvx)}{\Delta}\right]\\
		&=-\eta\sum_{m=1}^S\sum_{k=0}^{K-1}\E_r\left[\inp{\nabla F(\rvx)}{\rvg_{\pi_m}(\rvx_{m,k})}\right]\\
		&=-\eta\sum_{m=1}^S\sum_{k=0}^{K-1}\E_r\left[\E_r\left[\inp{\nabla F(\rvx)}{\rvg_{\pi_m}(\rvx_{m,k})}\mid \rvx_{m,k}\right]\right]\\
		&=-\eta\sum_{m=1}^S\sum_{k=0}^{K-1}\E_r\left[\inp{\nabla F(\rvx)}{\nabla F_{\pi_m}(\rvx_{m,k})}\right]\\
		&=-\eta SK\E_r\left[\inp{\nabla F(\rvx)}{\frac{1}{S}\sum_{m=1}^S\frac{1}{K}\sum_{k=0}^{K-1}\nabla F_{\pi_m}(\rvx_{m,k})}\right]\\
		&=-\eta SK\E_r\left[\inp{\nabla F(\rvx)}{\frac{1}{S}\sum_{m=1}^S\frac{1}{K}\sum_{k=0}^{K-1}\left( \nabla F_{\pi_m}(\rvx_{m,k}) - \nabla F_{\pi_m}(\rvx) + \nabla F_{\pi_m}(\rvx) \right)}\right]\\
		&= -\eta SK \norm{\nabla F(\rvx)}^2 -\eta SK\E_r\left[\inp{\nabla F(\rvx)}{\frac{1}{S}\sum_{m=1}^S\frac{1}{K}\sum_{k=0}^{K-1}\left( \nabla F_{\pi_m}(\rvx_{m,k}) - \nabla F_{\pi_m}(\rvx) \right)}\right]\\
		&\leq -\eta SK \norm{\nabla F(\rvx)}^2 + \frac{1}{2}\eta SK \norm{\nabla F(\rvx)}^2 + \frac{1}{2}\eta \sum_{m=1}^S\sum_{k=0}^{K-1}\E_r\norm{\nabla F_{\pi_m}(\rvx_{m,k}) - \nabla F_{\pi_m}(\rvx)}^2\\
		&\leq-\frac{1}{2}\eta SK \norm{\nabla F(\rvx)}^2 + \frac{1}{2}L^2\eta \sum_{m=1}^S\sum_{k=0}^{K-1}\E_r\norm{\rvx_{m,k} - \rvx}^2 ,
	\end{align*}
	where the second equality uses the law of total expectation, the third equality uses \cref{asm:stochasticity}, the sixth equality holds because the client sampling is unbiased, the first inequality uses $\abs{\inp{a}{b}} \leq \frac{1}{2}a^2+\frac{1}{2}b^2$ for any $a,b\in \R$, and the last inequality uses \cref{asm:smoothness}.
	
	For $\frac{1}{2}L\E_r\norm{\Delta}^2 $, substituting the global update $\Delta$ with \eqref{eq:SFL:complete update}, we obtain
	\begin{align}
		\frac{1}{2}L\E_r\norm{\Delta}^2
		&\leq 2L\eta^2\E_r\norm{\sum_{m=1}^S\sum_{k=0}^{K-1} \left(\rvg_{\pi_m}(\rvx_{m,k}) - \nabla F_{\pi_m}(\rvx_{m,k})\right)}^2 \nonumber\\
		&\quad+ 2L\eta^2\E_r\norm{\sum_{m=1}^S\sum_{k=0}^{K-1} \left(\nabla F_{\pi_m}(\rvx_{m,k}) - \nabla F_{\pi_m}(\rvx)\right)}^2 \nonumber\\
		&\quad+ 2L\eta^2\E_r\norm{\sum_{m=1}^S\sum_{k=0}^{K-1} \left(\nabla F_{\pi_m}(\rvx) - \nabla F(\rvx)\right)}^2 \nonumber\\
		&\quad+ 2L\eta^2\E_r\norm{\sum_{m=1}^S\sum_{k=0}^{K-1}\nabla F(\rvx)}^2.\label{eq:recursion-non-convex-1}
	\end{align}
	
	For the first term on the right hand side in \eqref{eq:recursion-non-convex-1},
	\begin{align*}
		\Term{1}{\eqref{eq:recursion-non-convex-1}}
		&= 2L\eta^2 \sum_{m=1}^S\sum_{k=0}^{K-1} \E_r\norm{\rvg_{\pi_m}(\rvx_{m,k}) - \nabla F_{\pi_m}(\rvx_{m,k})}^2\\
		&\leq 2L\eta^2SK\sigma^2,
	\end{align*}
	where we use \cref{lem:martingale difference property} and \cref{asm:stochasticity}.
	
	For the second term on the right hand side in \eqref{eq:recursion-non-convex-1},
	\begin{align*}
		\Term{2}{\eqref{eq:recursion-non-convex-1}}
		&\leq 2L\eta^2SK\sum_{m=1}^S\sum_{k=0}^{K-1}\E_r\norm{\nabla F_{\pi_m}(\rvx_{m,k}) - \nabla F_{\pi_m}(\rvx)}^2\\
		&\leq 2L^3\eta^2SK\sum_{m=1}^S\sum_{k=0}^{K-1}\E_r\norm{\rvx_{m,k} - \rvx}^2,
	\end{align*}
	where the first inequality uses Jensen's inequality and the second inequality uses \cref{asm:smoothness}.
	
	For the third and fourth terms on the right hand side in \eqref{eq:recursion-non-convex-1},
	\begin{align*}
		\Term{3+4}{\eqref{eq:recursion-non-convex-1}}
		&\leq 2L\eta^2S^2K^2\frac{M-S}{S(M-1)}\zeta^2 + 2L\eta^2S^2K^2\left(1+ \frac{M-S}{S(M-1)}\beta^2\right)\norm{\nabla F(\rvx)}^2,
	\end{align*}
	where we use Lemma~\ref{lem:simple random sampling} and Assumption~\ref{asm:heterogeneity1} for the last inequality. 
	
	Plugging the preceding bounds into \eqref{eq:recursion-non-convex-1}, we get the bound for $\frac{1}{2}L\E_r\norm{\Delta}^2$. Then, plugging the bounds for $\E_r\left[\inp{\nabla F(\rvx)}{\Delta}\right]$ and $\frac{1}{2}L\E_r\norm{\Delta}^2$ into \eqref{eq:SFL:non-convex:decomposition}, and using $\eta \leq \frac{1}{6LSK (1+\beta^2/S)}$, we obtain
	\begin{align*}
		&\E_r\left[F(\rvx+\Delta) - F(\rvx)\right] \\
		&\leq-\frac{1}{6}SK\eta\norm{\nabla F(\rvx)}^2 +  2L\eta^2SK\sigma^2 + 2L\eta^2S^2K^2\frac{M-S}{S(M-1)}\zeta^2
		+ \frac{5}{6}L^2\eta\sum_{m=1}^S\sum_{k=0}^{K-1}\E_r\norm{\rvx_{m,k}-\rvx}^2.
	\end{align*}
	The claim follows after recovering the superscripts.
\end{proof}
\subsubsection{Bounding the client drift with Assumption~\ref{asm:heterogeneity1}}
\begin{lemma}\label{lem:non-convex:drift}
	Let Assumptions~\ref{asm:smoothness}, \ref{asm:stochasticity}, \ref{asm:heterogeneity1} hold. If $\eta \leq \frac{1}{6LSK(1+\beta^2/S)}$, then
	\begin{align}
		E_r\leq \frac{9}{4}S^2K^2\eta^2\sigma^2+\frac{9}{4}S^2K^3\eta^2\zeta^2+\frac{9}{4}(\beta^2/S+1)S^3K^3\eta^2\Norm{\nabla F(\rvx^{(r)})}^2.\label{eq:SFL:non-convex:drift bound}
	\end{align}
\end{lemma}
\begin{proof}
 	In the following, we focus on a single training round, and hence we drop the superscripts $r$ for a while, e.g., replacing $\rvx_{m,k}^{(r)}$ with $\rvx_{m,k}$ and replacing $\rvx^{(r)}$ with $\rvx$.
	
	Using \cref{eq:SFL:partial update}, we obtain
	\begin{align}
		\E_r\norm{\rvx_{m,k} - \rvx}^2
		&\leq 4\eta^2\E_r\norm{\sum_{i=1}^{m}\sum_{j=0}^{b_{m,k}(i)} \left(\rvg_{\pi_i}(\rvx_{i,j}) - \nabla F_{\pi_i} (\rvx_{i,j})\right)}^2 \nonumber\\
		&\quad+ 4\eta^2\E_r\norm{\sum_{i=1}^{m}\sum_{j=0}^{b_{m,k}(i)}\left(\nabla F_{\pi_i} (\rvx_{i,j}) - \nabla F_{\pi_i} (\rvx)\right)}^2 \nonumber\\
		&\quad+4\eta^2\underbrace{\E_r\norm{\sum_{i=1}^{m}\sum_{j=0}^{b_{m,k}(i)}\left(\nabla F_{\pi_i} (\rvx) - \nabla F(\rvx)\right)}^2}_{T_{m,k}}\nonumber\\
		&\quad + 4\eta^2\E_r\norm{\sum_{i=1}^{m}\sum_{j=0}^{b_{m,k}(i)}\nabla F (\rvx)}^2.\label{eq:drift-non-convex-1}
	\end{align}
	For the first term on the right hand side in \eqref{eq:drift-non-convex-1},
	\begin{align*}
		\Term{1}{\eqref{eq:drift-non-convex-1}}&\leq 4\eta^2 \sum_{i=1}^m \sum_{j=0}^{b_{m,k}(i)}\E_r\norm{\rvg_{\pi_i}(\rvx_{i,j})-\nabla F_{\pi_i} (\rvx_{i,j})}^2\\ &\leq 4\eta^2\gB_{m,k}\sigma^2,
	\end{align*}
	where we use \cref{lem:martingale difference property} and \cref{asm:stochasticity}. 
	
	For the second term on the right hand side in \eqref{eq:drift-non-convex-1},
	\begin{align*}
		\Term{2}{\eqref{eq:drift-non-convex-1}} &\leq 4\eta^2\gB_{m,k}\sum_{i=1}^{m}\sum_{j=0}^{b_{m,k}(i)}\E_r\norm{\nabla F_{\pi_i} (\rvx_{i,j}) - \nabla F_{\pi_i} (\rvx)}^2\\
		&\leq4L^2\eta^2\gB_{m,k}\sum_{i=1}^{m}\sum_{j=0}^{b_{m,k}(i)}\E_r\norm{\rvx_{i,j} - \rvx}^2,
	\end{align*}
	where the first inequality uses Jensen's inequality and the second inequality uses \cref{asm:smoothness}.
	
	Then, plugging the bounds of $\E_r\norm{\rvx_{m,k} - \rvx}^2$ into $E_r$, we can get
	\begin{align*}
		E_r &\leq4\eta^2\sigma^2\sum_{m=1}^S\sum_{k=0}^{K-1}\gB_{m,k}+ 4L^2\eta^2\sum_{m=1}^S\sum_{k=0}^{K-1}\gB_{m,k}\sum_{i=1}^{m}\sum_{j=0}^{b_{m,k}(i)}\E_r\norm{\rvx_{i,j} - \rvx}^2\\ 
		&\quad +4\eta^2\sum_{m=1}^S\sum_{k=0}^{K-1}T_{m,k} + 4\eta^2\sum_{m=1}^S\sum_{k=0}^{K-1}\gB_{m,k}^2\norm{\nabla F(\rvx)}^2\,.
	\end{align*}	
	Noting that the third term can be bounded by Lemma~\ref{lem:sequential partial participation} with $\rvx_{\pi_i}=\nabla F_{\pi_i}(\rvx)$ and $\overline\vx=\nabla F(\rvx)$ and using $\sum_{m=1}^S\sum_{k=0}^{K-1}\gB_{m,k} \leq \frac{1}{2}S^2K^2$ and $\sum_{m=1}^S\sum_{k=0}^{K-1}\gB_{m,k}^2 \leq \frac{1}{3}S^3K^3$, we can simplify the preceding inequality as
	\begin{align*}
		E_r\leq2S^2K^2\eta^2\sigma^2+2L^2S^2K^2\eta^2E_r+2S^2K^3\eta^2\zeta^2 + 2(\beta^2/S+1)S^3K^3\eta^2\norm{\nabla F(\rvx)}^2
	\end{align*}
	After rearranging the preceding inequality and using $\eta \leq \frac{1}{6LSK(1+\beta^2/S)}$, we get
	\begin{align*}
		E_r\leq \frac{9}{4}S^2K^2\eta^2\sigma^2 + \frac{9}{4}S^2K^3\eta^2\zeta^2+\frac{9}{4}(\beta^2/S+1)S^3K^3\eta^2\norm{\nabla F(\rvx)}^2\,.
	\end{align*}
	The claim follows after recovering the superscripts.
\end{proof}
\subsubsection{Proof of non-convex case of Theorem~\ref{thm:SFL} and Corollary~\ref{cor:SFL}}
\begin{proof}
	Plugging \eqref{eq:SFL:non-convex:drift bound} into \eqref{eq:SFL:non-convex:recursion}, taking unconditional expectation and using $\eta \leq \frac{1}{6LSK(1+\beta^2/S)}$, we obtain
	\begin{align*}
		\E \left[F(\rvx^{(r+1)})-F(\rvx^{(r)})\right]
		&\leq -\frac{1}{10}SK\eta\E\left[\Norm{\nabla F(\rvx^{(r)})}^2\right]+2L\eta^2SK\sigma^2+ 2L\eta^2S^2K^2\tfrac{M-S}{S(M-1)}\zeta^2\\
		&\quad+ \frac{15}{8}L^2\eta^3S^2K^2\sigma^2 + \frac{15}{8}L^2\eta^3S^2K^3\zeta^2\,.
	\end{align*}
	Using $\tilde \eta = \eta SK$, subtracting $F^\ast$ from both sides and then rearranging the terms, we obtain
	\begin{align*}
		\E \left[F(\rvx^{(r+1)})- F^\ast\right] &\leq \E\left[F(\rvx^{(r)}) - F^\ast\right] - \frac{\tilde\eta}{10}\E\left[\Norm{\nabla F(\rvx^{(r)})}^2\right]\\
		&\quad+ \frac{2L\tilde\eta^2\sigma^2}{SK}+2L\tilde\eta^2\frac{M-S}{S(M-1)}\zeta^2+ \frac{15}{8}\frac{L^2\tilde\eta^3\sigma^2}{SK} + \frac{15}{8}\frac{L^2\tilde\eta^3\zeta^2}{S}\,.
	\end{align*}
	Then applying Lemma~\ref{lem:general convex:tuning learning rate} with $t=r$ ($T=R$), $\gamma=\tilde\eta$, $r_{t} = \E\left[F(\rvx^{(r)})-F^\ast\right]$, $b=\frac{1}{10}$, $s_t = \E\left[\Norm{\nabla F(\rvx^{(r)})}^2\right]$, $w_t=1$, $c_1 = \frac{2L\sigma^2}{SK}+2L\frac{M-S}{S(M-1)}\zeta^2$, $c_2 = \frac{15}{8}\frac{L^2\sigma^2}{SK} + \frac{15}{8}\frac{L^2\zeta^2}{S}$ and $\frac{1}{d}=\frac{1}{6L(1+\beta^2/S)}$ ($\tilde\eta=SK\eta\leq\frac{1}{6L(1+\beta^2/S)}$), we have
	\begin{align}
		&\frac{1}{R+1}\sum_{r=0}^{R} \E\left[\Norm{\nabla F(\rvx^{(r)})}^2\right] \nonumber\\
		&\leq \frac{10\left(F(\rvx^{0}) - F^\ast\right)}{\tilde\eta R} + \frac{20L\tilde\eta\sigma^2}{SK} +\frac{20L\tilde\eta\zeta^2(M-S)}{S(M-1)}+ \frac{75L^2\tilde\eta^2\sigma^2}{4SK} + \frac{75L^2\tilde\eta^2\zeta^2}{4S}\,.\label{eq:thm:non-convex:bound}
	\end{align}
	When $\tilde\eta \asymp\min \{\frac{1}{L(1+\beta^2/S)}, \frac{A^{1/2}}{c_1^{1/2} R^{1/2}}, \frac{A^{1/3}}{c_2^{1/3}R^{1/3}}\}$, $\frac{1}{R+1}\sum_{r=0}^{R} \E\left[\Norm{\nabla F(\rvx^{(r)})}^2\right]$ is bounded by
	\begin{align}
		\gO\left(\frac{ \left(L\sigma^2A\right)^{1/2}}{\sqrt{SKR}}+\sqrt{1-\frac{S}{M}}\cdot\frac{ \left(L\zeta^2A\right)^{1/2}}{\sqrt{SR}} + \frac{\left(L^2\sigma^2A^2\right)^{1/3}}{(SK)^{1/3}R^{2/3}} + \frac{\left(L^2\zeta^2A^2\right)^{1/3}}{S^{1/3}R^{2/3}} + \frac{LA(1+\beta^2/S)}{R}\right)\label{eq:thm:non-convex:bound2}
	\end{align}
	where $A = F(\rvx^{0})- F^\ast$. Ineqs.~\eqref{eq:thm:non-convex:bound},~\eqref{eq:thm:non-convex:bound2} are the bounds for the partial participation. When $S=M$, we get the bounds for the full participation. Recall that $\bar\rvx^{(R)}$ is sampled uniformly at random from $\{\rvx^{(r)}\}_{r=0}^{R}$, so we have $\E\left[\Norm{\nabla F(\bar \rvx^{(R)})}^2\right] = \frac{1}{R+1}\sum_{r=0}^{R} \E\left[\Norm{\nabla F(\rvx^{(r)})}^2\right]$.
\end{proof}

\newpage
\section{Proofs of Theorem~\ref{thm:PFL}}\label{sec:proof PFL}
Here we slightly improve the convergence guarantee for the strongly convex case by combining the works of \citet{karimireddy2020scaffold} and \citet{koloskova2020unified}. Moreover, we reproduce the guarantees for the general convex and non-convex cases for completeness. The results are given in Theorem~\ref{thm:PFL}. We provide the proof of Theorem~\ref{thm:PFL} for the strongly convex, general convex and non-convex cases in Sections~\ref{subsec:PFL strongly convex}, \ref{subsec:PFL general convex} and \ref{subsec:PFL non-convex}, respectively.
\begin{theorem}\label{thm:PFL}
	Define $\bar\rvx^{(R)}$ as an iterate constructed from the iterates
	$\{\rvx^{(r)}\}_{r=0}^{R}$ generated by \cref{algorithm2}, with its
	specific construction given for each setting below. Define the effective learning rate $\tilde\eta \coloneqq \eta K$. Define $D\coloneqq\Norm{\rvx^{(0)}-\rvx^\ast}$ for the convex cases and $A \coloneqq F(\rvx^{(0)}) - F^\ast$ for the non-convex case. Then, we have the following upper bounds:
	\setlist[itemize]{label=}
	\begin{itemize}[leftmargin=0.5em]
		\item \textbf{Strongly convex}: Under \cref{asm:smoothness,asm:stochasticity,asm:heterogeneity2}, there exists $\tilde\eta \leq \frac{1}{6L}$, such that for $M\geq 2$, $K\geq 1$ and $R\geq 6\kappa$ ($\kappa \coloneqq L/\mu$),
		\begin{flalign*}
			\E\left[F(\bar\rvx^{(R)})-F(\rvx^\ast)\right] \leq \frac{9}{2}\mu D^2 \exp\left(-\frac{\mu\tilde\eta R}{2} \right)+\frac{12\tilde{\eta}\sigma^2}{MK}+\frac{18L\tilde{\eta}^2\sigma^2}{K}+12L\tilde{\eta}^2\zeta_\ast^2, &&
		\end{flalign*}
		where $\bar\rvx^{(R)} = \frac{1}{\sum_{r=0}^Rw_r} \sum_{r=0}^{R}w_r\rvx^{(r)}$ with $w_r=(1-\frac{\mu\tilde\eta}{2})^{-(r+1)}$.
		\item \textbf{General convex}: Under \cref{asm:smoothness,asm:stochasticity,asm:heterogeneity2}, there exists $\tilde\eta \leq \frac{1}{6L}$, such that for $M\geq 2$, $K\geq 1$ and $R\geq 1$,
		\begin{flalign*}
			\E\left[F(\bar\rvx^{(R)})-F(\rvx^\ast)\right] \leq \frac{3D^2}{\tilde\eta R}+\frac{12\tilde{\eta}\sigma^2}{MK}+\frac{18L\tilde{\eta}^2\sigma^2}{K}+12L\tilde{\eta}^2\zeta_\ast^2, &&
		\end{flalign*}
		where $\bar\rvx^{(R)} = \frac{1}{\sum_{r=0}^Rw_r} \sum_{r=0}^{R}w_r\rvx^{(r)}$ with $w_r=1$.
		\item \textbf{Non-convex}: Under Assumptions~\ref{asm:smoothness},~\ref{asm:stochasticity} and~\ref{asm:heterogeneity1}, there exists $\tilde\eta\leq \frac{1}{6L(1+\beta^2)}$, such that for $M\geq 2$, $K\geq 1$ and $R\geq 1$,
		\begin{flalign*}
			\E\left[\Norm{\nabla F(\bar \rvx^{(R)})}^2\right] \leq \frac{8A}{\tilde\eta R} + \frac{16L\tilde\eta\sigma^2}{MK}+ \frac{15L^2\tilde\eta^2\sigma^2}{K} + 10L^2\tilde\eta^2\zeta^2, &&
		\end{flalign*}
		where $\bar\rvx^{(R)}$ is sampled uniformly at random from $\{\rvx^{(r)}\}_{r=0}^{R}$, and the expectation is taken over both the algorithmic randomness and the additional randomness introduced by this sampling procedure.
	\end{itemize}
\end{theorem}
\begin{corollary}\label{cor:PFL}
	By choosing an appropriate learning rate (see the proofs of Theorem~\ref{thm:PFL} in Appendices~\ref{subsec:PFL strongly convex},~\ref{subsec:PFL general convex},~\ref{subsec:PFL non-convex}) for the results of Theorem~\ref{thm:PFL}, we can obtain the upper bounds of PFL:
	\setlist[itemize]{label=}
	\begin{itemize}[leftmargin=0.5em]
		\item \textbf{Strongly convex}: When choosing $\tilde\eta=\eta K \asymp  \min \{\frac{1}{L},\frac{1}{\mu R} \}$ for Theorem~\ref{thm:PFL}, then for $R\gtrsim \kappa$
		\begin{flalign*}
			\E\left[F(\bar\rvx^{(R)})-F(\rvx^\ast)\right] = \tilde\gO\left(\frac{\sigma^2}{\mu MKR}+ \frac{L\sigma^2}{\mu^2KR^2} + \frac{L\zeta_\ast^2}{\mu^2R^2}+\mu D^2 \exp\left(\frac{-\mu R}{L}\right)\right)\,.&&
		\end{flalign*}
		\item \textbf{General convex}: When choosing $\tilde\eta=\eta K \asymp  \min \{\frac{1}{L},\frac{D}{c_1^{1/2}R^{1/2}},\frac{D^{2/3}}{c_2^{1/3}R^{1/3}} \}$ with $c_1 \asymp \frac{\sigma^2}{MK}$ and $c_2 \asymp \frac{L\sigma^2}{K} + L\zeta_\ast^2$ for Theorem~\ref{thm:PFL}, then
		\begin{flalign*}
			\E\left[F(\bar\rvx^{(R)})-F(\rvx^\ast)\right] = \gO\left(\frac{\sigma D}{\sqrt{MKR}}+\frac{\left(L\sigma^2D^4\right)^{1/3}}{K^{1/3}R^{2/3}} + \frac{\left(L\zeta_\ast^2D^4\right)^{1/3}}{R^{2/3}} + \frac{LD^2}{R}\right). &&
		\end{flalign*}
		\item \textbf{Non-convex}: When choosing $\tilde\eta=\eta K \asymp  \min \{\frac{1}{L(1+\beta^2)},\frac{A^{1/2}}{c_1^{1/2}R^{1/2}},\frac{A^{1/3}}{c_2^{1/3}R^{1/3}} \}$ with $c_1 \asymp \frac{L\sigma^2}{MK}$ and $c_2 \asymp \frac{L^2\sigma^2}{K} + L^2\zeta^2$ for Theorem~\ref{thm:PFL}, then
		\begin{flalign*}
			\E\left[\Norm{\nabla F(\bar \rvx^{(R)})}^2\right] = \gO\left(\frac{ \left(L\sigma^2A\right)^{1/2}}{\sqrt{MKR}} + \frac{\left(L^2\sigma^2A^2\right)^{1/3}}{K^{1/3}R^{2/3}} + \frac{\left(L^2\zeta^2A^2\right)^{1/3}}{R^{2/3}} + \frac{L(\beta^2+1) A}{R}\right). &&
		\end{flalign*}
	\end{itemize}
\end{corollary}
In the following proofs, we consider the partial client participation setting. Specifically, in each training round $r$, given a complete permutation of the clients, $\pi^{(r)}=(\pi_1^{(r)},\pi_2^{(r)},\ldots,\pi_M^{(r)})$, only the first $S$ clients $\pi_1^{(r)},\pi_2^{(r)},\ldots,\pi_S^{(r)}$ participate in training. The permutation $\pi^{(r)}$ is sampled uniformly at random and independently across training rounds.

According to Algorithm~\ref{algorithm2}, the global updates of PFL after one complete training round (with $S$ clients selected for training) are
\begin{align}
	\Delta^{(r)} \coloneqq \rvx^{(r+1)}-\rvx^{(r)} =-\frac{\eta}{S}\sum_{m=1}^S\sum_{k=0}^{K-1}\rvg_{\pi_m^{(r)}}(\rvx_{\pi_m,k}^{(r)}).\label{eq:PFL:complete update}
\end{align}
The local update of PFL from $\rvx^{(r)}$ to $\rvx_{m,k}^{(r)}$ is
\begin{align}
	\rvx_{m,k}^{(r)}-\rvx^{(r)}=-\eta \sum_{j=0}^{k-1} \rvg_m(\rvx_{m,j}^{(r)}).\label{eq:PFL:partial update}
\end{align}
Define $\E_r[\cdot]\coloneqq \E[\cdot \mid \rvx^{(r)}]$. The ``client drift'' \citep{karimireddy2020scaffold} in PFL is defined as follows:
\begin{align}
	\gE_r \coloneqq \frac{1}{M}\sum_{m=1}^M\sum_{k=0}^{K-1} \E_r\left[\Norm{\rvx_{m,k}^{(r)} - \rvx^{(r)}}^2\right]\label{eq:PFL client drift}
\end{align}
For clarity, we will use ``Term$_n$'' to denote the $n$-th term on the right hand side in some equation in the following proofs.
\subsection{Strongly convex case}\label{subsec:PFL strongly convex}
\subsubsection{Find the per-round recursion}
\begin{lemma}\label{lem:PFL strongly convex:per-round recursion}
	Let Assumptions~\ref{asm:smoothness}, \ref{asm:stochasticity}, \ref{asm:heterogeneity2} hold and assume that all the local objectives are $\mu$-strongly convex. If the learning rate satisfies $\eta \leq \frac{1}{6LK}$, then it holds that
	\begin{align}
		\E_r\left[\Norm{\rvx^{(r+1)}-\rvx^\ast}^2\right] &\leq \left(1-\frac{1}{2}\mu K\eta\right)\Norm{\rvx^{(r)}-\rvx^\ast}^2+\frac{4K\eta^2\sigma^2}{S}+4K^2\eta^2\frac{M-S}{S(M-1)}\zeta_\ast^2 \nonumber\\
		&\quad-\frac{2}{3}K\eta D_F(\rvx^{(r)},\rvx^\ast)+\frac{8}{3}L\eta\frac{1}{M}\sum_{m=1}^M\sum_{k=0}^{K-1}\E_r\left[\Norm{\rvx_{m,k}^{(r)} - \rvx^{(r)}}^2\right]\label{eq:lem:PFL strongly convex:per-round recursion}
	\end{align}
\end{lemma}
\begin{proof}
	In the following, we focus on a single training round, and hence we drop the superscripts $r$ for a while, e.g., replacing $\rvx_{m,k}^{(r)}$ with $\rvx_{m,k}$ and replacing $\rvx^{(r)}$ with $\rvx$.
	
	We start from the following equation:
	\begin{align}
		\E_r\big[\Vert\rvx + \Delta-\rvx^\ast\Vert^2\big] =\E_r\big[\Norm{\rvx-\rvx^*}^2\big] + 2\E_r\big[\langle\rvx-\rvx^*,\Delta\rangle\big]+\E_r\big[\Norm{\Delta}^2\big].\label{eq:PFL:single round decomposition}
	\end{align}
	
	For $2\E_r\big[\langle\rvx^{(r)}-\rvx^*,\Delta\rangle\big]$, substituting the overall updates $\Delta$, we have
	\begin{align*}
		&\hspace{-1em}2\E_r\big[\langle\rvx^{(r)}-\rvx^*,\Delta\rangle\big] \\
		&=  -\frac{2\eta}{S}\sum_{m=1}^S\sum_{k=0}^{K-1}\E_r\big[\langle\rvx-\rvx^*, \rvg_{\pi_m}(\rvx_{\pi_m,k}) \rangle\big]  \\
		&=  -\frac{2\eta}{M}\sum_{m=1}^M\sum_{k=0}^{K-1}\E_r\big[\langle\rvx-\rvx^*, \rvg_{m}(\rvx_{m,k}) \rangle\big]  \\
		&=  -\frac{2\eta}{M}\sum_{m=1}^M\sum_{k=0}^{K-1}\E_r\Big[\E_r\big[\langle\rvx-\rvx^*, \rvg_{m}(\rvx_{m,k}) \rangle \mid \rvx_{m,k} \big] \Big] \\
		&=  -\frac{2\eta}{M}\sum_{m=1}^M\sum_{k=0}^{K-1}\E_r\big[\langle\rvx-\rvx^*, \nabla F_{m}(\rvx_{m,k}) \rangle \big]  \\
		&\leq -\frac{2\eta}{M}\sum_{m=1}^M\sum_{k=0}^{K-1} \E_r\left[F_{m}(\rvx)-F_{m}(\rvx^\ast) + \frac{\mu}{4}\Norm{\rvx-\rvx^\ast}^2 - L\Norm{\rvx_{m,k}-\rvx}^2\right]\\
		&\leq -2\eta K D_F(\rvx,\rvx^\ast) - \frac{1}{2}\mu \eta K\Norm{\rvx-\rvx^\ast}^2 + \frac{2L\eta}{M}\sum_{m=1}^M\sum_{k=0}^{K-1}\E_r\big[\Norm{\rvx_{m,k}-\rvx}^2\big],
	\end{align*}
	where the second equality holds because the randomness of local updates is independent of the randomness of client sampling and the client sampling is unbiased, the third equality uses the law of total expectation, the fourth equality uses \cref{asm:stochasticity}, and the first inequality uses \cref{lem:perturbed strong convexity} with $\vx=\rvx_{m,k}$, $\vy=\rvx^\ast$, $\vz=\rvx$ and $h = F_{\pi_m}$.
	
	For $\E_r\big[\norm{\Delta}^2\big]$, substituting the overall updates $\Delta$, we have
	\begin{align}
		\E_r\big[\norm{\Delta}^2\big]
		&\leq 4\eta^2\E_r\norm{\frac{1}{S}\sum_{m=1}^S\sum_{k=0}^{K-1} \left(\rvg_{\pi_m}(\rvx_{\pi_m,k}) - \nabla F_{\pi_m}(\rvx_{\pi_m,k})\right)}^2 \nonumber\\
		&\quad+ 4\eta^2\E_r\norm{\frac{1}{S}\sum_{m=1}^S\sum_{k=0}^{K-1} \left(\nabla F_{\pi_m}(\rvx_{\pi_m,k}) - \nabla F_{\pi_m}(\rvx)\right)}^2 \nonumber\\
		&\quad+ 4\eta^2\E_r\norm{\frac{1}{S}\sum_{m=1}^S\sum_{k=0}^{K-1} \left(\nabla F_{\pi_m}(\rvx) - \nabla F_{\pi_m}(\rvx^\ast)\right)}^2 \nonumber\\
		&\quad+ 4\eta^2\E_r\norm{\frac{1}{S}\sum_{m=1}^S\sum_{k=0}^{K-1} \nabla F_{\pi_m}(\rvx^\ast)}^2.\label{eq3:proof:lem:PFL strongly convex:per-round recursion}
	\end{align}
	
	For the first term on the right hand side in \eqref{eq3:proof:lem:PFL strongly convex:per-round recursion}, we have
	\begin{align*}
		\Term{1}{\eqref{eq3:proof:lem:PFL strongly convex:per-round recursion}} &= 4\eta^2\frac{1}{S^2}\sum_{m=1}^S \E_r\norm{\sum_{k=0}^{K-1} \left(\rvg_{\pi_m}(\rvx_{\pi_m,k}) - \nabla F_{\pi_m}(\rvx_{\pi_m,k})\right)}^2\nonumber\\
		&=4\eta^2\frac{1}{S^2}\sum_{m=1}^S\sum_{k=0}^{K-1} \E_r\norm{ \rvg_{\pi_m}(\rvx_{\pi_m,k}) - \nabla F_{\pi_m}(\rvx_{\pi_m,k})}^2\nonumber\\
		&\leq \frac{4\eta^2K\sigma^2}{S},
	\end{align*}
	where the first equality uses the fact that the local updates across clients are independent, the second equality uses \cref{lem:martingale difference property} and the last inequality uses \cref{asm:stochasticity}.
	
	For the second term on the right hand side in \eqref{eq3:proof:lem:PFL strongly convex:per-round recursion}, we have
	\begin{align*}
		\Term{2}{\eqref{eq3:proof:lem:PFL strongly convex:per-round recursion}}
		&\leq 4\eta^2\frac{1}{S^2}SK\sum_{m=1}^S\sum_{k=0}^{K-1}\E_r\norm{\nabla F_{\pi_m}(\rvx_{\pi_m,k}) - \nabla F_{\pi_m}(\rvx)}^2\\
		&= 4\eta^2K\frac{1}{M}\sum_{m=1}^M\sum_{k=0}^{K-1}\E_r\norm{\nabla F_{m}(\rvx_{m,k}) - \nabla F_{m}(\rvx)}^2 \\
		&\leq 4\eta^2L^2K\frac{1}{M}\sum_{m=1}^M\sum_{k=0}^{K-1}\E_r\norm{\rvx_{m,k} - \rvx}^2,
	\end{align*}
	where the first inequality uses Jensen's inequality, the second equality holds because the randomness of local updates is independent of the randomness of client sampling and the client sampling is unbiased, and the second inequality uses \cref{asm:smoothness}.
		
	For the third term on the right hand side in \eqref{eq3:proof:lem:PFL strongly convex:per-round recursion}, we have
	\begin{align*}
		\Term{3}{\eqref{eq3:proof:lem:PFL strongly convex:per-round recursion}}
		&\leq 4\eta^2\frac{1}{S^2}SK\sum_{m=1}^S\sum_{k=0}^{K-1}\E_r\left[\norm{\nabla F_{\pi_m}(\rvx) - \nabla F_{\pi_m}(\rvx^\ast)}^2\right]\nonumber\\
		&\overset{\eqref{eq:smooth+convex:bregman lower bound}}{\leq} 8L\eta^2\frac{1}{S^2}SK\sum_{m=1}^S\sum_{k=0}^{K-1}\E_r\left[D_{F_{\pi_m}}(\rvx, \rvx^\ast)\right] \\
		&= 8L\eta^2K^2D_{F}(\rvx,\rvx^\ast),
	\end{align*}
	where the first inequality uses Jensen's inequality and the equality holds because the client sampling is unbiased.
	
	The fourth term on the right hand side in \eqref{eq3:proof:lem:PFL strongly convex:per-round recursion} can be bounded by Lemma~\ref{lem:simple random sampling} as follows:
	\begin{align*}
		\Term{4}{\eqref{eq3:proof:lem:PFL strongly convex:per-round recursion}} \leq  4\eta^2K^2\frac{M-S}{S(M-1)}\zeta_\ast^2.
	\end{align*}
	
	Plugging the preceding bounds into \eqref{eq3:proof:lem:PFL strongly convex:per-round recursion}, we obtain
	\begin{align*}
		\E_r\norm{\Delta}^2 &\leq \frac{4\eta^2K\sigma^2}{S} + 8L\eta^2K^2D_{F}(\rvx, \rvx^\ast) + 4\eta^2K^2\frac{M-S}{S(M-1)}\zeta_\ast^2\\
		&\quad+4L^2\eta^2K\frac{1}{M}\sum_{m=1}^M\sum_{k=0}^{K-1}\E_r\norm{\rvx_{m,k} - \rvx}^2.
	\end{align*}
	
	Plugging the bounds for $\E_r\big[\langle\rvx^{(r)}-\rvx^*,\Delta\rangle\big]$ and $\E_r\big[\norm{\Delta}^2\big]$ into \eqref{eq:PFL:single round decomposition}, we obtain
	\begin{align*}
		\E_r\norm{\rvx+\Delta-\rvx^\ast}^2 &\leq \left(1-\tfrac{\mu \eta K}{2}\right)\norm{\rvx-\rvx^\ast}^2+\frac{4K\eta^2\sigma^2}{S}+4K^2\eta^2\frac{M-S}{S(M-1)}\zeta_\ast^2 \nonumber\\
		&\;-2K\eta(1-4LK\eta)D_F(\rvx, \rvx^\ast)+2L\eta(1+2LK\eta)\frac{1}{M}\sum_{m=1}^M\sum_{k=0}^{K-1}\E_r\norm{\rvx_{m,k} - \rvx}^2\nonumber\\
		&\leq \left(1-\tfrac{\mu K\eta}{2}\right)\norm{\rvx-\rvx^\ast}^2+\frac{4K\eta^2\sigma^2}{S}+4K^2\eta^2\frac{M-S}{S(M-1)}\zeta_\ast^2 \nonumber\\
		&\quad-\frac{2}{3}K\eta D_F(\rvx, \rvx^\ast)+\frac{8}{3}L\eta\frac{1}{M}\sum_{m=1}^M\sum_{k=0}^{K-1}\E_r\norm{\rvx_{m,k} - \rvx}^2,
	\end{align*}
	where we use the condition that $\eta\leq \frac{1}{6LK}$ in the last inequality. The claim of this lemma follows after recovering the superscripts and taking unconditional expectation.
\end{proof}
\subsubsection{Bounding the client drift with Assumption~\ref{asm:heterogeneity2}}\label{subsubsec:PFL strongly convex:client drift}
\begin{lemma}\label{lem:PFL strongly convex:client drift}
	Let Assumptions~\ref{asm:smoothness}, \ref{asm:stochasticity}, \ref{asm:heterogeneity2} hold and assume that all the local objectives are $\mu$-strongly convex. If the learning rate satisfies $\eta \leq \frac{1}{6LK}$, then the client drift is bounded:
	\begin{align}
		\gE_r\leq \frac{9}{4}K^2\eta^2\sigma^2 + \frac{3}{2}K^3\eta^2\zeta_\ast^2 + 3LK^3\eta^2D_{F}(\rvx^{(r)},\rvx^\ast)\label{eq:lem:PFL strongly convex:client drift}
	\end{align}
\end{lemma}
\begin{proof}
	
	In the following, we focus on a single training round, and hence we drop the superscripts $r$ for a while, e.g., replacing $\rvx_{m,k}^{(r)}$ with $\rvx_{m,k}$ and replacing $\rvx^{(r)}$ with $\rvx$.
	
	The term $\E_r\norm{\rvx_{m,k} - \rvx}^2$ can be bounded by
	\begin{align}
		\E_r\norm{\rvx_{m,k} - \rvx}^2
		&\leq 4\eta^2\E_r\norm{\sum_{j=0}^{k-1} \left(\rvg_m(\rvx_{m,j}) - \nabla F_{m} (\rvx_{m,j})\right)}^2 \nonumber\\
		&\quad+ 4\eta^2\E_r\norm{\sum_{j=0}^{k-1}\left(\nabla F_{m} (\rvx_{m,j}) - \nabla F_{m} (\rvx)\right)}^2 \nonumber\\
		&\quad+ 4\eta^2\E_r\norm{\sum_{j=0}^{k-1}\left(\nabla F_{m} (\rvx) - \nabla F_{m} (\rvx^{\ast})\right)}^2 \nonumber\\
		&\quad+ 4\eta^2\E_r\norm{\sum_{j=0}^{k-1}\nabla F_{m} (\rvx^{\ast})}^2\,.\label{eq:proof:lem-PFL-drift-1}
	\end{align}
	
	For the first term on the right hand side in \eqref{eq:proof:lem-PFL-drift-1},
	\begin{align*}
		\Term{1}{\eqref{eq:proof:lem-PFL-drift-1}} &= 4\eta^2 \sum_{j=0}^{k-1} \E_r\normsq{ \rvg_m(\rvx_{m,j}) - \nabla F_{m} (\rvx_{m,j}) }\\
		& \leq 4\eta^2k \sigma^2,
	\end{align*}
	where we use \cref{lem:martingale difference property} and \cref{asm:stochasticity}.
	
	For the second term on the right hand side in \eqref{eq:proof:lem-PFL-drift-1},
	\begin{align*}
		\Term{2}{\eqref{eq:proof:lem-PFL-drift-1}} &\leq 4\eta^2 k\sum_{j=0}^{k-1}\E_r\norm{\nabla F_{m} (\rvx_{m,j}) - \nabla F_{m} (\rvx)}^2\\ &\leq 4L^2\eta^2 k\sum_{j=0}^{k-1}\E_r\norm{\rvx_{m,j} - \rvx}^2,
	\end{align*}
	where the first inequality uses Jensen's inequality and the second inequality uses \cref{asm:smoothness}.
	
	For the third term on the right hand side in \eqref{eq:proof:lem-PFL-drift-1},
	\begin{align*}
		\Term{3}{\eqref{eq:proof:lem-PFL-drift-1}} &\leq 4\eta^2k\sum_{j=0}^{k-1}\E_r\norm{\nabla F_{m} (\rvx) - \nabla F_{m} (\rvx^{\ast})}^2\\
		&\overset{\eqref{eq:smooth+convex:bregman lower bound}}{\leq} 8L\eta^2k\sum_{j=0}^{k-1}D_{F_m} (\rvx, \rvx^\ast) ,
	\end{align*}
	where the first inequality uses Jensen's inequality.
	
	Thus, we have
	\begin{align*}
		\E_r\norm{\rvx_{m,k} - \rvx}^2 &\leq 4\eta^2k \sigma^2 + 4L^2\eta^2 k\sum_{j=0}^{k-1}\E_r\norm{\rvx_{m,j} - \rvx}^2 \\
		&\quad+ 8L\eta^2k\sum_{j=0}^{k-1}D_{F_m} (\rvx, \rvx^\ast) + 4\eta^2k^2 \norm{\nabla F_m(\rvx^\ast)}^2\,.
	\end{align*}
	Then returning to $\gE_r\coloneqq \frac{1}{M}\sum_{m=1}^M\sum_{k=0}^{K-1} \E_r\left[\norm{\rvx_{m,k} - \rvx}^2\right]$, we have
	\begin{align*}
		\gE_r &\leq 4\eta^2\sigma^2\sum_{k=0}^{K-1}k + 4L^2\eta^2\frac{1}{M}\sum_{m=1}^M\sum_{k=0}^{K-1}k\sum_{j=0}^{k-1}\E_r\norm{\rvx_{m,j}-\rvx}^2\\
		&\quad+8L\eta^2\sum_{k=0}^{K-1}k^2D_{F}(\rvx, \rvx^\ast) + 4\eta^2\sum_{k=0}^{K-1}k^2\zeta_\ast^2\,.
	\end{align*}
	Using the facts $\sum_{k=1}^{K-1}k = \frac{(K-1)K}{2} \leq \frac{K^2}{2}$ and $\sum_{k=1}^{K-1}k^2 = \frac{(K-1)K(2K-1)}{6} \leq \frac{K^3}{3}$, we can simplify the preceding inequality:
	\begin{align*}
		\gE_r &\leq2K^2\eta^2\sigma^2 + 2L^2K^2\eta^2\frac{1}{M}\sum_{m=1}^M\sum_{j=0}^{K-1}\E_r\norm{\rvx_{m,j}-\rvx}^2+\frac{8}{3}LK^3\eta^2D_{F}(\rvx, \rvx^\ast) + \frac{4}{3}K^3\eta^2\zeta_\ast^2\,.
	\end{align*}
	After rearranging the preceding inequality, we get
	\begin{align*}
		(1-2L^2K^2\eta^2)\gE_r \leq 2K^2\eta^2\sigma^2 + \frac{4}{3}K^3\eta^2\zeta_\ast^2 + \frac{8}{3}LK^3\eta^2D_{F}(\rvx,\rvx^\ast)\,.
	\end{align*}
	Finally, using the condition that $\eta \leq \frac{1}{6LK}$, which implies $1-2L^2K^2\eta^2 \geq \frac{8}{9}$, we have
	\begin{align*}
		\gE_r\leq \frac{9}{4}K^2\eta^2\sigma^2 + \frac{3}{2}K^3\eta^2\zeta_\ast^2 + 3LK^3\eta^2D_{F}(\rvx,\rvx^\ast).
	\end{align*}
	The claim follows after recovering the superscripts.
\end{proof}
\subsubsection{Proof of strongly convex case of Theorem~\ref{thm:PFL}}
\begin{proof}[Proof of strongly convex case of Theorem~\ref{thm:PFL}]
	Substituting \eqref{eq:lem:PFL strongly convex:client drift} into \eqref{eq:lem:PFL strongly convex:per-round recursion}, taking unconditional expectation and using $\eta \leq \frac{1}{6LK}$, we obtain
	\begin{align*}
		\E\left[\Norm{\rvx^{(r+1)}-\rvx^*}^2\right]
		&\leq \left(1-\frac{1}{2}\mu K\eta\right)\E\left[\Norm{\rvx^{(r)}- \rvx^\ast}^2\right] - \frac{1}{3}K\eta\E\left[D_F(\rvx^{(r)}, \rvx^\ast)\right] \nonumber\\
		&\quad+\frac{4K\eta^2\sigma^2}{S} +4K^2\eta^2\frac{M-S}{S(M-1)}\zeta_\ast^2+6LK^2\eta^3\sigma^2+4LK^3\eta^3\zeta_\ast^2
	\end{align*}
	Using $\tilde{\eta} = \eta K$, we have
	\begin{align}
		\E\left[\Norm{\rvx^{(r+1)}-\rvx^*}^2\right] &\leq \left(1-\frac{\mu \tilde{\eta}}{2}\right)\E\left[\Norm{\rvx^{(r)}- \rvx^\ast}^2\right] - \frac{\tilde{\eta}}{3}\E\left[D_F(\rvx^{(r)}, \rvx^\ast)\right]\nonumber\\
		&\quad+\frac{4\tilde{\eta}^2\sigma^2}{SK}+\frac{4\tilde{\eta}^2(M-S)\zeta_\ast^2}{S(M-1)}+\frac{6L\tilde{\eta}^3\sigma^2}{K}+4L\tilde{\eta}^3\zeta_\ast^2\label{eq1:thm:proof:PFL strongly convex}
	\end{align}
	
	Applying Lemma~\ref{lem:stongly convex:tuning learning rate} with $t=r$ ($T=R$), $\gamma=\tilde\eta$, $r_{t} = \E\left[\norm{\rvx^{(r)}- \rvx^*}^2\right]$, $a = \frac{\mu}{2}$, $b=\frac{1}{3}$, $s_t = \E\left[D_F(\rvx^{(r)}, \rvx^\ast)\right]$, $w_t=(1-\tfrac{\mu\tilde\eta}{2})^{-(r+1)}$, $c_1 = \frac{4\sigma^2}{SK}+\frac{4(M-S)\zeta_\ast^2}{S(M-1)}$, $c_2 = \frac{6L\sigma^2}{K} + 4L\zeta_\ast^2$ and $\frac{1}{d}=\frac{1}{6L}$ ($\tilde\eta=\eta K\leq\frac{1}{6L}$), we obtain
	\begin{align}
		&\E\left[F(\bar\rvx^{(R)})-F(\rvx^\ast)\right] \nonumber\\
		&\leq \frac{1}{\sum_{r=0}^Rw_r}\sum_{r=0}^Rw_r\E\left[F(\rvx^{(r)}) - F(\rvx^\ast)\right]\nonumber\\
		&\leq \frac{9}{2}\mu\Norm{\rvx^{(0)}- \rvx^*}^2 \exp\left(-\tfrac{1}{2}\mu\tilde\eta R\right)+\frac{12\tilde{\eta}\sigma^2}{SK}+\frac{12\tilde{\eta}(M-S)\zeta_\ast^2}{S(M-1)}+\frac{18L\tilde{\eta}^2\sigma^2}{K}+12L\tilde{\eta}^2\zeta_\ast^2,\label{eq2:thm:proof:PFL strongly convex}
	\end{align}
	where $\bar\rvx^{(R)} = \frac{1}{\sum_{r=0}^Rw_r}\sum_{r=0}^Rw_r\rvx^{(r)}$ and we use Jensen's inequality ($F$ is convex) in the first inequality. Tuning the learning rate carefully, we get
	\begin{align}
		\E\left[F(\bar\rvx^{(R)})-F(\rvx^\ast)\right] = \tilde\gO\left(\mu D^2 \exp\left(-\frac{\mu R}{12L}\right)+\frac{\sigma^2}{\mu SKR}+\frac{(M-S)\zeta_\ast^2}{\mu SR(M-1)} + \frac{L\sigma^2}{\mu^2KR^2} + \frac{L\zeta_\ast^2}{\mu^2R^2}\right)\label{eq3:thm:proof:PFL strongly convex}
	\end{align}
	where $D= \norm{\rvx^{(0)}- \rvx^\ast}$. Eq.~\eqref{eq2:thm:proof:PFL strongly convex} and Eq.~\eqref{eq3:thm:proof:PFL strongly convex} are the upper bounds for partial client participation. In particular, when $S=M$, we can get the claim of the strongly convex case of Theorem~\ref{thm:PFL} and Corollary~\ref{cor:PFL}.
\end{proof}
\subsection{General convex case}\label{subsec:PFL general convex}
\subsubsection{Proof of general convex case of Theorem~\ref{thm:PFL} and Corollary~\ref{cor:PFL}}
\begin{proof}[Proof of general convex case of Theorem~\ref{thm:PFL}]
Letting $\mu=0$ in \eqref{eq1:thm:proof:PFL strongly convex}, we get the simplified per-round recursion of general convex case,
\begin{align*}
	\E\left[\Norm{\rvx^{(r+1)}-\rvx^*}^2\right] &\leq \E\left[\Norm{\rvx^{(r)}- \rvx^\ast}^2\right] - \frac{\tilde{\eta}}{3}\E\left[D_F(\rvx^{(r)}, \rvx^\ast)\right]\nonumber\\
	&\quad+\frac{4\tilde{\eta}^2\sigma^2}{SK}+\frac{4\tilde{\eta}^2(M-S)\zeta_\ast^2}{S(M-1)}+\frac{6L\tilde{\eta}^3\sigma^2}{K}+4L\tilde{\eta}^3\zeta_\ast^2
\end{align*}
Applying Lemma~\ref{lem:general convex:tuning learning rate} with $t=r$ ($T=R$), $\gamma=\tilde\eta$, $r_{t} = \E\left[\norm{\rvx^{(r)}- \rvx^*}^2\right]$, $b=\frac{1}{3}$, $s_t = \E\left[D_F(\rvx^{(r)}, \rvx^\ast)\right]$, $w_t=1$, $c_1 = \frac{4\sigma^2}{SK}+\frac{4(M-S)\zeta_\ast^2}{S(M-1)}$, $c_2 = \frac{6L\sigma^2}{K} + 4L\zeta_\ast^2$ and $\frac{1}{d}=\frac{1}{6L}$ ($\tilde\eta=K\eta\leq\frac{1}{6L}$), we obtain
\begin{align}
	&\E\left[F(\bar\rvx^{(R)})-F(\rvx^\ast)\right] \nonumber\\
	&\leq \frac{1}{\sum_{r=0}^Rw_r}\sum_{r=0}^Rw_r\E\left[F(\rvx^{(r)}) - F(\rvx^\ast)\right]\nonumber\\
	&\leq \frac{3\norm{\rvx^{(0)}- \rvx^*}^2}{\tilde\eta R}+\frac{12\tilde{\eta}\sigma^2}{SK}+\frac{12\tilde{\eta}(M-S)\zeta_\ast^2}{S(M-1)}+\frac{18L\tilde{\eta}^2\sigma^2}{K}+12L\tilde{\eta}^2\zeta_\ast^2\label{eq1:thm:proof:PFL general convex}
\end{align}
where $\bar\rvx^{(R)} = \frac{1}{\sum_{r=0}^Rw_r}\sum_{r=0}^Rw_r\rvx^{(r)}$ and we use Jensen's inequality ($F$ is convex) in the first inequality. After tuning the learning rate, we obtain
\begin{align}
	&\E\left[F(\bar\rvx^R)-F(\rvx^\ast)\right] \nonumber\\
	&= \gO\left(\frac{\sigma D}{\sqrt{SKR}}+\sqrt{1-\frac{S}{M}}\cdot \frac{\zeta_\ast D}{\sqrt{SR}} + \frac{\left(L\sigma^2D^4\right)^{1/3}}{K^{1/3}R^{2/3}} + \frac{\left(L\zeta_\ast^2D^4\right)^{1/3}}{R^{2/3}} + \frac{LD^2}{R}\right)\label{eq2:thm:proof:PFL general convex}
\end{align}
where $D = \Norm{\rvx^{(0)}- \rvx^\ast}$. Eq.~\eqref{eq1:thm:proof:PFL general convex} and Eq.~\eqref{eq2:thm:proof:PFL general convex} are the upper bounds for partial client participation. In particular, when $S=M$, we can get the claim of the strongly convex case of Theorem~\ref{thm:PFL} and Corollary~\ref{cor:PFL}.
\end{proof}
\subsection{Non-convex case}\label{subsec:PFL non-convex}
\begin{lemma}\label{lem:PFL non-convex:per-round recursion}
	Let Assumptions~\ref{asm:smoothness}, \ref{asm:stochasticity}, \ref{asm:heterogeneity1} hold. If the learning rate satisfies $\eta \leq \frac{1}{6LK (1+\beta^2)}$, then
	\begin{align}
		\E_r\left[F(\rvx^{(r+1)}) - F(\rvx^{(r)})\right] &\leq -\frac{1}{6}\eta K\Norm{\nabla F(\rvx^{(r)})}^2 + \frac{2L\eta^2 K\sigma^2}{S} + 2L\eta^2K^2 \frac{M-S}{S(M-1)}\zeta^2\nonumber\\
		&\quad+\frac{5}{6}L^2\eta K \frac{1}{M} \sum_{m=1}^M \frac{1}{K}\sum_{k=0}^{K-1} \E_r\left[\Norm{ \rvx_{m,k}^{(r)} - \rvx^{(r)}}^2\right]\label{eq:lem:PFL non-convex:per-round recursion}
	\end{align}
\end{lemma}
\begin{proof}
	In the following, we focus on a single training round, and hence we drop the superscripts $r$ for a while, e.g., replacing $\rvx_{m,k}^{(r)}$ with $\rvx_{m,k}$ and replacing $\rvx^{(r)}$ with $\rvx$.
	
	Starting from the smoothness of $F$, we have
	\begin{align*}
		\E_r\left[F(\rvx+\Delta) - F(\rvx)\right]\leq \E_r\left[\inp{\nabla F(\rvx)}{\Delta}\right] + \frac{L}{2}\E_r\norm{\Delta }^2.
	\end{align*}
	For $\E_r\left[\inp{\nabla F(\rvx)}{\Delta}\right]$, after substituting the global update $\Delta$ with \eqref{eq:PFL:complete update}, we obtain
	\begin{align*}
		\E_r\left[\inp{\nabla F(\rvx)}{\Delta}\right]
		&= -\eta \frac{1}{S}\sum_{m=1}^S\sum_{k=0}^{K-1}\E_r\left[\inp{\nabla F(\rvx)}{\rvg_{\pi_m}(\rvx_{\pi_m,k})}\right]\\
		&= -\eta \frac{1}{M}\sum_{m=1}^M\sum_{k=0}^{K-1}\E_r\left[\inp{\nabla F(\rvx)}{\rvg_{m}(\rvx_{m,k})}\right]\\
		&= -\eta \frac{1}{M}\sum_{m=1}^M\sum_{k=0}^{K-1}\E_r\left[\E_r\left[\inp{\nabla F(\rvx)}{\rvg_{m}(\rvx_{m,k})}\mid \rvx_{m,k}\right]\right]\\
		&= -\eta \frac{1}{M}\sum_{m=1}^M\sum_{k=0}^{K-1}\E_r\left[\inp{\nabla F(\rvx)}{\nabla F_{m}(\rvx_{m,k})}\right]\\
		&= -\eta K\E_r\left[\inp{\nabla F(\rvx)}{\frac{1}{M}\sum_{m=1}^M\frac{1}{K}\sum_{k=0}^{K-1}\nabla F_{m}(\rvx_{m,k})}\right] \\
		&\leq-\frac{1}{2}\eta K\normsq{\nabla F(\rvx)} +\frac{1}{2}\eta K \E_r\norm{\frac{1}{M}\sum_{m=1}^M\frac{1}{K}\sum_{k=0}^{K-1}\nabla F_{m}(\rvx_{m,k}) - \nabla F(\rvx)}^2\\
		&\leq-\frac{1}{2}\eta K\normsq{\nabla F(\rvx)} +\frac{1}{2}\eta K \frac{1}{M}\sum_{m=1}^M\frac{1}{K}\sum_{k=0}^{K-1}\E_r\norm{\nabla F_{m}(\rvx_{m,k}) - \nabla F_m(\rvx)}^2\\
		&\leq-\frac{1}{2}\eta K\normsq{\nabla F(\rvx)} +\frac{1}{2}\eta L^2K \frac{1}{M}\sum_{m=1}^M\frac{1}{K}\sum_{k=0}^{K-1}\E_r\norm{\rvx_{m,k} - \rvx}^2,
	\end{align*}
	where the second equality holds because the randomness of local updates is independent of the randomness of client sampling and the client sampling is unbiased, the third equality uses the law of total expectation, the fourth equality uses \cref{asm:stochasticity}, the first inequality uses $\inp{a}{b} = \frac{1}{2}\norm{a}^2 + \frac{1}{2}\norm{b}^2 - \frac{1}{2}\norm{a-b}^2 \geq \frac{1}{2}\norm{a}^2 - \frac{1}{2}\norm{a-b}^2 $, the second inequality uses Jensen's inequality, and the last inequality uses \cref{asm:smoothness}.
	
	For $\frac{L}{2}\E_r\norm{\Delta}^2$, after substituting the global update $\Delta$ with \eqref{eq:PFL:complete update}, we obtain
	\begin{align}
		\frac{L}{2}\E_r\norm{\Delta}^2 &\leq 2L\eta^2\E_r\norm{\frac{1}{S}\sum_{m=1}^S\sum_{k=0}^{K-1} \left( \rvg_{\pi_m}(\rvx_{\pi_m,k}) - \nabla F_{\pi_m}(\rvx_{\pi_m,k})\right) }^2 \nonumber\\
		&\quad+ 2L\eta^2\E_r\norm{\frac{1}{S}\sum_{m=1}^S\sum_{k=0}^{K-1} \left(\nabla F_{\pi_m}(\rvx_{\pi_m,k}) - \nabla F_{\pi_m}(\rvx)\right)  }^2\nonumber\\
		&\quad+2L\eta^2\E_r\norm{\frac{1}{S}\sum_{m=1}^S\sum_{k=0}^{K-1} \left(\nabla F_{\pi_m}(\rvx) - \nabla F(\rvx) \right)}^2 \nonumber\\
		&\quad+ 2L\eta^2\E_r\norm{\frac{1}{S}\sum_{m=1}^S\sum_{k=0}^{K-1}\nabla F(\rvx)}^2.\label{eq:proof:PFL-non-convex:recursion-1}
	\end{align}
	
	For the first term on the right hand side in \eqref{eq:proof:PFL-non-convex:recursion-1},
	\begin{align*}
		\Term{1}{\eqref{eq:proof:PFL-non-convex:recursion-1}} &= 2L\eta^2 \frac{1}{S^2} \sum_{m=1}^S \sum_{k=0}^{K-1} \E_r\normsq{\rvg_{\pi_m}(\rvx_{\pi_m,k}) - \nabla F_{\pi_m}(\rvx_{\pi_m,k})}\\ &\leq  \frac{2L\eta^2 K\sigma^2}{S},
	\end{align*}
	where we use \cref{lem:martingale difference property} and \cref{asm:stochasticity}.
	
	For the second term on the right hand side in \eqref{eq:proof:PFL-non-convex:recursion-1},
	\begin{align*}
		\Term{2}{\eqref{eq:proof:PFL-non-convex:recursion-1}}&\leq 2L \eta^2 K^2 \frac{1}{S} \sum_{m=1}^S\frac{1}{K} \sum_{k=0}^{K-1} \E_r\norm{\nabla F_{\pi_m}(\rvx_{\pi_m,k}) - \nabla F_{\pi_m}(\rvx) }^2  \\
		&= 2L\eta^2 K^2 \frac{1}{M} \sum_{m=1}^M \frac{1}{K}\sum_{k=0}^{K-1} \E_r\norm{ \nabla F_m(\rvx_{m,k}) - \nabla F_m(\rvx)}^2 \\
		&\leq 2L^3\eta^2 K^2 \frac{1}{M} \sum_{m=1}^M \frac{1}{K}\sum_{k=0}^{K-1} \E_r\norm{\rvx_{m,k} - \rvx}^2,
	\end{align*}
	where the first inequality uses Jensen's inequality, the second equality holds because the randomness of local updates is independent of the randomness of client sampling and the client sampling is unbiased, and the last inequality uses \cref{asm:smoothness}.
	
	For the third term on the right hand side in \eqref{eq:proof:PFL-non-convex:recursion-1},
	\begin{align*}
		\Term{3}{\eqref{eq:proof:PFL-non-convex:recursion-1}}&= 2L\eta^2K^2 \frac{M-S}{S(M-1)} \frac{1}{M} \sum_{m=1}^M\norm{ \nabla F_m(\rvx) - \nabla F(\rvx) }^2  \\
		&\leq 2L\eta^2 K^2 \frac{M-S}{S(M-1)} \beta^2 \norm{\nabla F(\rvx)}^2 + 2L\eta^2 K^2 \frac{M-S}{S(M-1)} \zeta^2
	\end{align*}
	where the equality uses \cref{lem:simple random sampling} and the inequality uses \cref{asm:heterogeneity1}.
	
	Plugging these bounds into \eqref{eq:proof:PFL-non-convex:recursion-1}, we obtain
	\begin{align*}
		\frac{L}{2}\E_r\norm{\Delta}^2 &\leq \frac{2L\eta^2 K\sigma^2}{S} + 2L\eta^2K^2 \frac{M-S}{S(M-1)}\zeta^2 + 2L\eta^2 K^2 \left(\frac{(M-S)\beta^2}{S(M-1)}+1\right)  \norm{\nabla F(\rvx)}^2\\
		&\quad+2L^3\eta^2 K^2 \frac{1}{M} \sum_{m=1}^M \frac{1}{K}\sum_{k=0}^{K-1} \E_r\norm{ \rvx_{m,k} - \rvx}^2.
	\end{align*}
	Next, using the condition $\eta \leq \frac{1}{6LK (1+\beta^2)}$, we can get
	\begin{align*}
		\E_r\left[F(\rvx+\Delta) - F(\rvx)\right]&\leq -\frac{1}{6}\eta K\norm{\nabla F(\rvx)}^2 + \frac{2L\eta^2 K\sigma^2}{S} + 2L\eta^2K^2 \frac{M-S}{S(M-1)}\zeta^2\\
		&\quad+\frac{5}{6}L^2\eta  \frac{1}{M} \sum_{m=1}^M\sum_{k=0}^{K-1} \E_r\norm{ \rvx_{m,k} - \rvx}^2.
	\end{align*}
	The claim follows after recovering the superscripts.
\end{proof}
\subsubsection{Bounding the client drift with Assumption~\ref{asm:heterogeneity1}}
\begin{lemma}\label{lem:PFL non-convex:client drift}
	Let Assumptions~\ref{asm:smoothness}, \ref{asm:stochasticity}, \ref{asm:heterogeneity1} hold. If the learning rate satisfies $\eta \leq \frac{1}{6LK}$, then the client drift is bounded:	
	\begin{align}
		\gE_r\leq \frac{9}{4}K^2\eta^2\sigma^2 + \frac{3}{2}K^3\eta^2\zeta^2 + \frac{3}{2}K^3\eta^2(\beta^2+1)\Norm{\nabla F(\rvx^{(r)})}^2.\label{eq:lem:PFL non-convex:client drift}
	\end{align}
\end{lemma}
\begin{proof}
	In the following, we focus on a single training round, and hence we drop the superscripts $r$ for a while, e.g., replacing $\rvx_{m,k}^{(r)}$ with $\rvx_{m,k}$ and replacing $\rvx^{(r)}$ with $\rvx$.
	
	The term $\E_r\norm{\rvx_{m,k} - \rvx}^2$ can be bounded by
	\begin{align}
		\E_r\norm{\rvx_{m,k} - \rvx}^2
		&\leq 4\eta^2\E_r\norm{\sum_{j=0}^{k-1} \left(\rvg_m(\rvx_{m,j}) - \nabla F_{m} (\rvx_{m,j})\right)}^2 \nonumber\\
		&\quad+ 4\eta^2\E_r\norm{\sum_{j=0}^{k-1}\left(\nabla F_{m} (\rvx_{m,j}) - \nabla F_{m} (\rvx)\right)}^2 \nonumber\\
		&\quad+ 4\eta^2\E_r\norm{\sum_{j=0}^{k-1}\left(\nabla F_{m} (\rvx)-\nabla F(\rvx)\right)}^2 \nonumber\\
		&\quad+ 4\eta^2\E_r\norm{\sum_{j=0}^{k-1}\nabla F (\rvx)}^2.\label{eq:proof:lem-PFL-drift-2}
	\end{align}
	
	For the first term on the right hand side in \eqref{eq:proof:lem-PFL-drift-2},
	\begin{align*}
		\Term{1}{\eqref{eq:proof:lem-PFL-drift-2}} &= 4\eta^2 \sum_{j=0}^{k-1} \E_r\normsq{ \rvg_m(\rvx_{m,j}) - \nabla F_{m} (\rvx_{m,j}) } \\&\leq 4\eta^2k \sigma^2 ,
	\end{align*}
	where we use \cref{lem:martingale difference property} and \cref{asm:stochasticity}.
	
	For the second term on the right hand side in \eqref{eq:proof:lem-PFL-drift-2},
	\begin{align*}
		\Term{2}{\eqref{eq:proof:lem-PFL-drift-2}} &\leq 4\eta^2 k\sum_{j=0}^{k-1}\E_r\norm{\nabla F_{m} (\rvx_{m,j}) - \nabla F_{m} (\rvx)}^2\\ &\leq 4L^2\eta^2 k\sum_{j=0}^{k-1}\E_r\norm{\rvx_{m,j} - \rvx}^2 ,
	\end{align*}
	where the first inequality uses Jensen's inequality and the second inequality uses \cref{asm:smoothness}.
	
	The third term on the right hand side in \eqref{eq:proof:lem-PFL-drift-2} is bounded by $4\eta^2k^2\E_r\norm{\nabla F_{m} (\rvx) - \nabla F (\rvx)}^2$. The fourth term on the right hand side in \eqref{eq:proof:lem-PFL-drift-2} is bounded by $4\eta^2k^2\E_r\norm{\nabla F (\rvx)}^2$.
	
	Plugging the preceding bounds into \eqref{eq:proof:lem-PFL-drift-2}, and then plugging the resulting bound for $\E_r \normsq{\rvx_{m,k} - \rvx}$ into $\gE_r$, we have
	\begin{align*}
		\gE_r &\leq 4\eta^2\sigma^2\sum_{k=0}^{K-1}k + 4L^2\eta^2\frac{1}{M}\sum_{m=1}^M\sum_{k=0}^{K-1}k\sum_{j=0}^{k-1}\E_r\norm{\rvx_{m,j}-\rvx}^2\\
		&\quad+ 4\eta^2(\beta^2+1)\norm{\nabla F(\rvx)}^2\sum_{k=0}^{K-1}k^2 + 4\eta^2\zeta^2\sum_{k=0}^{K-1}k^2.
	\end{align*}
	Using $\sum_{k=1}^{K-1}k = \frac{(K-1)K}{2} \leq \frac{K^2}{2}$ and $\sum_{k=1}^{K-1}k^2 = \frac{(K-1)K(2K-1)}{6} \leq \frac{K^3}{3}$, we can simplify the preceding inequality:
	\begin{align*}
		\gE_r &\leq2K^2\eta^2\sigma^2 + 2L^2K^2\eta^2\frac{1}{M}\sum_{m=1}^M\sum_{j=0}^{K-1}\E_r\norm{\rvx_{m,j}-\rvx}^2+\frac{4}{3}K^3\eta^2(\beta^2+1)\norm{\nabla F(\rvx)}^2 + \frac{4}{3}K^3\eta^2\zeta^2
	\end{align*}
	After rearranging the preceding inequality, we get
	\begin{align*}
		(1-2L^2K^2\eta^2)\gE_r \leq 2K^2\eta^2\sigma^2 + \frac{4}{3}K^3\eta^2\zeta^2 + \frac{4}{3}K^3\eta^2(\beta^2+1)\norm{\nabla F(\rvx)}^2
	\end{align*}
	Finally, using the condition that $\eta \leq \frac{1}{6LK}$, which implies $1-2L^2K^2\eta^2 \geq \frac{8}{9}$, we have
	\begin{align*}
		\gE_r\leq \frac{9}{4}K^2\eta^2\sigma^2 + \frac{3}{2}K^3\eta^2\zeta^2 + \frac{3}{2}K^3\eta^2(\beta^2+1)\norm{\nabla F(\rvx)}^2
	\end{align*}
	The claim follows after recovering the superscripts.
\end{proof}
\subsubsection{Proof of non-convex case of Theorem~\ref{thm:PFL}}
\begin{proof}[Proof of non-convex case of Theorem~\ref{thm:PFL}]
	Substituting \eqref{eq:lem:PFL non-convex:client drift} into \eqref{eq:lem:PFL non-convex:per-round recursion}, taking unconditional expectation and then using $\eta\leq \frac{1}{6LK(\beta^2+1)}$, we obtain
	\begin{align*}
		\E \left[F(\rvx^{(r+1)})-F(\rvx^{(r)})\right]
		&\leq -\frac{1}{8}\eta K\E\left[\Norm{\nabla F(\rvx^{(r)})}^2\right]+\frac{2LK\eta^2\sigma^2}{S} + \frac{15}{8}L^2K^2\eta^3\sigma^2\\
		&\quad + 2L \eta^2 K^2\frac{M-S}{S(M-1)}\zeta^2 + \frac{5}{4}L^2\eta^3K^3\zeta^2.
	\end{align*}
	Using $\tilde \eta = \eta K$, subtracting $F^\ast$ from both sides and then rearranging the terms, we have
	\begin{align*}
		\E \left[F(\rvx^{(r+1)})- F^\ast\right] &\leq \E\left[F(\rvx^{(r)}) - F^\ast\right] - \frac{\tilde\eta}{8}\E\left[\Norm{\nabla F(\rvx^{(r)})}^2\right] \\
		&\quad+ \frac{2L\tilde\eta^2\sigma^2}{SK}+ \frac{15}{8}\frac{L^2\tilde\eta^3\sigma^2}{K} + 2L \tilde\eta^2\frac{M-S}{S(M-1)}\zeta^2 + \frac{5}{4}L^2\tilde \eta^3\zeta^2
	\end{align*}
	Then applying Lemma~\ref{lem:general convex:tuning learning rate} with $t=r$ ($T=R$), $\gamma=\tilde\eta$, $r_{t} = \E\left[F(\rvx^{(r)})-F^\ast\right]$, $b=\frac{1}{8}$, $s_t = \E\left[\norm{\nabla F(\rvx^{(r)})}^2\right]$, $w_t=1$, $c_1 = \frac{2L\sigma^2}{SK}+2L\frac{M-S}{S(M-1)}\zeta^2$, $c_2 = \frac{15}{8}\frac{L^2\sigma^2}{K} + \frac{5}{4}L^2\zeta^2$ and $\frac{1}{d}=\frac{1}{6L(\beta^2+1)}$ ($\tilde\eta=\eta K\leq\frac{1}{6L(\beta^2+1)}$), we have
	\begin{align}
		&\frac{1}{R+1}\sum_{r=0}^{R}\E\left[\Norm{\nabla F(\rvx^{(r)})}^2\right]\nonumber\\
		&\leq \frac{8\left(F(\rvx^{0}) - F^\ast\right)}{\tilde\eta R} + \frac{16L\tilde\eta\sigma^2}{SK}+\frac{16L\tilde\eta\zeta^2(M-S)}{S(M-1)}+ \frac{15L^2\tilde\eta^2\sigma^2}{K} + 10L^2\tilde\eta^2\zeta^2\label{eq1:thm:proof:PFL non-convex}.
	\end{align}
	After tuning the learning rate, we obtain
	\begin{align}
		&\frac{1}{R+1}\sum_{r=0}^{R}\E\left[\Norm{\nabla F(\rvx^{(r)})}^2\right] \nonumber\\
		&= \gO\left(\frac{ \left(L\sigma^2A\right)^{1/2}}{\sqrt{SKR}} + \sqrt{1-\frac{S}{M}}\cdot \frac{ \left(L\zeta^2A\right)^{1/2}}{\sqrt{SR}}+ \frac{\left(L^2\sigma^2A^2\right)^{1/3}}{K^{1/3}R^{2/3}} + \frac{\left(L^2\zeta^2A^2\right)^{1/3}}{R^{2/3}} + \frac{L(\beta^2+1) A}{R}\right)\label{eq2:thm:proof:PFL non-convex}
	\end{align}
	where $A = F(\rvx^{0})- F^\ast$. Eq.~\eqref{eq1:thm:proof:PFL non-convex} and Eq.~\eqref{eq2:thm:proof:PFL non-convex} are the bounds for the partial participation. When $S=M$, we get the bounds for the full participation. Recall that $\bar\rvx^{(R)}$ is sampled uniformly at random from $\{\rvx^{(r)}\}_{r=0}^{R}$, so we have $\E\left[\Norm{\nabla F(\bar \rvx^{(R)})}^2\right] = \frac{1}{R+1}\sum_{r=0}^{R} \E\left[\Norm{\nabla F(\rvx^{(r)})}^2\right]$.
\end{proof}

\newpage
\section{Simulations on quadratic functions}\label{sec:app:simulation}
\begin{figure}[htbp]
	\begin{subfigure}{0.325\linewidth}
		\centering
		\includegraphics[width=1\linewidth]{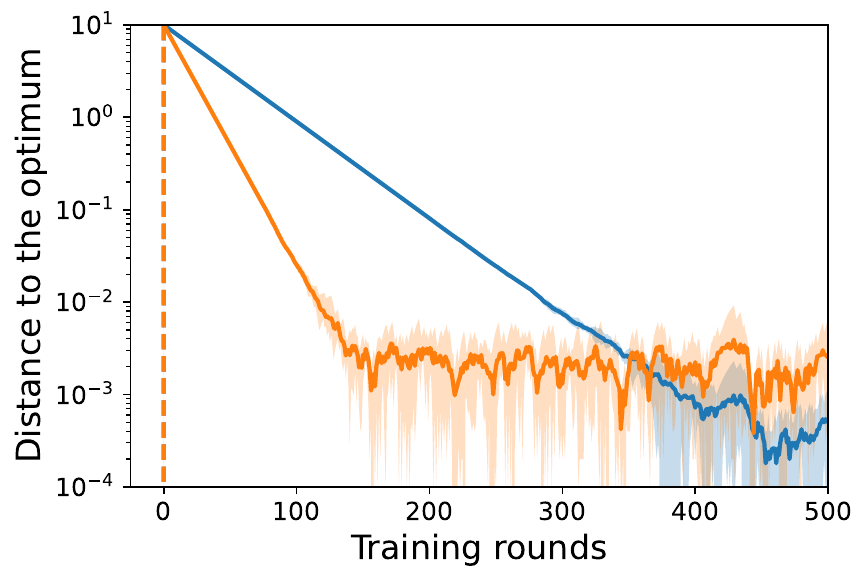}
		\caption{$\delta=0$, $\zeta_\ast=1$}
	\end{subfigure}
	\begin{subfigure}{0.325\linewidth}
		\centering
		\includegraphics[width=1\linewidth]{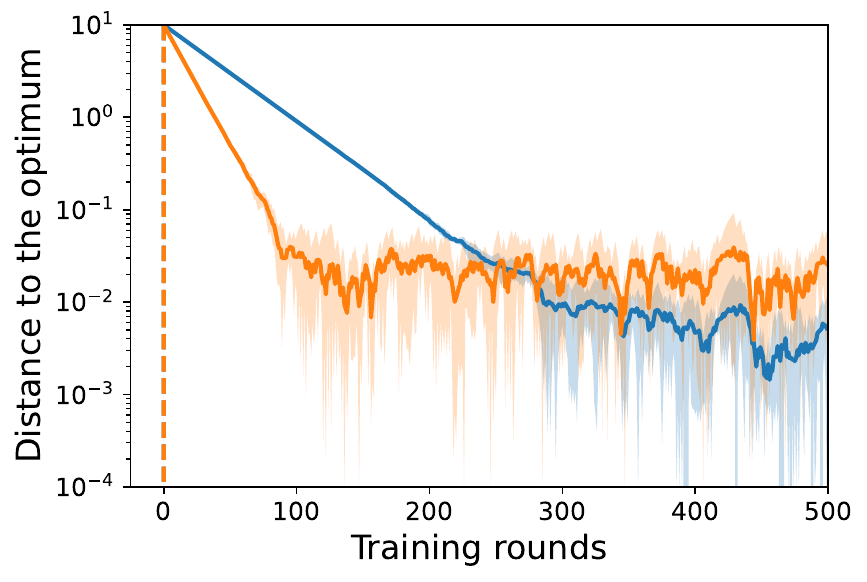}
		\caption{$\delta=0$, $\zeta_\ast=10$}
	\end{subfigure}
	\begin{subfigure}{0.325\linewidth}
		\centering
		\includegraphics[width=1\linewidth]{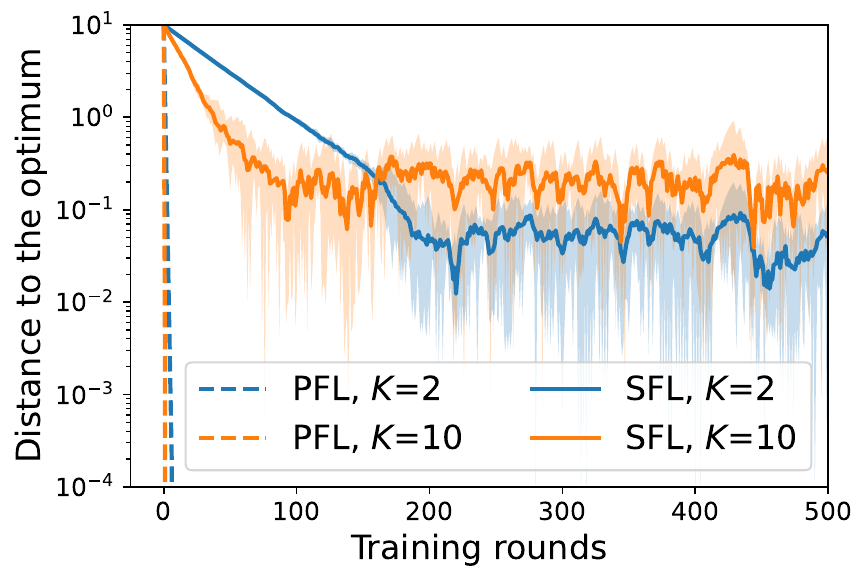}
		\caption{$\delta=0$, $\zeta_\ast=100$}
	\end{subfigure}
	\begin{subfigure}{0.325\linewidth}
		\centering
		\includegraphics[width=1\linewidth]{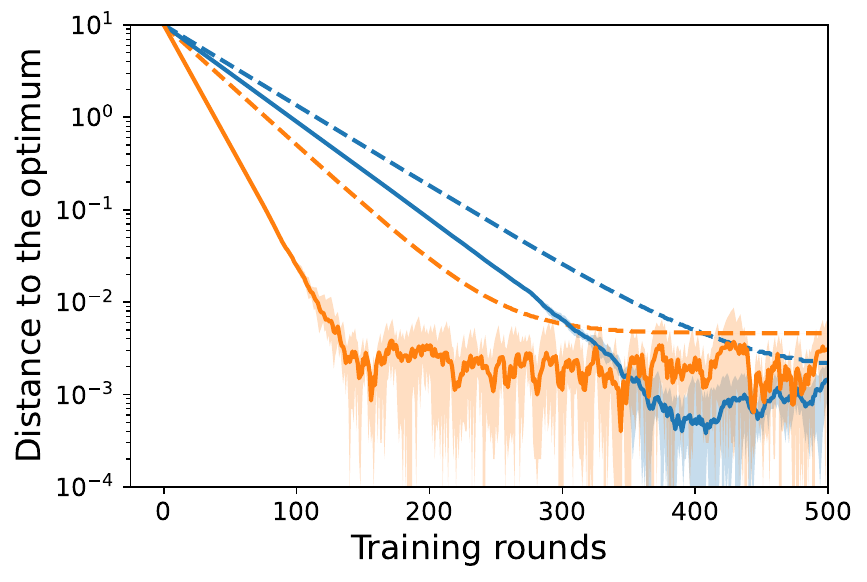}
		\caption{$\delta=\frac{1}{3}$, $\zeta_\ast=1$}
	\end{subfigure}
	\begin{subfigure}{0.325\linewidth}
		\centering
		\includegraphics[width=1\linewidth]{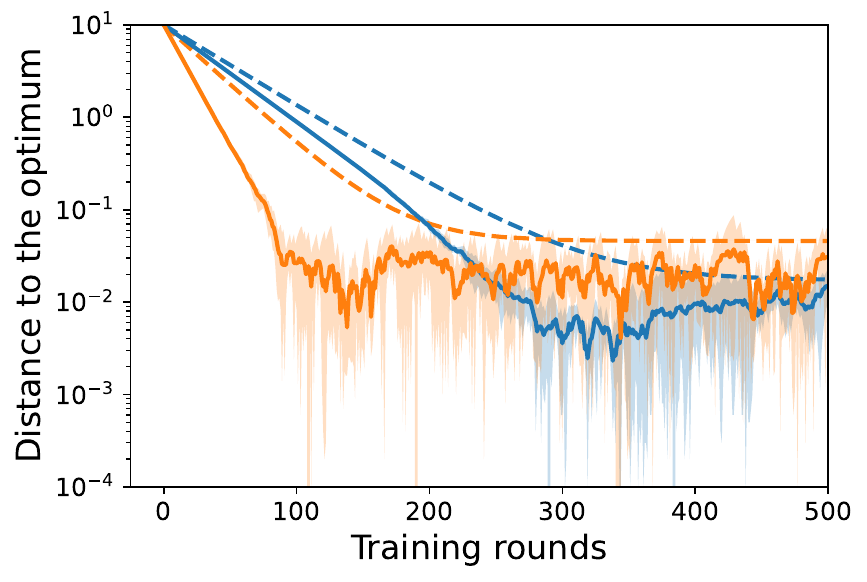}
		\caption{$\delta=\frac{1}{3}$, $\zeta_\ast=10$}
	\end{subfigure}
	\begin{subfigure}{0.325\linewidth}
		\centering
		\includegraphics[width=1\linewidth]{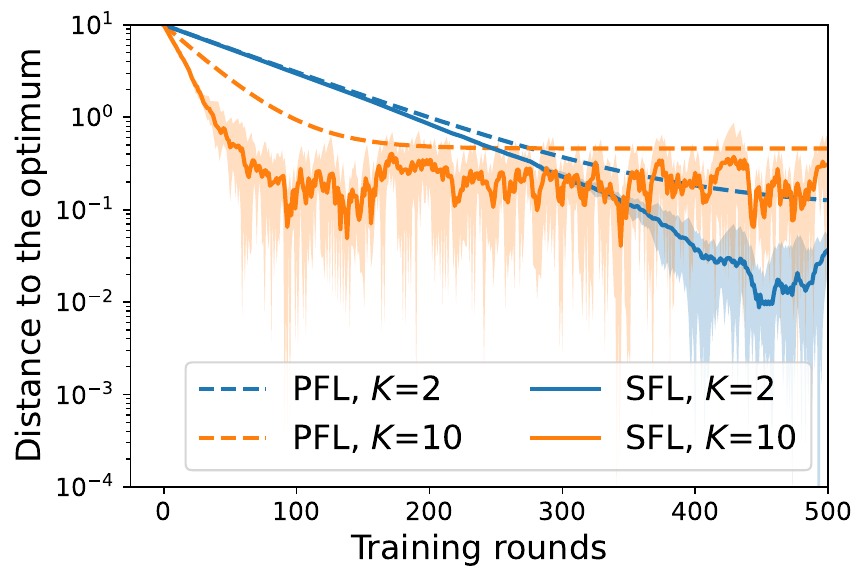}
		\caption{$\delta=\frac{1}{3}$, $\zeta_\ast=100$}
	\end{subfigure}
	\begin{subfigure}{0.325\linewidth}
		\centering
		\includegraphics[width=1\linewidth]{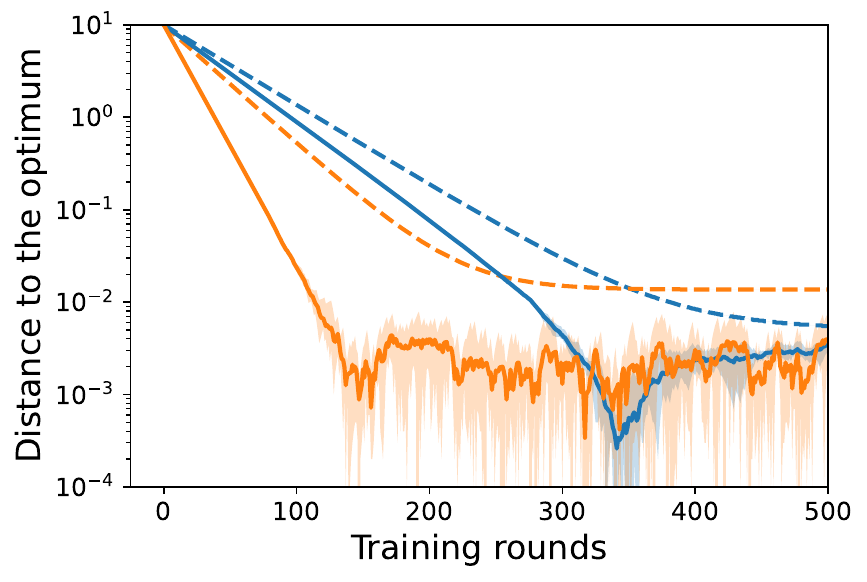}
		\caption{$\delta=1$, $\zeta_\ast=1$}
	\end{subfigure}
	\begin{subfigure}{0.325\linewidth}
		\centering
		\includegraphics[width=1\linewidth]{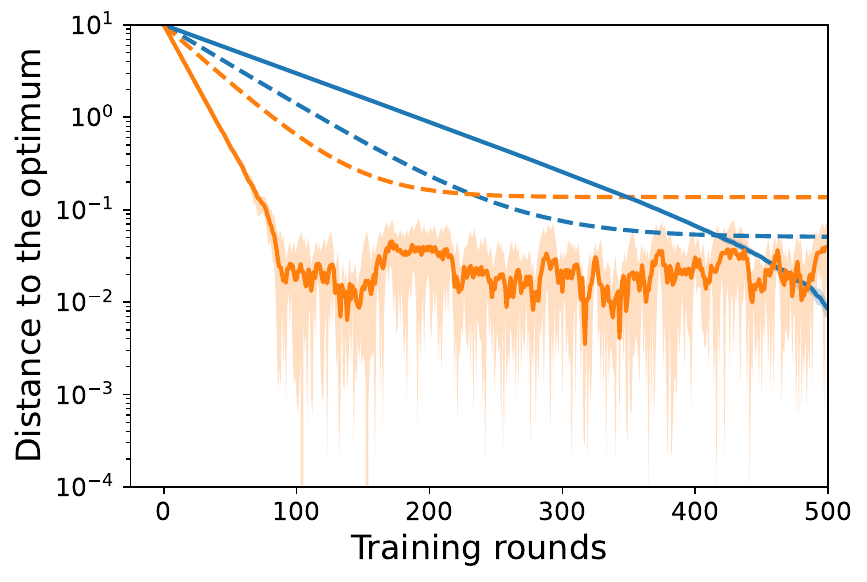}
		\caption{$\delta=1$, $\zeta_\ast=10$}
	\end{subfigure}
	\begin{subfigure}{0.325\linewidth}
		\centering
		\includegraphics[width=1\linewidth]{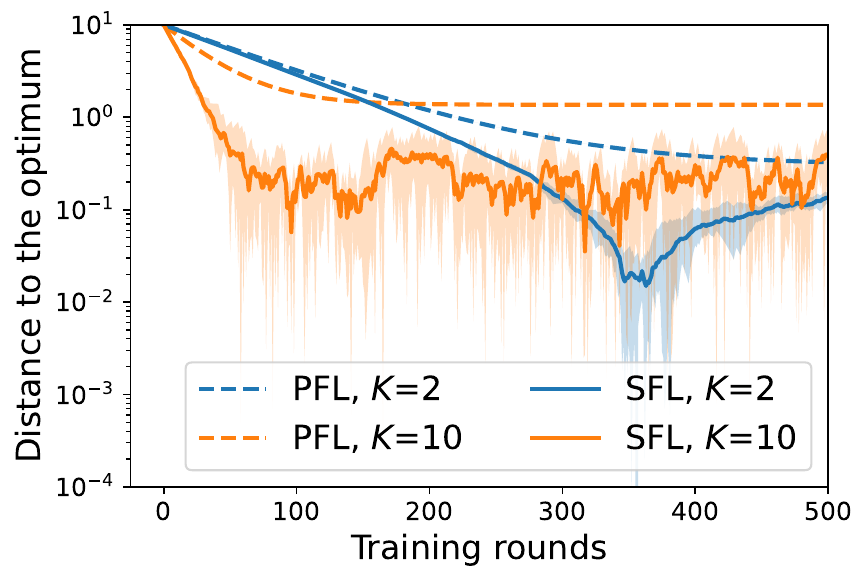}
		\caption{$\delta=1$, $\zeta_\ast=100$}
	\end{subfigure}
	\caption{Simulations on quadratic functions. The best learning rates are chosen from [0.003, 0.006, 0.01, 0.03, 0.06, 0.1, 0.3, 0.6] with grid search. We run each experiment for 5 random seeds. Shaded areas show the min-max values.}
	\label{fig:app:simulation}
\end{figure}

Nine groups of simulated experiments with various degrees of heterogeneity are provided in Table~\ref{tab:app:simulation settings} as an extension of the experiment in Subsection~\ref{subsec:simulation}. Figure~\ref{fig:app:simulation} plots the results of PFL and SFL with various combinations of $\delta$ and $\zeta_\ast$.
\begin{table}[h!]
	\renewcommand{\arraystretch}{1}
	\centering
	\caption{Settings of simulated experiments. Each setting has two local objectives (i.e., $M=2$) and shares the same global objective. Choosing large value of $\zeta_\ast$ and $\delta$ means higher heterogeneity. The definitions of $\zeta_\ast$ and $\delta$ can be found in Subsection~\ref{subsec:simulation}.}
	\label{tab:app:simulation settings}
	\normalsize{
		\begin{tabular}{clll}
			\toprule
			Settings &$\zeta_\ast=1$ &$\zeta_\ast=10$ &$\zeta_\ast=100$ \\
			\midrule
			$\delta=0$ &$\begin{cases}F_1(x)=\frac{1}{2}x^2 + x\\ F_2(x)=\frac{1}{2}x^2 - x\end{cases}$ &$\begin{cases}F_1(x)=\frac{1}{2}x^2 + 10x\\ F_2(x)=\frac{1}{2}x^2 - 10x\end{cases}$ &$\begin{cases}F_1(x)=\frac{1}{2}x^2 + 100x\\ F_2(x)=\frac{1}{2}x^2 - 100x\end{cases}$ \\
			\midrule
			$\delta=\frac{1}{3}$ &$\begin{cases}F_1(x)=\frac{2}{3}x^2 + x\\ F_2(x)=\frac{1}{3}x^2 - x\end{cases}$ &$\begin{cases}F_1(x)=\frac{2}{3}x^2 + 10x\\ F_2(x)=\frac{1}{3}x^2 - 10x\end{cases}$ &$\begin{cases}F_1(x)=\frac{2}{3}x^2 + 100x\\ F_2(x)=\frac{1}{3}x^2 - 100x\end{cases}$ \\
			\midrule
			$\delta=1$ &$\begin{cases}F_1(x)=x^2 + x\\ F_2(x) = -x\end{cases}$ &$\begin{cases}F_1(x)=x^2 + 10x\\ F_2(x) = -10x\end{cases}$ &$\begin{cases}F_1(x)=x^2 + 100x\\ F_2(x) = -100x\end{cases}$ \\
			\bottomrule
	\end{tabular}}
\end{table}

\newpage
\section{More experimental details}\label{sec:app:exp}
This section serves as a supplement and enhancement to Section~\ref{sec:exp}.
\subsection{Extended Dirichlet partition}\label{subsec:app:exdir}
\begin{figure}[htbp]
	\centering
	\begin{subfigure}{0.325\linewidth}
		\centering
		\includegraphics[width=1\linewidth]{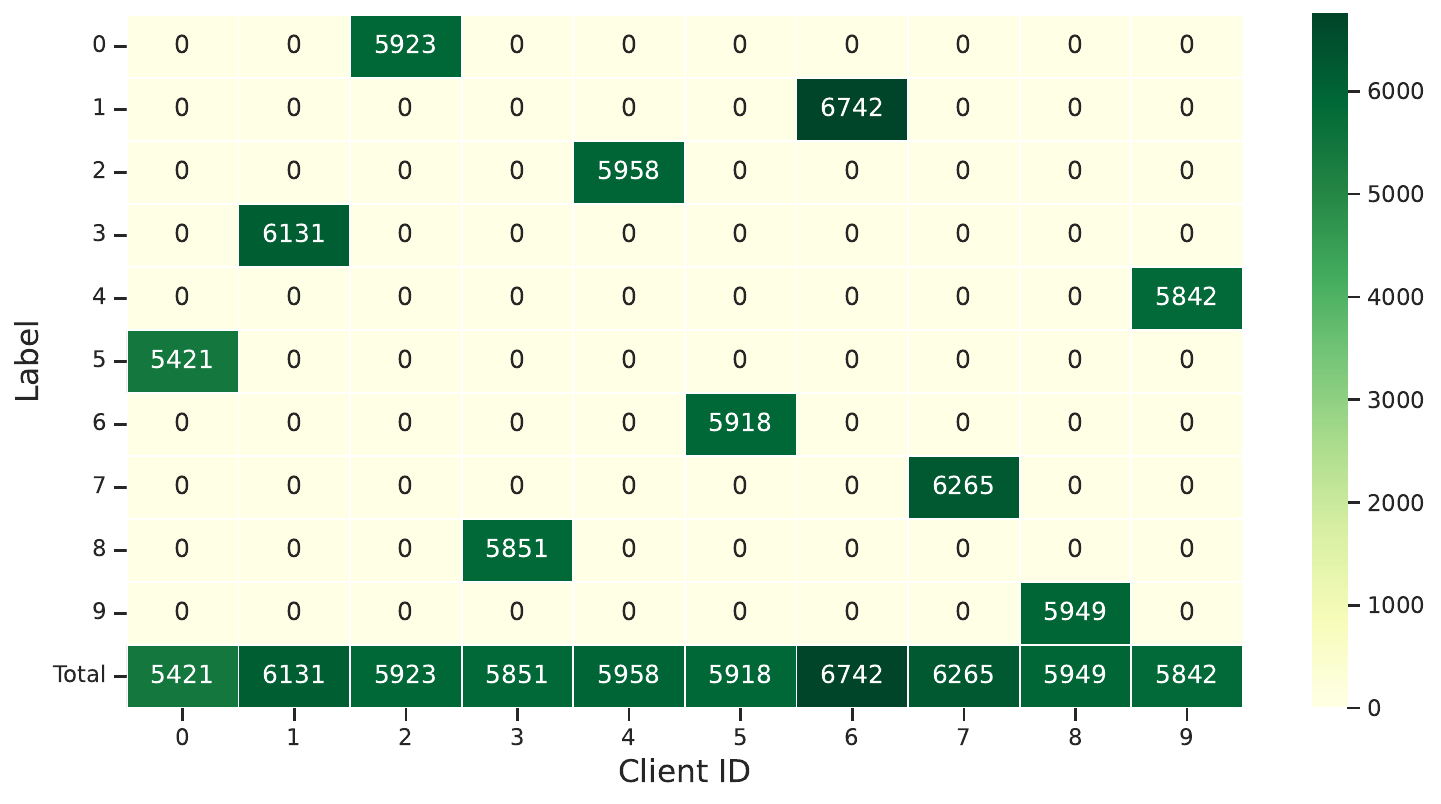}
		\caption{10 clients, $\text{ExDir}(1,100.0)$}
	\end{subfigure}
	\centering
	\begin{subfigure}{0.325\linewidth}
		\centering
		\includegraphics[width=1\linewidth]{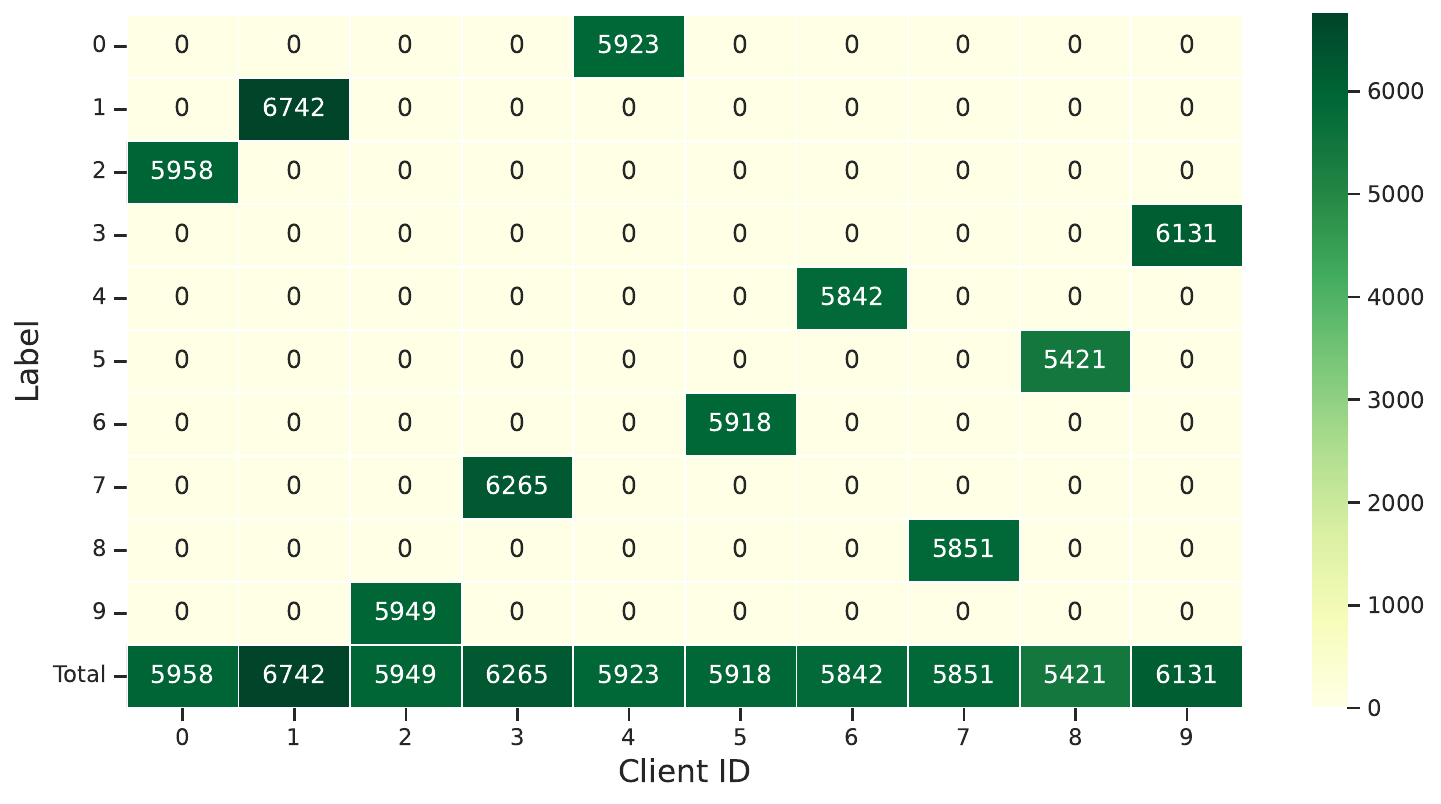}
		\caption{10 clients, $\text{ExDir}(1,10.0)$}
	\end{subfigure}
	\centering
	\begin{subfigure}{0.325\linewidth}
		\centering
		\includegraphics[width=1\linewidth]{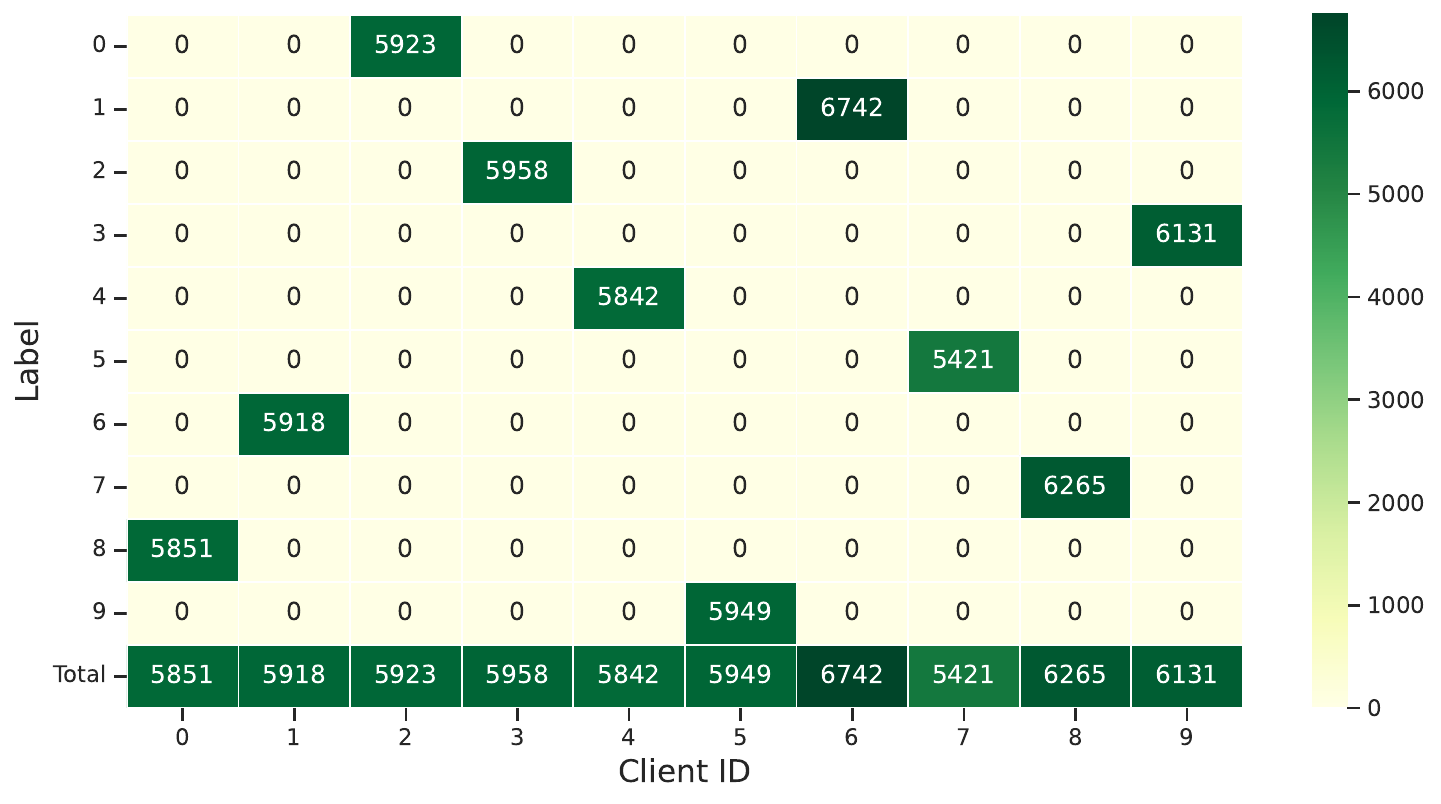}
		\caption{10 clients, $\text{ExDir}(1,1.0)$}
	\end{subfigure}	
	\centering
	\begin{subfigure}{0.325\linewidth}
		\centering
		\includegraphics[width=1\linewidth]{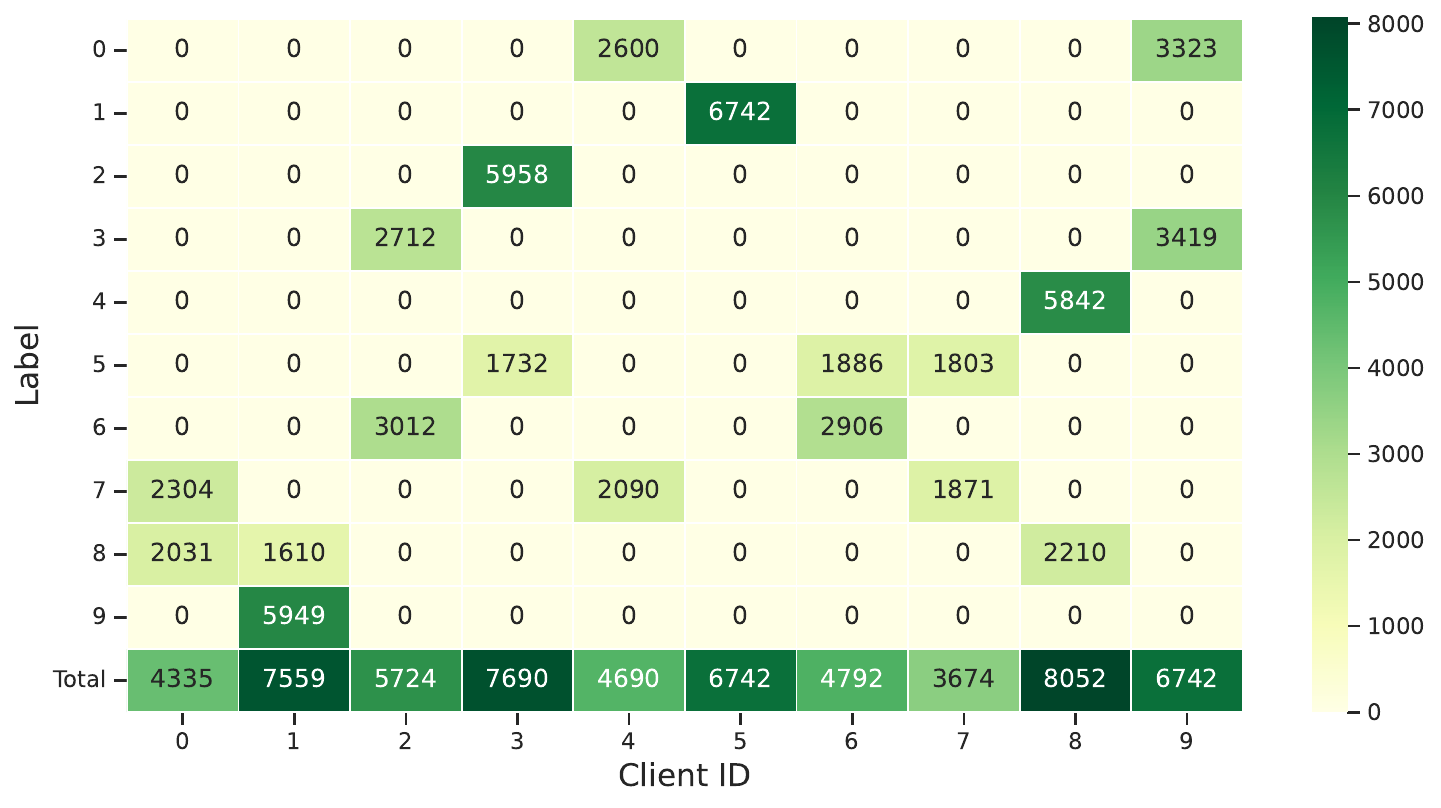}
		\caption{10 clients, $\text{ExDir}(2,100.0)$}
	\end{subfigure}
	\centering
	\begin{subfigure}{0.325\linewidth}
		\centering
		\includegraphics[width=1\linewidth]{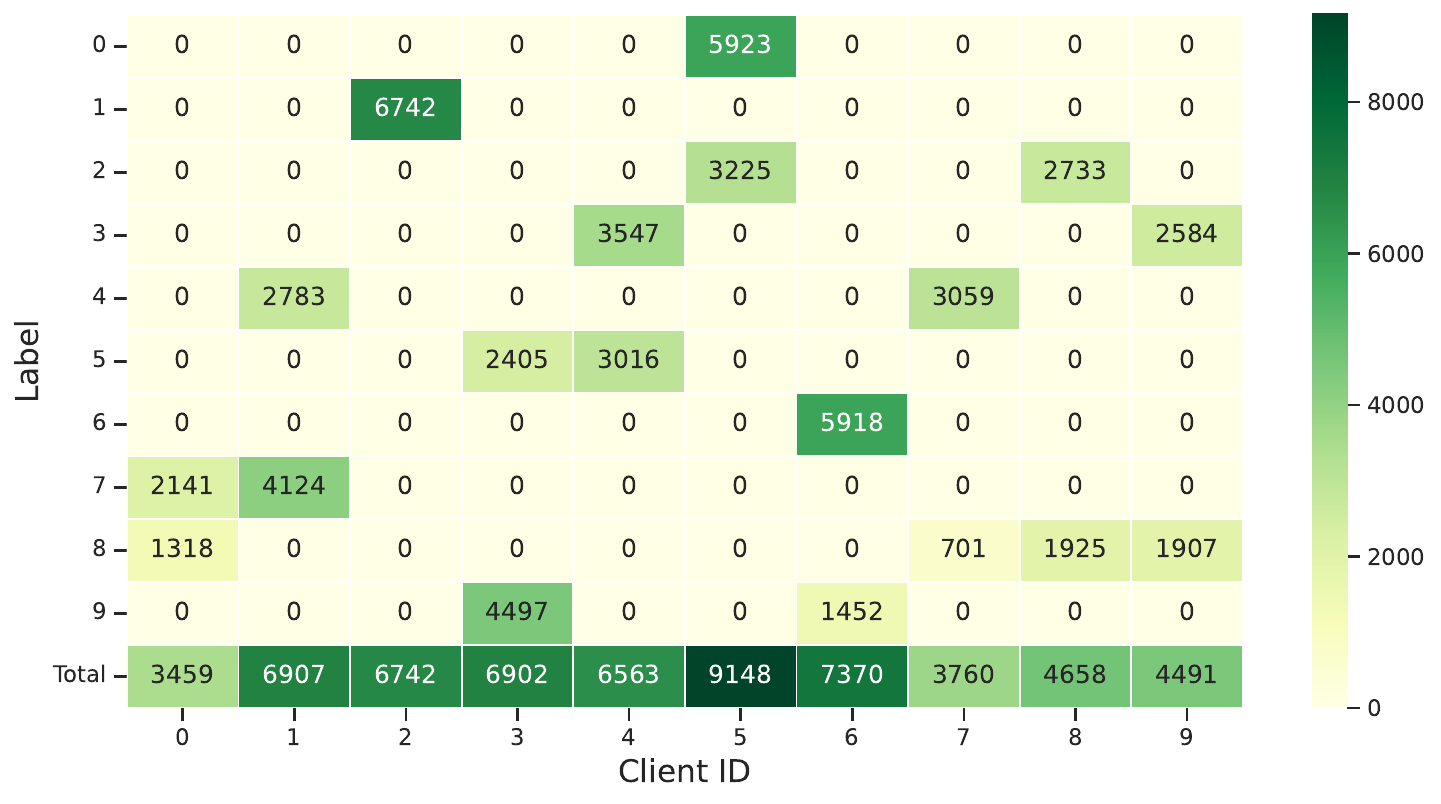}
		\caption{10 clients, $\text{ExDir}(2,10.0)$}
	\end{subfigure}
	\centering
	\begin{subfigure}{0.325\linewidth}
		\centering
		\includegraphics[width=1\linewidth]{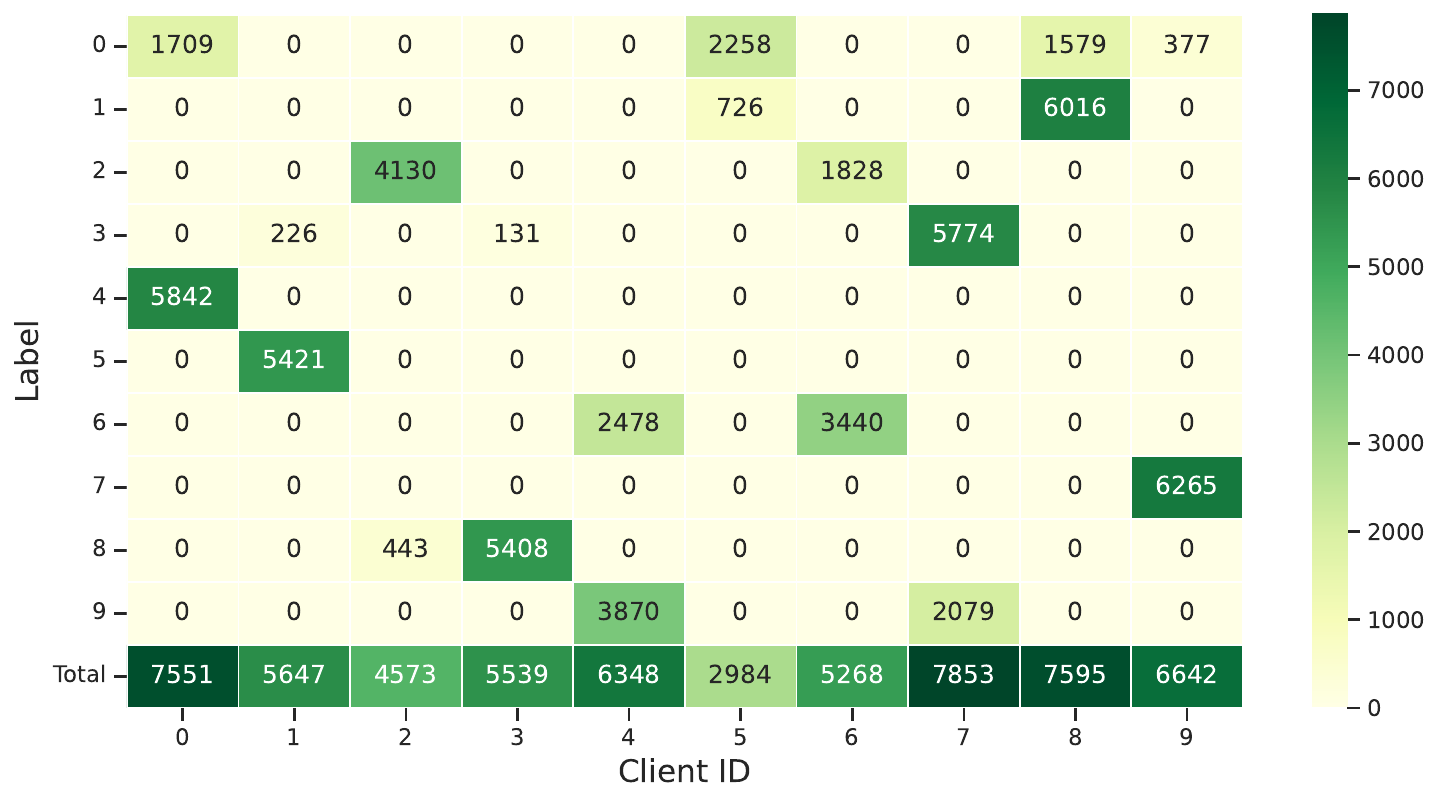}
		\caption{10 clients, $\text{ExDir}(2,1.0)$}
	\end{subfigure}	
	\centering
	\begin{subfigure}{0.325\linewidth}
		\centering
		\includegraphics[width=1\linewidth]{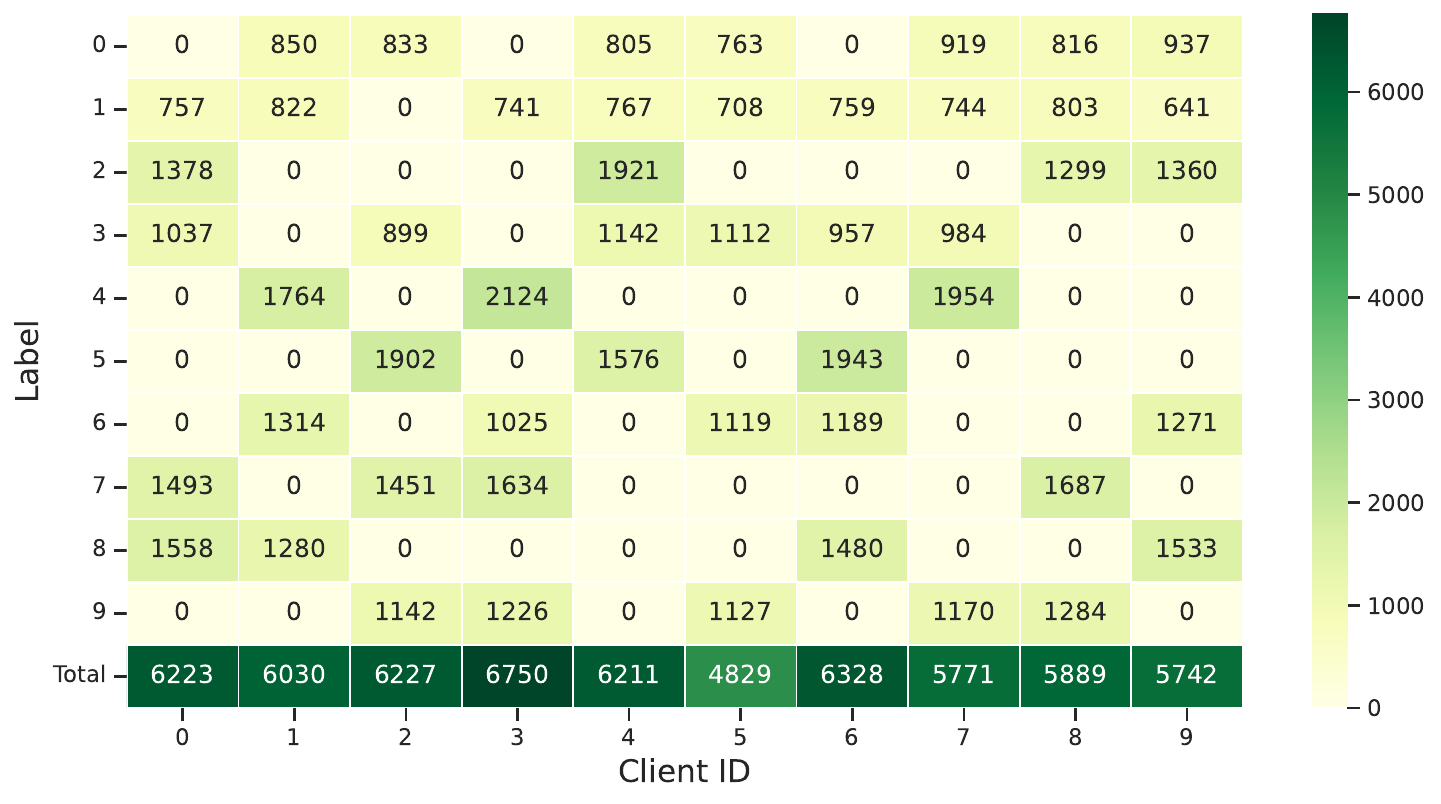}
		\caption{10 clients, $\text{ExDir}(5,100.0)$}
	\end{subfigure}
	\centering
	\begin{subfigure}{0.325\linewidth}
		\centering
		\includegraphics[width=1\linewidth]{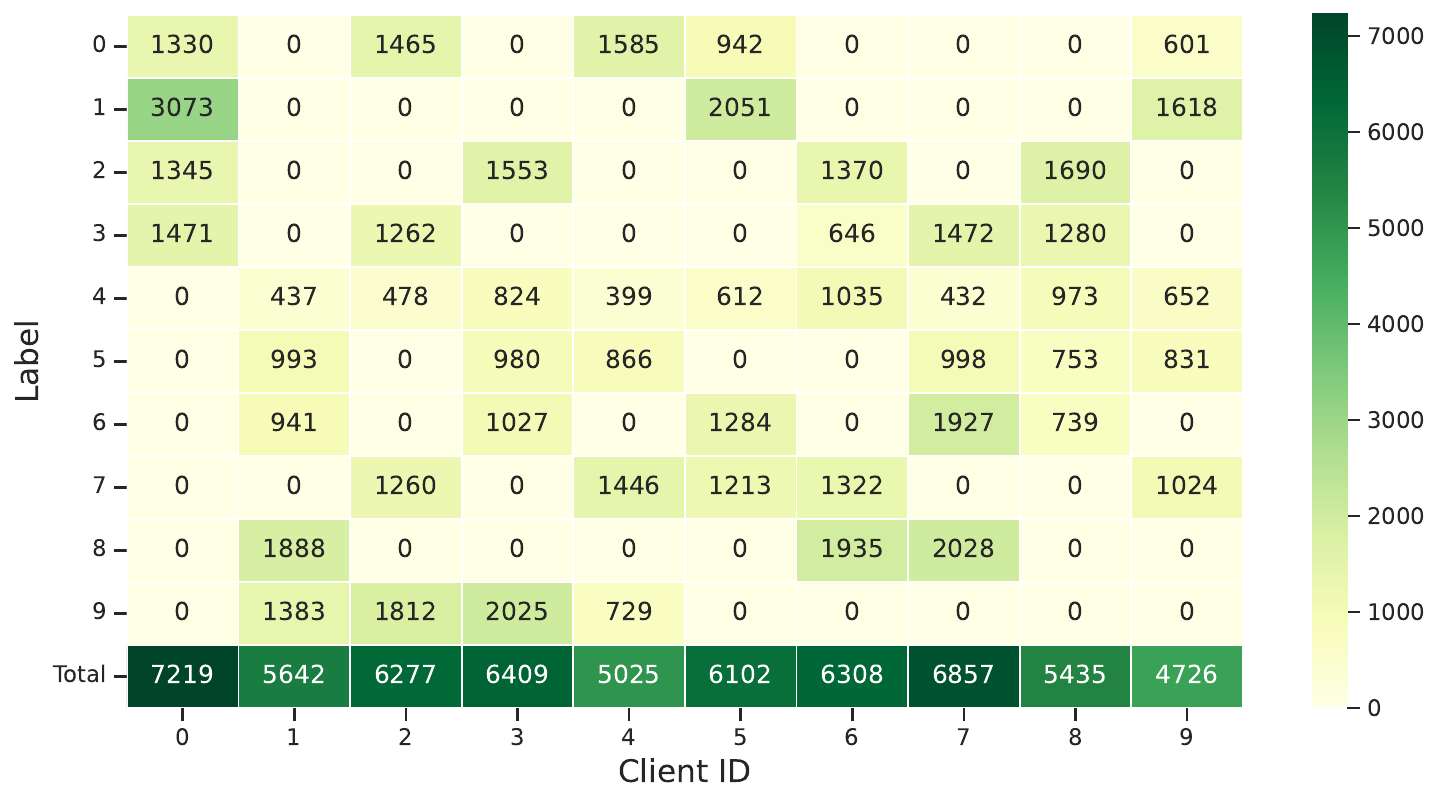}
		\caption{10 clients, $\text{ExDir}(5,10.0)$}
	\end{subfigure}
	\centering
	\begin{subfigure}{0.325\linewidth}
		\centering
		\includegraphics[width=1\linewidth]{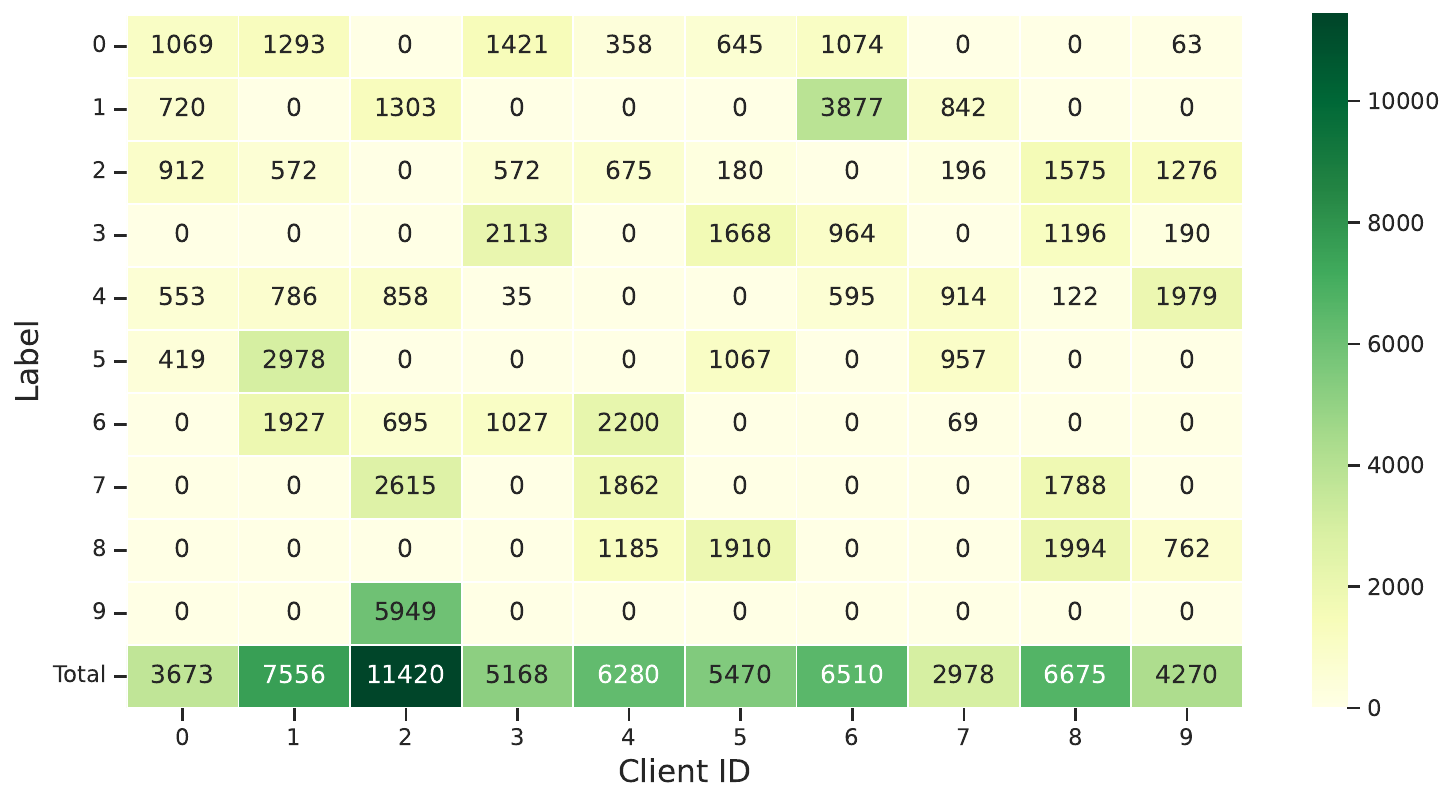}
		\caption{10 clients, $\text{ExDir}(5,1.0)$}
	\end{subfigure}	
	\centering
	\begin{subfigure}{0.325\linewidth}
		\centering
		\includegraphics[width=1\linewidth]{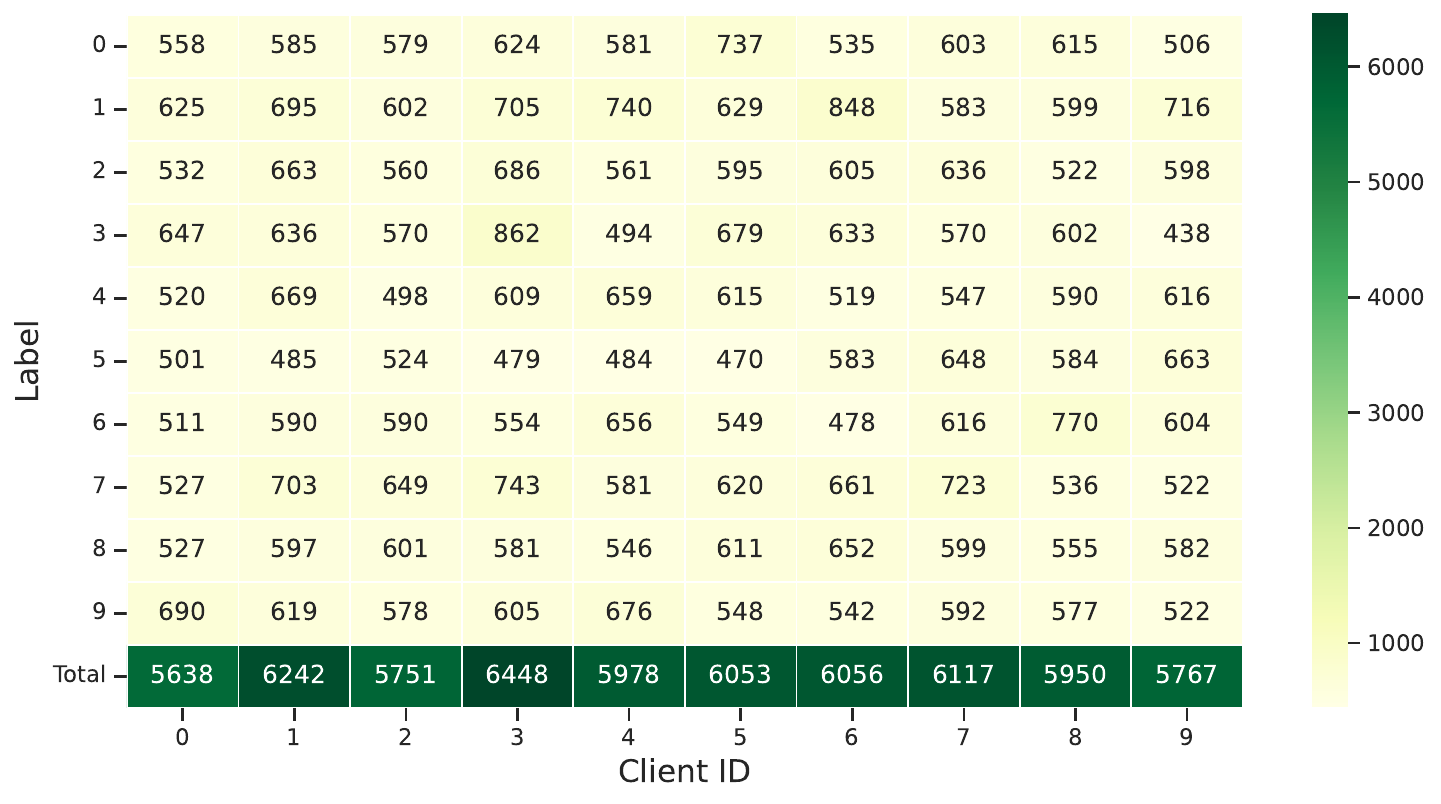}
		\caption{10 clients, $\text{ExDir}(10,100.0)$}
	\end{subfigure}
	\centering
	\begin{subfigure}{0.325\linewidth}
		\centering
		\includegraphics[width=1\linewidth]{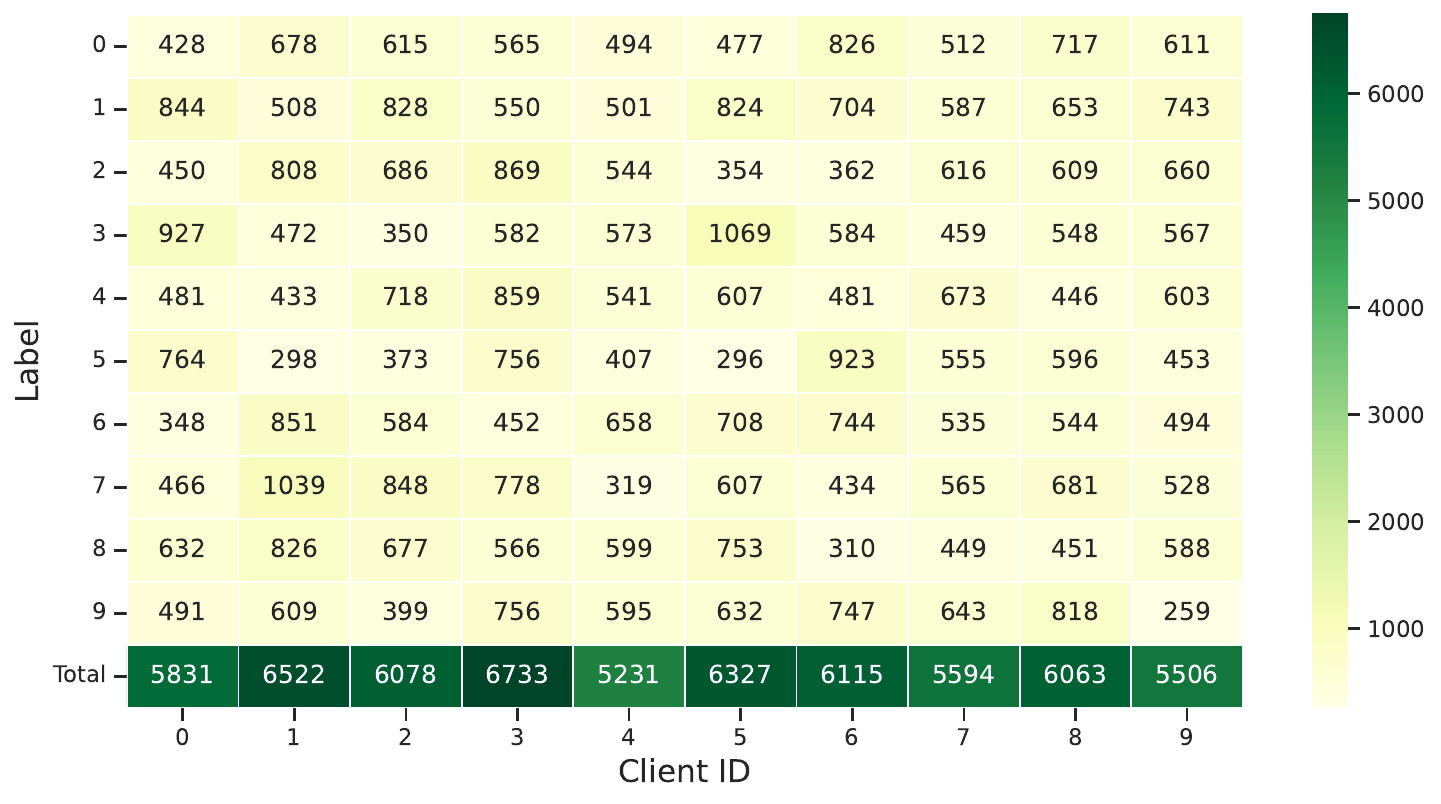}
		\caption{10 clients, $\text{ExDir}(10,10.0)$}
	\end{subfigure}
	\centering
	\begin{subfigure}{0.325\linewidth}
		\centering
		\includegraphics[width=1\linewidth]{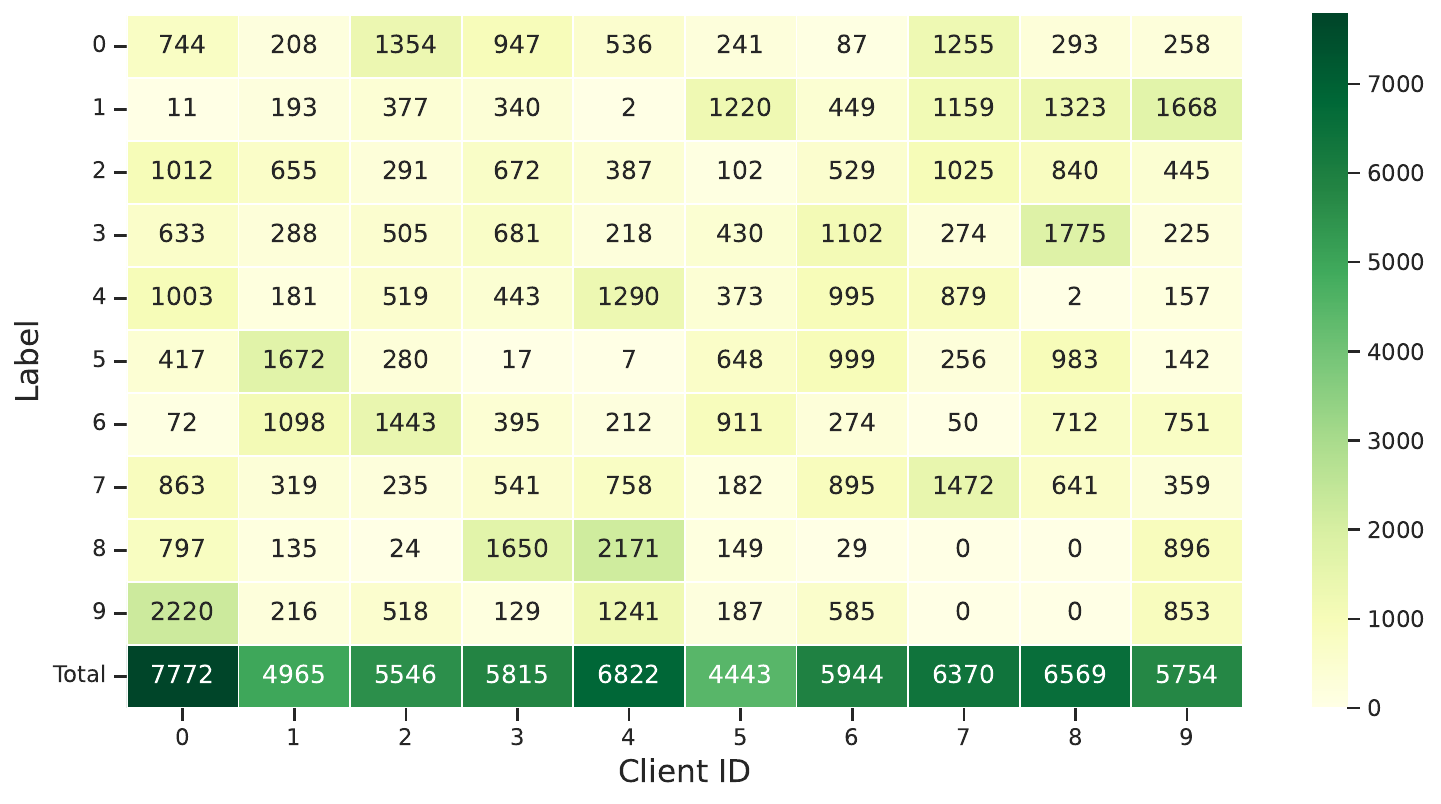}
		\caption{10 clients, $\text{ExDir}(10,1.0)$}
	\end{subfigure}
	\caption{Visualization of partitioning results on MNIST by Extended Dirichlet strategy. The $x$-axis indicates client IDs and the $y$-axis indicates labels. The value in each cell is the number of data samples of a label belonging to that client. For the first row, there are only one possible results in the case where each client owns one label with 10 clients and 10 labels in total, so these three partitions are the same. For the second, third and fourth rows, data heterogeneity increases from left to right.}
	\label{fig:extended dirichlet strategy}
\end{figure}
\textbf{Baseline.} There are two common partition strategies to simulate the heterogeneous settings in the FL literature. According to \cite{li2022federated}, they can be summarized as follows:
\begin{itemize}
	\item Quantity-based class imbalance: Under this strategy, each client is allocated data samples from a fixed number of classes. The initial implementation comes from \cite{mcmahan2017communication}, and has extended by \cite{li2022federated} recently. Specifically, \citeauthor{li2022federated} first randomly assign $C$ different classes to each client. Then, the samples of each class are divided randomly and equally into the clients which owns the class.
	\item Distribution-based class imbalance: Under this strategy, each client is allocated a proportion of the data samples of each class according to Dirichlet distribution. The initial implementation, to the best of our knowledge, comes from \cite{yurochkin2019bayesian}. For each class~$c$, \citeauthor{yurochkin2019bayesian} draw $\vp_c \sim \text{Dir}(\alpha \vq)$ and allocate a $\vp_{c,m}$ proportion of the data samples of class~$c$ to client~$m$. Here $\vq$ is the prior distribution, which is set to $\1$.
\end{itemize}

\begin{figure}[htbp]
	\centering
	\includegraphics[width=0.2\textwidth]{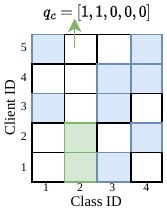}
	\caption{Allocating 2 classes (4 classes in total) to 5 clients.}
	\label{fig:allocate classes}
\end{figure}
\textbf{Extended Dirichlet strategy.} This is to generate arbitrarily heterogeneous data across clients by combining the two strategies above. The difference is to add a step of allocating classes (labels) to determine the number of classes per client (denoted by $C$) before allocating samples via Dirichlet distribution (with concentration parameter $\alpha$). Thus, the extended strategy can be denoted by $\text{ExDir}(C,\alpha)$. The implementation is as follows (one partitioning example is shown in Figure~\ref{fig:extended dirichlet strategy}):
\begin{itemize}
	\item Allocating classes: We randomly allocate $C$ different classes to each client. After assigning the classes, we can obtain the prior distribution $\vq_c$ for each class $c$ (see Figure~\ref{fig:allocate classes}).
	\item Allocating samples: For each class $c$, we draw $\vp_c\sim \text{Dir}(\alpha \vq_c)$ and then allocate a $\vp_{c,m}$ proportion of the samples of class $c$ to client $m$. For example, $\vq_c = [1, 1, 0, 0, \ldots,]$ means that the samples of class $c$ are only allocated to the first 2 clients.
\end{itemize}
This strategy has two levels, the first level to allocate classes and the second level to allocate samples. We note that \cite{reddi2021adaptive} use a two-level partition strategy to partition the CIFAR-100 dataset. They draw a multinomial distribution from the Dirichlet prior at the root ($\text{Dir}(\alpha)$) and a multinomial distribution from the Dirichlet prior at each coarse label ($\text{Dir}(\beta)$).

\subsection{Gradient clipping.}\label{subsec:app:gradient clipping}
\begin{figure}[htbp]
	\centering
	\includegraphics[width=0.245\linewidth]{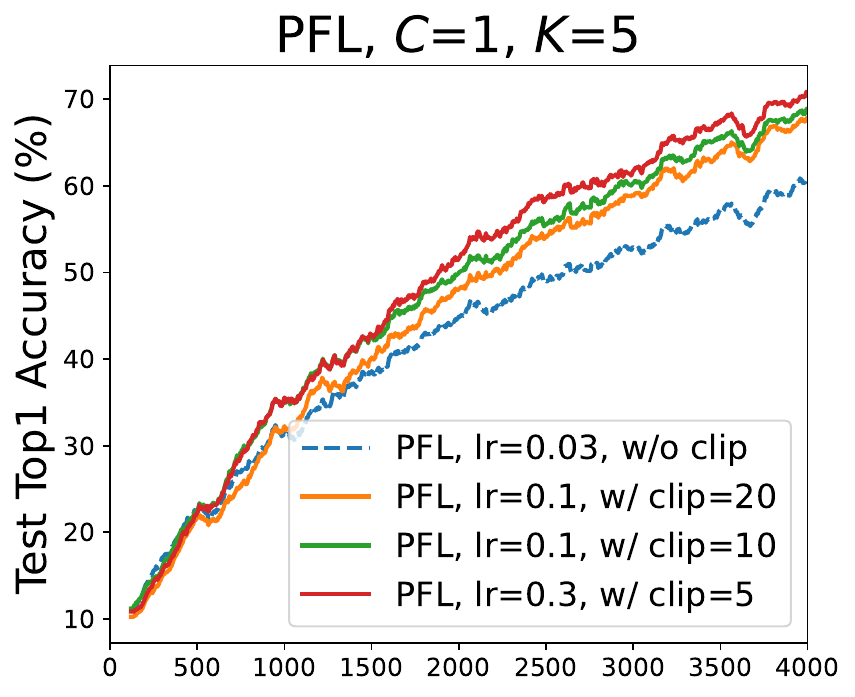}
	\includegraphics[width=0.245\linewidth]{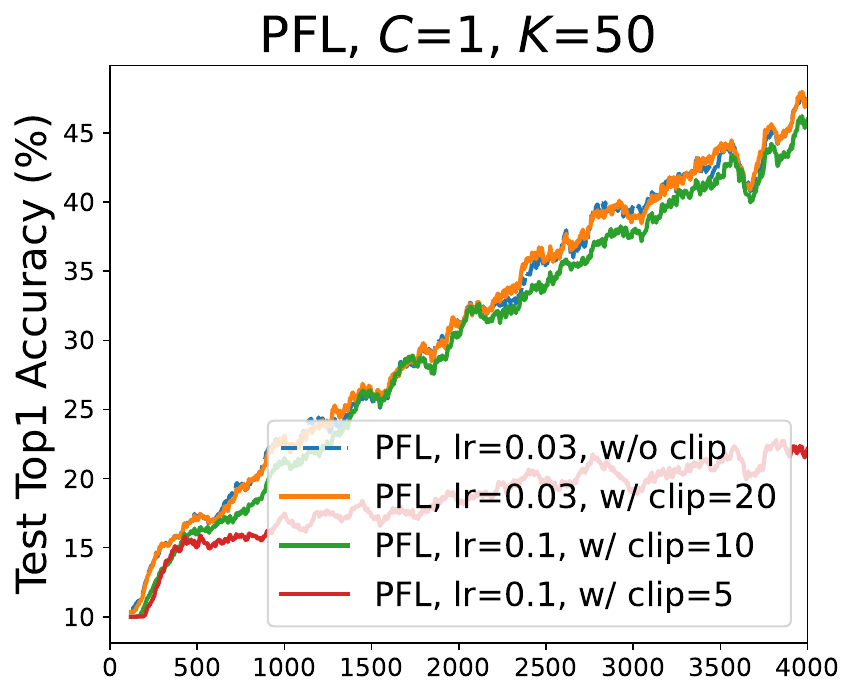}
	\includegraphics[width=0.245\linewidth]{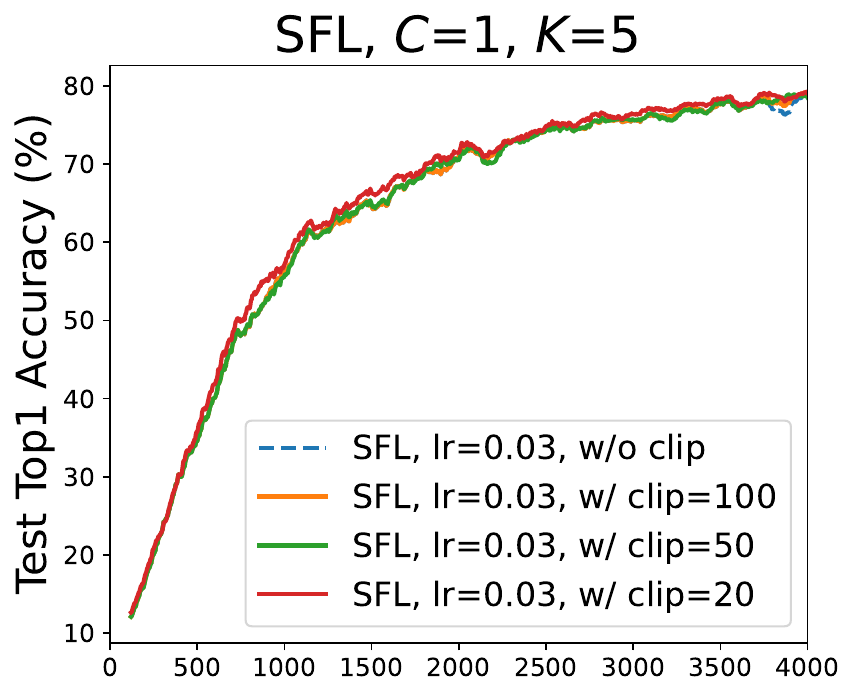}
	\includegraphics[width=0.245\linewidth]{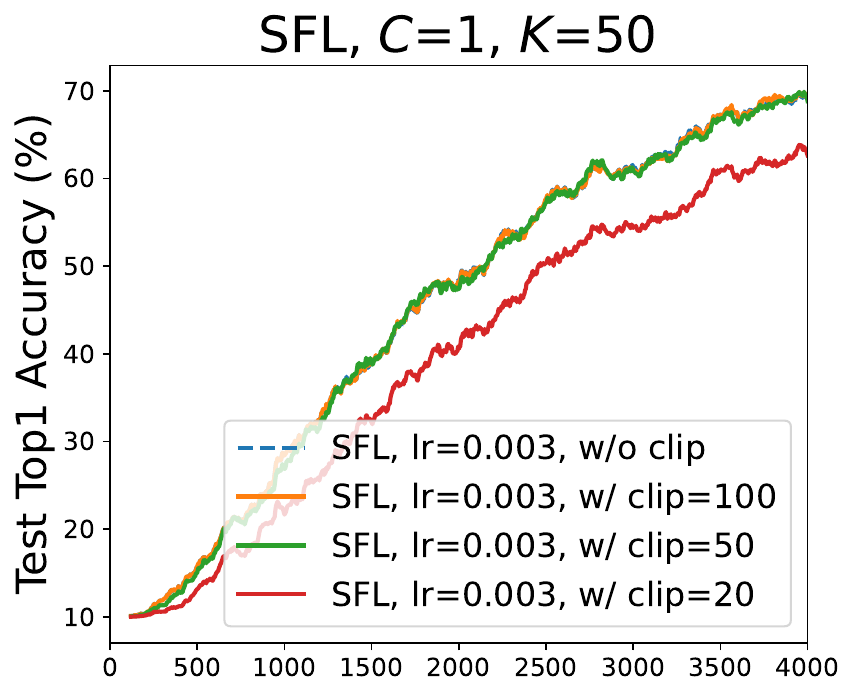}
	\includegraphics[width=0.245\linewidth]{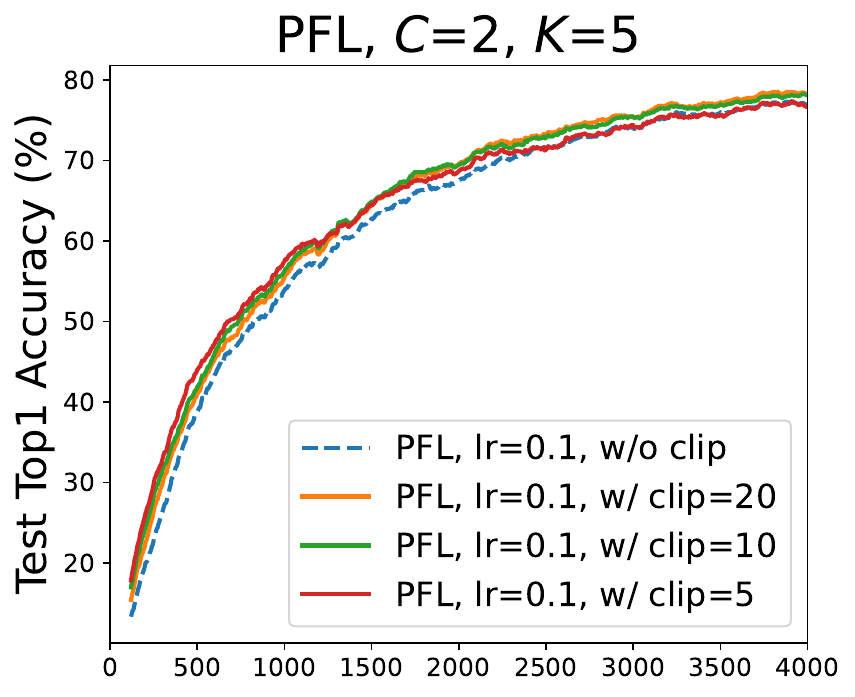}
	\includegraphics[width=0.245\linewidth]{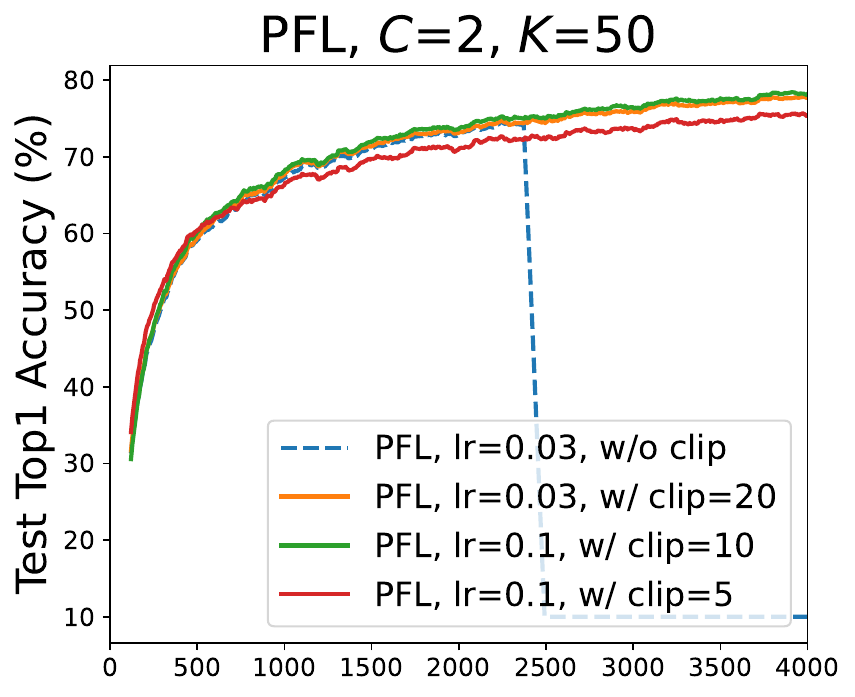}
	\includegraphics[width=0.245\linewidth]{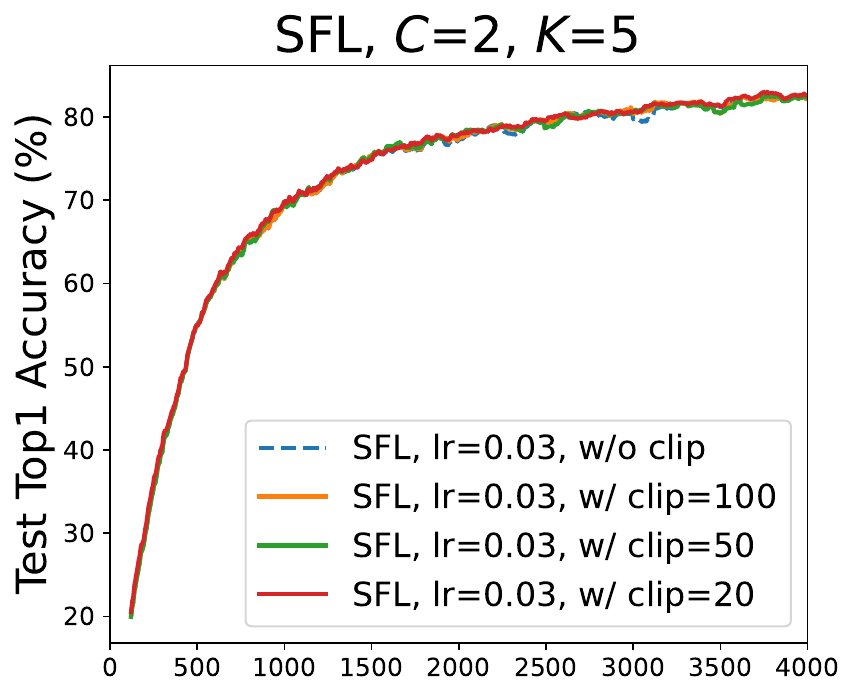}
	\includegraphics[width=0.245\linewidth]{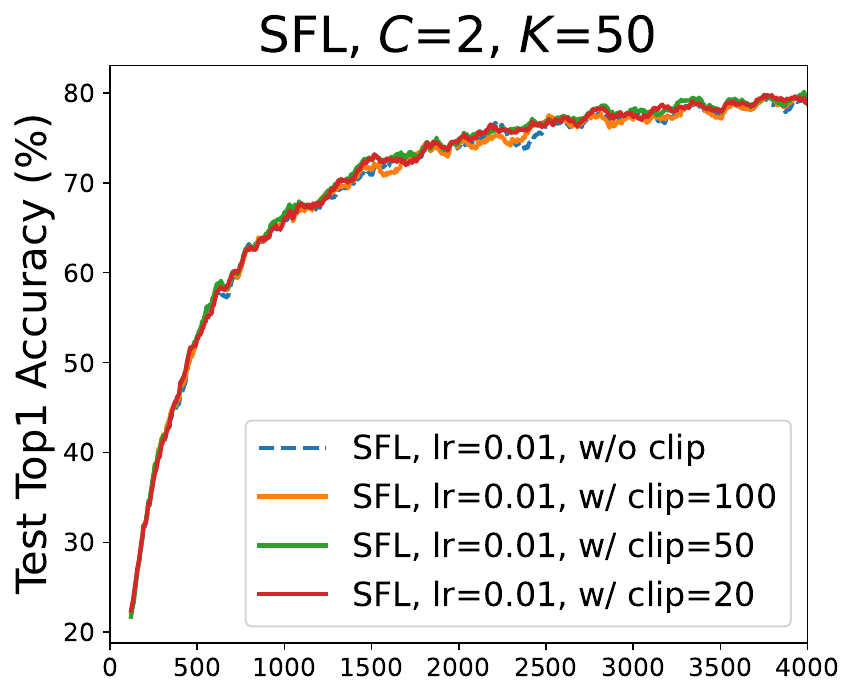}
	\caption{The corresponding training curves of Table~\ref{tab2:app:gradient clipping} (VGG-9 on CIFAR-10).}
	\label{fig:app:gradient clipping}
\end{figure}

Two partitions, the extreme setting $C=1$ (i.e., $\text{ExDir}(1, 10.0)$) and the moderate settings $C=2$ (i.e., $\text{ExDir}(2, 10.0)$) are used in the main body. For both settings, we use the gradient clipping to improve the stability of the algorithms as done in previous works \citep{acar2021federated, jhunjhunwala2023fedexp}. Further, we note that \textit{the gradient clipping is critical for PFL and SFL to prevent divergence} in the learning process on heterogeneous data, especially in the extreme setting. Empirically, we find that the fitting ``max norm'' of SFL is larger than PFL. Thus we trained VGG-9 on CIFAR-10 with PFL and SFL for various values of the max norm of gradient clipping to select the fitting value, and had some empirical observations for gradient clipping in FL. The experimental results are given in Table~\ref{tab:app:gradient clipping}, Table~\ref{tab2:app:gradient clipping} and Figure~\ref{fig:app:gradient clipping}. The empirical observations are summarized as follows:
\begin{enumerate}[leftmargin=2em, label=(\arabic*)]
	\item The fitting max norm of SFL is larger than PFL. When the max norm is set to be 20 for both algorithms, gradient clipping helps improve the performance of PFL, while it may even hurt that of SFL (e.g., see the 12-th row in Table~\ref{tab2:app:gradient clipping}).
	\item The fitting max norm for the case with high data heterogeneity is larger than that with low data heterogeneity. This is suitable for both algorithms. (e.g., see the 12, 24-th rows in Table~\ref{tab2:app:gradient clipping})
	\item The fitting max norm for the case with more local update steps is larger than that with less local update steps. This is suitable for both algorithms. (e.g., see the 4, 8, 12-th rows in Table~\ref{tab2:app:gradient clipping})
	\item Gradient clipping with smaller values of the max norm exhibits a preference for a larger learning rate. This means that using gradient clipping makes the best learning rate larger. This phenomenon is more obvious when choosing smaller values of the max norm (see Table~\ref{tab:app:gradient clipping}).
	\item The fitting max norm is additionally affected by the model architecture, model size, and so on.
\end{enumerate}
Finally, taking into account the experimental results and convenience, we set the max norm of gradient clipping to 10 for PFL and 50 for SFL for all settings in this paper.
\begin{table}[h]
	\centering
	\caption{Test accuracies when using gradient clipping with various values of the max norm for VGG-9 on CIFAR-10. Other settings without being stated explicitly are identical to that in the main body. The results are computed over the last 40 training rounds (with 1000 training rounds in total). The highest test accuracy among different learning rates is marked in \textcolor{cyan}{cyan} for both algorithms.}
	\label{tab:app:gradient clipping}
	\setlength{\tabcolsep}{0.15em}{
		\resizebox{\linewidth}{!}{
			\begin{tabular}{l|p{1.5em}cccc|p{1.5em}cccc}
				\toprule
				Setting &PFL &$10^{-2.0}$ &$10^{-1.5}$  &$10^{-1.0}$ &$10^{-0.5}$ &SFL &$10^{-2.5}$ &$10^{-2.0}$ &$10^{-1.5}$  &$10^{-1.0}$\\\midrule
				VGG-9, $C=1$, $K=5$, w/o clip  &$\infty$ &25.43 &{\color{cyan}30.95} &30.63 &10.00 &$\infty$ &43.93 &53.79 &{\color{cyan}57.69} &10.00\\
				VGG-9, $C=1$, $K=5$, w/ clip   &20 &21.92 &31.04 &{\color{cyan}32.41} &24.99 &100 &43.93 &53.79 &{\color{cyan}57.69} &10.00\\
				VGG-9, $C=1$, $K=5$, w/ clip   &10 &12.51 &25.67 &{\color{cyan}34.89} &28.77 &50 &43.79 &53.73 &{\color{cyan}58.63} &10.00\\
				VGG-9, $C=1$, $K=5$, w/ clip   &5 &10.54 &16.72 &27.01 &{\color{cyan}35.11} &20 &43.12 &53.17 &{\color{cyan}57.96} &10.00\\
				\midrule
				VGG-9, $C=1$, $K=20$, w/o clip &$\infty$ &24.23 &{\color{cyan}27.53} &26.91 &10.00 &$\infty$ &{\color{cyan}35.63} &10.00 &10.00 &10.00\\
				VGG-9, $C=1$, $K=20$, w/ clip  &20 &19.44 &{\color{cyan}27.60} &26.41 &15.00 &100 &{\color{cyan}35.63} &10.00 &10.00 &10.00\\
				VGG-9, $C=1$, $K=20$, w/ clip  &10 &11.51 &22.81 &{\color{cyan}28.79} &21.10 &50 &{\color{cyan}34.55} &27.11 &10.00 &10.00\\
				VGG-9, $C=1$, $K=20$, w/ clip  &5 &10.39 &14.56 &22.48 &{\color{cyan}27.26} &20 &{\color{cyan}30.49} &10.00 &10.00 &10.00\\
				\midrule
				VGG-9, $C=1$, $K=50$, w/o clip &$\infty$ &22.44 &{\color{cyan}23.70} &20.97 &10.00 &$\infty$ &{\color{cyan}25.11} &10.00 &10.00 &10.00\\
				VGG-9, $C=1$, $K=50$, w/ clip  &20 &17.82 &{\color{cyan}23.80} &20.72 &10.00 &100 &{\color{cyan}25.11} &10.00 &10.00 &10.00\\
				VGG-9, $C=1$, $K=50$, w/ clip  &10 &10.14 &20.58 &{\color{cyan}21.95} &10.00 &50 &{\color{cyan}23.41} &10.00 &10.00 &10.00\\
				VGG-9, $C=1$, $K=50$, w/ clip  &5 &10.31 &10.44 &{\color{cyan}18.25} &18.02 &20 &{\color{cyan}18.27} &10.00 &10.00 &10.00\\
				\midrule\midrule
				VGG-9, $C=2$, $K=5$, w/o clip  &$\infty$ &41.36 &51.34 &{\color{cyan}55.22} &10.00 &$\infty$ &58.33 &69.14 &{\color{cyan}71.58} &10.00\\
				VGG-9, $C=2$, $K=5$, w/ clip   &20 &41.46 &51.30 &{\color{cyan}56.54} &47.47 &100 &58.33 &69.14 &{\color{cyan}71.58} &10.00\\
				VGG-9, $C=2$, $K=5$, w/ clip   &10 &38.50 &50.99 &{\color{cyan}57.09} &53.46 &50 &58.35 &68.75 &{\color{cyan}71.24} &10.00\\
				VGG-9, $C=2$, $K=5$, w/ clip  &5 &26.07 &46.49 &{\color{cyan}58.26} &56.71 &20 &57.81 &69.17 &{\color{cyan}70.94} &10.00\\
				\midrule
				VGG-9, $C=2$, $K=20$, w/o clip &$\infty$ &55.64 &{\color{cyan}64.04} &10.00 &10.00 &$\infty$ &60.56 &{\color{cyan}67.70} &64.94 &10.00\\
				VGG-9, $C=2$, $K=20$, w/ clip  &20 &55.70 &64.14 &{\color{cyan}66.57} &10.00 &100 &59.48 &{\color{cyan}68.11} &64.94 &10.00\\
				VGG-9, $C=2$, $K=20$, w/ clip  &10 &54.70 &63.97 &{\color{cyan}68.07} &61.92 &50 &60.82 &{\color{cyan}67.93} &66.50 &10.00\\
				VGG-9, $C=2$, $K=20$, w/ clip  &5 &47.32 &61.61 &{\color{cyan}67.11} &63.56 &20 &58.36 &67.65 &{\color{cyan}67.74} &10.00\\
				\midrule
				VGG-9, $C=2$, $K=50$, w/o clip &$\infty$ &63.93 &{\color{cyan}67.85} &10.00 &10.00 &$\infty$ &61.93 &{\color{cyan}68.05} &61.77 &10.00\\
				VGG-9, $C=2$, $K=50$, w/ clip  &20 &63.94 &{\color{cyan}68.26} &67.15 &59.46 &100 &62.36 &{\color{cyan}67.01} &62.84 &10.00\\
				VGG-9, $C=2$, $K=50$, w/ clip  &10 &62.72 &67.57 &{\color{cyan}69.11} &64.21 &50 &62.27 &{\color{cyan}67.52} &62.59 &10.00\\
				VGG-9, $C=2$, $K=50$, w/ clip  &5 &58.52 &65.34 &{\color{cyan}66.77} &64.75 &20 &59.80 &{\color{cyan}68.72} &64.26 &38.22\\
				\bottomrule
	\end{tabular}}}
\end{table}
\begin{table}[h]
	\centering
	\caption{The best learning rate (selected in Table~\ref{tab:app:gradient clipping}) and its corresponding test accuracies in the short run (1000 training rounds) and in the long run (4000 training rounds). The results are computed over the last 40 training rounds in the short run (the 5-th, 9-th columns) and 100 training rounds in the long run (the 6-th, 10-th columns). That the algorithms diverge when without gradient clipping makes the result with $^\dagger$. The results that deviate from the vanilla case (w/o gradient clipping) considerably (more than 2\%) are marked in {\color{magenta}magenta} and {\color{teal}teal}.}
	\label{tab2:app:gradient clipping}
	\setlength{\tabcolsep}{0.15em}{
		\resizebox{\linewidth}{!}{
			\begin{tabular}{c|l|p{1.5em}cll|p{1.5em}cll}
				\toprule
				&Setting &PFL &Lr &Acc.  &Acc. &SFL &Lr &Acc.  &Acc. \\\midrule
				\multicolumn{10}{c}{\{CIFAR-10, $C=1$\}}\\\midrule
				1 &VGG-9, $K=5$, w/o clip &$\infty$ &$10^{-1.5}$ &30.95 &60.75 &$\infty$ &$10^{-1.5}$ &57.69 &78.56  \\
				2 &VGG-9, $K=5$, w/ clip &20 &$10^{-1.0}$ &32.41 &67.97 {\color{magenta}(+7.2)}  &100 &$10^{-1.5}$ &57.69 &78.75 \\
				3 &VGG-9, $K=5$, w/ clip &10 &$10^{-1.0}$ &34.89 {\color{magenta}(+3.9)}&69.10 {\color{magenta}(+8.4)} &50 &$10^{-1.5}$ &58.63 &78.56 \\
				4 &VGG-9, $K=5$, w/ clip &5 &$10^{-0.5}$ &35.11 {\color{magenta}(+4.2)} &71.07 {\color{magenta}(+10.3)}&20 &$10^{-1.5}$ &57.96 &79.06 \\
				\midrule
				5 &VGG-9, $K=20$, w/o clip &$\infty$ &$10^{-1.5}$ &27.53 &56.89 &$\infty$ &$10^{-2.5}$ &35.63 &72.90 \\
				6 &VGG-9, $K=20$, w/ clip  &20 &$10^{-1.5}$ &27.60 &57.01 &100 &$10^{-2.5}$ &35.63 &73.06\\
				7 &VGG-9, $K=20$, w/ clip &10 &$10^{-1.0}$ &28.79 &64.11 {\color{magenta}(+7.2)} &50 &$10^{-2.5}$ &34.55 &73.16  \\
				8 &VGG-9, $K=20$, w/ clip &5 &$10^{-0.5}$ &27.26 &62.31 {\color{magenta}(+5.4)} &20 &$10^{-2.5}$ &30.49 {\color{teal}(-5.1)} &69.66 {\color{teal}(-3.2)}  \\
				\midrule
				9 &VGG-9, $K=50$, w/o clip  &$\infty$ &$10^{-1.5}$ &23.70 &48.29 &$\infty$ &$10^{-2.5}$ &25.11 &69.10 \\
				10 &VGG-9, $K=50$, w/ clip  &20 &$10^{-1.5}$ &23.80 &47.64 &100 &$10^{-2.5}$ &25.11 &69.01 \\
				11 &VGG-9, $K=50$, w/ clip  &10 &$10^{-1.0}$ &21.95 &46.21 {\color{teal}(-2.1)} &50 &$10^{-2.5}$ &23.41 &68.71 \\
				12 &VGG-9, $K=50$, w/ clip  &5 &$10^{-1.0}$ &18.25 {\color{teal}(-5.5)} &22.58 {\color{teal}(-25.7)} &20 &$10^{-2.5}$ &18.27 {\color{teal}(-6.8)}  &62.70 {\color{teal}(-6.4)}\\
				\midrule
				\multicolumn{10}{c}{\{CIFAR-10, $C=2$\}}\\\midrule
				13 &VGG-9, $K=5$, w/o clip &$\infty$ &$10^{-1.0}$ &55.22 &76.98 &$\infty$ &$10^{-1.5}$ &71.58 &82.09  \\
				14 &VGG-9, $K=5$, w/ clip  &20 &$10^{-1.0}$ &56.54 &78.28 &100 &$10^{-1.5}$ &71.58 &82.17 \\
				15 &VGG-9, $K=5$, w/ clip &10 &$10^{-1.0}$ &57.09 &78.18 &50 &$10^{-1.5}$ &71.24 &82.18  \\
				16 &VGG-9, $K=5$, w/ clip &5 &$10^{-1.0}$ &58.26 {\color{magenta}(+3.0)} &76.69 &20 &$10^{-1.5}$ &70.94 &82.48  \\
				\midrule
				17 &VGG-9, $K=20$, w/o clip &$\infty$ &$10^{-1.5}$ &64.04 &77.21 &$\infty$ &$10^{-2.0}$ &67.70 &81.31 \\
				18 &VGG-9, $K=20$, w/ clip  &20 &$10^{-1.0}$ &66.57 {\color{magenta}(+2.5)} &78.87 &100 &$10^{-2.0}$ &68.11 &82.08\\
				19 &VGG-9, $K=20$, w/ clip &10 &$10^{-1.0}$ &68.07 {\color{magenta}(+4.0)} &78.85 &50 &$10^{-2.0}$ &67.93 &81.50  \\
				20 &VGG-9, $K=20$, w/ clip &5 &$10^{-1.0}$ &67.11 {\color{magenta}(+3.1)} &76.66 &20 &$10^{-1.5}$ &67.74 &77.59 {\color{teal}(-3.7)} \\
				\midrule
				21 &VGG-9, $K=50$, w/o clip  &$\infty$ &$10^{-1.5}$ &67.85 &10.00$^\dagger$ &$\infty$ &$10^{-2.0}$ &68.05 &79.30 \\
				22 &VGG-9, $K=50$, w/ clip  &20 &$10^{-1.5}$ &68.26 &77.83  &100 &$10^{-2.0}$ &67.01 &78.88 \\
				23 &VGG-9, $K=50$, w/ clip  &10 &$10^{-1.0}$ &69.11 &78.13 &50 &$10^{-2.0}$ &67.52 &79.42 \\
				24 &VGG-9, $K=50$, w/ clip  &5 &$10^{-1.0}$ &66.77 &75.42 &20 &$10^{-2.0}$ &68.72 &78.51 \\
				\bottomrule
	\end{tabular}}}
\end{table}
\clearpage
\subsection{Grid search}\label{subsec:app:grid search}
We use the grid search to find the best learning rate on one random seed ``1234''. Since we have observed that the best learning rate of SFL is smaller than PFL empirically, the grid for PFL is $\{10^{-2.0}, 10^{-1.5}, 10^{-1.0}, 10^{-0.5}\}$ (\{0.01, 0.03, 0.1, 0.3\} in fact) and the grid for SFL is $\{10^{-2.5}, 10^{-2.0}, 10^{-1.5}, 10^{-1.0}\}$ (\{0.003, 0.01, 0.03, 0.1\} in fact). We use these grids for all tasks in this paper, including MNIST and FMNIST in the next subsection.
One practical method used in \cite{jhunjhunwala2023fedexp} to find the best learning rate is running the algorithms for a fewer training rounds and then comparing the short-run results by some metrics (e.g.,  training accuracy) when the computation resources are restrictive and the task is complex (e.g., CIFAR-10). However, we should pay attention to whether the chosen learning rates are appropriate in the specific scenarios, as the best learning rate in the short run may not be the best in the long run. One alternative method is using the short-run results to find some alternatives (coarse-grained search) and then using the long-run results to find the best one (fine-grained search).
In this paper, for CIFAR-10 and CINIC-10, we run the algorithms for 1000 training rounds to find the candidate learning rates (with a less than 3\% difference to the best result in test accuracy), called as coarse-grained search; and then run the algorithms for 4000 training rounds with the candidate learning rates to find the best learning rate (with the highest test accuracy), called as fine-grained search. The max norm of gradient clipping is set to 10 for PFL and 50 for SFL for all settings (see subsection~\ref{subsec:app:gradient clipping}). Other hyperparameters are identical to that in the main body.

\begin{figure}[htbp]
	\vspace{-2ex}
	\centering
	\includegraphics[width=0.325\linewidth]{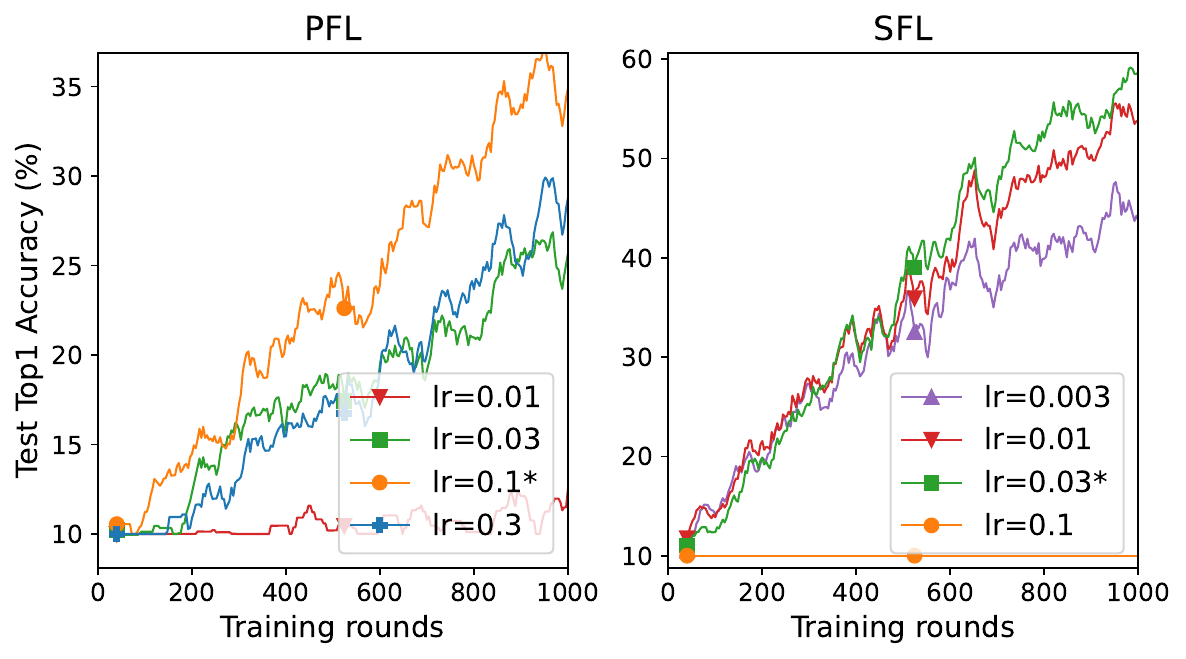}
	\includegraphics[width=0.325\linewidth]{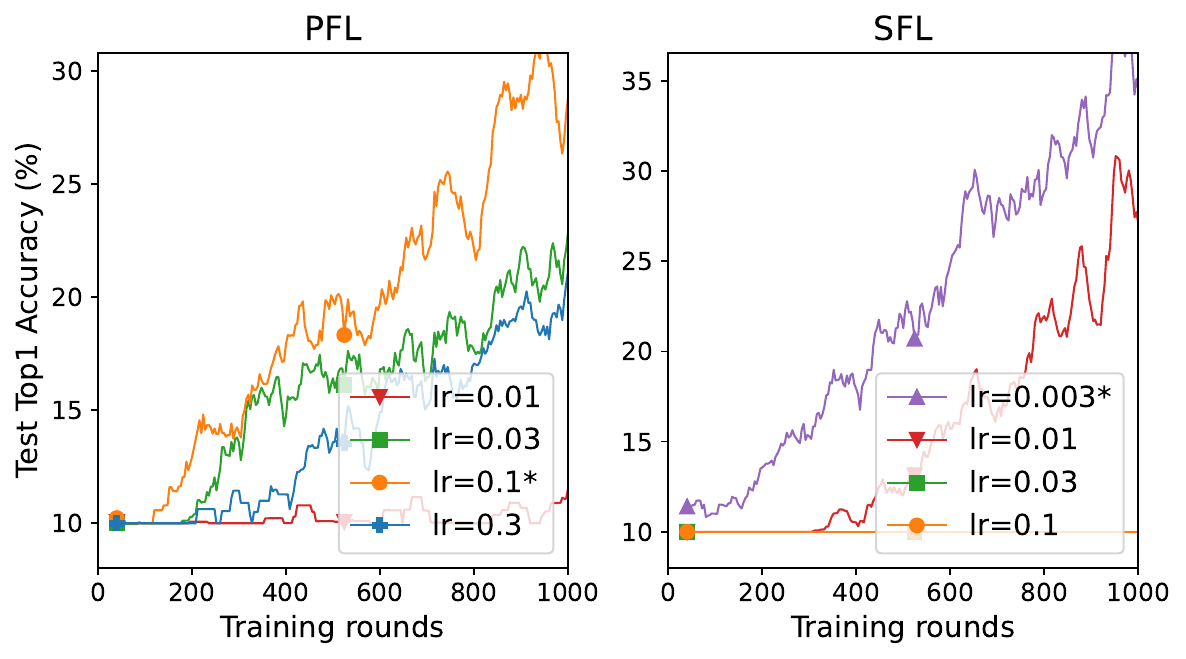}
	\includegraphics[width=0.325\linewidth]{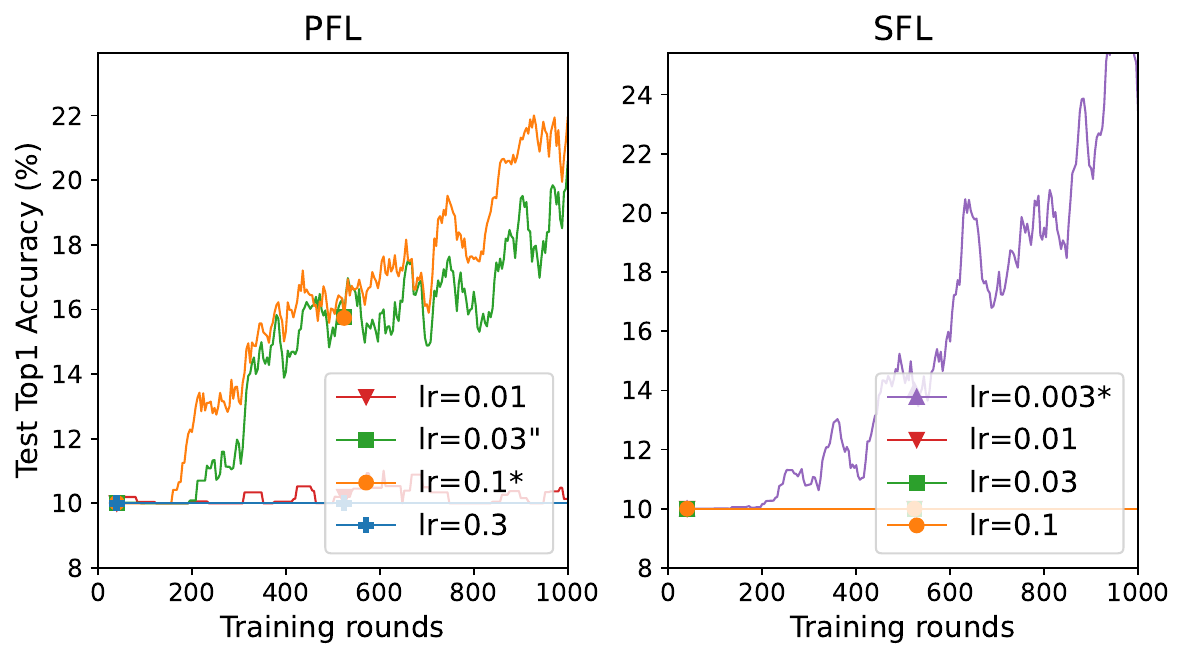}
	\includegraphics[width=0.325\linewidth]{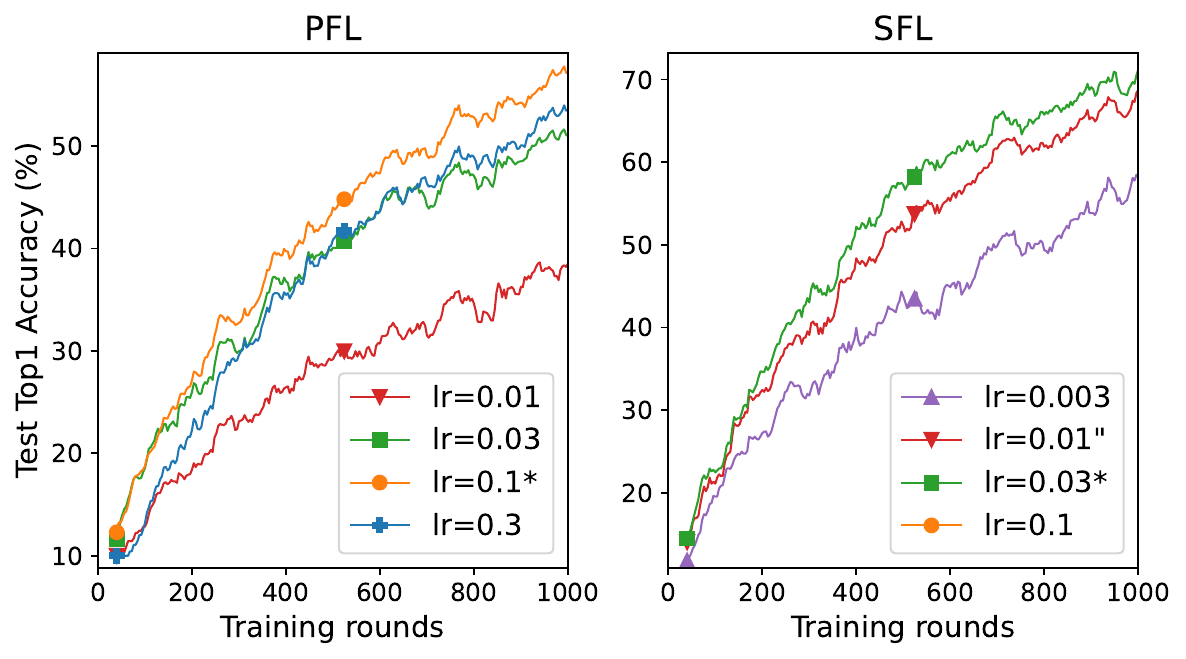}
	\includegraphics[width=0.325\linewidth]{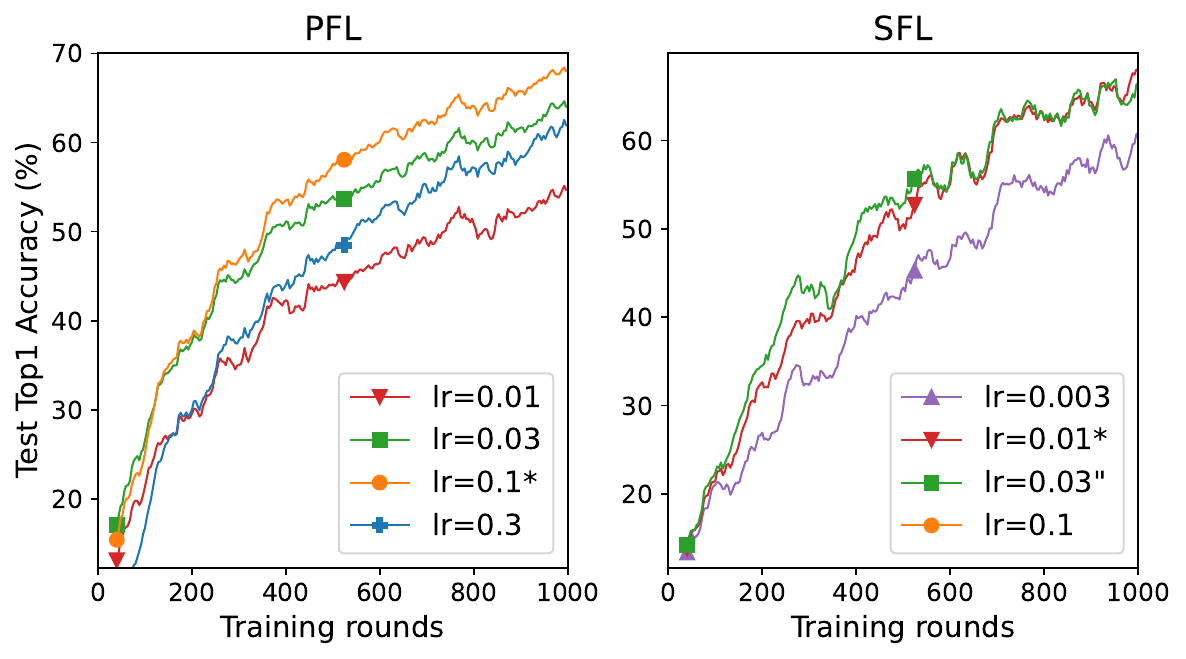}
	\includegraphics[width=0.325\linewidth]{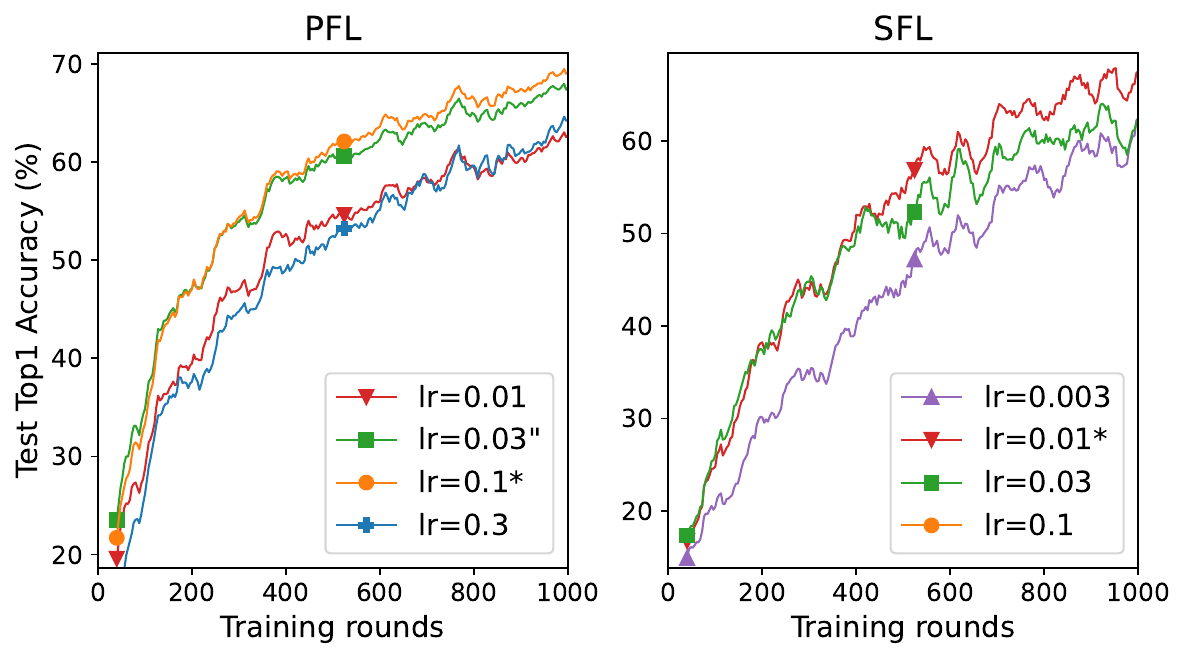}
	\caption{The corresponding training curves of Table~\ref{tab:app:fine-grained search}. We mark the best learning rate in the short run with ``*'' and other candidate learning rates with ``"'' in the legend. The top row shows the first three settings and the bottom row shows the last three settings in Table~\ref{tab:app:fine-grained search}.}
	\label{fig:app:fine-grained search}
\end{figure}

\begin{figure}[htbp]
	\vspace{-5ex}
	\centering
	\begin{subfigure}{\linewidth}
		\centering
		\includegraphics[width=0.3\linewidth]{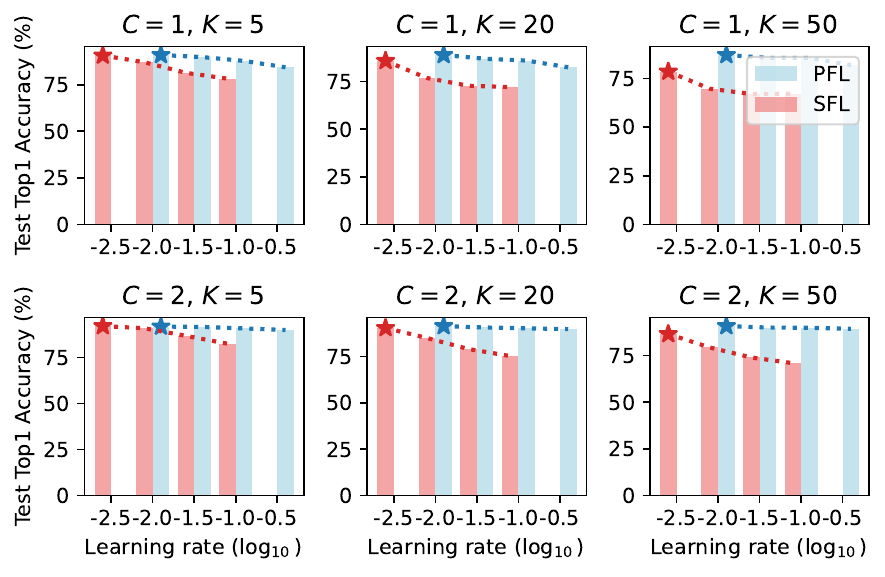}\hspace{1em}
		\includegraphics[width=0.3\linewidth]{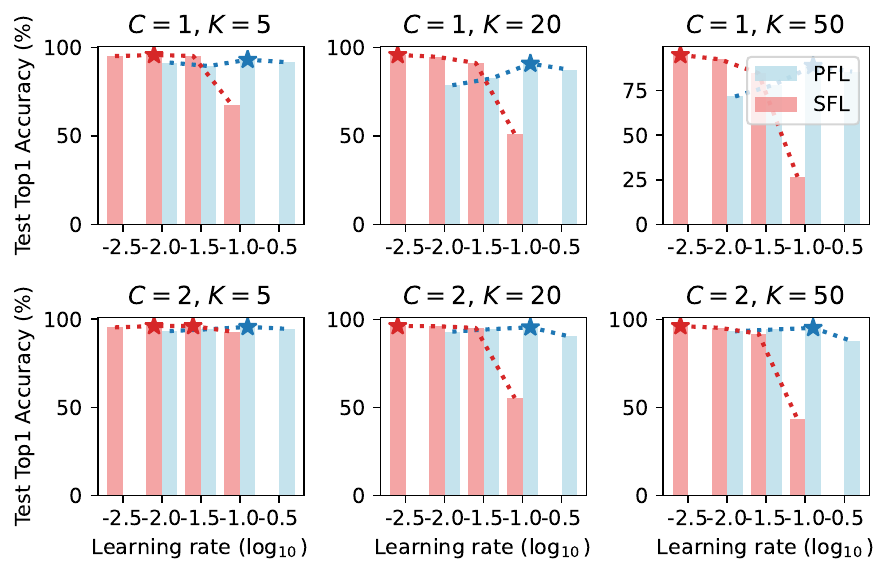}\hspace{1em}
		\includegraphics[width=0.3\linewidth]{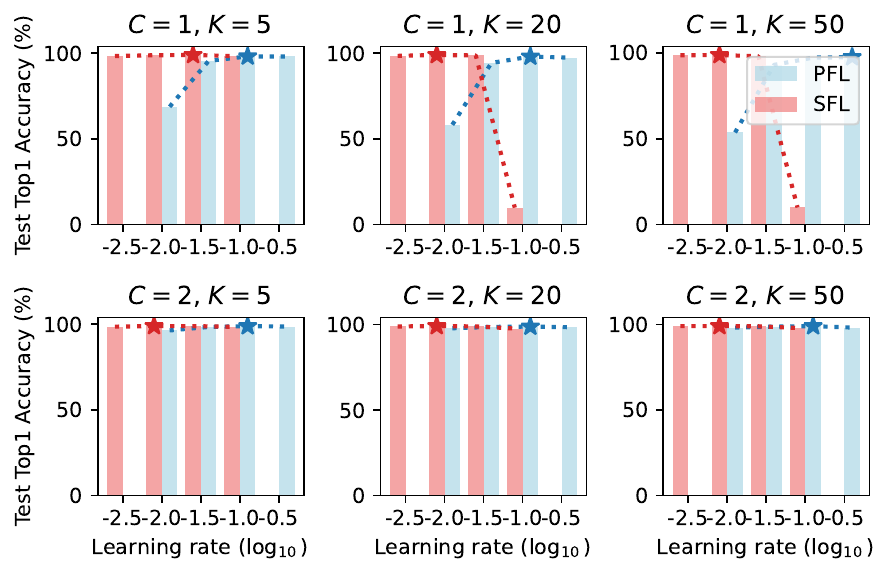}
		\caption{Left: Logistic/MNIST. Middle: MLP/MNIST. Right: LeNet-5/MNIST}
	\end{subfigure}
	\begin{subfigure}{\linewidth}
		\centering
		\includegraphics[width=0.3\linewidth]{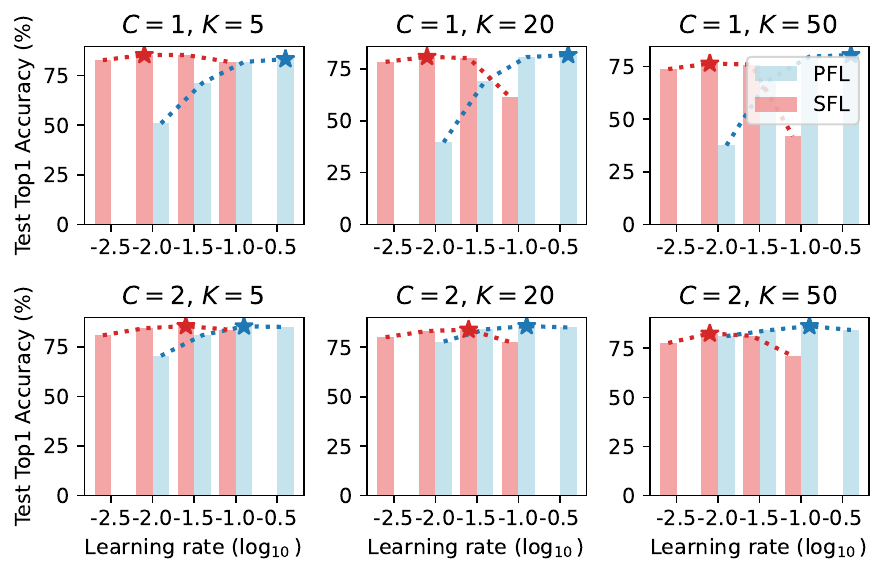}
		\hspace{1em}
		\includegraphics[width=0.3\linewidth]{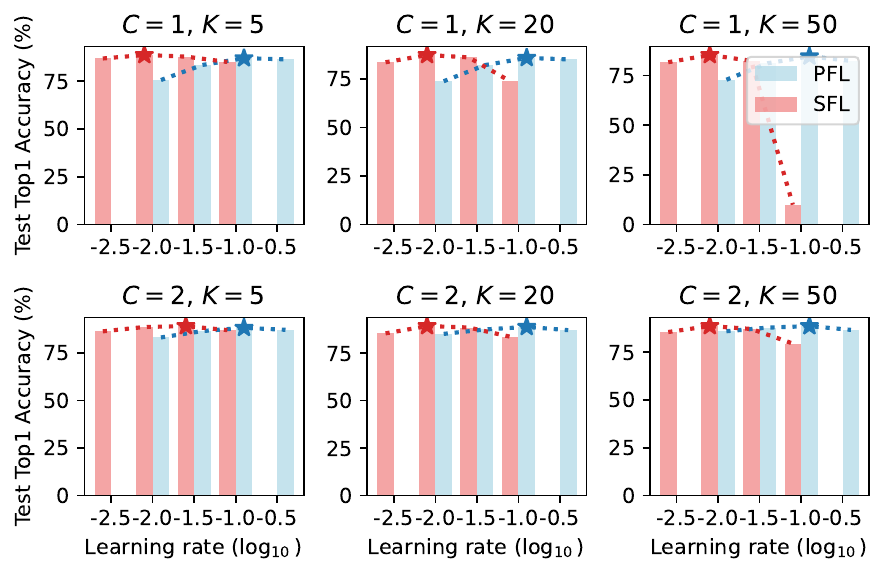}
		\caption{Left: LeNet-5/FMNIST. Right: CNN/FMNIST}
	\end{subfigure}
	\begin{subfigure}{\linewidth}
		\centering
		\includegraphics[width=0.3\linewidth]{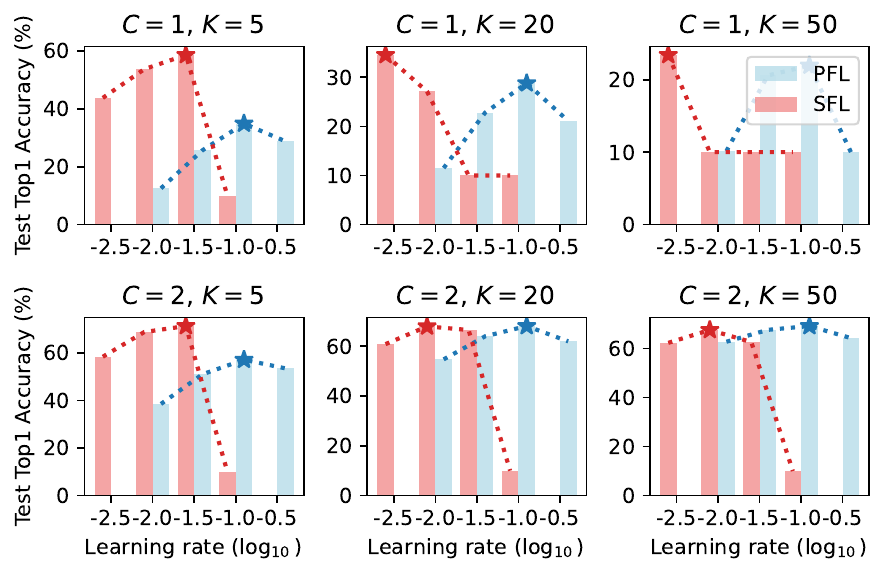}
		\hspace{1em}
		\includegraphics[width=0.3\linewidth]{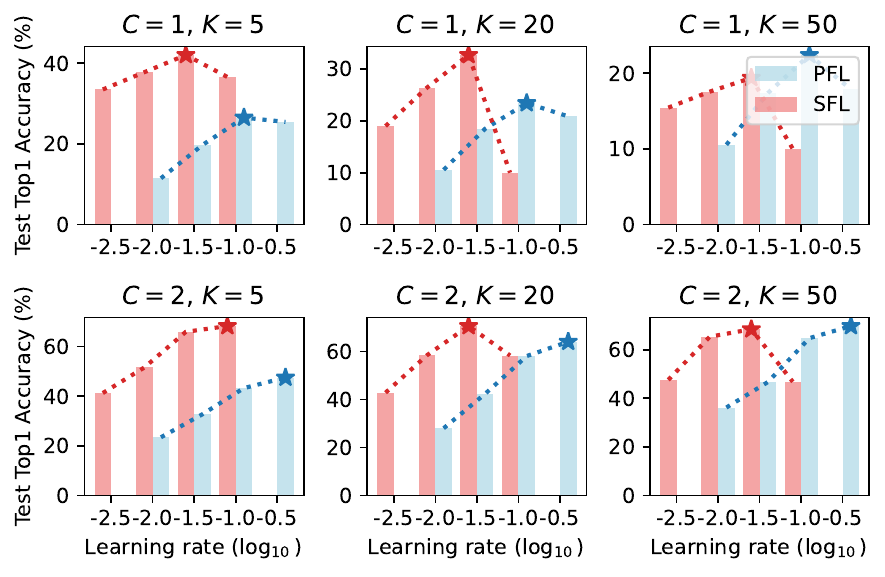}
		\caption{Left: VGG-9/CIFAR-10. Right: ResNet-18/CIFAR-10}
	\end{subfigure}
	\begin{subfigure}{\linewidth}
		\centering
		\includegraphics[width=0.3\linewidth]{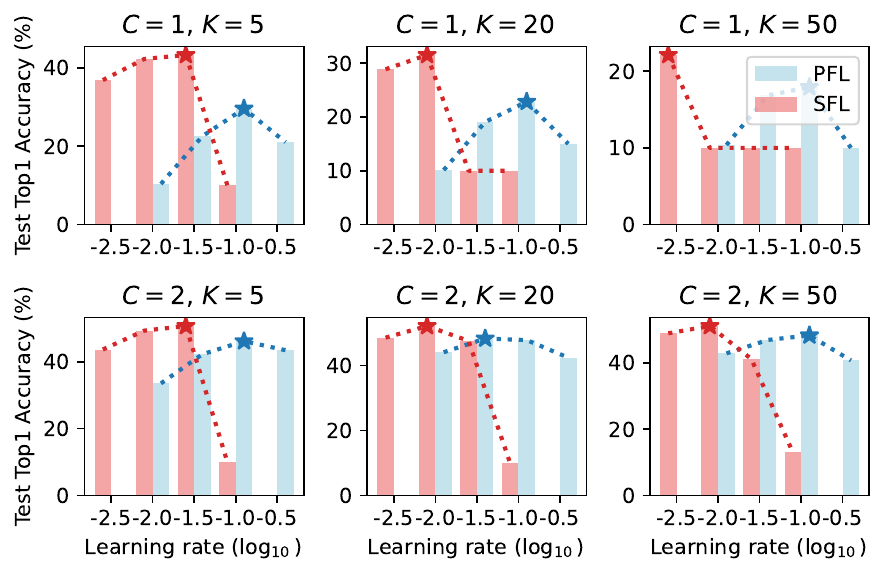}
		\hspace{1em}
		\includegraphics[width=0.3\linewidth]{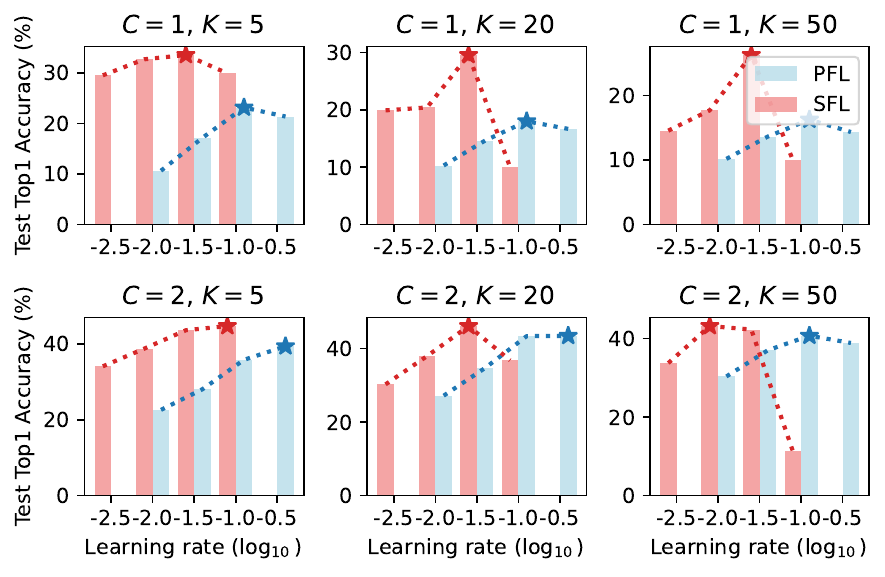}
		\caption{Left: VGG-9/CINIC-10. Right: ResNet-18/CINIC-10}
	\end{subfigure}
	\caption{Test accuracies after training for 1000 rounds for various settings. Details are in Table~\ref{tab:app:coarse-grained search}.}
	\label{fig:app:best learning rate}	
	\vspace{-1ex}
\end{figure}

The results of the coarse-grained search are collected in Figure~\ref{fig:app:best learning rate}, Table~\ref{tab:app:coarse-grained search}. Taking the setting training VGG-9 on CIFAR-10 as an example. We first find the candidate learning rates, whose short-run test accuracies are only 3\% or less below the best accuracy. The candidate learning rates are summarized in Table~\ref{tab:app:fine-grained search}. The training curves are in Figure~\ref{fig:app:fine-grained search}. We then find the best learning rate, whose long-run test accuracy is the highest among the candidate learning rates. The final best learning rates are in Table~\ref{tab:app:fine-grained search}. For fine-grained search of other settings, please refer to the code.
\begin{table}[h!]
	\renewcommand{\arraystretch}{1}
	\centering
	\caption{Best learning rates found by the fine-grained search. The candidate learning rates are collected in the cell and the correspond long-run test accuracies are in the parentheses. According to the long-run test accuracies, we keep the best learning rate and cross out the others.}
	\label{tab:app:fine-grained search}
	\resizebox{\linewidth}{!}{
		\begin{tabular}{lll}
			\toprule
			Settings &PFL &SFL \\
			\midrule
			CIFAR-10, VGG-9, $C=1$, $K=5$ &0.1 &0.03 \\
			CIFAR-10, VGG-9, $C=1$, $K=20$ &0.1 &0.003  \\
			CIFAR-10, VGG-9, $C=1$, $K=50$ &\st{0.03 (28.72)}, 0.1 (46.21) &0.003 \\
			CIFAR-10, VGG-9, $C=2$, $K=5$ &0.1 &\st{0.01 (82.05)}, 0.03 (82.18) \\
			CIFAR-10, VGG-9, $C=2$, $K=20$ &0.1 &0.01 (81.50), \st{0.03 (78.35)}  \\
			CIFAR-10, VGG-9, $C=2$, $K=50$ &\st{0.03 (76.14)}, 0.1 (78.13) &0.01 \\
			\bottomrule
	\end{tabular}}
\end{table}
\begin{table}[h]
	\centering
	\caption{Test accuracy results of grid searches for various settings. The results are computed over the last 40 training rounds (with 1000 training rounds in total). The highest test accuracy among different learning rates is marked in \textcolor{cyan}{cyan} for both algorithms.}
	\label{tab:app:coarse-grained search}
	\setlength{\tabcolsep}{0.45em}{
		\resizebox{\linewidth}{!}{
			\begin{tabular}{l|cccc|cccc}
				\toprule
				\multirow{2}{*}{Setting} &\multicolumn{4}{c}{PFL} &\multicolumn{4}{c}{SFL}\\\cmidrule(lr){2-5}\cmidrule(lr){6-9}
				&$10^{-2.0}$ &$10^{-1.5}$  &$10^{-1.0}$ &$10^{-0.5}$ &$10^{-2.5}$ &$10^{-2.0}$ &$10^{-1.5}$  &$10^{-1.0}$\\
				\midrule
				MNIST, Logistic, $C=1$, $K=5$ &{\color{cyan}90.95} &89.81 &87.78 &84.33 &{\color{cyan}90.68} &87.05 &81.38 &78.27\\
				MNIST, Logistic, $C=1$, $K=20$ &{\color{cyan}88.98} &87.00 &86.06 &82.49 &{\color{cyan}85.87} &76.88 &72.82 &72.14\\
				MNIST, Logistic, $C=1$, $K=50$ &{\color{cyan}86.99} &85.50 &85.52 &81.77 &{\color{cyan}78.45} &69.67 &67.00 &67.09\\
				MNIST, Logistic, $C=2$, $K=5$ &{\color{cyan}91.72} &91.48 &90.70 &89.84 &{\color{cyan}91.97} &90.79 &86.68 &82.43\\
				MNIST, Logistic, $C=2$, $K=20$ &{\color{cyan}91.31} &90.54 &90.10 &89.66 &{\color{cyan}90.29} &84.86 &78.94 &75.18\\
				MNIST, Logistic, $C=2$, $K=50$ &{\color{cyan}90.69} &89.89 &89.78 &89.20 &{\color{cyan}86.45} &79.24 &73.89 &70.85\\
				\midrule
				MNIST, MLP, $C=1$, $K=5$ &91.47 &89.51 &{\color{cyan}93.06} &91.68 &95.13 &{\color{cyan}95.74} &95.22 &67.44\\
				MNIST, MLP, $C=1$, $K=20$ &78.70 &82.53 &{\color{cyan}90.89} &87.46 &{\color{cyan}95.79} &94.79 &91.24 &51.06\\
				MNIST, MLP, $C=1$, $K=50$ &71.61 &78.20 &{\color{cyan}88.73} &85.52 &{\color{cyan}94.83} &92.46 &84.62 &26.71\\
				MNIST, MLP, $C=2$, $K=5$ &93.32 &94.35 &{\color{cyan}95.66} &94.58 &95.40 &{\color{cyan}96.14} &{\color{cyan}96.14} &93.00\\
				MNIST, MLP, $C=2$, $K=20$ &93.18 &94.59 &{\color{cyan}95.62} &90.42 &{\color{cyan}96.33} &96.25 &95.24 &55.47\\
				MNIST, MLP, $C=2$, $K=50$ &93.40 &94.40 &{\color{cyan}95.28} &87.78 &{\color{cyan}96.34} &95.36 &91.99 &43.36\\
				\midrule
				MNIST, LeNet-5, $C=1$, $K=5$ &68.72 &95.46 &{\color{cyan}98.07} &98.05 &98.28 &98.76 &{\color{cyan}98.91} &98.29\\
				MNIST, LeNet-5, $C=1$, $K=20$ &58.20 &94.30 &{\color{cyan}98.02} &97.41 &98.43 &{\color{cyan}98.94} &98.68 &9.84\\
				MNIST, LeNet-5, $C=1$, $K=50$ &53.61 &93.00 &97.23 &{\color{cyan}97.49} &98.60 &{\color{cyan}98.77} &97.98 &9.95\\
				MNIST, LeNet-5, $C=2$, $K=5$ &96.40 &98.43 &{\color{cyan}98.90} &98.67 &98.62 &{\color{cyan}98.95} &98.91 &98.55\\
				MNIST, LeNet-5, $C=2$, $K=20$ &97.88 &98.69 &{\color{cyan}98.86} &98.50 &98.94 &{\color{cyan}99.17} &98.96 &97.74\\
				MNIST, LeNet-5, $C=2$, $K=50$ &98.07 &98.65 &{\color{cyan}98.93} &98.12 &98.95 &{\color{cyan}98.99} &98.85 &97.88\\
				\midrule\midrule
				FMNIST, LeNet-5, $C=1$, $K=5$ &50.88 &71.24 &81.83 &{\color{cyan}83.26} &82.76 &{\color{cyan}85.34} &85.31 &82.02\\
				FMNIST, LeNet-5, $C=1$, $K=20$ &39.64 &69.26 &80.73 &{\color{cyan}81.74} &78.44 &{\color{cyan}80.84} &80.24 &61.50\\
				FMNIST, LeNet-5, $C=1$, $K=50$ &37.56 &68.40 &79.73 &{\color{cyan}80.53} &73.84 &{\color{cyan}76.51} &75.95 &42.07\\
				FMNIST, LeNet-5, $C=2$, $K=5$ &70.59 &81.00 &{\color{cyan}85.44} &85.16 &80.94 &84.51 &{\color{cyan}85.59} &83.83\\
				FMNIST, LeNet-5, $C=2$, $K=20$ &77.85 &84.00 &{\color{cyan}85.70} &85.00 &80.09 &83.20 &{\color{cyan}83.88} &77.65\\
				FMNIST, LeNet-5, $C=2$, $K=50$ &81.24 &84.17 &{\color{cyan}86.33} &84.36 &77.73 &{\color{cyan}82.53} &81.39 &71.13\\
				\midrule
				FMNIST, CNN, $C=1$, $K=5$ &75.57 &83.44 &{\color{cyan}86.98} &86.40 &86.83 &{\color{cyan}88.66} &87.82 &85.01\\
				FMNIST, CNN, $C=1$, $K=20$ &73.84 &82.17 &{\color{cyan}85.86} &85.11 &83.62 &{\color{cyan}87.37} &86.14 &73.85\\
				FMNIST, CNN, $C=1$, $K=50$ &72.94 &81.26 &{\color{cyan}84.67} &82.41 &81.75 &{\color{cyan}85.39} &82.42 &10.00\\
				FMNIST, CNN, $C=2$, $K=5$ &83.20 &86.58 &{\color{cyan}88.26} &87.24 &86.41 &88.71 &{\color{cyan}89.19} &87.06\\
				FMNIST, CNN, $C=2$, $K=20$ &85.12 &87.46 &{\color{cyan}88.80} &87.08 &85.39 &{\color{cyan}89.24} &88.48 &83.65\\
				FMNIST, CNN, $C=2$, $K=50$ &85.93 &87.82 &{\color{cyan}88.61} &86.64 &85.53 &{\color{cyan}88.63} &87.12 &79.50\\
				\midrule\midrule
				CIFAR-10, VGG-9, $C=1$, $K=5$ &12.51 &25.67 &{\color{cyan}34.89} &28.77 &43.79 &53.73 &{\color{cyan}58.63} &10.00\\
				CIFAR-10, VGG-9, $C=1$, $K=20$&11.51 &22.81 &{\color{cyan}28.79} &21.10 &{\color{cyan}34.55} &27.11 &10.00 &10.00\\
				CIFAR-10, VGG-9, $C=1$, $K=50$&10.14 &20.58 &{\color{cyan}21.95} &10.00 &{\color{cyan}23.41} &10.00 &10.00 &10.00\\
				CIFAR-10, VGG-9, $C=2$, $K=5$ &38.50 &50.99 &{\color{cyan}57.09} &53.46 &58.35 &68.75 &{\color{cyan}71.24} &10.00\\
				CIFAR-10, VGG-9, $C=2$, $K=20$&54.70 &63.97 &{\color{cyan}68.07} &61.92 &60.82 &{\color{cyan}67.93} &66.50 &10.00\\
				CIFAR-10, VGG-9, $C=2$, $K=50$&62.72 &67.57 &{\color{cyan}69.11} &64.21 &62.27 &{\color{cyan}67.52} &62.59 &10.00\\
				\midrule
				CIFAR-10, ResNet-18, $C=1$, $K=5$ &11.46 &19.72 &{\color{cyan}26.50} &25.45 &33.49 &37.77 &{\color{cyan}42.04} &36.59\\
				CIFAR-10, ResNet-18, $C=1$, $K=20$&10.56 &18.48 &{\color{cyan}23.41} &20.86 &18.99 &26.41 &{\color{cyan}32.70} &10.00\\
				CIFAR-10, ResNet-18, $C=1$, $K=50$&10.55 &17.39 &{\color{cyan}22.42} &17.86 &15.46 &17.57 &{\color{cyan}19.39} &10.00\\
				CIFAR-10, ResNet-18, $C=2$, $K=5$ &23.45 &32.71 &43.24 &{\color{cyan}47.45} &41.24 &51.51 &65.86 &{\color{cyan}68.17}\\
				CIFAR-10, ResNet-18, $C=2$, $K=20$&28.02 &42.36 &58.07 &{\color{cyan}64.03} &42.83 &58.43 &{\color{cyan}70.55} &58.29\\
				CIFAR-10, ResNet-18, $C=2$, $K=50$&36.22 &46.69 &64.75 &{\color{cyan}69.71} &47.57 &65.33 &{\color{cyan}68.40} &46.80\\
				\midrule\midrule
				CINIC-10, VGG-9, $C=1$, $K=5$ &10.36 &22.56 &{\color{cyan}29.58} &21.02 &36.87 &42.26 &{\color{cyan}43.27} &10.00\\
				CINIC-10, VGG-9, $C=1$, $K=20$&10.13 &19.09 &{\color{cyan}22.83} &15.00 &28.88 &{\color{cyan}31.54} &10.00 &10.00\\
				CINIC-10, VGG-9, $C=1$, $K=50$&10.37 &16.85 &{\color{cyan}17.92} &10.00 &{\color{cyan}22.12} &10.00 &10.00 &10.00\\
				CINIC-10, VGG-9, $C=2$, $K=5$ &33.52 &42.24 &{\color{cyan}46.12} &43.43 &43.64 &49.28 &{\color{cyan}50.64} &10.00\\
				CINIC-10, VGG-9, $C=2$, $K=20$&44.04 &{\color{cyan}48.26} &47.80 &42.40 &48.54 &{\color{cyan}52.10} &47.55 &10.00\\
				CINIC-10, VGG-9, $C=2$, $K=50$&42.87 &46.93 &{\color{cyan}48.29} &40.75 &48.93 &{\color{cyan}51.11} &41.19 &13.04\\
				\midrule
				CINIC-10, ResNet-18, $C=1$, $K=5$ &10.64 &17.16 &{\color{cyan}23.15} &21.37 &29.52 &32.70 &{\color{cyan}33.55} &29.92\\
				CINIC-10, ResNet-18, $C=1$, $K=20$&10.29 &14.65 &{\color{cyan}18.01} &16.68 &19.91 &20.42 &{\color{cyan}29.57} &10.00\\
				CINIC-10, ResNet-18, $C=1$, $K=50$&10.18 &13.65 &{\color{cyan}16.27} &14.35 &14.56 &17.72 &{\color{cyan}26.30} &10.00\\
				CINIC-10, ResNet-18, $C=2$, $K=5$ &22.59 &28.03 &35.78 &{\color{cyan}39.35} &34.06 &38.73 &43.57 &{\color{cyan}44.67}\\
				CINIC-10, ResNet-18, $C=2$, $K=20$&27.18 &34.74 &43.40 &{\color{cyan}43.46} &30.27 &37.96 &{\color{cyan}46.13} &36.81\\
				CINIC-10, ResNet-18, $C=2$, $K=50$&30.36 &37.05 &{\color{cyan}40.74} &38.90 &33.77 &{\color{cyan}43.20} &42.32 &11.35\\
				\bottomrule
	\end{tabular}}}
\end{table}
\subsection{More experimental results}\label{subsec:app:more results}
More experimental results are provided in this subsection. These include results on MNIST \citep{lecun1998gradient}, FMNIST \citep{xiao2017fashion} and additional results on CIFAR-10 and CINIC-10.

\textbf{Setup on MNIST and FMNIST.} We consider five additional tasks: 1) training Logistic Regression on MNIST , 2) training Multi-Layer Perceptron (MLP) on MNIST, 3) training LeNet-5 \citep{lecun1998gradient} on MNIST, 4) training LeNet-5 on FMNIST, 5) training CNN on FMNIST. We partition the training sets of both MNIST and FMNIST into 500 clients by extended Dirichlet strategy $C=1$ and $C=2$ (with $\alpha=10.0$) and spare the test sets for computing test accuracy. We apply gradient clipping with the max norm of 10 for PFL and 50 for SFL. We find the best learning rate by grid search. This is done by running algorithms for 1000 training rounds and choosing the learning rate that achieves the highest test accuracy averaged over the last 40 training rounds. Note that since tasks on MNIST/FMNIST are quite simpler than that on CIFAR-10/CINIC-10, we do not use the coarse, fine-grained search. The results of grid search are given in Table~\ref{tab:app:coarse-grained search}. Other hyperparameters without being stated explicitly are identical to that of CIFAR-10/CINIC-10 in the main body.

\begin{figure}[htbp]
	\centering
	\includegraphics[width=\linewidth]{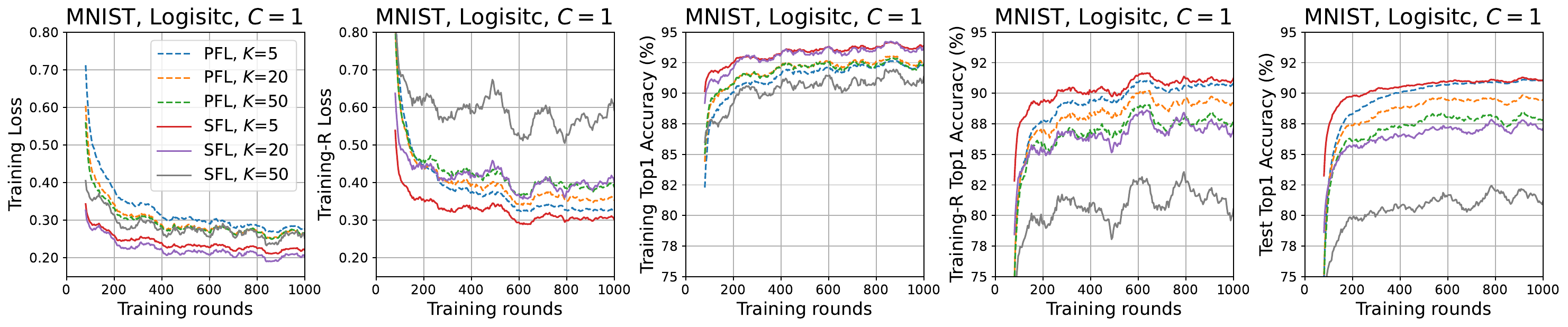}
	\includegraphics[width=\linewidth]{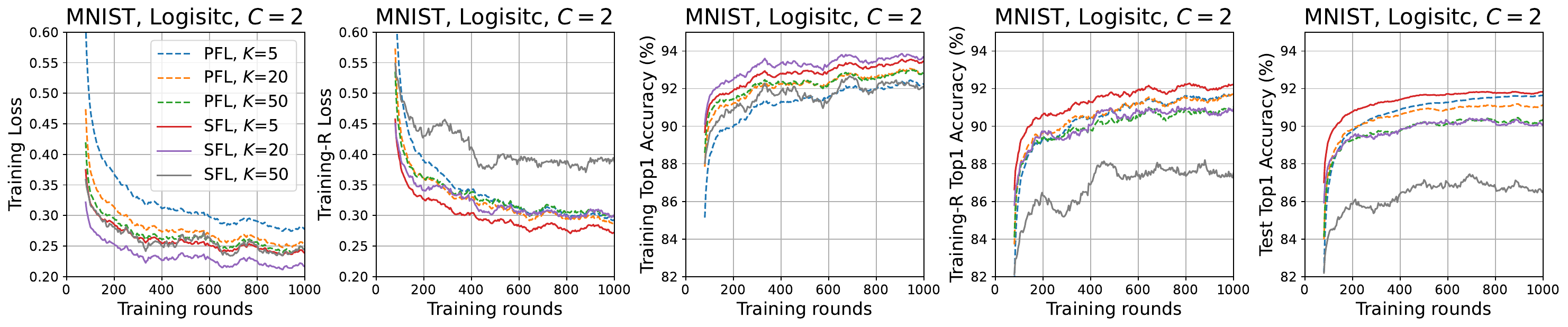}
	\includegraphics[width=\linewidth]{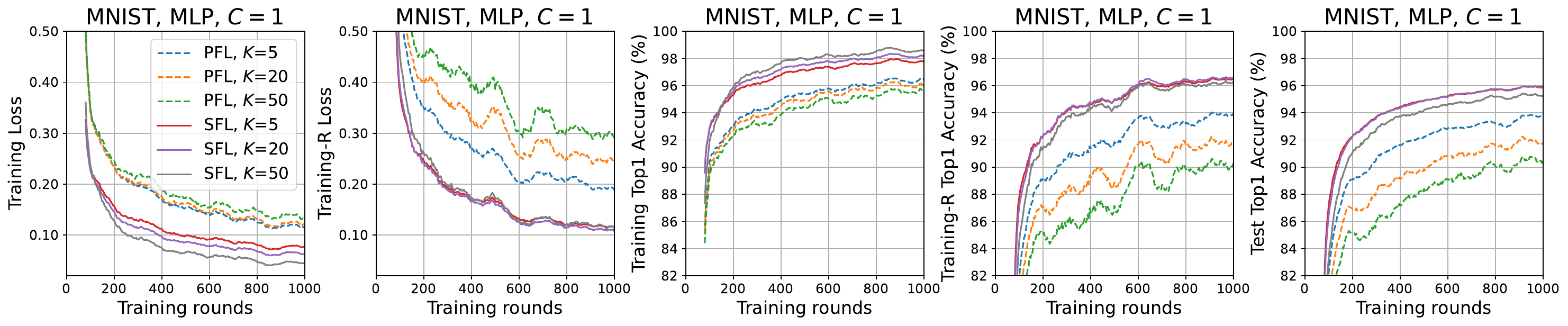}
	\includegraphics[width=\linewidth]{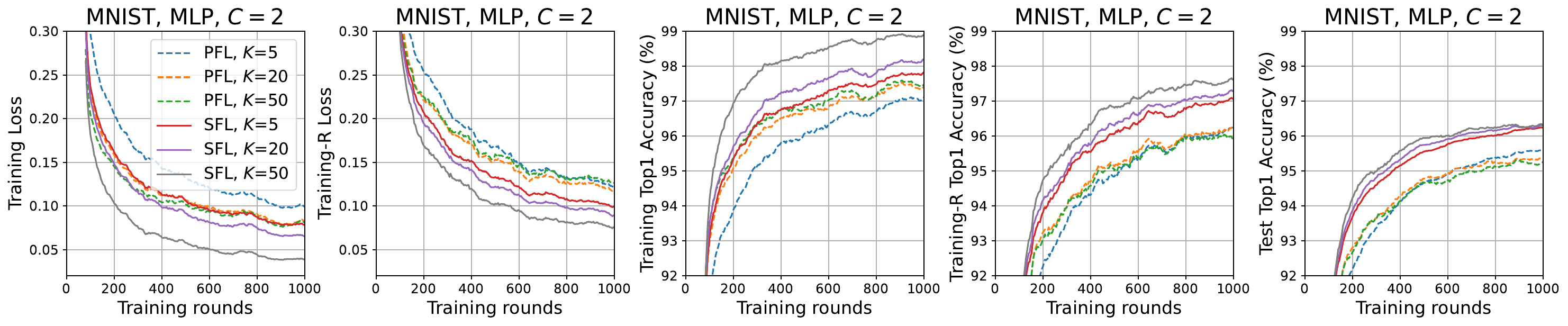}
	\includegraphics[width=\linewidth]{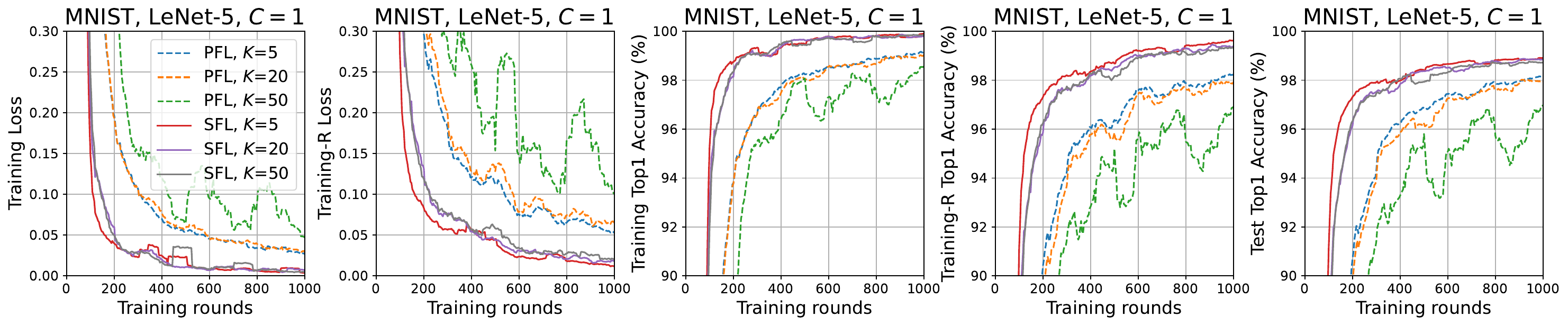}
	\includegraphics[width=\linewidth]{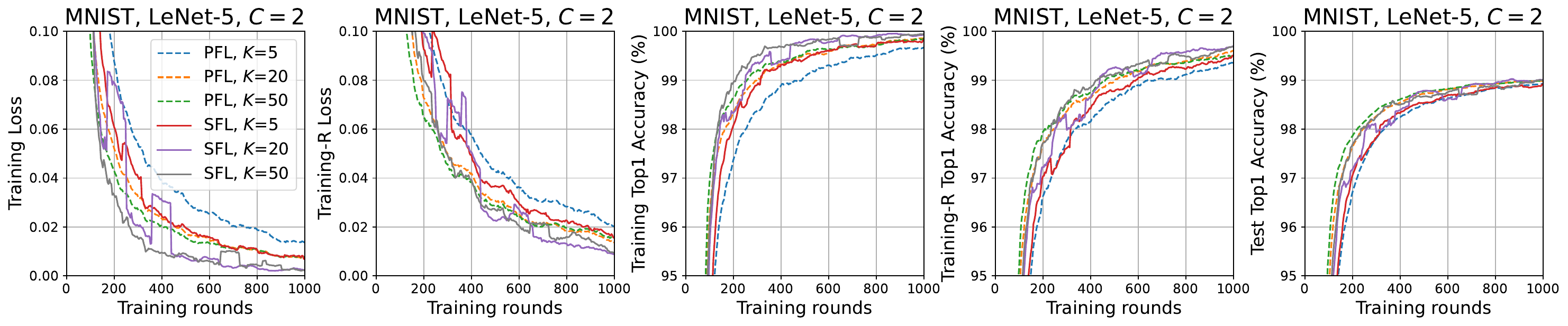}
	\caption{Experimental results on MNIST. For the best viewing experience, we apply moving average over a window length of 8\% of the data points. Note that ``Training Loss/Accuracy'' are computed over the training data of participating clients, ``Training-R Loss/Accuracy'' are computed over the training data of random clients and ``Test Accuracy'' are computed over the original test set.}
	\label{fig:app:mnist exp}
\end{figure}

\begin{figure}[htbp]
	\centering
	\includegraphics[width=\linewidth]{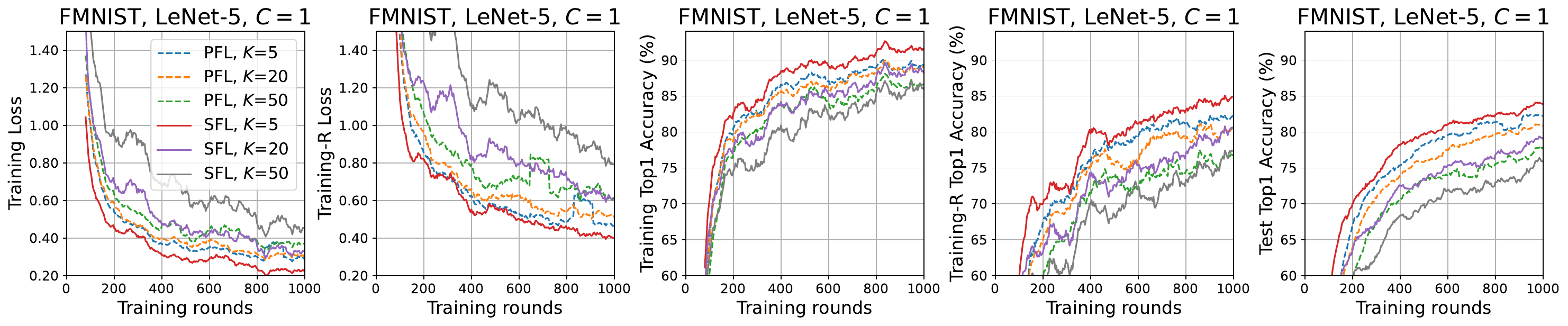}
	\includegraphics[width=\linewidth]{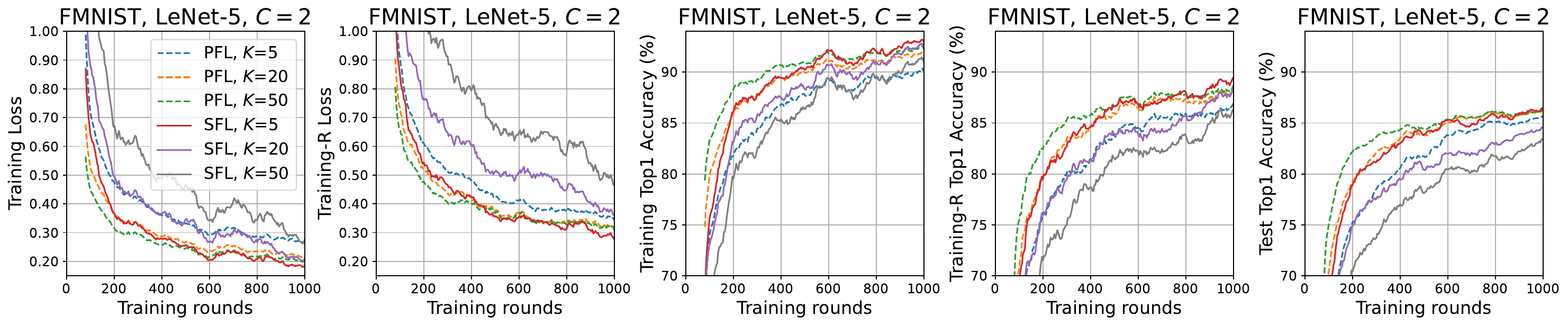}
	\includegraphics[width=\linewidth]{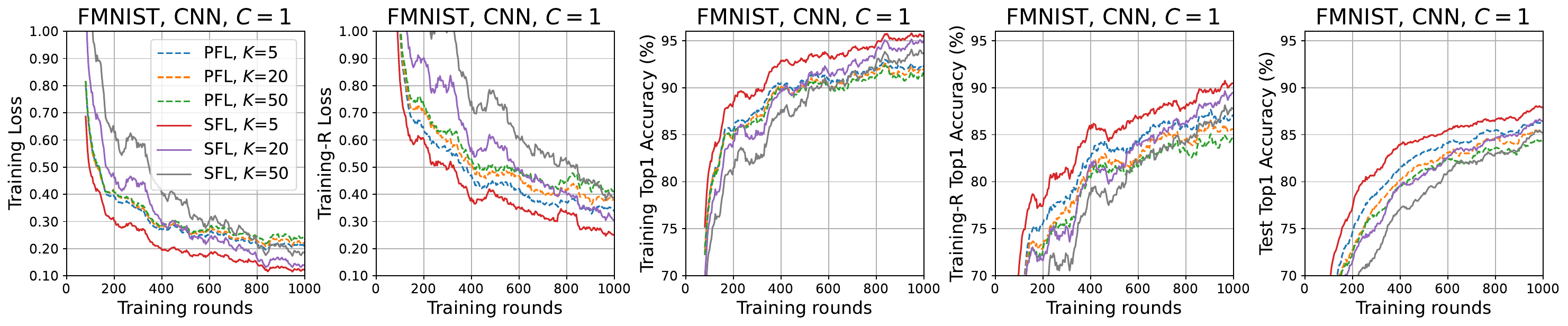}
	\includegraphics[width=\linewidth]{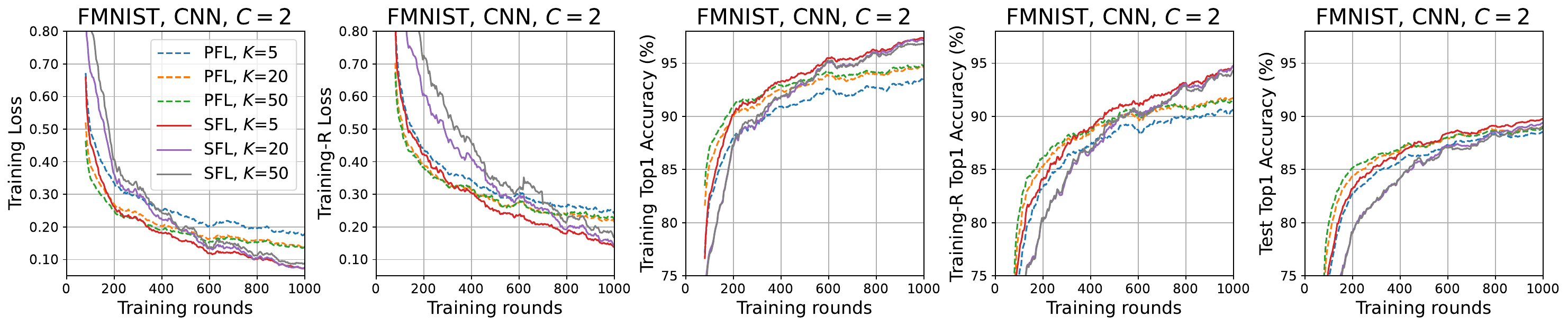}
	\caption{Experimental results on FMNIST. For the best viewing experience, we apply moving average over a window length of 8\% of the data points. Note that ``Training Loss/Accuracy'' are computed over the training data of participating clients, ``Training-R Loss/Accuracy'' are computed over the training data of random clients and ``Test Accuracy'' are computed over the original test set.}
	\label{fig:app:fashionmnist exp}
\end{figure}

\textbf{Results of MNIST and FMNIST.} The results of these five tasks are in Figures~\ref{fig:app:mnist exp}, \ref{fig:app:fashionmnist exp} and Table~\ref{tab:app:cross-device settings}. In the tasks MNIST/MLP, MNIST/LeNet-5, FMNIST/CNN, the performance of SFL is better when $C=1$, which is consistent with the analysis in \cref{subsec:cross-device exp}. However, we note that SFL shows worse even when $C=1$ in simpler tasks MNIST/Logistic and FMNIST/LeNet-5, especially when $K$ is large. This may be because the (objective function) heterogeneity on simple models and datasets is limited even with extreme data distributions (i.e., $C=1$). Thus, more extensive experiments are still required before drawing conclusions, which is beyond the scope of this paper.
\begin{table}[ht]
	\renewcommand{\arraystretch}{1}
	\centering
	\caption{Test accuracy results for various settings. We run PFL and SFL for 1000 training rounds for MNIST and FMNIST and 4000 training rounds for CIFAR-10 and CINIC-10 with 3 different random seeds. The results are computed over the random seeds and the last 40 training rounds for MNIST and FMNIST and the last 100 training rounds for CIFAR-10 and CINIC-10. The better results (with more than 1\% advantage for MNIST/FMNIST and 2\% advantage for CIFAR-10/CINIC-10) between \textcolor{teal}{PFL} and \textcolor{magenta}{SFL} in each setting are marked in color.}
	\label{tab:app:cross-device settings}
	\setlength{\tabcolsep}{0.15em}{
		\resizebox{\linewidth}{!}{
			\begin{tabular}{llllll@{\hspace{1em}}lll}
				\toprule
				\multicolumn{3}{c}{\multirow{1}{*}{Setup}} &\multicolumn{3}{c}{$C=1$} &\multicolumn{3}{c}{$C=2$} \\\midrule
				Dataset &Model &Method  &$K=5$ &$K=20$ &$K=50$ &$K=5$ &$K=20$ &$K=50$ \\\midrule
				\multirow{6}{*}{MNIST} 
				&\multirow{2}{*}{Logistic} 
				&PFL &91.10\tiny{$\pm$0.35} &\textcolor{teal}{89.46}\tiny{$\pm$1.20} &\textcolor{teal}{87.82}\tiny{$\pm$1.98} &91.69\tiny{$\pm$0.17} &\textcolor{teal}{91.19}\tiny{$\pm$0.47} &\textcolor{teal}{90.46}\tiny{$\pm$0.86} \\
				& &SFL &91.09\tiny{$\pm$0.67} &87.11\tiny{$\pm$2.09} &80.94\tiny{$\pm$3.70} &91.89\tiny{$\pm$0.32} &90.16\tiny{$\pm$1.10} &86.52\tiny{$\pm$2.41} \\[2ex]
				&\multirow{2}{*}{MLP} 
				&PFL &93.61\tiny{$\pm$1.42} &91.84\tiny{$\pm$2.20} &90.27\tiny{$\pm$3.17} &95.65\tiny{$\pm$0.38} &95.46\tiny{$\pm$0.47} &95.34\tiny{$\pm$0.51} \\
				& &SFL &\textcolor{magenta}{95.91}\tiny{$\pm$0.33} &\textcolor{magenta}{95.90}\tiny{$\pm$0.44} &\textcolor{magenta}{95.25}\tiny{$\pm$0.76} &96.26\tiny{$\pm$0.22} &96.35\tiny{$\pm$0.22} &\textcolor{magenta}{96.35}\tiny{$\pm$0.32} \\[2ex]
				&\multirow{2}{*}{LeNet-5} 
				&PFL &98.21\tiny{$\pm$0.40} &98.02\tiny{$\pm$0.63} &97.21\tiny{$\pm$1.55} &98.94\tiny{$\pm$0.09} &98.97\tiny{$\pm$0.10} &98.98\tiny{$\pm$0.11} \\
				& &SFL &98.90\tiny{$\pm$0.18} &98.87\tiny{$\pm$0.19} &\textcolor{magenta}{98.79}\tiny{$\pm$0.19} &98.91\tiny{$\pm$0.11} &99.07\tiny{$\pm$0.12} &98.99\tiny{$\pm$0.10} \\
				\midrule
				\multirow{4}{*}{FMNIST} 
				&\multirow{2}{*}{LeNet-5} 
				&PFL &82.57\tiny{$\pm$2.03} &\textcolor{teal}{81.09}\tiny{$\pm$3.19} &\textcolor{teal}{78.22}\tiny{$\pm$4.38} &85.86\tiny{$\pm$0.87} &\textcolor{teal}{86.35}\tiny{$\pm$1.12} &\textcolor{teal}{86.58}\tiny{$\pm$0.88} \\
				& &SFL &\textcolor{magenta}{83.97}\tiny{$\pm$2.42} &79.39\tiny{$\pm$2.59} &76.21\tiny{$\pm$2.95} &86.52\tiny{$\pm$1.67} &84.69\tiny{$\pm$2.26} &83.58\tiny{$\pm$2.55} \\[2ex]
				&\multirow{2}{*}{CNN} 
				&PFL &86.61\tiny{$\pm$1.62} &85.40\tiny{$\pm$2.07} &84.62\tiny{$\pm$2.18} &88.61\tiny{$\pm$0.93} &89.16\tiny{$\pm$0.77} &89.10\tiny{$\pm$0.89} \\
				& &SFL &\textcolor{magenta}{88.03}\tiny{$\pm$1.28} &\textcolor{magenta}{86.75}\tiny{$\pm$1.39} &85.44\tiny{$\pm$1.66} &\textcolor{magenta}{89.91}\tiny{$\pm$0.96} &89.60\tiny{$\pm$1.01} &89.05\tiny{$\pm$1.27} \\
				\midrule
				\multirow{4}{*}{CIFAR-10} 
				&\multirow{2}{*}{VGG-9} 
				&PFL &67.61\tiny{$\pm$4.02} &62.00\tiny{$\pm$4.90} &45.77\tiny{$\pm$5.91} &78.42\tiny{$\pm$1.47} &78.88\tiny{$\pm$1.35} &78.01\tiny{$\pm$1.50} \\
				& &SFL &\textcolor{magenta}{78.43}\tiny{$\pm$2.46} &\textcolor{magenta}{72.61}\tiny{$\pm$3.27} &\textcolor{magenta}{68.86}\tiny{$\pm$4.19} &\textcolor{magenta}{82.56}\tiny{$\pm$1.68} &\textcolor{magenta}{82.18}\tiny{$\pm$1.97} &79.67\tiny{$\pm$2.30} \\[2ex]
				&\multirow{2}{*}{ResNet-18} 
				&PFL &52.12\tiny{$\pm$6.09} &44.58\tiny{$\pm$4.79} &34.29\tiny{$\pm$4.99} &80.27\tiny{$\pm$1.52} &82.27\tiny{$\pm$1.55} &79.88\tiny{$\pm$2.18} \\
				& &SFL &\textcolor{magenta}{83.44}\tiny{$\pm$1.83} &\textcolor{magenta}{76.97}\tiny{$\pm$4.82} &\textcolor{magenta}{68.91}\tiny{$\pm$4.29} &\textcolor{magenta}{87.16}\tiny{$\pm$1.34} &\textcolor{magenta}{84.90}\tiny{$\pm$3.53} &79.38\tiny{$\pm$4.49} \\
				\midrule
				\multirow{4}{*}{CINIC-10} 
				&\multirow{2}{*}{VGG-9} 
				&PFL &52.61\tiny{$\pm$3.19} &45.98\tiny{$\pm$4.29} &34.08\tiny{$\pm$4.77} &55.84\tiny{$\pm$0.55} &53.41\tiny{$\pm$0.62} &52.04\tiny{$\pm$0.79} \\
				& &SFL &\textcolor{magenta}{59.11}\tiny{$\pm$0.74} &\textcolor{magenta}{58.71}\tiny{$\pm$0.98} &\textcolor{magenta}{56.67}\tiny{$\pm$1.18} &\textcolor{magenta}{60.82}\tiny{$\pm$0.61} &\textcolor{magenta}{59.78}\tiny{$\pm$0.79} &\textcolor{magenta}{56.87}\tiny{$\pm$1.42} \\[2ex]
				&\multirow{2}{*}{ResNet-18} 
				&PFL &41.12\tiny{$\pm$4.28} &33.19\tiny{$\pm$4.73} &24.71\tiny{$\pm$4.89} &57.70\tiny{$\pm$1.04} &55.59\tiny{$\pm$1.32} &46.99\tiny{$\pm$1.73} \\
				& &SFL &\textcolor{magenta}{60.36}\tiny{$\pm$1.37} &\textcolor{magenta}{51.84}\tiny{$\pm$2.15} &{\color{magenta}{44.95}}\tiny{$\pm$2.97} &\textcolor{magenta}{64.17}\tiny{$\pm$1.06} &\textcolor{magenta}{58.05}\tiny{$\pm$2.54} &\textcolor{magenta}{56.28}\tiny{$\pm$2.32} \\
				\bottomrule
	\end{tabular}}}
\end{table}
\begin{figure}[htbp]
	\centering
	\includegraphics[width=\linewidth]{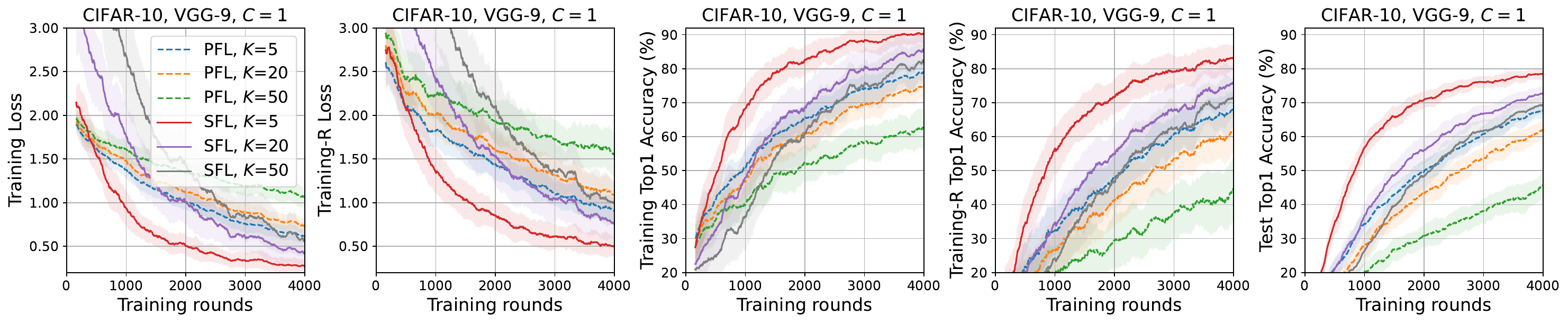}
	\includegraphics[width=\linewidth]{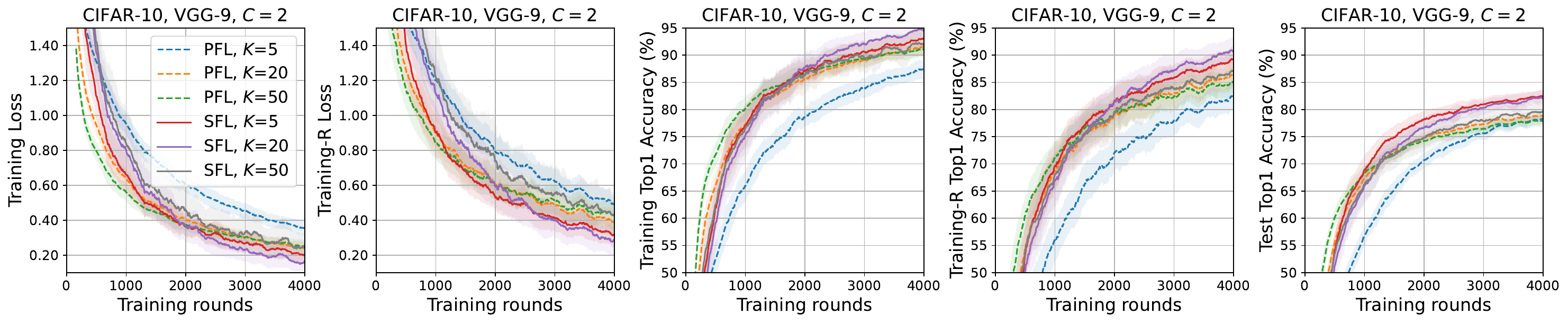}
	\includegraphics[width=\linewidth]{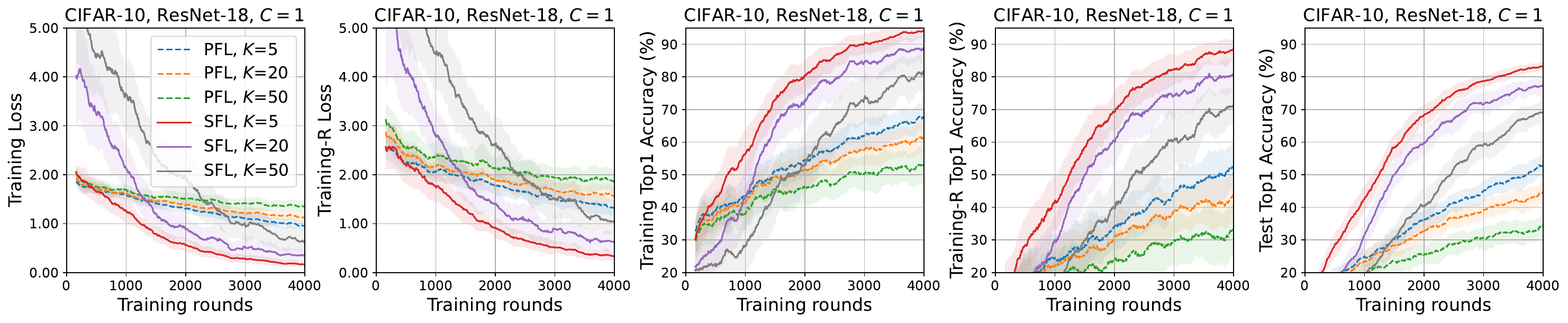}
	\includegraphics[width=\linewidth]{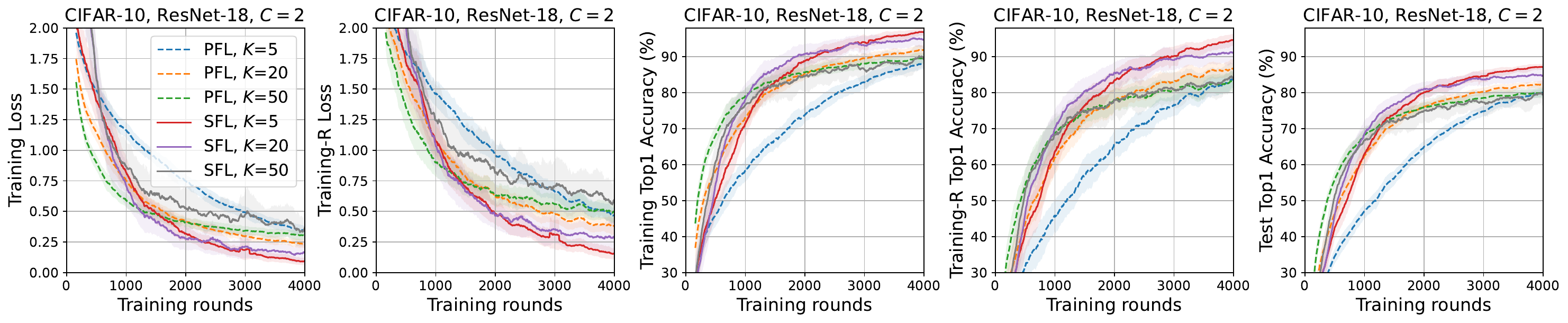}
	\caption{Experimental results on CIFAR-10. For the best viewing experience, we apply moving average over a window length of 4\% of the data points. Note that ``Training Loss/Accuracy'' are computed over the training data of participating clients, ``Training-R Loss/Accuracy'' are computed over the training data of random clients and ``Test Accuracy'' are computed over the original test set. The shaded areas show the standard deviation with 3 random seeds.}
	\label{fig:app:cifar10 exp}
\end{figure}
\begin{figure}[htbp]
	\centering
	\includegraphics[width=\linewidth]{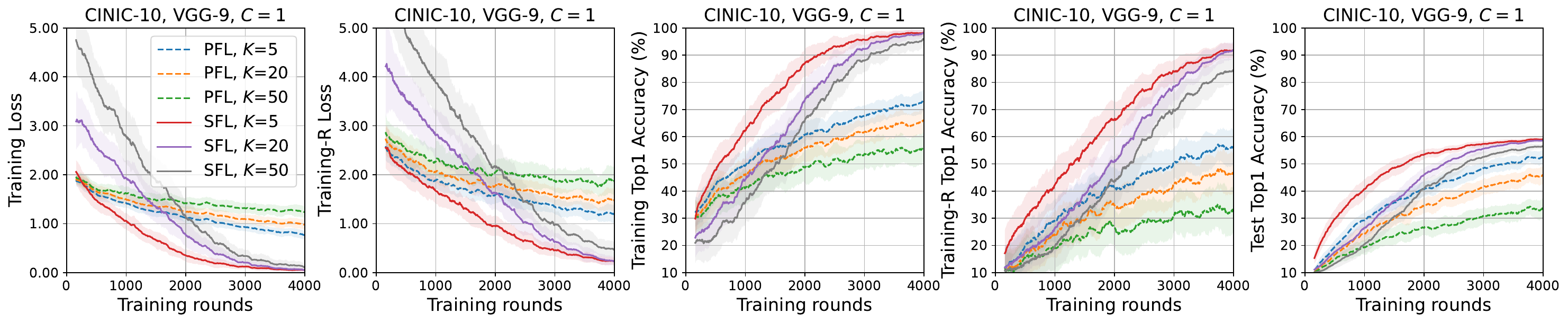}
	\includegraphics[width=\linewidth]{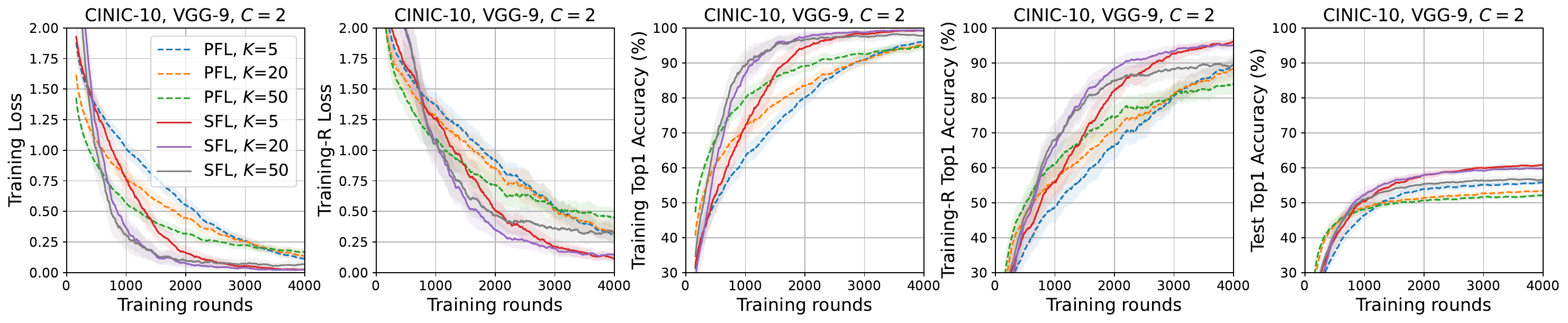}
	\includegraphics[width=\linewidth]{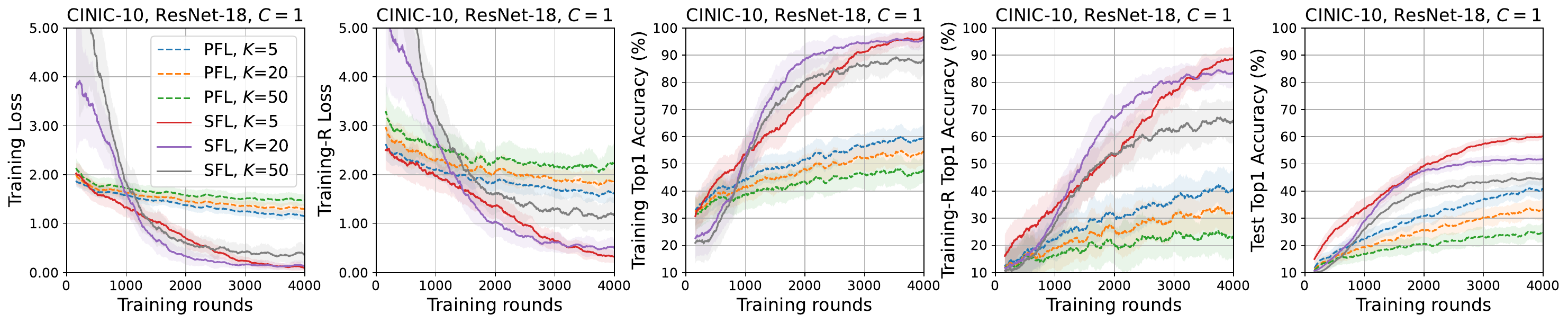}
	\includegraphics[width=\linewidth]{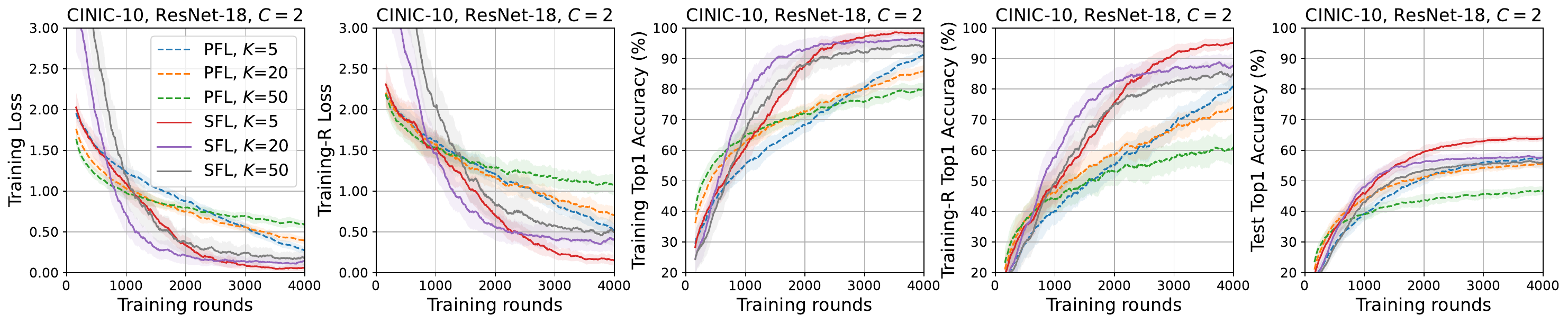}
	\caption{Experimental results on CINIC-10. For the best viewing experience, we apply moving average over a window length of 4\% of the data points. Note that ``Training Loss/Accuracy'' are computed over the training data of participating clients, ``Training-R Loss/Accuracy'' are computed over the training data of random clients and ``Test Accuracy'' are computed over the original test set. The shaded areas show the standard deviation with 3 random seeds.}
	\label{fig:app:cinic10 exp}
\end{figure}

\end{document}